\documentclass[12pt,fleqn]{ucithesis}

\usepackage{amsmath}
\usepackage{amsthm}
\usepackage{array}
\usepackage{graphicx}
\usepackage{natbib}
\usepackage{relsize}
\usepackage{color}
\usepackage{pdflscape}
\usepackage{amssymb}
\usepackage{array}
\usepackage{booktabs}
\usepackage{epigraph}
\usepackage{algorithmic}
\usepackage{mdframed}
\usepackage{xcolor}
\usepackage{booktabs}
\usepackage{lscape}
\usepackage{longtable}
\usepackage{graphics}
\usepackage[normalem]{ulem}
\useunder{\uline}{\ul}{}
\usepackage{wrapfig}

\usepackage{caption}
\usepackage{subcaption}  
\usepackage{multirow}
\usepackage{tabularx}
\usepackage{latexsym, array,epsfig, comment , MnSymbol, multirow, enumitem, verbatim, booktabs, amsfonts, wasysym}

\usepackage{mwe}

\usepackage{graphicx}
\usepackage[plainpages=false,pdfborder={0 0 0}]{hyperref}

\usepackage{algorithm}
\renewcommand{\listalgorithmname}{\protect\centering\protect\Large LIST OF ALGORITHMS}

\thesistitle{Health State Estimation}

\degreename{Doctor of Philosophy}

\degreefield{Computer Science}

\authorname{Nitish Nag}

\committeechair{Professor Ramesh C. Jain}
\othercommitteemembers
{
  Professor Michael J. Carey\\
  Professor Gopi Meenakshisundaram
}

\degreeyear{2020}

\copyrightdeclaration
{
  {\copyright} {\Degreeyear} \Authorname
}

\dedications
{ \textit {To my inspiring advisor Ramesh Jain, exceptional parents Bishwajit and Sesha, my illuminating partner Tabya, and the shoulders of all the giants we stand upon.}
}

\acknowledgments
{
Professor Ramesh Jain has been the single most influential figure in my academic development, not just as a graduate student, but in my ability to think about life in its entirety. His aspirations of changing the world in a meaningful way have been vital in the development of my thought processes. We both share this same vibration together and thus have been exceptionally synergistic in our dream of making society a truly healthier and happier place. Our relationship has been multi-faceted in persevering towards this aim. I am jovial to know that will continue in the coming years. Ramesh's approach to always have fun, love what you do, and fuel the best intentions reverberate in my mind daily. I aspire to pass forward his level of enthusiasm and optimism to future generations beyond my own life. Through challenges and breakthroughs, Ramesh has always been a genuine best friend and trusted mentor. Every moment I was able to spend with him since the first day I walked into his office is treasured. I look forward to continuing the growth in our relationship long into the future and thank him for his unwavering belief in me.

Next, I would like to thank my thesis committee members, Professor  Gopi Meenakshisundaram and Professor Michael J. Carey. Professor Meenakshisundaram has been thoughtful and supportive in shaping this work towards helping humanity. Professor Carey has equally been thoughtful in questioning the technical approach to make this work polished and have greater clarity.

I want to thank my lab mates, Vaibhav Pandey, Preston Putzel, Hari Bhimaraju, Siripen Pongpaichet, Hyungik Oh, Mengfan Tang, and Pranav Agrawal for the stimulating discussions and their support of my work. Additionally, I would like to thank Srikanth Krishnan, Daniel Azzam, Matthew Wade, Mars Wang, and Abhisaar Sharma for their collaboration towards advancing ideas in this thesis. I want to extend my appreciation for Vaibhav's close cooperation in a large portion of the research progress presented, and for his companionship as a labmate.

My parents and brother have been the backbone of my upbringing through their unwavering support. Most importantly, they have always given me the freedom and means to pursue my passions freely and to the best of my ability. I hope to embody their virtues as I continue in life. My partner Tabya Sultan has been one of the most significant influences in my total lifespan. She has shaped much of the thoughts in this work. Her strength in believing in my potential is unrivaled, and I kindle this strength in my pursuit of excellence.

Finally, I want to extend my gratefulness for everyone who has impacted me in my life journey to this point. Each one of these people will always be a part of shaping who I am and will be. Thank you.

}

\newcommand{\mypubentry}[3]{
  \begin{tabular*}{1\textwidth}{@{\extracolsep{\fill}}p{4.5in}r}
    \textbf{#1} & \textbf{#2} \\ 
    \multicolumn{2}{@{\extracolsep{\fill}}p{.95\textwidth}}{#3}\vspace{6pt} \\
  \end{tabular*}
}

\curriculumvitae
{
\textbf{EDUCATION}
  
  \begin{tabular*}{1\textwidth}{@{\extracolsep{\fill}}lr}
    \textbf{Doctor of Medicine} & \textbf{2023*} \\
    \vspace{6pt}
    University of California, Irvine (UCI) \textit{*expected} & \emph{Irvine, CA} \\
    \textbf{Doctor of Philosophy in Computer Science} & \textbf{2020} \\
    \vspace{6pt}
    University of California, Irvine (UCI) & \emph{Irvine, CA} \\
     \textbf{Private Pilot} & \textbf{2013} \\
    \vspace{6pt}
    California Airways & \emph{Hayward, CA} \\
    \textbf{Bachelor of Science in Nutrition Physiology and Metabolism} & \textbf{2011} \\
    \vspace{6pt}
    University of California, Berkeley (UCB) & \emph{Berkeley, CA} \\
    \textbf{Bachelor of Arts in Integrative Biology} & \textbf{2011} \\
    \vspace{6pt}
    University of California, Berkeley (UCB) & \emph{Berkeley, CA} \\
    \textbf{Bachelor of Arts in Molecular and Cell Biology: Biochemistry} & \textbf{2011} \\
    \vspace{6pt}
    University of California, Berkeley (UCB) & \emph{Berkeley, CA} \\
  \end{tabular*}

\vspace{12pt}

\textbf{RESEARCH EXPERIENCE}
  
  \begin{tabular*}{1\textwidth}{@{\extracolsep{\fill}}lr}
    \textbf{Graduate Student Researcher} & \textbf{2015--2020} \\
    \vspace{6pt}
    University of California, Irvine (UCI) \textit{*expected} & \emph{Irvine, CA} \\
    \textbf{Scientist} & \textbf{2011--2013} \\
    \vspace{6pt}
    Renovel Discoveries Inc. & \emph{Fremont, CA} \\
     \textbf{Undergraduate Researcher} & \textbf{2009--2011} \\
    \vspace{6pt}
     UCB, \textit{Hellerstein Metabolism Lab} & \emph{Berkeley, CA} \\
     \textbf{Undergraduate Researcher} & \textbf{2009--2011} \\
    \vspace{6pt}
     UCB, \textit{Brooks Exercise Physiology Lab} & \emph{Berkeley, CA} \\
     \textbf{Undergraduate Researcher} & \textbf{2007--2009} \\
    \vspace{6pt}
     UCB, \textit{Sul Nutrition Lab} & \emph{Berkeley, CA} \\
     \textbf{Research Assistant} & \textbf{2004--2007} \\
    \vspace{6pt}
     Bexel Pharmaceuticals Inc. & \emph{Union City, CA} \\
  \end{tabular*}

\pagebreak

\textbf{PUBLICATIONS}

\mypubentry{A Novel Epidemiological Approach to Geographically Mapping Dry Eye Disease in the United States through Google Trends}{2020}{Daniel Azzam, Nitish Nag, Julia Tran, Lauren Chen, Kaajal Visnagra, Matthew Wade, American Society of Cataract and Refractive Surgery}

\mypubentry{Continuous Health Interface Event Retrieval}{2020}{Vaibhav Pandey, Nitish Nag, Ramesh Jain, Proceedings of the 2020 ACM on International Conference on Multimedia Retrieval}

\mypubentry{A Navigational Approach to Health: Actionable Guidance for Improved Quality of Life}{2019}{Nitish Nag, Ramesh Jain, IEEE Computer}

\mypubentry{Flavour Enhanced Food Recommendation}{2019}{Nitish Nag, Aditya Narendra Rao, Akash Kulhalli, Kushal Samir Mehta, Nishant Bhattacharya, Pratul Ramkumar, Aditya Bharadwaj, Dinkar Sitaram, Ramesh Jain, Proceedings of the 5th International Workshop on Multimedia Assisted Dietary Management}

\mypubentry{HealthMedia'19: 4th International Workshop on Multimedia for Personal Health and Health Care}{2019}{Susanne Boll, Jeannie S Lee, Jochen Meyer, Nitish Nag, Noel E O'Connor, Proceedings of the 27th ACM International Conference on Multimedia}

\mypubentry{Synchronizing Geospatial Information for Personalized Health Monitoring}{2019}{Nitish Nag, Vaibhav Pandey, Likhita Navali, Prateek Mohan, Ramesh Jain, arXiv preprint arXiv:1907.10594}

\mypubentry{Cross-Modal Health State Estimation}{2018}{Nitish Nag, Vaibhav Pandey, Preston J Putzel, Hari Bhimaraju, Srikanth Krishnan, Ramesh Jain, 2018 ACM Multimedia Conference on Multimedia Conference}

\mypubentry{Ubiquitous Event Mining to Enhance Personal Health}{2018}{Vaibhav Pandey, Nitish Nag, Ramesh Jain, Proceedings of the 2018 ACM International Joint Conference and 2018 International Symposium on Pervasive and Ubiquitous Computing and Wearable Computers}

\mypubentry{Surface Type Estimation from GPS Tracked Bicycle Activities}{2018}{Nitish Nag, Vaibhav Pandey, Aishwarya Manjunath, Avinash Vaka, Ramesh Jain, arXiv preprint arXiv:1809.09745}

\mypubentry{Intrinsic and Extrinsic Motivation Modeling Essential for Multi-Modal Health Recommender Systems}{2018}{Nitish Nag, Mathias Lux, Ramesh Jain, arXiv preprint arXiv:1808.06467}

\mypubentry{Endogenous and Exogenous Multi-Modal Layers in Context Aware Recommendation Systems for Health}{2018}{Nitish Nag, Vaibhav Pandey, Ramesh Jain, arXiv preprint arXiv:1808.06468}

\mypubentry{Live Personalized Nutrition Recommendation Engine}{2017}{Nitish Nag, Vaibhav Pandey, Ramesh Jain, Proceedings of the 2nd International Workshop on Multimedia for Personal Health and Health Care}

\mypubentry{Pocket Dietitian: Automated Healthy Dish Recommendations by Location}{2017}{Nitish Nag, Vaibhav Pandey, Abhisaar Sharma, Jonathan Lam, Runyi Wang, Ramesh Jain, International Conference on Image Analysis and Processing}

\mypubentry{Health Multimedia: Lifestyle Recommendations Based on Diverse Observations}{2017}{Nitish Nag, Vaibhav Pandey, Ramesh Jain, Proceedings of the 2017 ACM on International Conference on Multimedia Retrieval}

\mypubentry{Cybernetic Health}{2017}{Nitish Nag, Vaibhav Pandey, Hyungik Oh, Ramesh Jain, arXiv preprint arXiv:1705.08514}

\mypubentry{Fibroblast Growth Factor 21 Is Not Required for the Reductions in Circulating Insulin-Like Growth Factor-1 or Global Cell Proliferation Rates in Response to Moderate Calorie Restriction in Adult Mice}{2014}{Airlia CS Thompson, Matthew D Bruss, Nitish Nag, Alexei Kharitonenkov, Andrew C Adams, Marc K Hellerstein, PloS one}

}

\thesisabstract
{
Life's most valuable asset is health. Continuously understanding the state of our health and modeling how it evolves is essential if we wish to improve it. Given the opportunity that people live with more data about their life today than any other time in history, the challenge rests in interweaving this data with the growing body of knowledge to compute and model the health state of an individual continually.

This dissertation presents an approach to build a personal model and dynamically estimate the health state of an individual by fusing multi-modal data and domain knowledge. The system is stitched together from four essential abstraction elements: 1. the events in our life, 2. the layers of our biological systems (from molecular to an organism), 3. the functional utilities that arise from biological underpinnings, and 4. how we interact with these utilities in the reality of daily life. Connecting these four elements via graph network blocks forms the backbone by which we instantiate a digital twin of an individual. Edges and nodes in this graph structure are then regularly updated with learning techniques as data is continuously digested. Experiments demonstrate the use of dense and heterogeneous real-world data from a variety of personal and environmental sensors to monitor individual cardiovascular health state.

State estimation and individual modeling is the fundamental basis for many revolutionary efforts in health. Continuous access to the individual's state allows a departure from disease-oriented approaches, to a total health continuum paradigm. Precision in predicting health requires a detailed understanding of an individual and their trajectory. By encasing this state estimation within a navigational approach, a user can use a systematic guidance framework to plan actions for each moment to transition their current state towards a desired one. This work concludes by presenting this framework of combining the health state and personal graph model to perpetually plan and assist us in living life towards our goals.
}

\hypersetup{
	pdftitle={\Thesistitle},
	pdfauthor={\Authorname},
	pdfsubject={\Degreefield},
}

\newcounter{infobox}
\renewcommand{\theinfobox}{\arabic{infobox}}
\newenvironment{infobox}[1][]{%
    \refstepcounter{infobox}
    \begin{mdframed}[%
        frametitle={Box \theinfobox: #1},
        backgroundcolor=black!6,
        nobreak=true,
    ]%
}{%
    \end{mdframed}
}

\begin{document}

\preliminarypages

\chapter{Introduction}

\epigraph{It is health that is real wealth and not pieces of gold and silver.}{\textit{Mahatma Gandhi}}
\vspace{12pt}


Vibrant health lays the foundation of high-quality living. If we define quality as a function of positive experiences over time, then we want to optimize those experiences and extend their duration. Any life experiences rely upon a subset of general abilities to move, think, sense, and maintain homeostasis of our biology. These abilities arise from a complex interactive network of molecular machines that are always in flux. Maintaining and expanding the capacity of these abilities paves the way for enhancing health and, therefore, quality of life. When these abilities suffer from dysfunction, we realize we must take care of them. Rather than address health when a person has already fallen ill, the challenge for the next century and beyond is to ensure long lives of high performing health for each individual, thus enabling people to flourish in their life goals.

Traditionally, health has been viewed with the lens of avoiding sickness rather than pushing towards health optimization. There are exceptions with athletics and military, where improved health status directly equates towards the success in the field. Why have the health sciences generally placed a large number of resources in research and application towards avoiding sickness rather than optimizing health? Based on the historical trend of communicable disease, it was \textit{was} a great need.

\begin{figure}[H]
  \centering \centerline{\epsfig{figure=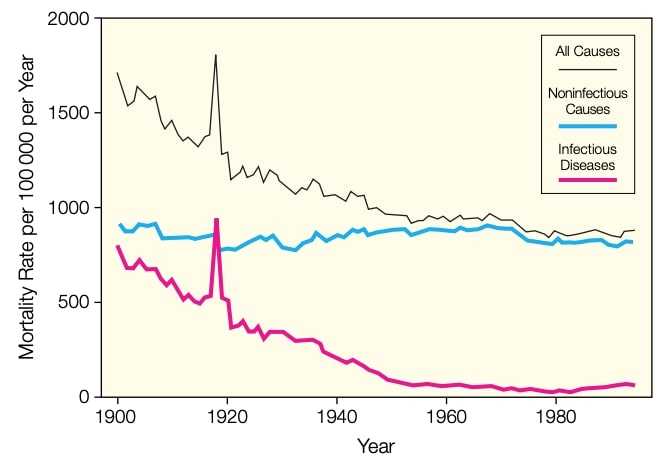,width=14cm}}
  \caption{\textbf{Crude Mortality Rates for All Causes.} In the latter half of the twentieth century, the non-infectious disease continues to be a significant cause of mortality, reflecting the need for new strategies in health sciences.}
    \label{fig:mortality}
\end{figure}

Until recently in human history, a significant contributor to mortality and morbidity was infectious disease and injury. Progress in health sciences during the last century resulted in a significant decrease in communicable diseases, as shown in Figure \ref{fig:mortality}. Antibiotics reflected one side of the coin, by treating patients in a reactive manner, once an infection had already become a health concern. On the flip side, advances in preventing infections through population-level data offered insight into ways to improve public hygiene and contain infectious agents, while vaccines biologically prevented infections. Continuing progress in battling infectious diseases today remains a valiant effort in sciences and medicine, but it cannot scale as a general approach for all human health. Infection is mostly an episodic problem, in which diagnoses are made after a patient is under attack and arrives at a medical facility. Treatment is usually prescribed based on evidence-based rules to solve the problem, and monitoring of the patient no longer occurs. The process described is an \textbf{episodic} approach towards an individual's health. Episodic refers to \textit{occasional measurement and awareness of health state} and actions to change the state towards a better one.

Globally chronic diseases, which include cardiovascular disease (CVD), cancers, chronic respiratory disease, and diabetes, have emerged as the primary contributor to degrading quality of health. We see this including impoverished areas of the world, with 80 percent of chronic disease burden occurring in developing countries \cite{Abegunde2007TheCountries}. The WHO explains that all regions of the world have a similar relationship between the major modifiable risk factors and the main chronic diseases \cite{Organization2020FacingFactors}. Modern medicine relies mostly on reactive health systems that take care of individuals only after they become sick due to the established paradigm of episodic medicine. This means that healthcare professionals are systematically responding to acute health problems and are symptom-focused. The overall health is not a priority within healthcare systems \cite{GovernmentofOntario2007PreventingCare}. Prescribing medicine is the hallowed strategy medical experts have taken toward treating disease, just as they approached infectious disease. It is necessary to move beyond this episodic health model to tackle chronic disease and significantly improve quality of life. When comparing chronic disease to infectious disease, there are fundamental differences in how the situation arises. There is no single event that leads to the disease, rather slow changes in the state of the body. Risk factors, including raised blood pressure and glucose levels, abnormal blood lipids, and weight (overweight and obesity), express the causes of chronic diseases. Major modifiable risk factors alongside non-modifiable risk factors (including age and genetics) are the root causes of these diseases over time \cite{Organization2020FacingFactors}. If we are to apply a control input to keep people on a healthy trajectory via feedback loops, we must be able to estimate the health state continuously \cite{Wiener1962CyberneticsMachine}. Control systems inherently need to know the difference between the current state versus the goal state. This approach is taken in acute moments in the hospital after entering an undesirable health state, such as during anesthesia or Intensive Care Unit (ICU) monitoring. Once again, these are episodic moments in a person's life. Hence there is a clear distinction between episodic versus life-long continuous estimation problems in health.

\begin{figure}[H]
  \centering \centerline{\epsfig{figure=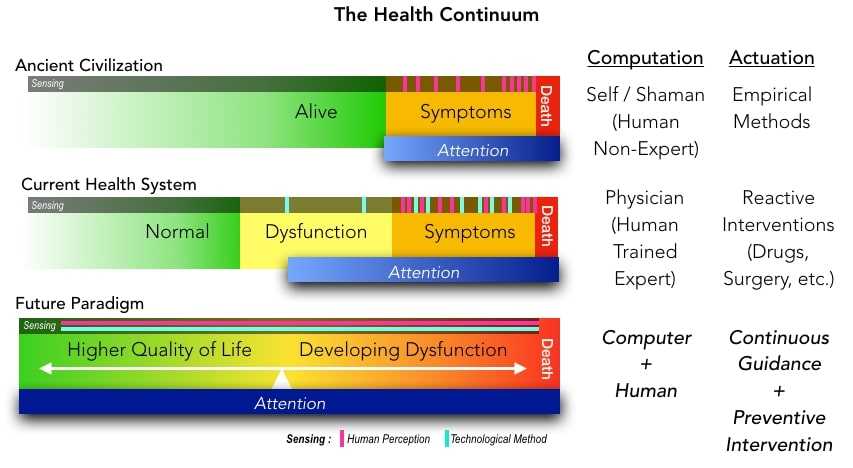,width=16cm}}
  \caption{\textbf{The Health Continuum.} Attention to health status occurs when there is a detectable signal. Ancient civilizations could only use subjective symptoms and signs as an indicator of the health state. Today, we have episodic measurements of a few criteria to understand our health state, such as blood pressure or lipids on an annual basis. Beyond those measurements, we still primarily rely on symptoms as the primary signal. Both of these signals are sparse, and most individuals do not even collect symptom data. The next revolution is to have continuous sensing of the health state so we can always have attention to improving health through the means of ubiquitous computing and modeling at the individual scale.}
    \label{fig:continuum}
\end{figure}

Health is a dynamic state that is continually changing based on a specific individual's biology, environment, and lifestyle. The state space of health is complex and multidimensional, but for the sake of simplicity, in Figure \ref{fig:continuum}, we describe a one-dimensional continuum ranging from optimal health to death. Everyone is somewhere on this continuum at every moment, but the quality of life increases if there is a shift towards the optimal (left or green side). Technology that senses a user's position on the health continuum can drastically improve human health. In prehistoric times, only symptoms (perceptual indications) guided people's actions. Ancient civilizations started to document medical conditions with basic human sensory abilities, such as how people visually appeared, their pulse rhythm, or coughing sounds. In clinical language, symptoms (what a person feels) or signs (what is observable by another person) are unfortunately latent signals of an already dysfunctional biological state. Today, advanced methods detect changes in health, but only in a professional health-care setting after patient visits due to severe enough symptoms or on an annual basis. Reducing the lag time in recognizing health state changes requires transitioning from the current measurement approach to a new high-resolution multi-modal continuous-sensing paradigm.

A general characteristic of humans and computing systems is the intelligibility and clarity of understanding a situation, allowing for the best long term, planned decisions for a future outcome. In today's society, we see this clearly in economics. The tangible and quantitative nature of financial capital allows humans or computers to make the decisions they see fit for their intentions and plan for a quality of life they aspire to have. The difference between finances and health is that live quantitative health status is virtually invisible to us, especially when we are healthy. In financial matters, people can quantitatively see how much they are saving or spending at any given moment. Market data updates every second across the globe. If people become aware of their biology, they will see how daily life is affecting their health and make informed decisions accordingly. Health decisions are too important to be episodic. They must be an intrinsic part of daily life, giving us the motivation behind proposing the state estimation and navigation perspective in this work.

Ultimately, individuals, clinicians, and in general cybernetic systems (Figure \ref{fig:continuum}), make decisions based on determining the health state of the individual. Because people are unable to feel their health change over time from the multitude of factors affecting them, we need to develop methods to quantify and report health status using continuously collected multi-modal data sources. If we can track changes before biological dysfunction, we may be able to correct course, prevent, or delay the onset of chronic diseases. We can use the same principle to change that health state towards a more optimal direction if the user wishes.

Computing efforts regarding state estimation in other fields have advanced society drastically in the last several decades. For example, the Global Positioning System (GPS) launched in 1973 by the United States military for location state estimation. It was opened to the public in 1983 by a bill signed by President Ronald Reagan with limited functionality and was subsequently offered with full, accurate functionality by a bill signed in 2000 by President Bill Clinton \cite{McDuffie2017WhyPublic}. In the last 20 years, this allowed for systems to take a user's location to provide them with services such as understanding their position on a map, tagging photos, providing navigation, tracking their movements, hailing a taxi, or ordering pizza to the home doorstep. The state space of location included four dimensions of latitude, longitude, altitude, and time. Our work aims to lay the groundwork for similar tasks in the context of health.

So how can we compute the health state continuously? The primary fuel arises from the diverse data streams produced about an individual daily. With the rapidly increasing availability of low-cost sensors in the last decade, there has been an explosion in the amount of continuously collected multi-modal data. We see the relevance in the field of health with the advent of wearable, Internet of Things (IoT), and ambient sensors, human-computer interactions, images, video, audio, and other digital sensors. Many of these low-cost sensors produce continuous streams of data usable in a wide range of scenarios. The challenge rests in leveraging this data to provide perpetual insight into the health of individuals. While this allows measurement of the user state continuously, the downside of these sensors is that measurements can be very noisy, cause information overload, produce non-actionable metrics, or are not directly related to the attribute we want to measure. These limitations have so far proven to be a barrier in using low-cost sensors for real-world health state estimation (HSE) and decision making for health. Furthermore, enabling data assimilation will require a causal understanding of how sensor information relates to health status. To transform the data into usable information, existing medical and biological scientific domain knowledge must be used.

There are many challenges in providing a high-resolution continuous and individualized state estimation for health. This dissertation aims to establish the preliminary groundwork to address issues for the general health state concept and methods by which to apply it at an individual scale. 

In the next section, we will look at the thought process transition from traditional medicine to state-of-the-art approaches towards improving health and state estimation, giving insight as to what components are emerging needs in computing the health state.

\section{Overview of Health and Medicine}
The history of significant trends are important for us to understand health today and its future developments, especially in understanding patterns that cause progress. From ancient times until today, humans pay most attention to their health with the lens of healing from illness. From a traditional point-of-view, the health of the individual only became a concern when negative impact could be perceived, causing the development of systems that operate in the following steps. First, the individual acts as the primary sensor that their health state is not the same as earlier. Second, a doctor or domain expert would conduct further tests to gain an understanding of the causal reason behind the change in the health state. Third, the doctor or health expert would prescribe medication or procedure that would hopefully return the individual to their former health state. After this point, the individual is no longer monitored until then next issue arises. The impacts on the health state occur due to several critical situations. Situations that are significant impacts on health are listed below. Each situation has a particular state estimation challenge and a unique set of actions to keep health and homeostasis.

\begin{enumerate}
    \item \textbf{Infections} have historically been a threat to health. The resulting invasion from different species battles with natural immune defenses, medical agents such as antibiotics, or physical removal.
    \item \textbf{Cancer} occurs due to errors in the genetic information of a cell. There is a natural rate of error for DNA replication enzymes, which causes errors during each division of a cell \cite{GSBrush1996DNAMechanisms}, a component of natural life and the process by which evolution occurs. We, therefore, see age as a factor for cancer. Additional errors in the DNA structure happen by insults from carcinogens, inflammation, or stress.
    \item \textbf{Physiological function} of the person determines their operation from cellular to organismal level. This includes the pumping of the heart to supply blood flow, the strength of bones to provide support, the absorption of nutrition in the digestive system, and the capacity to see. Many chronic diseases are a result of physiological dysfunction.
    \item \textbf{Injury} can occur from either chronic issues or acute events. Overuse or improper use may eventually result in injury. Acute injuries are caused by unintentional accidents and or intentional combat. The severity can range from mild to traumatic.
    \item \textbf{Neurological and mental health} is a complex topic and traverses a broad range of sciences. Inherently there are some components of this aspect that are objective and others that are subjective, leading to unique challenges in scientific progress.
    \item \textbf{Developmental} disorders are caused by various factors determined by cellular growth from fertilization until adulthood. As examples, these can be encoded in genetic errors (chromosomal errors at birth), cell signaling biology (receptor dysfunction), or deficiency during growth (low bone mass from low dietary calcium).
\end{enumerate}

Before the twentieth century, infectious disease was a key determining factor for mortality. With emerging infections, finding a cure, or reactively treating the disease that had already occurred became commonplace. This is why health was primarily determined by the absence of illness or disease \cite{Armstrong1999TrendsCentury}. The progress in the dramatic reduction of infectious disease came through significant advances in state detection and actionable steps. The state detection improved through the use of culturing the infection and microscopy to identify the pathogen. Culturing amplifies the signal from the infectious agent. We use these techniques even today, although further advances in gene sequencing and other assays promise to speed diagnosis time and precision. Actionable steps to address the infection are usually antibiotics or antiviral medications.

\begin{figure}[H]
  \centering \centerline{\epsfig{figure=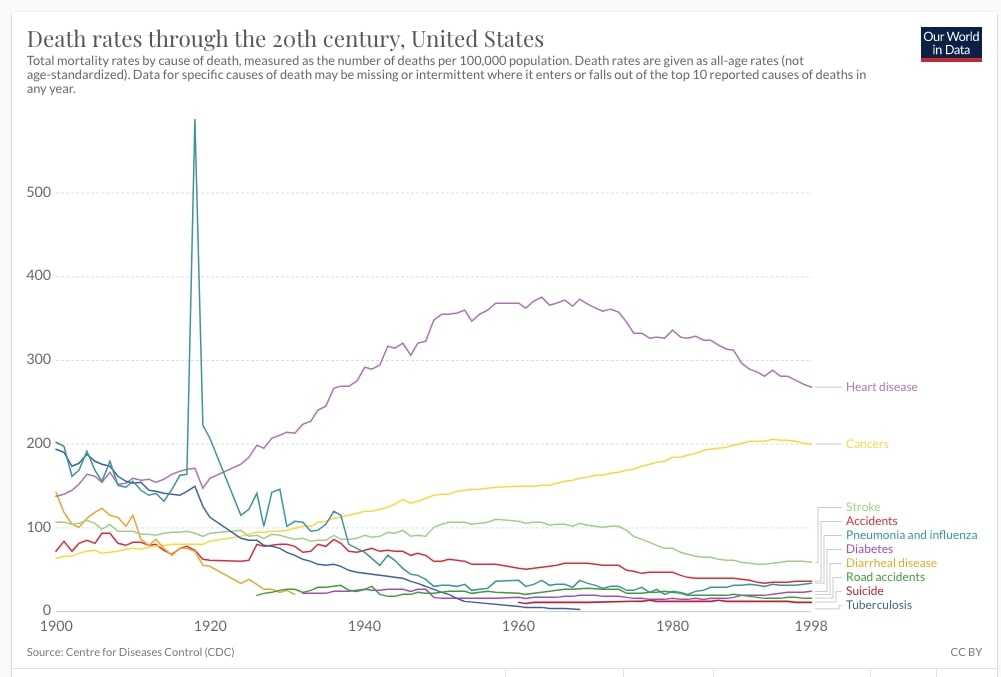,width=16cm}}
  \caption{\textbf{Death Rates.} This graph shows the death rates in the 20th Century United States \cite{MaxRoser2019CancerData}.}
    \label{fig:chronic disese}
\end{figure}

Adversarial pathogens became less of a fixture in medical research over the last century, and chronic diseases became the center point of health challenges. These chronic diseases are primarily a result of physiological function change. Currently, six in 10 adults in the United States have a chronic disease; 4 in 10 adults have two or more. This makes chronic disease the leading cause of death and disability \cite{NationalCenterforChronicDiseasePreventionandHealthPromotion2019ChronicAmerica}. The major chronic disease that dominated epidemiological data since 1940s was and continues to be cardiovascular diseases as shown in Figure \ref{fig:chronic disese}, with 647,000 Americans dying from heart disease each year (or 1 in 4 deaths) \cite{CDC2020HeartCdc.gov}. Heart disease is, therefore, the leading cause of death in the United States. These above situations are mostly modeled using large-scale population data by aggregating individuals to produce statistically significant outcomes due to the lack of high-resolution personal data. Thus the state estimation for chronic disease is based on a population model that predicts the health outcomes of an individual. Measurements on the individual are taken generally on an annual basis to update the state. Medications are given to individuals if the state changes to a new classification, given population model thresholds. Overall for chronic disease, patient state estimation is conducted through general modeling, updated on an episodic basis.

While genetics and mutations play a role, CVD and a significant portion of cancers are largely a function of external factors that can be addressed by factors extrinsic to the biological risk, such as lifestyle and the environment. Cancer develops from both random intrinsic DNA replication errors (or ``bad luck") and from factors not related to genetics. Cancer causation is multi-factorial and the modification of risk factors is critical in cancer prevention \cite{Wu2018EvaluatingFactors}. If external and environmental risk factors dominate, then lifestyle choice, which reduces risk exposure, is crucial for this reduction. One-third of cancers are attributable to environmental factors or inherited predispositions \cite{Tomasetti2015VariationDivisions}. The approach to chronic diseases has been similar to infectious diseases in that treatment has been a reaction to symptoms. Introductions to medication continue; however, the rate of disease decrease has stabilized, as shown in Figure \ref{fig:chronic disese}. This motivates the need for more robust methods to address these diseases than sole reliance on medication developments and reactive approaches for addressing when an individual is already in a state of disease. Furthermore, tracking the lifestyle and environmental events give substantial clues into the evolving health state of the person in addition to medications or procedures that assist in reducing risk for disease. In summary, the actionable items to address the chronic disease are widespread and occurring at every moment of the day.

Health science disciplines usually still focus on these reactive approaches today. Integrating the knowledge from this is essential, as the illness lens highlights regions of our health state that are undesirable. This thesis work will primarily focus on physiological function concerning cardiovascular function and disease, as it is the primary determinant of major chronic diseases. 

\section{Modern Trends in Health Sciences}
There is a need for better state estimation and actionable steps for future progress. As reactive methods for addressing disease is no longer sufficient for a global decrease in rates of chronic disease, we examine new models. These approaches require a true understanding of the individual state of health and predicting and preventing entry to poor health state regions. Personalized methods also acknowledge that individual state must be uniquely observed, requiring high quality and quantity of data about an individual.

Projections show that medicine and the health sciences will move from a reactive to a proactive discipline over the next few decades — a discipline that is predictive, preventive, precise, personalized, and participatory \cite{Hood2011PredictiveMedicine,Sagner2017TheHealthspan}. There must be a holistic approach to disease, emerging technologies, and analytical tools to ensure proactive advancements. Holistic approaches refer to the total connected network of the individual's health. Emerging technologies include the proliferation of sensors and computing power, and analytical tools refer to the manipulation of large amounts of information to produce meaningful and actionable insight. Other advances we will also review include developments the growing wellness industry which are currently outside the scope of the healthcare industry. Wellness focuses on moving a patient towards the optimal state of health, as shown in the future health continuum in Figure \ref{fig:continuum}. In contrast, healthcare focuses on moving the patient from the red area of dysfunctional health to the level of mediocre health. We will take a deeper dive into these new approaches as they will be an integral part of the design requirements of this research.

\subsection{Personalized Health}
\begin{figure}[H]
  \centering \centerline{\epsfig{figure=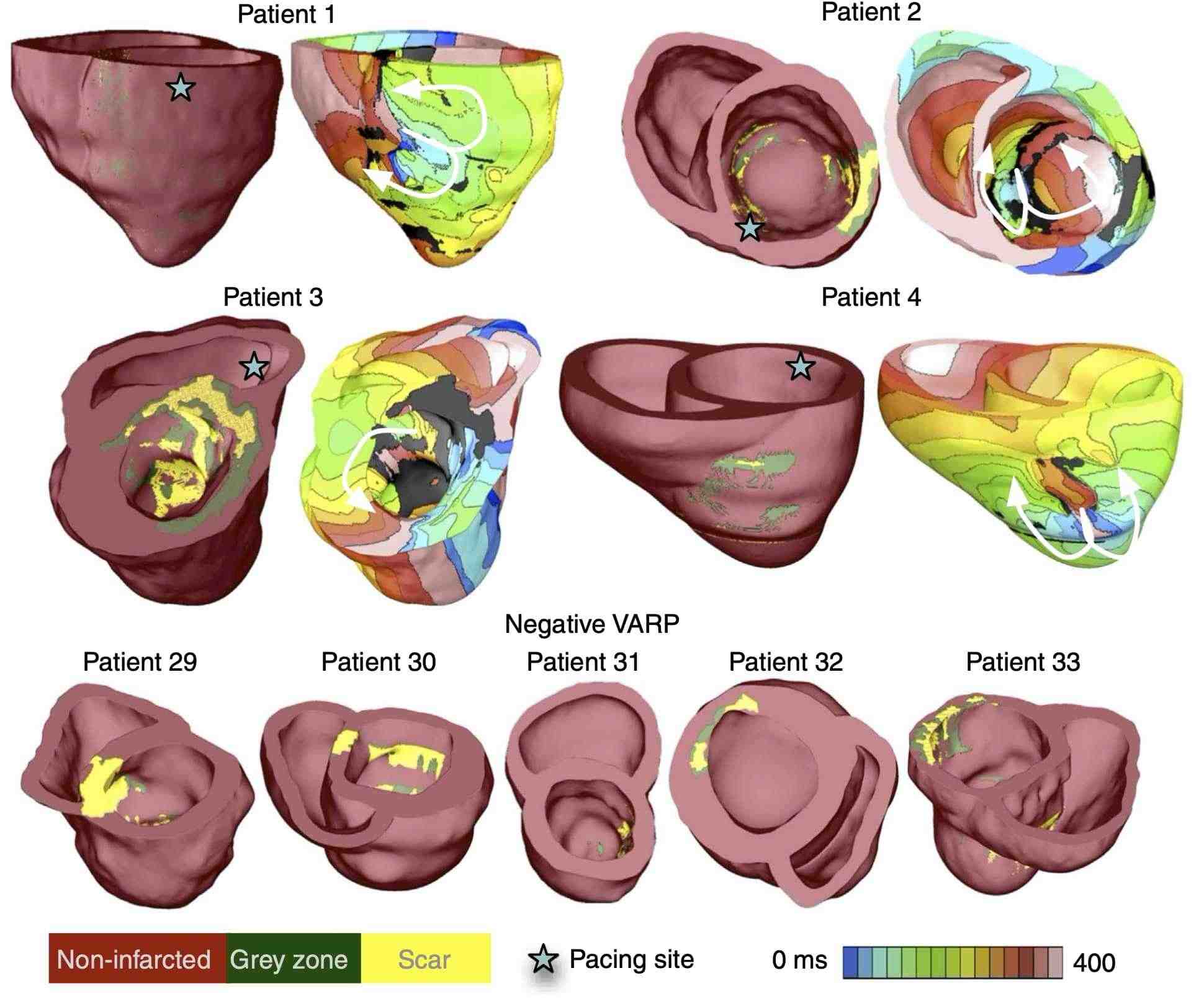,width=14cm}}
  \caption{\textbf{Personalized Heart Models.} Patients’ clinical magnetic resonance imaging data constructs personalized three-dimensional computer models of post-infarction hearts. These are assessed for the propensity of each model to develop arrhythmia. In a proof-of-concept retrospective study, the virtual heart test significantly outperformed several existing clinical metrics in predicting future arrhythmic events. The non-invasive personalized virtual heart risk assessment is a form oh highly specialized individual health state modelling\cite{Arevalo2016ArrhythmiaModels}.}
    \label{fig:arevalo}
\end{figure}

\textit{Personalized health refers to building individual models to represent the health state, and tailoring inputs for a specific individual.} Personalization includes two key sub-models. The first is understanding of an individual's preferences and life situation, encapsulating the subjective attributes and autonomy of the user. On the other hand personalization for the biological machinery is an objective model. As an example, in Figure \ref{fig:arevalo}, the individual heart electrical patterns are modeled \cite{Arevalo2016ArrhythmiaModels}. Data from ``omic" sources are incorporated to understand the complexity of an individual's biology, which includes genetics, genomics, microbiome, proteomics, and metabolomics \cite{Nicholson2005GutCare}. The word ``oma" refers to a mass in Latin, and in the context of omic data refers to the large mass of information. There must be a combination of both of these models (subjective and objective) for many use cases. Examples include food preferences for a diet change and exercise preferences for modulating cardiovascular health.

Personalized medicine, which is more specific in the treatment of disease, is used in the context of focusing on a person's unique molecular profile to help them be less susceptible to disease. Genetics plays a large role in personalized medicine to understand one's unique traits and shape the biological model through this knowledge. The Human Genome Project had an impact on producing treatments that could be better tailored to the individual. Reading the sequence of the DNA will allow one to predict how an individual will respond to a particular drug and to receive earlier diagnoses and risk assessments for more appropriate treatment \cite{Vogenberg2010PersonalizedTheranostics}. Objective personalization alone is not enough, as the person must be part of the whole loop \cite{Coote2015IsWorld}. The integration of contextual components relevant to the user's wishes and preferences is just as important. As mentioned earlier, augmenting the data and model with subjective data from the user also has unique challenges. From this model, we can create predictions, precisely guide the individual, and prevent entry into undesirable regions of the health state space.

\subsection{Predictive Health}
\textit{Prediction refers to estimating the probability of a future event.} Improvements in this require the above personal model. Predictive medicine is a branch of medicine that focuses on risk identification for a disease, such as a future unwanted biological event, including a heart attack. The goal is to identify patients at risk that could then enable prevention or early treatment. This can be done by either single or multiple analyses as markers for future disease \cite{Bach1996PredictiveMedicine}. Risk calculation of particular diseases and obtaining characteristics of individuals could help medical professions to predict a patient's risk for an outcome of interest and also determine what treatment may be most effective. The Emory Predictive Health and Society Strategic Initiative and its Center for Health Discovery and Well Being define health in the context of the entire human experience. They identify measurable variables that describe and predict a healthy state and to use that knowledge to design health-focused interventions that are affordable and effective \cite{Brigham2010PredictiveCare}. Predictive health utilizes data sources from a diverse array including genomics and epidemiology in order to best predict the risks for an individual. From a computational perspective, this requires data mining to find patterns that pull in knowledge from population and individual components, while also employing learning techniques for data-driven pattern discovery.

\subsection{Precision Health}
\textit{Precision health refers to more accurate approaches to understanding the health state and optimizing inputs to produce the ideal state.} For example, while doctors may inquire about family history of disease to understand hereditary risk, a more precise approach would be to sequence one's DNA in order to analyze the genetics data for inherited disease risk. This greater detail in understanding allows for a more targeted approach in the estimation and input selection. Precision will contribute to improving HSE and inputs through various avenues that we will discuss below. 

\begin{itemize}
\item \textbf{Biological} understanding with more precision, allows for better state estimation and input approaches. At an organ level, each individual has unique biology. In the case of cardiovascular function, models shown in Figure \ref{fig:arevalo} are also precise, in that they try to pinpoint precisely the nature of the cardiovascular electrical activity \cite{Arevalo2016ArrhythmiaModels}. At a molecular level, the details of signaling pathways and cellular processes allow for better health and medicine. A classic example in medicine is in the Cyclooxygenase (COX) inhibitor drug class. Ibuprofen is a drug that inhibits all COX enzymes, which help in reducing pain but also cause problems for COX enzymes, which are part of general functions in the gut. The classification of Ibuprofen is thus a non-specific COX inhibitor. The discovery of COX-1 and COX-2 specific sub-types gave rise to a new class of medications that are specific for COX-2 (the component responsible for pain), allowing for more the development of more advanced pain medications (Figure \ref{fig:cox2}). This improvement in the precision of molecular targets continues to be an active area of science. Lastly, genetics greatly influences precise and personal biological understanding, which we will review next.

\begin{figure}[H]
  \centering \centerline{\epsfig{figure=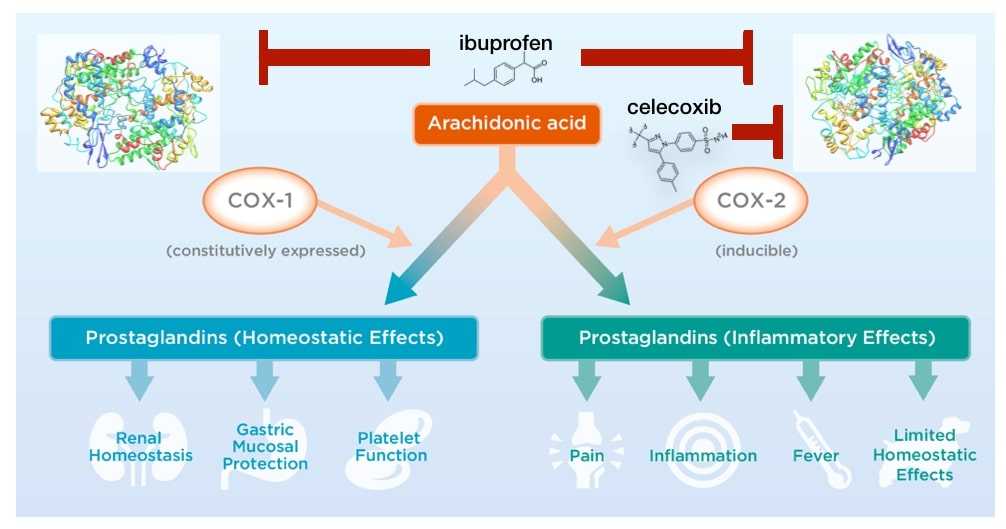,width=16cm}}
  \caption{\textbf{Precision in Biological Understanding.} Selective COX-2 inhibitors like celecoxib are a type of Non-Steroidal Anti Inflammatory Drugs (NSAID) that directly targets COX-2, an enzyme responsible for inflammation and pain. Targeting selectivity for COX-2 reduces the risk of peptic ulceration and is the main feature of the drug celecoxib. This type of precision arises from more in-depth biological understanding.}
  \label{fig:cox2}
\end{figure}

\item \textbf{Genetics} plays a fundamental role in having precise information for molecular and cellular modeling and targeting for an individual. Even more cutting edge in this field is the use of immune therapy, in which a person's own white blood cells are trained to precisely identify and destroy cancer cells in the body rather than delivering poisons that impact both healthy and cancer tissue \cite{Wolchok2009GuidelinesCriteria}. Furthermore, targeting specific antibodies can more precisely treat an autoimmune disease. These are examples of biological targeting using genetics. Risk modelling and predictive efforts may also use genetics. For example, the BRCA gene is a determinant of breast cancer risk. Patients having high-risk BRCA genes may elect to have mastectomies or get screened more frequently, given this genetic risk. In summary, genetics allows for many aspects of state estimation or technologies to provide a more accurate outcome for an individual.

\item \textbf{Location} precision on the body is a general concept that may improve many areas of health. The traditional crude approach is to give a general pain killer medication that is distributed to the whole body via the blood and has adverse side effects due to the lack of a specific target. A precision approach would understand where the inflammation is in the knee and deliver a medication targeted to only the knee tissue and molecular pathways of relevance to the disease pathology. For example, a cream or injection is more precise than an oral pill. Topical NSAIDs may be better for localized pain near the surface of the skin, such as pain in the hands and knees, while oral NSAIDs may be better for pain deep within the body. Similarly, location for surgery can be more precise using laparoscopic methods, rather than an invasive open procedure. Precision in surgical methods also enables providers to be more accurate in their procedures, making fewer mistakes and damage to healthy tissue. Laser cataract surgery is needed to remove a cataract. In this procedure, a small probe is inserted into the eye, emitting ultrasound waves to break up the cataract into pieces. In laser surgery procedures, lasers are used to make incisions to create a more precise opening to the lens capsule \cite{Center2020LaserFAQs}.

\item \textbf{Dosage} of inputs, such as drugs, can be more precise to the needs of an individual. Context, the personal model, and the components mentioned above characterize precise needs. When an individual is in high altitude, there is an adverse effect on coagulation in patients taking Warfarin \cite{TissotvanPatot2006RiskM}. If the dosage of the medication could be controlled depending on the location of the individual, this could be more precise in controlling the coagulation more appropriately. Cytochrome P450 is an enzyme that metabolizes the breakdown of drugs; individuals have different rates of breakdown depending on their specific enzyme type. Knowing the exact sensitivity and breakdown methods will better allow for precise drug dosage. Extending beyond drugs, we can understand there is a precise quantity of inputs that produce the optimal outcome of the individual's health state. Knowing the precise amount for each input (such as food, exercise, drugs, environment, sleep) is a precision challenge.

\end{itemize}

Precision medicine is a specific subset of precision health focused on optimizing the inputs to better alter the health state. Medicine is an input that is purposefully designed to change the health state in a certain circumstance. The NIH defines precision medicine by how genetics, environment, and lifestyle can help determine the best approach to prevent or treat disease \cite{NIH2020WhatReference}. More medical institutions such as Stanford Medicine are using precision medicine via unique biology and life circumstances for diagnostics \cite{Ritcher2015PrecisionIt}. Former President Barack Obama launched the Precision Medicine Initiative to accelerate these advancements, emphasizing how precision medicine will enable health care providers to tailor treatments to better suit patients through the use of gene sequencing and lifestyle factors. To get there, we need data about the patient collected by health professionals and the patients themselves \cite{House2020PrecisionInitiative}. From a computational perspective, this requires an in-depth database for capturing the relevant domain knowledge that matches precisely with each individual's context and personal model. Capturing this high resolution of data and having the individual take action on the best inputs requires them to be a part of the cybernetic loop.

\subsection{Participatory Health}
\textit{Participatory health refers to an individual directly engaging with understanding their health state and providing inputs for change.} Traditionally healthcare providers were the key individuals who understood patient health state and decided how to alter it. The providers and patients generally held all information had little interaction with their health state (other than through following directions on medications). However, more recently, healthcare providers and patients are working as partners in health by using communication tools to increase active patient participation in medical decisions. This approach uses physical and physiological tracking devices, ``easy-to-use software that records one's data, analyzes them, interprets them graphically, and allows one to share and compare them with others. This is expected to enable more reliable self-monitoring and more self-management of health care" \cite{Martin-Sanchez2013BiomedicalHealth}. Ultimately, participatory medicine refers to the patient taking actions that change their health state, rather than depending upon the healthcare team to address the changing health state of an individual. From a computational perspective, this requires development in human-computer-interaction (HCI) for engagement, integrating into relevant services for the user, and learning patterns from the event streams, which increase the probability of the user completing specific actions. 

\subsection{Preventive Health}
\textit{Prevention refers to altering the probability of a future event / or avoid region of interest.} Preventive health behavior has been described in 1966 by Kasl and Cobb, as ``any activity undertaken by a person who believes himself to be healthy for the purpose of preventive disease or detecting disease in an asymptomatic stage, in contrast with illness behavior for an individual who is ill, defining the stage of health and discovering remedy" \cite{Kasl1966HealthBehavior.}. Today, preventive medicine is a medical specialty recognized by the American Board of Medical Specialties (ABMS), focusing on promoting health and well-being and preventing disease, disability, and death. Physicians in the preventive medicine specialty practice in the sub-specialty areas of public health and general preventive medicine, occupational medicine, and aerospace medicine. While officially, preventive medicine is practiced medically with these specific guidelines, the field of preventive medicine continues to ameliorate individual risk for disease by altering the probability of future risk. It represents a health state as a compilation of risks and seeks actions that reduce this risk based on evidence. Based on the predictive component, preventive measures are taken, such as the common practice of reducing risk factors for heart disease. These risk factors include lowering cholesterol, blood pressure, and other modifiable components. Knowing the heart health state, however, requires much more depth. From a computational perspective, this requires mapping the regions which are undesirable in the health state space, understanding the user health state, and providing inputs that increase the distance to the undesirable regions.

\subsection{Healthcare Systems}
Healthcare defined in this work includes all services and products that are within the scope of health insurance coverage. The economic size of this field globally is 7.3 trillion US Dollars \cite{GlobalWellnessInstitute2016StatisticsInstitute}. This field is evolving to encompass new frontiers that will impact individual health. The healthcare industry is attempting to address the complexity of individual health via technological advancements such as digital therapeutics, pharmacy supply chain, virtual primary care, artificial intelligence, genomics, and advancements in consumer health and wellness. There are also developments for providers including diagnostics, and drug research and development to ultimately enhance their services \cite{CBInsights2019Global2019}. Trends in the healthcare industry show that consumers have access to more personalized healthcare options that provide a concierge level of service and are expanding convenience through services like video conferencing. There is also greater transparency of healthcare services, given the level of engagement individuals are now having with the health system. All together, healthcare is growing in domains for the individual, providers, and systems at large. The primary computing advancements for HSE will be in using more advanced diagnostics that are too expensive or exclusive for the patient to procure on their own.

\subsection{Performance and Wellness}
A large developing area within modern approaches to holistic health is through the modes of ``wellness," the consideration of inputs of lifestyle, the environment, and social and perceptual factors. Together, these bring about a more robust consideration of health in daily living and behavioral considerations, and an improvement in health performance, rather than an emphasis on eliminating or reducing disease.  The economic size of this field globally is 4.5 trillion US Dollars \cite{GlobalWellnessInstitute2016StatisticsInstitute}. The wellness industry is targeting improvement in various components of an individual's life, as outlined by CBInsights \cite{CBInsights2019WellnessTech_MarketMap1.png1333888}. Many of the services provided involve tracking and developing methods for changing the health state. High growth areas which include areas for technological advancements for individual health include: 

\begin{itemize}
\item Personalized nutrition such as food and beverages, vitamins/supplements, and active nutrition.
\item Beauty and personal care.
\item Fitness and active living, including gym access applications, activity tracking, and measurement of physiological changes.
\item Mental wellness including behavioral change programs, educational platforms for mindfulness/meditation.
\item Corporate wellness as provided by an individual's employer, that provides comprehensive platforms for interactive with health such as by measuring, tracking, understanding and making changes to one's state in multiple domains of their health.
\item Sleep technologies to track, change behaviors, and understand the best methods for sleep improvement specific to an individual.
\item E-commerce for the sale of online health specific products.
\item Feminine care to address women's unique health needs.
\item Travel/hospitality to provide modes of experiencing wellness via retreats and getaways.
\item Clothing and fit unique for one's active performance.
\item Functional health or primary care medical practice with a holistic approach, combining nutrition, lifestyle, and medical testing.
\end{itemize}

New forms of wellness demonstrate the convergence of personalized medicine with digital health, artificial intelligence, systems biology, social networks, big data analysis, and precision medicine. This field is known as \textit{scientific wellness}. ``The idea of scientific wellness is a quantitative approach that includes improving the health of individuals, creating personalized treatments, reversing disease transitions and reducing costs — distinct from the current wellness trend focusing primarily on behaviors such as diet and lifestyle" \cite{Sullivan2018TheWellness}. People will then be able to use data and tools to modify habits for diet, exercise, sleep, and more to optimize personal health.

Looking historically and at current advancements, progress in health depends on estimating the health state in order to realize improvements in the treatment of disease or helping individuals live well. Thus the state estimation and health measurements over time are critical for scientific progress. 

\section{Health State Estimation is Critical for Progress}

Even though the World Health Organization (WHO) for 74 years has defined a preliminary health state, there continue to be inferior methods by which to understand the individual health state. The first page of Google Scholar with the search term ``health state estimation" (as of January 1st 2020) primarily displays results in the health state of batteries, drilling, machines, ball bearing, and aircraft engines. The challenge remains in developing a generalized framework for estimating the health state of humans. Health is estimated in most detail when we are in our worst state in the Intensive Care Unit (ICU) at a hospital. Humans doctors with minimal historical knowledge or elementary quantitative modeling of the individual are making judgments about the state of health for their health decisions. Using this judgment, they provide episodic actions to re-direct the health state to an acceptable level, not an optimal level.

Methods for measurement have been developed through time to estimate particular components of health more accurately and more often. Microscopy and the identification of microorganisms detected specific pathogens, highlighting the presence of what pathogens determined disease. The Framingham Heart Study, which started in 1948, has reviewed a diverse population over time to analyze patterns, trends, and outcomes that may apply to the general population. One finding was the discovery of cholesterol and its impact on CVD. From measurements at a population level, the relationship between cholesterol and the progression of CVD was established \cite{Institute2018FraminghamNHLBI}. Thus, cholesterol now has become an indicator of CVD and cardiorespiratory fitness (CRF), as described by the American Heart Association (AHA) \cite{Ross2016ImportanceAssociation,Lee2010MortalityFitness.}.

The more health estimation moves towards real-time, it will open doors for controlling health more precisely. For example, the first glucose monitor was introduced in the 1970s and then began the self-monitoring of blood glucose. Advancements in continuous glucose monitoring, which read blood glucose monitors more frequently, gave more precise information to manage diabetes. The accuracy of the sensors has been improving over time, and the frequency for measurements has increased \cite{HirschIrlB2018HistoryMonitoring}. Quantified Self, a movement with the motto “self-knowledge through numbers” embodies the importance of continuous health measurements. The movement aids individuals in using different technologies to collect data on their daily living (i.e., food intake, physiology) to analyze and improve their health state. Sensors are also more commonly used to collect genetic, environmental, and phenotype information and storage in the personal health record \cite{Martin-Sanchez2013BiomedicalHealth}. Objective Self takes the quantified self concept towards a more computationally robust representation of the individual by establishing a basic model about the individual \cite{jain2014objective}.

What we need is a means by which we can continuously know our health state at all times in high temporal and spatial resolution. A parallel example is accessing GPS location on-demand, anytime. In a total health state space, an individual can see where they are in their state space and choose their location within the state space. Disease is just a subset of this state space of regions that are undesirable, similar to not driving a car into the middle of the ocean. What one cares about are the \textit{places to be}, and the health state required to be there. Ultimately, why is the HSE needed? 

\begin{quote}
    \textit{Measurement is the first step that leads to control and eventually to improvement. If you can’t measure something, you can’t understand it. If you can’t understand it, you can’t control it. If you can’t control it, you can’t improve it.}
    \newline
    - H. James Harrington
\end{quote}

\section{Contributions}
\begin{figure}[H]
  \centering \centerline{\epsfig{figure=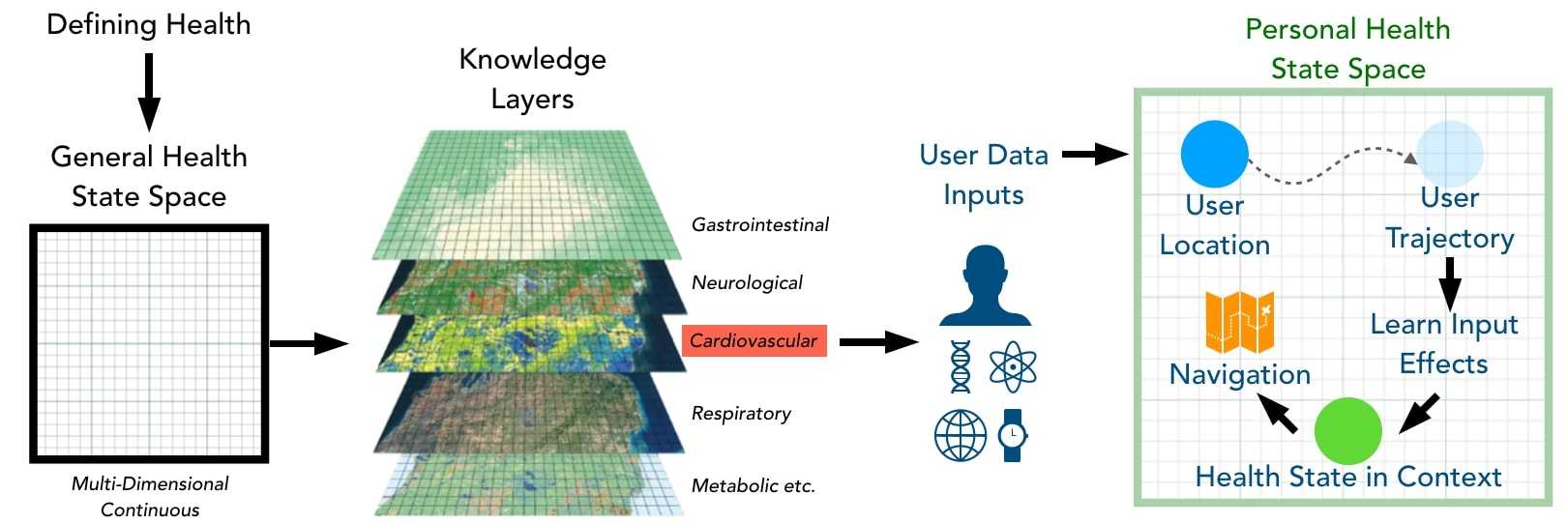,width=16cm}}
  \caption{\textbf{Visualization of Contributions.}}
    \label{fig:contributions}
\end{figure}

This dissertation contributes towards the advancement of dynamically computing the health state of an individual throughout every moment in life. The evolving state of the individual depends on the integration of both domain knowledge and data-driven discovery frameworks that facilitate computation of how data streams and event streams are related. The connected nature of life is modeled computationally through graph-based abstraction and combinatorial generalization. This state estimation paradigm allows individuals or entities to use the continually computed state as a foundation to build systems that can perturb the state towards a chosen goal. An application of this is done through a cybernetic and navigational approach. The focus will be through the physiological function lens, especially of CVD and CRF. Below the contributions are distilled into two main subgroups of general concepts and personalization concepts, visualized in Figure \ref{fig:contributions}.

The general concepts include, but are not limited to:

\begin{enumerate}
\item \textit{\textbf{Precisely defining health.}} Because health is defined in many different terms, we focus on the definition at the scale of an individual. A precise definition sets the best stage for any computation and measurement.
    \item \textbf{\textit{Establishing the dimensions of health.}} The complex nature of health requires high-level abstraction to represent critical features that make up the health of the individual. In this work, we use utilities as abstracted microservices to represent the dimensions of health.
    \item \textbf{\textit{Representing the general state space of health.}} The state space of health represents all the possibilities that can exist for human health and provides a foundation map base layer that allows embedding of other layers.
    \item\textbf{\textit{Structuring the model of inputs to the health state.}} Each moment in life causes some small change in the health state of the individual. Capturing these moments and translating them to inputs into the health state give the next time iteration of the health state.
    \item\textbf{\textit{Integration of domain knowledge and data-driven approaches.}} In order for the estimation and modelling to be done in an understandable way, depending on a single top-down or bottom-up approach will not suffice. Each has advantages and disadvantages. In this work, we use a combinatorial generalization artificial intelligence technique of graph network blocks (GNB) to meld the two approaches together.
\end{enumerate}

The personalized concepts include, but are not limited to:

\begin{enumerate}
    \item \textbf{\textit{Creating a personal health state space relevant to the individual.}} The true health state space that is reachable for a single person is a subset of the general state space. Multiple factors such as genetics and age, carve out the state space for the individual.
    \item \textbf{\textit{Current health state of the individual within their personal state space.}} Knowing the current state is the primary objective of this dissertation. Using combinatorial generalization, we can establish a global set of attributes to give a current state estimate of the individual. Computing is done by instantiating a graph of an individual's utility and biological microservices, and estimating how the microservices change. These microservices in aggregate determine the global set of health state attributes.
    \item \textbf{\textit{Predicting the trajectory of the state at future time points with a personal model.}} By using the GNB, we can simulate future time points of inputs and produce their outputs as the predicted future health state.
    \item \textbf{\textit{Learning the input effect on a given state for a particular individual, editing the personal model.}} Given a large quantity of data produced by an individual, we use modern learning techniques to produce a model at the N=1 level, allowing us to more accurately model. We observe how inputs to a node in the graph produce results in the real world setting by modifying domain knowledge-based connections to suit a particular person.
    \item \textbf{\textit{Given a desired direction or goal, reducing the state space dimensions and presenting the current state in the context of the user intent.}} People interact with their health at a high level of abstraction and also have abstract goals. By decomposing this level of user interaction into the relevant dimensions, we can provide a user interface to understand their health state.
    \item \textbf{\textit{Providing a structure for health navigation.}} By understanding what inputs produce certain state changes in the individual, we present a method by which perpetual planning, used to guide an individual to alter their current health state towards their desired state. The above goal decomposition guides which inputs are most relevant and useful for the individual and presents each moment as a cybernetic control loop for the navigation system to control each step.
\end{enumerate}

\section{Thesis Outline}
Chapter 2 introduces the driving concepts of cybernetics and navigation for health computing and the flow by which the health state of each individual is dynamically changing. It reviews the literature to define a comprehensive view of ``health," how this health is continually changing, and how certain design elements are essential to produce continuous personal state estimation. Chapter 3 dives into a specific literature review of sensing, estimation, and modeling techniques used in this work. It concludes by presenting the scope of techniques for this dissertation. Chapter 4 presents the HSE framework through the detailed assembly of the design requirements into a cohesive system. Chapter 5 describes the experiments that illustrate HSE at an individual scale using multi-modal data and knowledge integration. Chapter 6 discusses the application of HSE by describing the methods of personal health navigation systems and cybernetics. Chapter 7 discusses the future opportunities in this field and related fields of work and concludes this dissertation.

\chapter{Understanding Health Computing}

\epigraph{To live effectively is to live with adequate information.}{\textit{Norbert Wiener}}
\vspace{12pt}








\section{Cybernetics: Inspiration for this Work}
\begin{figure}[H]
  \centering
  \includegraphics[width=0.8\textwidth]
  {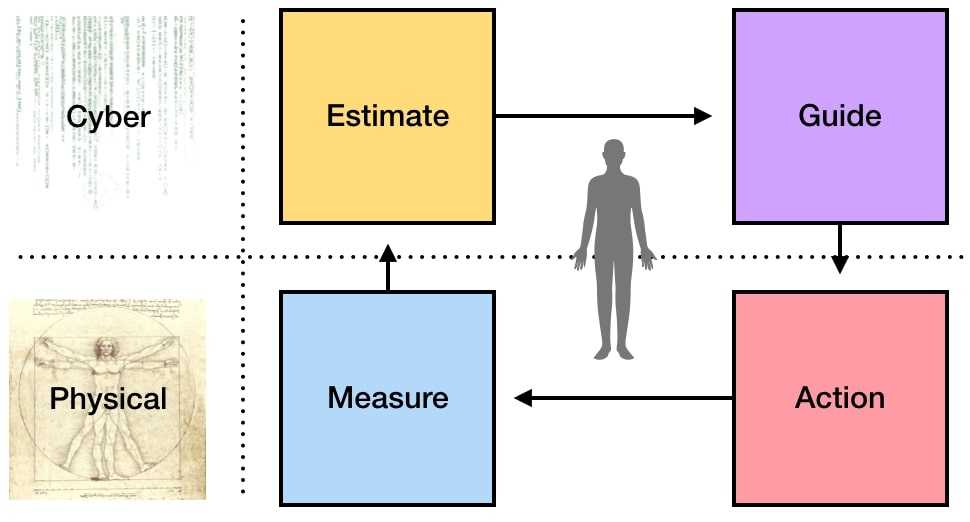}
  \caption{\textbf{Cyber-Physical Connection.} Cybernetics connects physical reality with computing information for control. In the real world, sensors measure various aspects of the world and translate these observations into data. The system must take the data and produce an estimate of the system state. Based on this computed estimation and a control goal, the system produces guidance for the real world. The actions taken in the real world may or may not be the same as the guidance due to disturbance, noise, or other factors. Thus, we take continuous measurements to see how the system state has been perturbed based on the last iteration of guidance. System components can be a human entity such as a doctor or a computing machine.}
  \label{fig:cyberphysical}
\end{figure}

In the simplest terms, cybernetics is about setting goals and devising action sequences to accomplish and maintain those goals in the presence of noise and disturbances. The Greek origins refer to the `art of steering', and the term was introduced into the literature by Norbert Wiener, a mathematician at the Massachusetts Institute of Technology, USA \cite{Wiener1962CyberneticsMachine}. His seminal publication ``Cybernetics, Control and Communication in the Animal and Machine" described what we know today as control theory for dynamic systems. Cybernetic principles are a universal mathematical theme in both natural and artificial machines. The availability of sensors that estimate the system state from observation enables communication and control. This state is perpetually compared to the goal state so the system can perform the best actions to maintain the desired goal state. In parallel, biological science also discovered control mechanisms as ``homeostasis" via a series of physiology scientists, including Claude Bernard, Walter Cannon, and Joseph Barcroft \cite{Gordon.2016AnatomyPhysiology}. Homeostasis referred to the stable internal environment at any given moment. It was first described in French literature originally in 1854 as milieu intérieur by Claude Bernard in \cite{BernardC.TransHoffHEGuilleminR1974LecturesPlants.}. Norbert Wiener's work in cybernetics generalized these principles beyond biology. There are four fundamental principles in connecting the information from the physical world to the virtual ``cyber" world, as shown in Figure \ref{fig:cyberphysical}.

The mathematical model at the core of classic cybernetic systems theory states that:

\begin{equation} \label{truecyberneticstate}
    X[k+1] = A[k]X[k] + B[k]U[k]
\end{equation}

\begin{equation} \label{measuredcyberneticstate}
    Y[k] = C[k]X[k] + D[k]U[k]
\end{equation}

\begin{figure}[H]
  \centering
  \includegraphics[width=0.8\textwidth]
  {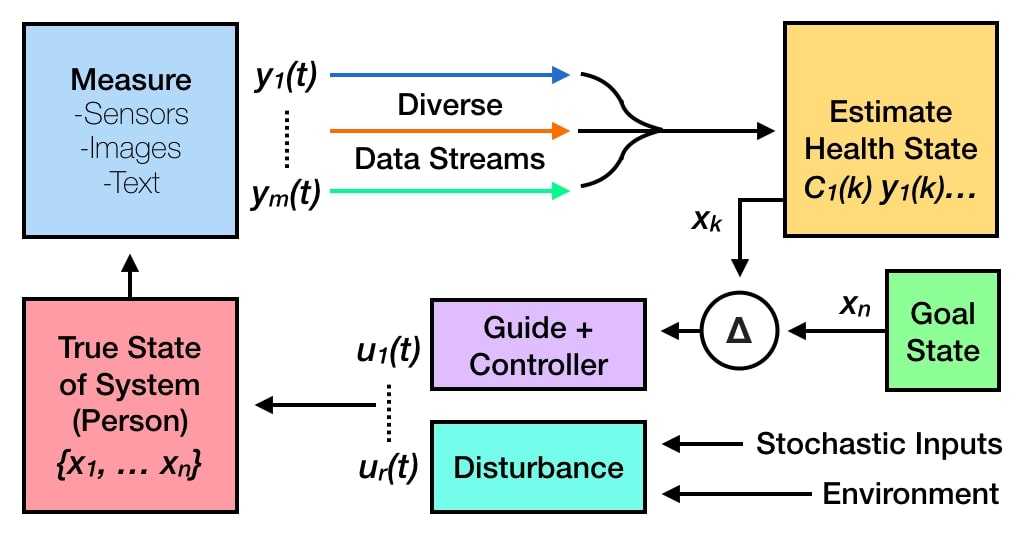}
  \caption{\textbf{Basic Cybernetic Loop.} Cybernetic control occurs through feedback loops. The loop begins with measurements from one or more sources, compiled into a computational representation of the system state. This estimated state is compared to the goal state. The delta between the goal and current state is used to give guidance to operate the controllers appropriately. These controllers partake in actions in the real-world to perturb the true system. Disturbances may also perturb the system from the environment or stochastic inputs. At the next time interval, this loop cycles again with a fresh set of measurements.}
  \label{fig:cybernetics}
\end{figure}

Where \textit{X}, \textit{U}, and \textit{Y} are the system true state, inputs, and measured output vectors respectively. \textit{A}, \textit{B}, \textit{C}, and \textit{D} are matrices that provide the appropriate transformation of these variables at a given time \textit{k}. The state system describes human health. The previous state and the inputs into the system play a role in determining health at time \textit{k+1}. Inputs into the human cybernetic system are anything that changes gene expression or physical actions in the body (from molecular interactions to coarse action). Thus a person is continuously exposed to these inputs. The inputs beyond the control of an individual are external disturbances, and the rest are controllable inputs \textit{u}. The true health state of an individual at a time \textit{k} is represented by \textit{X}. The health state is difficult to obtain and is always estimated. What we do get are the observable output variables. The state estimation challenge is in interpreting the observable data to understand the true underlying state. If we solely focus on this, the challenge of state estimation is in matrix \textit{C}, with observable components as \textit{Y} and our unknown state as \textit{X}. This representation is visualized in Figure \ref{fig:cybernetics}.

Two subsets of inputs determine the true health state of an individual. The first set of inputs is beyond the control of the individual, such as circumstances of where an individual is born and how they are raised, the genetic composition and biological processes occurring within the body, resources available, and the ambient environment. These inputs are usually not within the ability of an individual to manipulate. We consider these as noise or disturbances to the system, as shown in Figure \ref{fig:cybernetics} The other significant subset of inputs into the system that determines health is traditionally known as ``modifiable risk factors." Modifiable refers to an ability to decide the control input for the individual. This usually encompasses every decision moment the person lives but can extend to any situation in which a person has a choice (i.e., choosing to take a medication, buying a home in a particular neighborhood). Lifestyle modifications particularly are currently seen as the most promising approach for inputs that lead to healthier living \cite{Ioannidis2019NeglectingScience}. This is how we will model controllable inputs in the above equation as \textit{u}.

``No one notices when things go right." This simple homily underlies the fact that 95 percent of the US healthcare economy is for direct medical care, and only 5 percent for health improvement \cite{Bortz2005BiologicalHealth}. This allocation of resources occurs due to the ongoing challenge of estimating the health state of individuals who seem to be healthy. The challenges arise due to 1) poor sensing ability to develop adverse outcomes with current human perception and methods, and 2) the long lead time for developing chronic disease. By the time current clinical measurements such as cholesterol, blood pressure, or glucose metabolism are beyond the normal range, the user has already been in a dysfunctional health state for quite some time. Clinical researchers refer to this as the \textit{prodromal} state. Capturing the change in health state always (before significant dysfunction begins) is paramount to keeping people healthy states and preventing them from slipping into a diseased state.

The target of this thesis work aims at structuring data and knowledge so we can have continuous HSE. It is essential to consider the representation of the system state, as all computing components of the loop depend on this representation. In general, representation of the state and state space in addition to searching through this state space are two key fundamental challenges in any intelligent system \cite{Zadeh2008Linear-}. In simple control systems such as thermostats or even complex systems such as military jets, the representation of the system is quite well understood \cite{Dorf2000ModernEDITION}. A thorough investigation of \textit{defining what the true health state is} must be the first step towards building a robust and universal computational health state representation.


\section{Defining Health}
There is a multitude of definitions of health. We will first review two general approaches to health, the first being continuous versus classification approaches and the second being objective and subjective approaches. Following this, we take the various lenses of health definitions and amalgamate central principles of each. These lenses include biological, environmental, functional, and perceptual views on defining health. In order to build a comprehensive and robust definition of health, we must consider all aspects for a complete understanding. This work will primarily focus on the integration of these definitions into a single unified approach at the individual level definition of health.


\subsubsection{Classification vs Continuum}
\begin{figure}[H]
  \centering \centerline{\epsfig{figure=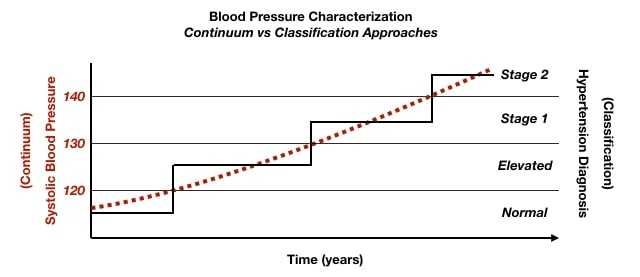,width=15cm}}
  \caption{\textbf{Blood Pressure Characterization.} Doctors use a classification approach with threshold values to have predetermined categories of defining health. The nature of health operates on a continuum basis, as shown in red. The state of resting blood pressure for each individual is always changing over time dynamically. Regardless of the approach, the individual is unable to sense their state position at any time, highlighting the need for continuous sensing.}
    \label{fig:bp}
\end{figure}

Before 1948, the WHO defined health as the absence of disease or infirmity \cite{World1946PreambleConstitution}. The original definition indicated that health is static or unilateral (disease or no disease). This method is a classification approach for having a definitive understanding of health. Medical systems and approaches in modern-day continue to classify health based on disease, conditions, and various metrics to define the state of one's health. For example, coronary heart disease, high blood pressure, and heart failure are all cardiovascular diseases that are considered diseases based on a range or threshold. These thresholds are recorded in the International Classification for Disease, which is used as the universal basis for the health field \cite{Organization2004InternationalProblems}. It has the word \textit{classification} in the middle of the name. If an individual does not meet the threshold, they do not have the disease. This has large implications. As an example, the American Heart Association formed ranges for cardiovascular risk, and medical professionals take into consideration ranges to determine if an individual is normal, elevated, or at risk, as shown in the blood pressure diagram in Figure \ref{fig:bp}. The thresholds are usually pre-determined for the entire population, assuming in the model that each person is identical to another and represented by a static system.

On the other hand, some classification approaches rely on the presence or absence of an attribute. A few examples include how an individual can either be positive or negative for an infection, have had a pregnancy or not, has an artificial limb or not, or has a chromosomal set of XX or XY. In summary, the classification approach to health is a combination of two subsets, the first subset of classifications use thresholds, and a second subset is similar to Boolean or categorical attributes \cite{Warshall1962AMatrices}.

In 1948, the WHO updated the definition of health to encompass health as a ``state of complete physical, mental and social well-being" \cite{World1946PreambleConstitution}. This shifted the thought towards defining health on a spectrum beyond disease and takes into consideration the full physical, mental, and social state of an individual. The WHO did not go further to quantitatively define what this means. In 1972, a physician named John Travis proposed a concept of health, referred to as the ``illness-wellness continuum," which states that a medical approach relying on the presence or absence of disease to demonstrate health is insufficient \cite{Travis2004WellnessVitality}. This concept introduces the idea that an individual falls somewhere on the spectrum of dysfunction and wellness. Where they are at each moment would be their health state on this single axis. The states in which dysfunction is perceivable is reflected by signs, symptoms, and disability. This approach to health is the ``health continuum," also described in chapter 1 Figure \ref{fig:continuum}. The health continuum implies that individuals would wish to strive for a higher level of optimal living through the means of health, not just as a unilateral classification of being labeled with a disease. This modern outlook of health acknowledges its dynamic nature and subject to constant change within an axis of illness and wellness. ``Both health and illness are dynamic processes, and each person is located on a graduated scale or continuous spectrum (continuum) ranging from wellness and optimal functioning in every aspect of one's life, at one end, to illness culminating in death, at the other" \cite{Svalastog2017ConceptsSociety}. However, it is important to note that the continuum may be interpreted as static if observing a single point in time or location within the spectrum. Regardless, this concept proposes a single dimension concept of state and paves the way for a continuous and quantitative approach. 

Clinically, resting blood pressure illustrates the continuum approach. People are unable to feel the biological change as it happens over time, and thus do not even know they are being affected. Second, the blood pressures vary due to a complex impact from both daily actions and environmental exposures, not just a single source or pathogen. Third, these diseases are not truly categorical in nature but are declines in physiologic function over time. In the example of hypertension, the medical practice uses cutoff thresholds to decide when to change the labeled blood pressure status of an individual, when in reality, the average pressure is increasing over time, as shown in Figure \ref{fig:bp}. The thresholds provide a basic region mapping of the blood pressure health state, but this does is not based on a precise model for each individual, instead of a population model.


\subsubsection{Objective vs Subjective}
In the literature, health has an expanded definition through the introduction of objective and subjective components. Objective components are quantifiable and externally visible, and subjective components are within an individual’s internal state of mind. The objective states have four components: the \textbf{biological} ability to exist, the \textbf{environmental} impact and handling of this impact on an individual, and the ability or capacity to \textbf{function}. The subjective component generally is understood through the individual’s \textbf{perception} of their quality of life.

\begin{table}[t]
\centering
\begin{tabular}{@{}lllll@{}}
\toprule
\textbf{Author} & \textbf{Biological} & \textbf{Environmental} & \textbf{Functional} & \textbf{Perceptual} \\ \midrule
Dubois                    & X & X & X & X \\
Dunn                      &   & X &   &   \\
Galen                     & X & X &   & X \\
Hippocrates               & X & X &   &   \\
Lalonde Report            & X & X & X &   \\
Ohlansky                  & X &   & X &   \\
Institute of Medicine     & X &   &   & X \\
Parsons                   &   &   & X & X \\
Sartorius                 & X & X & X & X \\
Simon et al.              &   &   & X & X \\
Spencer                   &   & X &   &   \\
Stokes et al.             & X & X & X & X \\
Travis                    &   &   & X & X \\
World Health Organization & X &   & X & X \\
Nag                       & X & X & X & X \\ \bottomrule
\end{tabular}
\caption{\textbf{Literature Review of Health Definitions.} Throughout health research history, various attempts have been made to define health. Here is an overview of these attempts.}
\label{tab:DefineHealthChart}
\end{table}


Various reports integrate objective and subjective components as a critical part of defining the scope of health:

a) Two prominent figures had established definitions of health used in modern medicine today. Hippocrates of the fifth century BCE, considered to be the “Father of Modern Medicine,” established the relationship between environmental and personal cleanliness and the origin of disease. He had classified health based on disease, resulting from internal physical imbalances. In the Roman Empire, Galen expanded on Hippocrates’s definition by stating that bodily fluids also determined temperament and personality, \cite{Badash2017RedefiningCare}. These combined outlined health as the idea of physical, environmental, and behavioral factors.

b) The updated WHO definition to expand the realm of health beyond disease and include mental and social well-being: “Health is a state of complete physical, mental and social well-being and not merely the absence of disease or infirmity” \cite{World1946PreambleConstitution}. 

c) The 1974 Lalonde Report from Canada introduces the concept of ``health field concept," in which identified determinants of health is through lifestyle (personal decisions), environment, human biology (all components of the body influenced by genetics), and health care systems \cite{Glouberman2003EvolutionCanada}. 

d) The Croatian Medical Journal describes health as three types: 1) absence of disease, 2) a state allowing the individual to cope with demands of daily life adequately, and 3) a state of balance, an equilibrium that an individual has establishes within himself and between himself and his social and physical environment \cite{Sartorius2006ThePromotion}.

e) The Center for Educational Development Health (CEDH) at Boston University and the Association of Teachers of Preventive Medicine Foundation (ATPMF) define health as a state characterized by anatomic integrity, including ``performing personally valued family, work, and community roles; ability to deal with physical, biological and social stress; a feeling of well-being; and a freedom from the risk of disease and untimely death" \cite{Stokes1982DefinitionMedicine}. Furthermore, the health status estimates of individual's health from ``one or more anatomic, functional, adaptive, and subjective indices."

f) A qualitative study by Simon et al. understands health through an individual's self-assessment, which is largely linked to one's physical health, functional capacity, health behavior, and other psychological components. Health also encompasses other aspects including health comparison, health transcendence, externally focused, non-reflective, social role activities, and social relationships \cite{Simon2005HowHealth}.

g) U.S. Institute of Medicine Committee on Health and Behavior: Research, Practice, and Policy in Washington, D.C. describes the need to integrate all the above approaches. ``Acceptance of the fact that stress is linked to cardiovascular disease or to other health problems has become commonplace. However, research also reveals many reciprocal links among the central nervous system, which recognizes and records experiences; the endocrine system, which produces hormones that govern many body functions; and the immune system, which organizes responses to infections and other challenges." The Committee illustrates how perception and biological health are tightly coupled, hence the call for an integrative approach to defining health
\cite{InstituteofMedicineUSCommitteeonHealthand2001IndividualsInterventions}.

The above approaches are compared in Figure \ref{tab:DefineHealthChart}. In the following sections within this chapter, we will first explore the individual components of objective health in biological, environmental, functional aspects. The subjective and social components of health will be discussed as a perceptual component. It is also important to note that there are large health systems that are part of the ecosystem that contributes to health for the individual.

\subsection{Biological}
\textit{Biological health is the sustenance of life through the homeostasis of natural systems.} Working biological health supports the basic living functions of life. The literature consensus of what these tenets are: Organization of Structure, Metabolism, Homeostasis, Growth, Reproduction, Response to Stimuli, and Evolution. The only main contention in general literature of living biology is in the case of viruses. Viruses do not conduct their metabolism and thus do not maintain any homeostatic state. They also cannot reproduce on their own.

Human biological health has been quantified by looking at lifespan or by the calculation of healthy aging and life expectancy \cite{Brussow2013WhatHealth}. The 2010 Global Burden of Disease Study provides quantifiable health metrics for calculating healthy life expectancy. Lozano et al. estimated life tables and annual number of death for 187 countries from 1970 to 2010. The estimation included number and rate of deaths by age and sex to determine the burden of disease in order to constrain estimates of cause-specific mortality \cite{Lozano2012Global2010}. A look at lifespan is only a primary attribute of biological existence. The biological understanding of a human is understood more deeply by the genetic composition and biological processes occurring within the body. Illness or disease negatively impacts the overall homeostasis of biological processes and therefore reduces the lifespan.

With stress from its surroundings, an individual must remain stable internally for its biological success. This stress is the particular reason behind the ``response to stimuli" factor of biological health. ``Under normal conditions, the external environment changes constantly in an unpredictable manner. In order to survive and to continue to function effectively the organism must make adaptive responses to the modifications. It must, as well as possible, repair destructive tissue damage and restore its own internal environment" \cite{Dubos1987MirageChange}. The ability to adapt and hold resilience is therefore critical for survival when introduced to external stresses or threats, such as disease. Biological health can, therefore, include the definition as ``the perfect, continuing adjustment of an organism within its biology to its environment" \cite{Wylie1970TheDisease.}.

\begin{figure}[H]
  \centering \centerline{\epsfig{figure=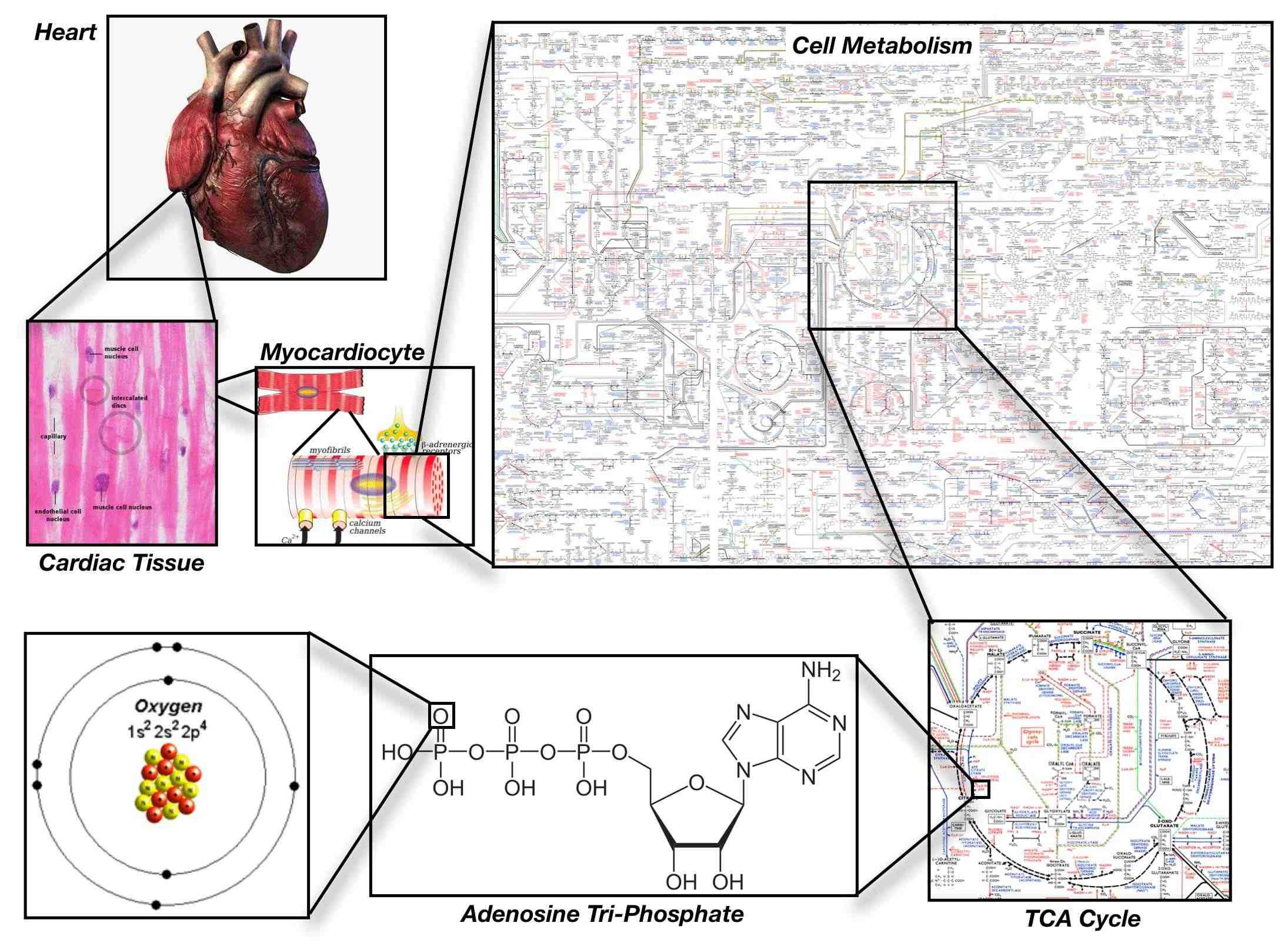,width=16cm}}
  \caption{ \textbf{Biological health is at the level of inanimate molecules interacting to provide the organism with the fundamental components of life.} Metabolic pathways provide the methods by which energy and materials can be synthesized to support these components. Here is an example of biological health at various abstraction levels. We zoom into the heart. There is the tissue of the heart, each heart cell (myocardiocyte), the metabolism network in this cell, a particular central pathway called the Tri-Carboxylic Acid Cycle (TCA Cycle), Adenosine Tri-Phosphate (ATP) which is an energy currency in biological systems produced by the TCA Cycle, and the oxygen atom within this ATP molecule.}

    \label{fig:metabolism}
\end{figure}

Biological health operates through the complicated yet harmonious interaction of non-living molecules to maintain the steady-state of living functions. A narrow slice of this complex interaction network can be visualized through the basic chart of metabolism, as seen in Figure \ref{fig:metabolism}. The health of this network to sustain itself in a stable state can be considered the generalized approach to biological health. A full biological health state framework would require a system that can incorporate all the abstraction levels from atomic to the whole organism, as shown in Figure \ref{fig:metabolism}, and the ability for computation at any level of choice.

\subsection{Functional}
\textit{Functional health is defined as the performance capacity to provide a service or utility.} This definition is applies to any abstraction level on the biological spectrum. At a whole human level, the function may be how well a person can run. At an organ level, the function can be how well the heart circulates blood, and at a molecular level, it may be how well a cell regulates a specific hormone such as insulin. Functional health assumes that health is a resource that allows individuals to be able to use their bodies and minds to perform specific tasks. The function abstracts away biology from health. A biological bone and a titanium bone both provide a service of structure to the body. A natural pancreas and artificial digital pancreas both provide a service of glucose regulation. A biological sinoatrial node in the heart and an electronic pacemaker both provide heartbeat regulation as a service. Functional health fully encompasses biological health, but can extend far beyond biological systems for defining health.

\begin{figure}[H]
  \centering \centerline{\epsfig{figure=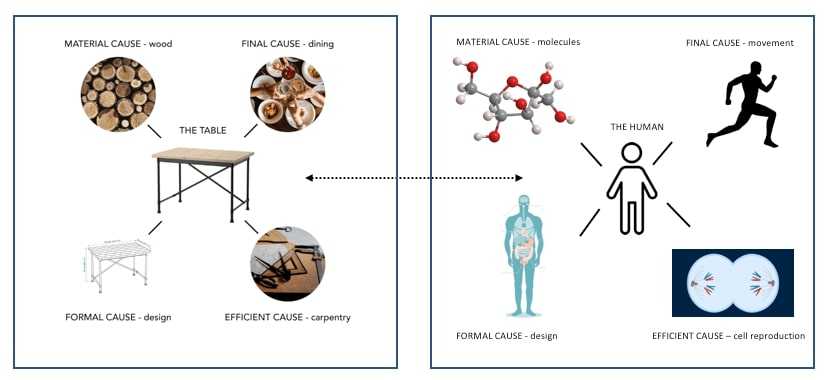,width=16cm}}
  \caption{\textbf{Aristotle's Four Causes.} Aristotle's Four Causes refers to the various lenses we can understand why something exists. Aristotle argues that the final cause is by far the most important since without it, the other causes may not have a direct purpose to exist.}
    \label{fig:Aristotle}
\end{figure}

Aristotle's ``Four Causes" is a realm of thought that explores the reason for existence \cite{Falcon2006AristotleCausality}. This concept can help answer ``why" our biological health truly matters. In Aristotle's telos philosophy, the end goal or purpose of anything is what function comes about. We can use this in making a parallel to how health is ultimately a utility function. Figure \ref{fig:Aristotle} visually describes his classical example of the ``cause" of a table. Material cause refers to the physical matter that makes an object. Wood makes up a table, and molecules make up a human. Formal cause refers to the design of an entity. A table may have a simple design, and a human has an intricate design of interacting parts. Efficient cause refers to how an object comes into existence. Carpentry makes the table and reproduction and cell growth from a single cell (zygote) makes up a human. Final cause, also known as telos, refers to the reason for existence. A table exists to dine upon, and a human exists to exhibit certain capacities such as movement and thinking.

Both Aristotle and modern-day organizations, like WHO, have contributed to the evolution of defining health. The WHO identified health as a ``resource for everyday life, not the objective of living.`` Health includes social and personal resources, as well as physical capabilities. in the first International Conference for Health Promotion in 1986 \cite{Williamson2009HealthConceptualization} Furthermore, it encompasses ``the \textit{ability} to lead a socially and economically productive life." Health is also viewed as the \textit{ability} to perform valued social roles, emphasizing functional views and the concept of health to include personal beliefs and values by a society \cite{Stokes1982DefinitionMedicine}. This means that a loss of a leg is grieved just as much because this means a loss of mobility, and thus, a loss of self. If an individual were to receive a prosthetic for a leg as a replacement, the mobility would be maintained and fully functional. However, the biological basis would alter to a mechanical one. The same can be said for porcelain teeth-replacing biological teeth for cutting food. However, to an individual, there is functional value due to the mobility or cutting and, thus, would no longer be a loss of self.

While the maximum lifespan may have increased, there may not be enough emphasis on the healthspan of the population. While there is a longer lifespan by reduction of child mortality rates, elimination of diseases through modern medicine, the human biological design is not for long-term use (i.e., maintaining healthy functioning knees, heart) \cite{Olshansky2018FromHealthspan}. This increased lifespan now becomes an issue of function; to address the functional component there must be an emphasis on the healthspan and the quality of life, not just the length. The WHO has quantified the Disability Adjusted Life Year (DALY) as one lost year of healthy life due to disability or poor health \cite{Organization2014WHODALY}. DALYs for a disease or health condition are calculated as the sum of the Years of Life Lost (YLL) due to premature mortality in the population and the Years Lost due to Disability (YLD) for people living with the health condition or its consequences. This metric indicates that disability or poor health is a loss of function that ultimately leads to a shorter lifespan. Individuals may also define themselves as healthy because they are functioning as well as expected in their age and gender group of the general population. Functional health is, therefore, interwoven with perceptual health.

With living comes the capability to respond to stressful situations. There is a functional capacity to cope and be resilient to the demands of daily living \cite{Sartorius2006ThePromotion}. From a cybernetic perspective, this is the stability of the system to perturbation to the external environment. The diminishing function of an individual in their ability to perform tasks that they would have otherwise been able to can be directly related to disease or illness, much of which can be caused by the surroundings.


\subsection{Environmental}
\textit{Environmental health encompasses external factors outside of an individual that impact the state within the individual.} The environment plays a significant role in the exposures that an individual has over time, and thus is inevitably a major component of one's health. Dubois explains that ``health and happiness cannot be absolute and permanent values, however careful the social and medical planning. Evolutionary fitness requires never-ending efforts of adaptation to the total environment, which is ever-changing" \cite{Tait1982DUBOSHEALTH}. Because of this continuous change in an individual's environment, this theory supports the health continuum method in which health is dynamic. While the environment is continually changing, and thus human health is continually changing, humans attempt to create equilibrium within themselves and the social and physical environment via self-maintenance systems \cite{Sartorius2006ThePromotion}. This theory supports that health relies on an adaptation mechanism to the external environment to produce homeostasis at various layers of biology and function.

The “Health Grid Model” was developed by Dunn et al. to show how external factors impact the quality of life \cite{DUNN1959High-levelSociety.}. The graph below shows peak performance as an optimal state of health based on the individual’s age and makeup but highly linked to a favorable and unfavorable environment. Dunn called for a more “integrated approach to human health” via the integration of environmental factors. There may be favorable and unfavorable environments that may positively and negatively impact health. Individuals may be exposed to various risk factors and inputs to their health through the environment (i.e., pollution, water sanitation, chemicals, occupational risks, climate change, radiation, agricultural practices). Individuals may also have exposures to external risk factors through their social environment (i.e., family, friends, society, decisions based on social circumstances). The social environment that an individual surrounds themselves with has a significant impact on daily lifestyle factors, for example, food choices and physical activities.

Understanding an individual's environment furthermore gives the livability of the location of how healthy a particular area may be for the individual. The Economist Intelligence Unit's Where-to-be-born Index and Mercer's Quality of Living reports are measuring the livability of a given city or nation \cite{Mercer2020QualityMercer,Economist2005Quality-of-life1}. This environmental component is a combination of subjective measurements such as life satisfaction surveys and objective determinants such as divorce rates, safety, and infrastructure. These studies review the livability for populations at large but directly links to predicting an individual's health state based on a chosen environment. A part of this local environment includes any system built in regards to health.

\begin{figure}[H]
  \centering \centerline{\epsfig{figure=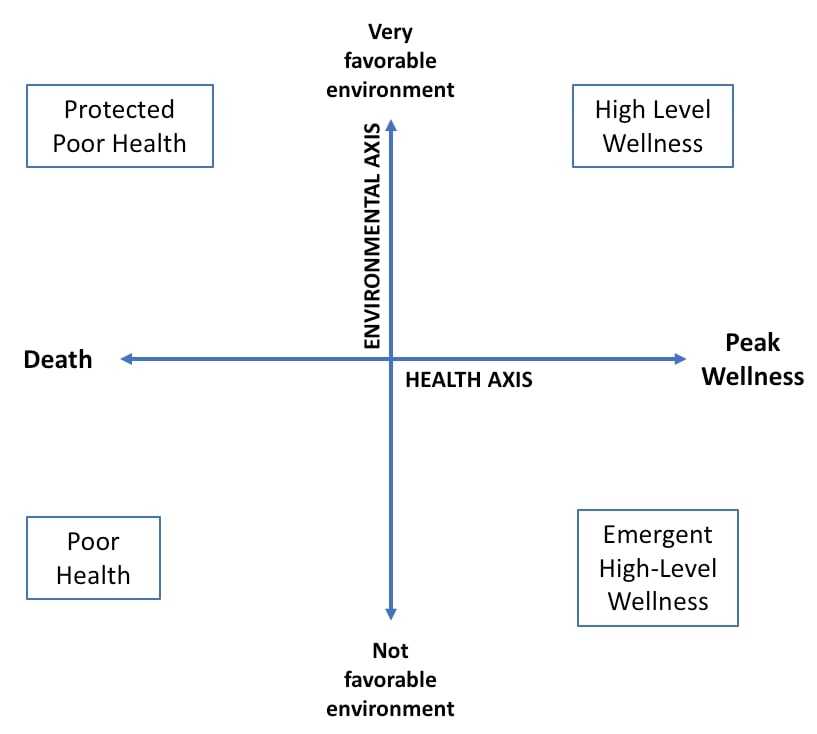,width=12cm}}
  \caption{\textbf{Dunn's Health Grid Modeling: External environment and its relationship to wellness.} Dunn's model of health highlights the strong relationship between a favorable environment and achieving peak wellness. If a good environment or good health exist exclusively, one cannot achieve an optimal level of health.}
    \label{fig:Health Grid}
\end{figure}


\subsection{Systems}
\textit{Systems health is the overarching structure for providing health to individuals at a population level.} Systems for providing health services largely contribute to the outcome of health for individuals. The effectiveness of these systems is tracked by reviewing populations at large. Health systems play an economic role in providing a positive return to individuals. Williamson et al. conceptualize health as a type of capital that invested in many individuals and institutions for positive health returns that reflect higher individual capacity. Health cannot be exchanged or sold for goods and services. Instead, ``as a type of capital, health is a stock of biopsychosocial resources that people can draw on to participate in society” \cite{Williamson2009HealthConceptualization}. A  modern approach to health, such as value-based healthcare, puts what patients value at the center of healthcare. It helps ensure that they receive the care that can provide them with outcomes they think are essential and that limited resources are focused on high-value interventions. In order to do this, there is a need for flexible definitions of health, personalized and tailored to patient values \cite{Gentry2017DefiningSystems}. Another system approach that is becoming more prevalent is the continuum of care. The continuum f care concept involves an integrated system that guides and tracks a patient over time through a comprehensive array of health services spanning all levels of intensity of care. The array of services includes promotion, prevention, treatment, and recovery as an individual goes through the process of receiving healthcare services \cite{Sagner2017TheHealthspan}. While systems are of relevance in defining health, the economic and population level of health will not be part of the scope of this work. Instead, the focus will be on computing how a system may impact an individual as part of their environment.


\subsection{Perceptual}
\textit{Perceptual health is an individual's outlook on their health state and judgment of their quality of life.} The view of one's self has become a large component of subjective determinants of health, including one's emotional state, social connectedness, intellectual stimulation, self-awareness, and development of lifestyle behaviors. With perception is an individual's ability to think and feel in the way of good health, as understood by their internal state. In addition to ``anatomic, functional and adaptive dimensions, much of health is simply about feeling good" \cite{Stokes1982DefinitionMedicine}. 

Health is of relevance to the individual concerning their defined goals and value in living. Aristotle's Telos philosophy indicates there is an end or purpose, in ``teleology" or the study of purpose within biology. Telos, which means ``end", ``purpose" or ``goal" as shown in Figure \ref{fig:Aristotle}. Dubois in the \textit{Mirage of Health} has also defined ``good health as the condition best suited for each individual to reach his or her personal and social goals" \cite{Tait1982DUBOSHEALTH}. The concept of personalized goals and values in living becomes critical to the state of health because it links to how an individual perceives the state of their life. Individuals act and function in alignment with their own values. Stokes has described Health as with the incorporation of the ``ability to perform personally valued family, work, and community roles" and dealing with psychological and social stress \cite{Stokes1982DefinitionMedicine}.

Kimball et al. at the University of Michigan, along with prominent thought leaders such as Janet Yellen and Daniel Kahneman, put together a concept of \textit{Utility and Happiness} where they connect how the perception of living happily relates to the utility for the individual \cite{Kimball2006UtilityHappiness}. They have separated short-term happiness as ``elation" due to relative change in utility and long-run happiness as the ``baseline mood" for the utility. Thaler's work in connecting behavioral psychology with economics shows how the ``baseline mood" (i.e., influenced by sleep, exercise, eating habits, genes, time with friends) is linked to the true health utility \cite{Kimball2006UtilityHappiness}. Both of these examples show how the quality of life is related to psychology bridged with economic principles of utility to an individual.

Increasingly described in the literature are understanding one’s self, motivations, and goals as a foundation for health. The health continuum also highlights that awareness of self, education, self-growth, and adopting healthy lifestyles are critical in moving towards the wellness end of the spectrum and reaching a more optimal level of health. In addition to values, self-perception is another measure for differentiating a healthy state from a diseased state and how this can be dependent on an individual’s outlook on the situation and how they feel internally (or subjective well-being) \cite{Bowling2005MeasuringHealth}. Physical and psychological “wellness” of an individual plays a role in the health continuum in the constant strive for improvement. The domains of wellness can include social  (lifestyle behaviors), emotional (stress, resilience), physical (activities), spiritual (meaning, purpose), occupational, intellectual components \cite{Institute2018FraminghamNHLBI}. The combination of these domains creates a wellness state that incorporates both the mind and the body.

Maslow’s ``Hierarchy of Needs" indicate that humans are built upon physiologic needs as a baseline, but are ultimately striving for self-actualization. This approach sees physiologic needs as the core need to be satisfied before needing safety, love/belonging, esteem, and self-actualization. This work is quite prevalent in the literature in psychological sciences.

\begin{figure}[H]
  \centering \centerline{\epsfig{figure=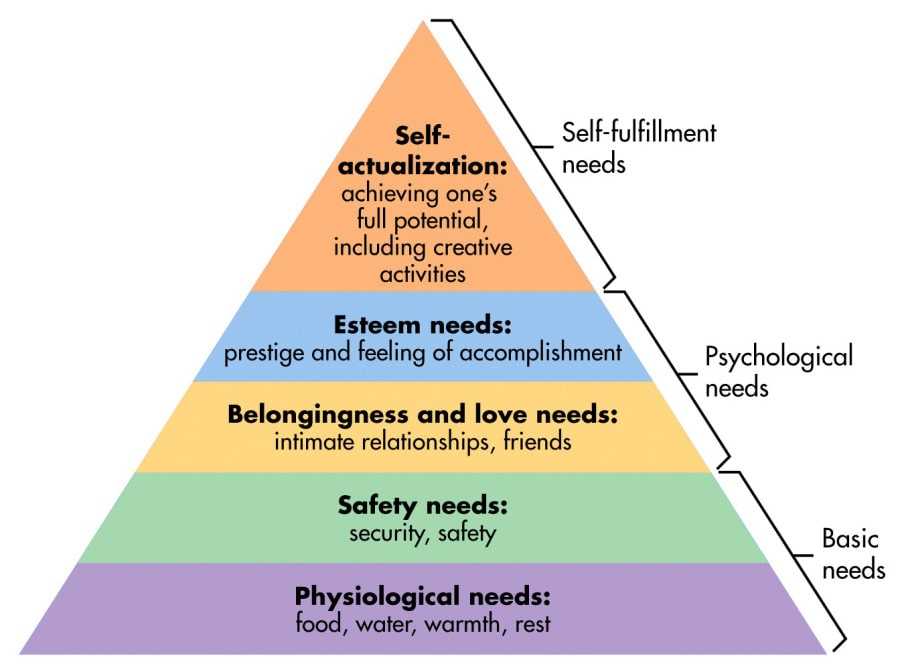,width=12cm}}
  \caption{\textbf{Maslow's Hierarchy of Needs.} There are base physiologic and safety needs (food, shelter, security) as a fundamental requirement, psychological needs such as a sense of belonging and esteem, and finally needs for self-fulfillment such as reaching one's full potential.}
    \label{fig:Maslow's Hierarchy of Needs}
\end{figure}

\begin{figure}[H]
  \centering \centerline{\epsfig{figure=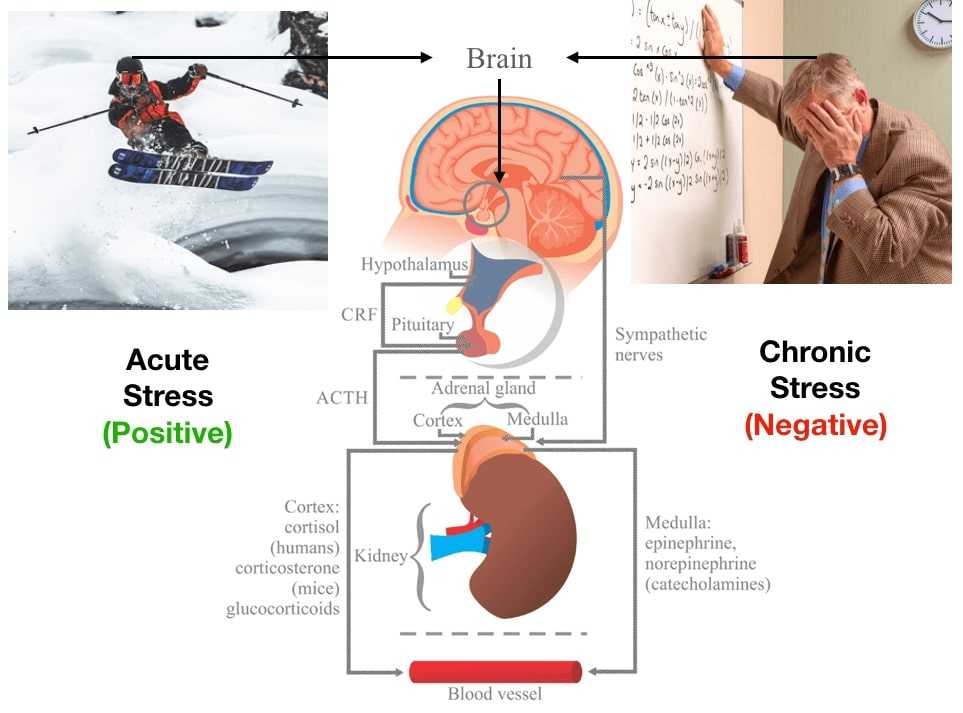,width=14cm}}
  \caption{\textbf{Mind - Body Connection: How Stress Impacts Physiology.} How a person perceives their environment has a direct impact on their physiology. When an individual perceives stress, the evolutionary encoded response of increasing blood pressure occurs to deliver high-performance blood delivery for a ``fight or flight" response. This can be both a positive or negative impact on the health state, depending on the context. For example, a skier negotiating a racecourse may have stress from turning at high speeds. This stress allows the muscles to function at high forces and serves as a positive acute stressor. Blood pressure returns to normal after the race.
In contrast, a professor who is worried about getting funding approved may have this stress effect chronically rather than acutely. In this context, the stress response may damage parts of the vessels and organs. Thus stress is a highly context-dependent component of computing the health state.}
    \label{fig:stress}
\end{figure}

The Iceberg model of health and illness, as presented by Travis in the illness-wellness continuum \cite{Travis2004WellnessVitality}, describes the state of health to be built upon the foundation of self-understanding, psychological motivations, and lifestyle and behavior. For both Travis and Maslow, the presentation of health state regards perceptual components as integral considerations that layer on top of high-functioning biological machinery. In this perspective, biological health directly impacts the perceptual health. On the flip side, perceptual health can directly impact biological health. As shown in Figure \ref{fig:stress}, mental states directly impact the biological components of the entire rest of the body. Thus, there is a need to build a network representation of these two views with a bi-directional relationship. The body impacts the mind, and the mind impacts the body.

We can also quantify the perception of how an individual perceives themselves and their quality of life (QoL). QoL is a term for the quality of various life domains and is measured as a standard that considers what a good life is for an individual. Many of the factors determining the quality of life may be driven by values and goals and encompassing of emotional, physical, and social well-being. The factors ultimately determine one’s life satisfaction or how an individual feels about his or her life. Measurements can be seen globally in the World Happiness Report of 2019, which explores how happy citizens consider themselves to be, which directly impacts their emotional and social well-being.

Ultimately, all the above definitions of health mention important components of health for an individual. We must consider these definitions in a generalized computing framework for computing the health state. These components are also highly interdependent, and this must be clear in the representation. The next phase of research describes how we can pull these components together in a common framework.


\subsection{Integrating a Holistic Approach}
As an individual, we want to consider all aspects of our health when making life choices, a holistic approach. For this reason, this thesis uses an integrated approach to the definition of health by integrating the four components as described previously: biological, environmental (including systems), functional, and perceptual. All four components contribute to the quality of living and a total health state computing framework. We will utilize the approach of the health continuum view and recognize that health is evolving and dynamic by nature on this continuum. This method still allows for classification approaches through thresholds and categories. There is also a nesting of various layers that build a single individual model. Nesting may occur for various domains such as biology, as shown in Figure \ref{fig:metabolism}. Depending on the abstraction level from a molecular level to a whole person level, various connected components rely on the functioning of the lower-level components. Measurements may come at various layers of abstraction, and the user or domain expert such as a doctor or intelligent agent may want to perturb or interact at a particular abstraction layer. A holistic model must accommodate this ability to zoom in and out in a fluid manner. Coupling these concepts also requires an understanding of what causal orientation they are related to each other in the real world. The dissertation goal is to bring all components of defining health together holistically in order to best estimate the health state. In the next section, we go into the depth of how this holistic approach translates into detailed design requirements.


\section{Design Requirements for Health State Estimation}

After considering the nature of different health definitions in this chapter, and the emerging frontiers in chapter one, we can establish the following design principles for a health estimation framework that can support a variety of complex applications. We will describe these requirements below.

\begin{enumerate}
    \item \textbf{Personalized Modeling.}
    To predict the future, provide guidance, and understand the preferences and particularities of an individual, researchers must build personal models. These models, which establish the premise that each individual is a unique system, are needed to best estimate the user’s health state and how various inputs uniquely move them in the state space. Models are accessed for state estimation or providing guidance. The personal model can build knowledge and predict many aspects of an individual’s life, such as how an individual reacts to different stimuli under specific conditions or physiological changes from an intervention. To model a person, a combination of long and short term information must interact. Long-term models of an individual can come from the genome or the history of event patterns and behavior. Biological or high-resolution continuous sensors can capture short-term models. These models are not static; they continuously change with an individual’s age, life events, and other life parameters, which makes model-building a dynamic and causal understanding process. Models must have a general representation that can at least address the following health attributes: 
    
    \textit{Physiologic function.} Dynamic health attributes include biological health, such as organ function and capacity (CRF, bone strength, and metabolism). Models may have various time and size levels in terms of detail, depending on the application. Short-term models can already predict how electrical activity in the heart may produce arrhythmia (such as in Figure \ref{fig:arevalo} while a long term model may predict the impact of long-term lifestyle habits on developing atherosclerosis. Cellular or gross organ-level processes can be modeled separately or in combination, as shown in Figure \ref{fig:metabolism}. These functions constantly change based on daily factors that include sleep, nutrition, medication, or stress.
    
    \textit{Psychological function.} Extrapolating and modeling mental health will be a complex challenge that requires advanced sensing mechanisms such as affective computing. Mental health components could include various time resolutions of interest, such as overall mental stability or minute-by-minute affect analysis. Evaluating risks for erroneous modeling before field deployment will be essential for the users’ safety. Understanding user preferences, influences, and motivations will be essential to close the loop when humans are involved as actuators in a participatory approach. The system uses this information to tailor guidance in a way that makes a healthy lifestyle pleasurable and convenient, enhancing the quality of life. For effective modeling, irrational behavior, social tendencies, and situations in which short term pleasure overrides long-term goals must be included. Risky behaviors may warrant preventive help to reduce harmful events such as drunk driving. 
    
    \textit{Risk.} Researchers can model risk assessment for reversible and irreversible physiologic dysfunction, undesired events, or regions of the health state space a person wishes to avoid. \textit{Disease risk} refers to the health state entering an undesirable region of the state space. An example would be the chance that an individual would develop metabolic syndrome, a condition that medical professionals could reverse with an early-stage intervention. \textit{Damage risk} refers to the risk in which an individual is negatively permanently altered. An example of this would be the likelihood of metabolic syndrome turning into type 1 diabetes, resulting in permanent pancreatic damage. Risk modeling will also include behavior and mental condition models to understand the propensity for self-harm, accidents, or harm to others, as mentioned previously in the psychological function section. For the greatest value,  we must prioritize resources and investments based on evaluating the likelihood and consequence of combined risks. Risk models of the individual are directly related to prediction and prevention.
    
The personal model is essentially a virtual twin of the individual, which allows for modeling the current and future state of the person in various aspects of their biology and psychology. The model's state space must be compatible with continuum and classification approaches.

    \item \textbf{Predictive and Preventive Capability.} The above model should be able to simulate what the health state will be at various future time points. Given a set of chosen inputs, we can test various scenarios to perturb the health state model to see how the current state changes. These simulations should be able to help predict a trajectory path to compare which input set would produce the most desirable health state for the user. A prediction that falls in an undesirable region can advise preventive actions to steer the state away from the region. 
    
    \item \textbf{Precision through Understanding and Learning.} A health estimation framework inherently will have errors in the outputs that may not be compared to a ground truth measurable factor, making it challenging to compare various models and judge performance of the model, state estimation, and predictions. The system must have a method by which to give an estimated error range so the user or computer can understand the quality of the estimation. Precision should improve in the model and estimation with increasing information. The two main approaches for this would be bottom-up data-driven methods and top-down knowledge-driven methods.
    
    \item \textbf{Participation with User.} The system must be able to interact with various types of users. These users include the individual themselves, any healthcare professionals, domain and research experts in any respective field, or other computing clients that access the health state for further services provided to the user. Facilitated human interaction with the framework requires representation in a tangible human-understandable form to bridge the gap between the physical and cyber worlds. The HCI should have easy to understand interfaces, a simple way to query the health state and model, and explainable to the human. Developers that wish to create software using the health state and personal model should have secure yet easy access to Software Development Kits (SDK) that can create such HCI and services.
    
    \item \textbf{Encompass Holistic Definition of Health.} As we explored above, there are various lenses and dimensions all part of a total health state for an individual. Any complete HSE framework must be able to include these lenses. Biological mechanisms give rise to functional and perceptual experiences and how the environment and systems impact this. Systems can be clumped as part of the environment. Based on the work presented by Kimball et al., we can also view perception through a utility lens, and thus can model this as a functional component of health. Thus, the requirements for the holistic definition must include the biological, functional, and environmental aspects.
    \item \textbf{Capture Connectedness.} As shown in Figure \ref{fig:metabolism} and Figure \ref{fig:stress}, health state is tightly connected to many different parameters in a networked web. This connectedness must be captured to give the best quality estimate of health.
    \item \textbf{Causal and Explainable.} Given a large number of connections, many of these may be spurious or artifacts of data. Understanding why something is connected and characterizing its true effect is key for a high-quality system. For trust by users, scientists, health experts, and computation outcomes, this explainable aspect is also essential. Many modern artificial intelligence techniques lack in this requirement \cite{Gunning2017Explainablexai.}. Current approaches in using data and machine learning to estimate health are falling short due to their non-understandable black-box nature. Knowing why an individual has a certain health status will be paramount for explaining why a recommendation or treatment is best. Purely data-driven methods can be components in a large system but will have a difficult time explaining the reason for the classification without the incorporation of relevant domain knowledge.
   \item \textbf{Domain Knowledge Integration.} The best source of understanding causal relationships and connectedness rests within the vast array of domain knowledge already in the literature. Leveraging this domain knowledge is essential, especially as a cold start, to build an explainable system. Any new domain knowledge discovered should also have the ability to integrate. Hard-wiring the domain knowledge is thus not an option. Domain knowledge also is organized at various levels of study (molecular to societal sciences), and the best approach would be to take advantage of the vast knowledge in all these layers. 
    \item \textbf{Zoom Fluidly Between Abstraction Layers.} As mentioned in the previous design requirement, the ability to zoom in and out of abstraction layers should be part of the system. This gives domain knowledge various entry points to participate in the framework and allows for the user or domain expert to understand a layer of interest. Looking at Figure \ref{fig:metabolism}, the framework should allow jumping between any box layer of interest, while still preserving the connectivity.
    \item \textbf{Flexible and Modular Architecture.} Because health can be abstracted away from biology through the functional lens, the framework must allow for flexibility through a modular approach. Imagine a patient who receives a kidney transplant. This new kidney has a whole new set of genetics and biology from the patient's original kidney. If the kidney is replaced with a mechanical kidney, this would have an entirely different operational structure. A flexible approach would allow the appropriate module, which modeled the kidney, to swap out for the current one. This ensures that no matter how technology adapts and situations change, the framework continues to be robust for the future. 
    \item \textbf{Diverse Data Fusion and Transformation.} In addition to taking advantage of domain knowledge of all abstraction layers, the system should be able to take advantage of any data produced about the individual. This could include a vast array of sources, types, and frequencies of data and event streams. Any type of data should be able to get processed in a way that is then useful to update the health state or improve the model. By acquiring data from different types of sensors, we can create a more precise understanding of how to model the individual health state. This process needs to collect, aggregate, and process heterogeneous data from multiple sources to infuse into the model. Data transformation methods are needed to convert data to a representation that is suitable for digestion into the personal health model framework.
    \item \textbf{Perpetually Updating.} As pointed several times in the last two chapters, the health state is in a dynamic flux. With data and domain knowledge, the system needs to continually update the health state of an individual regardless of the situation to give the current state. Similar to how a GPS is always refreshing its alignment with the signal satellites. As an example, the highest performance GPS receivers can update at 50 Hertz (1 Hertz = 1 cycle per second). For the health state, various applications will require a minimum update rate in order to perform. 
\end{enumerate}

Using these above requirements for building a state estimation framework, we will dive into the current day literature to see how the research community is approaching these requirements. We will review the various methods for collecting data, approaching state estimation, and constructing a model for individual health.
\chapter{Literature Review}

\epigraph{Science is organized knowledge. Wisdom is organized life.}{\textit{Immanuel Kant}}
\vspace{12pt}


Given the goal of this work is to have a model for computing the individual health state, we explore a literature review of methods for modeling and state representation on human health. The organization of the literature review includes four sections. First, we look at an overview of methods for personal data acquisition to obtain measurements about an individual. Second, we review how to estimate the health state from various measurements. Third, we survey models developed for representing human health. Fourth, we look at relevant techniques by which models can be made intelligent and dynamic. The chapter concludes with challenges that remain in the field of HSE and how this dissertation aims to address these challenges in a novel method.


\section{Health Data Acquisition}
\begin{figure}[H]
  \centering \centerline{\epsfig{figure=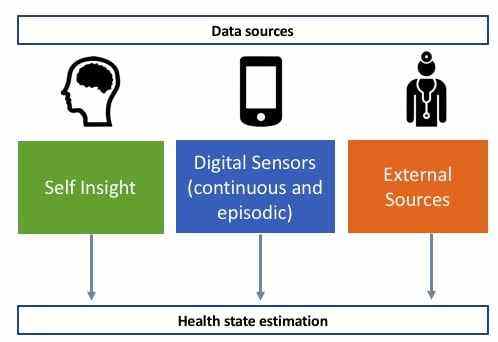,width=10cm}}
  \caption{\textbf{Sensor categories for acquiring health data.} }
    \label{fig:dataacquisition}
\end{figure}

To begin any endeavor in estimation, we must first gather measurements in the real world that we can use towards computing the estimate. These measurements are gathered through a sensor. Sensors bridge the computing world with the physical world, as shown in Figure \ref{fig:cyberphysical}. The cost and accuracy of sensing mechanisms vary, so data stream continuity and availability will always be different depending on the modality. For example, a wearable heart-rate monitor is an inexpensive and continuous way to measure heart rate but will be less insightful than cardiac perfusion testing with a computerized tomography scanner to evaluate heart attack risk or tissue oxygen delivery. Intuitively, an estimation system that can ingest the most types of data streams will perform better due to capturing a more well-rounded picture of a given situation. This highlights the need for the estimation framework to have a versatile ability to ingest data types. In order to ensure the constructed estimation framework best meets this need, we will review a wide variety of literature demonstrating how personal health data can be acquired through sensors, as summarized in Figure \ref{fig:dataacquisition}.

\subsection{Self Insight}
The most primitive and natural sensor is the person's biological signals and cognition. Two sets of signals influence our understanding of our own mental and physical health. First, \textbf{sensory organs} capture sensory signals about the world external to our skin through sight, hearing, touch (including sensory pain, temperature, texture), taste, and smell. These natural sensors measure various attributes to understand the individual's physical environment or directly perceivable attributes. Internal to our skin, we then use \textbf{internal cognizance} to make interpretations about the external signals and integrate them with signals about our biology. The interpretation occurs through feelings of perceived pain, gut motility, energy, proprioception, and mood, to name a few. Humans use both these abilities to understand their health state and how their environment is impacting them. The ways our natural biological sensors unify, store, quantify, and use these data are not easily compatible with modern computing. Tedious methods that do not capture data continuously, such as questionnaires and surveys, are used to capture these signals in most scientific fields that study these signals. We go over traditional and emerging techniques for extracting self-insight computationally.

\textbf{Sensory Organs} capture data streams (sight, hearing, touch, smell, taste) that give us the ability to garner information about the external world. There are several mechanisms to understand insight computationally, capturing a human's experience with its outside world. These sensors mimic and measure human sensory experience. Here we review sensors developed to mimic the body's unimpaired natural functionality.

Concerning touch, tactile gloves serve as pressure sensors applied by the hand. High-resolution pressure sensing and mapping enable understanding of human movements, comfort, and ergonomics through visual feedback of hand interactions \cite{Patents2002FlexibleSensor}. Electronic instruments, including the electronic tongue, measure and compare tastes. These instruments can detect an array of dissolved organic and inorganic compounds like the human taste receptors via reactions for each sensor. They generate electric signals as potentiometric variations, to relate the signal strength to the concentration of these molecules \cite{Maezawa2008SomatosensoryElectrodes}. Microphones can easily capture the audio experiences of an individual and are ubiquitous in today's world through mobile phones, computers, wearables, and IoT devices. Cameras and videos provide visual representations of what a human sees. Google Glass and Snap Glasses are both wearable cameras that are worn as a pair of glasses but can record or observe all the information that an individual sees directly\cite{Muensterer2014GoogleStudy}. Lifelogging cameras have thus been in research for several years now \cite{Wolf2014Lifelogging:Camera}. An example of cameras understanding human perception includes the three-dimensional gesture recognition vision systems in which video of the user's gestures are understood for certain meaning and actions \cite{Lv2014Hand-freeGlass}. These sensory signals give us detailed information about our experiences. Understanding experiences is one set of clues towards HSE. For example, if an event is relaxing, it may impact the health state positively versus a stressful event impacting the state negatively.

\textit{Perception and Internal Cognizance}
Experiences combined with internal signals that interpret health is another avenue by which to collect data about the individual. The self-perception theory indicates that people will know their emotions, attitudes, and other internal states based on their own understanding of their behavior and the circumstances at hand. ``If these signals are weak or non-interpretable, the individual becomes an observer and must rely on cues to understand their inner state" \cite{Bem1972Self-PerceptionTheory}. This theory implies that internal sensory signals are learned and understood from making interpretations about their behavioral and physical circumstances.

Furthermore, our internal receptors and neurons enable us to comprehend our sense of position and movement. Proprioception is the sense of position and movement of the limbs, occurring by activity in sensory neurons located in the skin, muscles, and joint tissues. Proprioceptors (sensory receptors) play a pivotal role \cite{Grigg1994PeripheralProprioception}. As these receptors are sensitive to both tension and pressure, they relay information concerning muscle dynamics to the conscious and subconscious parts of the central nervous system. They provide the brain with proprioceptive information at the subconscious level, about the body's positioning with respect to gravity\cite{Physiopedia2020MuscleSpindles}.

Perception data may give us one's view regarding their illness, and thus play a major role in the outcome of disease. The treatment of depression in CVD patients may benefit from a psychological intervention focused on patients' illness representations, self-efficacy beliefs, and their perceived social support \cite{Greco2014PredictingMediators,Goldston2008DepressionApproaches}. People who are feeling susceptible to an illness, expect to benefit if they change their behavior, and perceive their social environment as encouraging the change, but if they lack a belief that they can indeed change, their efforts are not likely to succeed. ``Substantial empirical evidence suggests that self-efficacy beliefs (and the related concept of optimism) are reliable predictors of behavior, and that they mediate the effects of intervention on behavior change, including a number of health-related behaviors" \cite{Health2001IndividualsBookshelf}.

The importance of identifying internal signals is to note the link between how internal cognizance and sensory experiences directly impact physiology. Two commonly known examples are the stress-response and the gut-brain connection. The common example of cognitive stress impacting physiology is in Figure \ref{fig:stress} within chapter 2. About 100 trillion bacteria in the gut make neuro-active compounds, including the neurotransmitter serotonin, that regulate our emotions. The neurotransmitter can transmit signals directly to the gastrointestinal system to stimulate or suppress digestion \cite{Washington2018TheHealth}. The ``gut-wrenching feeling" that people have historically is a person's stomach, or intestinal distress may cause anxiety or stress. Psycho-social situations directly influence gastrointestinal physiology and symptoms \cite{Washington2018TheHealth}.

The methods for measuring or capturing information about sensory signals and internal cognizance is a computing challenge, as the information is difficult to attain continuously and unobtrusively. Currently, methods for capturing information is done through self-tracking, querying the user through surveys, yet there is no integration at a broader scale of other objective data streams. There is an increasing number of ``Quantified-Self" individuals who diligently track many kinds of data about themselves, via data collection tools and self-experimentation methods \cite{Choe2014UnderstandingData}. There may be several obstacles to this self-tracking, including tracking too much information and not tracking triggers with sufficient context. The current state of the field is to use Ecological-Momentary-Assessment (EMA) to have context-aware triggers that prompt the surveys \cite{Shiffman2008EcologicalAssessment,Dunton2016FeasibilityActivity}.

Capturing the natural senses with the tools mentioned above helps to feed data towards the estimation, but there is an obvious need to go beyond this. The next section describes data capture methods that go beyond our human senses.

\subsection{Digital Sensors}
Throughout history, quantifying health at a higher resolution has always changed what it means to be considered healthy. The stethoscope, blood pressure cuff, and microscope all modified the definition because of their ability to sense a change in the health state beyond human senses. Modern continuous sensors, digital interactions, and biological measurement devices produce vast amounts of data. \textbf{Continuous digital sensors} including computers, smartphones, and wearables, are relatively inexpensive in cost, energy, and computing power. The significant progress in this technology makes it widely available for monitoring individual health data streams continuously \cite{Majumder2019SmartphoneDiagnosis}. While continuous sensing gives us information readily, non-continuous or \textbf{episodic sensors} are still crucial for attaining high-resolution information with known levels of accuracy. Non-continuous methods that medical experts use with professional equipment give a deeper understanding and verification of ground truth. These methods include imaging techniques such as functional magnetic resonance imaging (fMRI), electrical signal capture such as electroencephalogram (EEG) and electrokardiogram (EKG), and other clinical methods that provide one time-slice high-resolution information. The noncontinuous methods give a good baseline of an individual's state. There is a need for merging information from both continuous and episodic sensors for the best quality estimation over time.

Transitioning further into the digital sensor age, we will explore both continuous and episodic digital sensors. The four key sensor themes for continuous sensors include: perceptual, biologic, environmental, and event sensors. \textbf{Perceptual quantification} records human sensory information and provides continuous streams of data about emotions, sensory inputs, and brain activity. \textbf{Biological sensors} capture physiologic and molecular changes in the biology of the individual. Modern biological sensing mechanisms include “omic” data arrays, wearables, IoT devices, bio-markers, advanced imaging, and electrical signals. \textbf{Environmental sensors} track the external environment to understand a person’s surroundings and social situations. \textbf{Event sensors} track events or outcomes continuously that can be done virtually or physically. Virtual events include how we use our computers, phones, wearables, and other computing devices into a single collection called a screenome \cite{Reeves2020TimeProject}. Physical events include the real-world daily life events of individuals, such as those as described by Nobel laureate Daniel Kahneman These are captured in personal life-logging applications such as personicle \cite{Kahneman2004AMethod,Oh2017FromChronicles}. There are various modalities for acquiring information about these themes. These include sensing mechanisms such as imaging, electrical sensing, optical sensing, HCI, audio, radio-frequency, and strain gauges.

\textbf{Perceptual quantification} of human sensory information can be observed by methods of observing sensory inputs and brain activity. It is a computational challenge to find methods for quantifying human perception via sensors; however, there are growing mechanisms to compute behavior and perception. 

Galvanic skin resistance (GSR) refers to the electrical resistance between electrodes when a weak current passes between them. The electrodes are placed apart and the resistance will vary based on the emotional state of the subject due to variations in sweat gland fluid production. This sensor can be useful for the skin, as the skin serves as a conductor of electricity in situations when internal or external stimuli are physiologically arousing. ``Arousal is widely considered to be one of the two main dimensions of an emotional response. Measuring arousal is not the same as measuring emotion, but is an important component of it. Arousal has been found to be a strong predictor of attention and memory" \cite{MITMediaLab2020FrequentlyQuestions}. Stress sensors have been built based on GSR to measure test stress responses and detect stress states of an individual  \cite{Villarejo2012AZigBee}.

The advent of the smartphone and wearable devices has increased the possibilities of embedding sensors such as the accelerometer, GPS, optical sensors, and microphone. The integration of the sensors can collectively analyze patterns to understand an individual's mental health. We can use passive data collected via devices (such as call history, text messages) or by active interpretation methods, such as analyzing voices in the microphone to measure stress. We can capture data on physical activity and movement during sleep from an accelerometer, optical heart rate, and GPS for a location to determine context. The combination of all of these gives a picture of an individual's mental state \cite{Majumder2019SmartphoneDiagnosis}.

The detection of emotions is becoming an increasingly growing area of research and development via emotion recognition methods. Corive et al. describes two channels of human interaction, one transmitting explicit messages and other transmitting implicit messages about speakers. The explicit channel is understood by linguistics and technology, but implicit messaging, such as understanding emotions, is not well understood \cite{Corive2001EmotionInteraction}. Techniques by which to process imaging and audio helps establish pattern recognition to uncover human emotions. The techniques may include signals from bio-sensors, such as Haag's method for training computers to recognize the emotions of the wearers\cite{Haag2004EmotionSystem}. ``Biosensors may be used based on the synchronization between audiovisual stimuli and the corresponding physiological responses" \cite{Hui2018CoverageBiosensors}. Emotion recognition techniques may also include analyzing speech signals (such as pitch and energy) \cite{KwonOhWookChangKwokleungHaoJiucangLee2003EurospeechAbstract}. 

In a broader sense of computing intelligence, The Turing test grades the ability of a computer to exhibit intelligent behavior indistinguishable from a real person \cite{Turing2009ComputingIntelligence,Shieber1994LessonsTest}. Human health services such as a human doctor have a heightened sensory ability to understand the above perceptual aspects about a person, and hence are still sought after to deliver healthcare to people. If a computer can understand these same factors, we may be able to scale health services to more people with better performance. Hence, the above perception and emotion-sensing research and related challenges are vital if computing intelligence aims towards improving human health. Human doctors do not solely use this mode of sensing. They also incorporate objective information about a person's biology.

\textbf{Biological} sensors measure internal changes in biological and physiologic function over time. There currently exists a vast selection of sensors that exist for capturing biological information beyond the scope of this thesis. We focus on devices such as wearables and smartphones that act as biological sensors in a continuous manner \cite{Majumder2019SmartphoneDiagnosis}.We review a few modalities for biological sensing below:

a) Optical sensors within a wearable device continuously and unobtrusively monitor heart rates over a wearer's wrist or head \cite{Holz2017Glabella}. Optical sensors also detect oxygen saturation in the blood, and current research is conducted on blood pressure and glucose monitoring. The optical sensors work by using a reflected light pattern scatter from skin surface contact to estimate then the parameter of choice similar to a radar mechanism.

b) Imaging modalities are useful when screening blood for hemoglobin through cameras to understand a patient's response to iron supplement treatment \cite{Wang2016HemaApp:Cameras}. With a light source shining through a person's finger, they have performed a chromatic analysis to analyze the color of their blood, estimating the hemoglobin level. Here a camera is directly collecting light, rather than looking for a return signal of scatter. Traditional images of the skin can also be used to detect changes in texture for wrinkles, color for changes in melanin and jaundice \cite{DeGreef2014BiliCam:Jaundice}.

c) Audio modalities such as the microphone can serve as a spirometer to measure lung function. Home spirometry is more widely accepted for its ability to  continuously detect pulmonary exacerbation and improve outcomes of chronic lung ailments. Larson et al. presented SpiroSmart, a low-cost mobile phone application that performs spirometry sensing using the built-in microphone \cite{Larson2012SpiroSmart:Phone}. Audio methods also capture lung, heart, and bowel sounds though digital stethoscopes \cite{Leng2015TheStethoscope}.

d) Strain gauge sensors measure the amount of force on the system. These are prevalent in many applications such as power meters, accelerometers, and the old-fashioned weight scale. Strain gauges work on the principle of Hooke's Law as given in Equation \ref{hooke}.

\begin{equation} \label{hooke}
    F = kx
\end{equation}

Where \textit{F} is the force applied, \textit{k} is the spring constant, and \textit{x} is the distance the spring changes. The most classic strain gauge sensor, yet still valuable, is the simple weight scale to measure body weight. These have been combined with electrical impedance sensors to get both body weight and body adipose tissue content \cite{Bera2014BioelectricalReview}. Flexible skin-attachable strain-gauge sensors are an essential component in the development of artificial systems that can mimic the complex characteristics of the human skin. Pang et al. describe a strain sensor that enables the detection of pressure, shear, and torsion.  Strain-gauge sensors are used to monitor signals from human heartbeats \cite{Pang2012ANanofibres}. We can also measure respiration rate and volume measurements with wearable strain sensors and are useful for monitoring patients with chronic respiratory diseases \cite{Chu2019RespirationSensors}. 

The strain gauge sensor is also used in bicycle ergometers and crank arms to measure the torque applied onto the pedals \cite{Cycling2020TechnologyAmerica,Bini2011AOutput}. Strain gauges can measure general movement activity (such as through wearable accelerometers) and aerobic exercise work capacity as CRF. CRF has a much stronger established relationship with true cardiovascular health \cite{Williams2001PhysicalMeta-analysis}. Widespread use of standard wearable accelerometers that measure steps or higher semantic activities like walking, jogging, biking are indicative of activity only, hence the need for using wearables that can give estimates of CRF for better HSE. Wearable devices have estimated energy expenditure through various computational approaches such as deep learning \cite{Zhu2015UsingSensors}, knowledge-based regression \cite{LUCIA2002KineticsCyclists}, and data filtering and segmentation techniques \cite{Albinali2010UsingEstimation}. Energy expenditure provides insight into the total amount of activity performed by an individual but does not provide maximal work output to estimate CRF.

\begin{figure}[H]
  \centering \centerline{\epsfig{figure=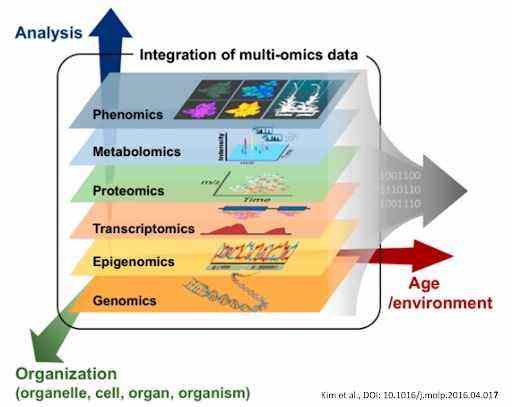,width=12cm}}
  \caption{\textbf{Multi-Omics.} ``Oma" means mass in Latin. Each layer of omic data is a massive array of data parameters regarding the specific layer. The advent of omic data mechanisms allows for high-resolution understanding of the biological state at various abstraction levels. Figure adapted from \cite{Kim2016TowardResearch}.}
    \label{fig:omics}
\end{figure}

Computational research in CRF prediction began in the 1970s with the formulation of exercise stress scoring metrics based on the newly available chest strap based heart rate monitors \cite{Banister1980PlanningTraining.,Borresen2009ThePerformance}. Recently, contextual understanding improved the performance of heart rate based CRF estimation and was further refined by calibrating custom algorithmic parameters for a particular user \cite{Altini2016CardiorespiratorySensors,Altini2015PersonalizedModels}. 
Heart rate data incorporates additional features, such as vagal tone (commonly referred to as heart rate variability) and respiratory rate, to provide regression analysis more features for prediction \cite{Smolander2008AWorkers}. Improvements in accelerometer-based CRF prediction have been achieved through body placement optimization \cite{Parkka2007EstimatingLocations}. The only known research that has attempted the use of multi-modal data for CRF prediction is by Firstbeat Corporation. They use both heart rate and speed information with a proprietary algorithm to filter periods of heart rate that are indicative of steady-state metabolism \cite{Contents2014AutomatedData}.

e) Omics, including genomic, microbiomic, metabolomic, proteomic, transcriptomic, and epigenomic data, is more widely used in understanding the biological model and state. This data can be used given a particular scope of the individual that is being reviewed. A genome consists of all genetic material, microbiome (all microbiota or bacteria that live in and on the body), metabolome (all metabolites), transcriptome (all  RNA transcripts for gene expression), and epigenome (all epigenetic material or changes to DNA not based on genetics). There are two broad types of omic data: static data and dynamic data. Genetic information is more static due to the fixed measurements that give information on risk and a person's unique model. For example, given the same sun exposure, two different individuals may have different responses to the sun due to their genetics. One may be sunburned, and another may not be. The genetics information gives insight into the relationship between sun exposure and the skin type, and thus the risks associated with genetics. Genetics is, therefore, a system model rather than a determinant for the state. Companies like 23andMe and Helix are focused on providing precise reports for genetics directly for the user via DNA genotyping methods done by mass sequencing technologies for individuals to understand their health model \cite{23andMe2020DNA23andMe,Helix2020HelixGenomics}. Dynamic omics data, including data from the metabolome, transcriptome, microbiome, and epigenome, is always in flux with a high rate of change. These give us a systems state with continuous information over time. For example, glucose is a metabolite that is constantly changing. With the transcriptome, each exercise event an individual may partake in rapidly changes one's total RNA and gene expression. The research challenge with dynamic omic data is two-fold. Time-series data is needed to understand how the parameters are changing, and the data still is not available at a low-cost to both user intrusiveness and burden. Omic data research is an active area focused on further understanding of one's precise state \cite{Wu2017-OmicMedicine,Integrative2014TheAuthor,Suravajhala2016Multi-omicWelfare}. Overall, the use of various omic techniques is referred to as a multi-omics approach, as shown in Figure \ref{fig:omics}.

In general, low-cost continuous sensors, computing methods, and lifestyle integration of cardiovascular health hence provide a unique opportunity towards advances in HSE. A robust framework to integrate vast amounts of temporal data and knowledge from biological sensors in cardiovascular health has yet to be published today.

\textbf{Environmental}
Environmental sensors allow for understanding the surroundings of an individual and their exposures. Many environmental sensors are measured directly (such as pressure and temperature sensors) or by sensor networks. Wireless sensor networks sense noise pollution in urban areas, and are available continuously. Traditionally noise pollution measurement has been involved and expensive. Santini et al. present a tool known as tinyLAB, a Matlab-based tool, enabling real-time acquisition, processing, and visualization of data collected in wireless sensor networks \cite{Santini2008FirstMonitoring}. 

While public data regarding the quality of the environment is readily available, it is not incorporated and tracked at the individual level. An open-source software platform, called EventShop, ingests and assimilates different data streams \cite{Gao2012EventShop:Control}. It combines different environmental data streams ranging from climate data, air quality, pollen counts, and micro-blogs (like Instagram and Twitter) to understand how the environment is evolving. Also, the GPS allows us to track an individual's location data and via geospatial data to understand exposures over time \cite{Dias2014ModellingApproach}.

\begin{figure}[H]
  \centering \centerline{\epsfig{figure=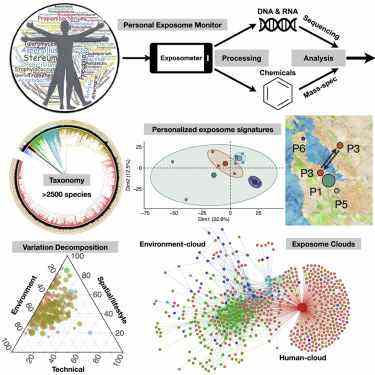,width=12cm}}
  \caption{\textbf{Personal Exposome.} The exposome consists of both biological agents we are exposed to, and physical components such as light, sound, and chemicals. These exposures are related to a certain location and time. Figure adapted from \cite{Jiang2018DynamicMonitoring}.}
    \label{fig:exposome}
\end{figure}

The concept of generalizing environmental impacts on a person is known as the ``exposome," or one's exposure history. The United States
Centers for Disease Control and Prevention (CDC) defines exposome as the ``measure of all the exposures of an individual in a lifetime and how those exposures relate to health" \cite{CDC2014CDCTopic}. This includes a longitudinal exposure throughout one's life since conception. The environmental impact, in conjunction with one's biology, creates the personal exposome, as shown in Figure \ref{fig:exposome}.

This concept of personal exposome is noted in several studies \cite{Wild2005ComplementingEpidemiology,CDC2014CDCTopic,Vrijheid2014ETheHealth,Siroux2016TheResearch,Maitre2019HealthChemicals}. Wild describes the exposome in three domains: 1) general external environment including the urban environment, education, climate factors, social capital, stress, 2) a specific external environment with specific contaminants, radiation, infections, lifestyle factors (e.g. tobacco, alcohol), diet, physical activity, etc., and 3) an internal environment to include internal biological factors such as metabolic factors, hormones, gut microflora, inflammation, oxidative stress \cite{Wild2005ComplementingEpidemiology}. He describes that the exposome requires consideration for both the nature of exposures and their changes over time. Jiang et al. developed a sensitive method to monitor personal airborne biological and chemical exposures. They found that human exposomes are diverse, dynamic, spatiotemporally-driven interaction networks with the potential to impact human health \cite{Jiang2018DynamicMonitoring}. The exposome allows us to gather information specifically about an individual and be tracking their exposure history and how this may impact their health from a personalized level. Listed below are various tools used for measuring the exposome \cite{Wild2005ComplementingEpidemiology}.  

Measuring changing environmental pollutants can help model human exposure to micro-environments. Having the ability to measure the local environment of each individual will give insight into how they are affected by unnoticeable factors.  For example, air and water pollution increase the risk of diabetes \cite{Rajagopalan2012AirInsights}. Long term exposure to particulate matter in the air can activate pathophysiological responses that can induce insulin resistance \cite{Zheng2013ExposureModel,Liu2014ExaggeratedMellitus}. A methodology developed and implemented by Dias provides the time-sequence of the exposure events, making possible association of the exposure with the individual activities and delivers main statistics on an individual’s air pollution exposure with high spatiotemporal resolution \cite{Dias2014ModellingApproach}. The data combined with air pollutant concentrations for traffic pollution, estimate the micro-environments given a probabilistic model. 

\begin{table}[t]
\centering
\resizebox{\textwidth}{!}{%
\begin{tabular}{@{}ll@{}}
\toprule
\textbf{Approach}                  & \textbf{Tools}                                                   \\ \midrule
Biomarkers (omics)                 &                                                                  \\
General                            & Genomics, transcriptomics, proteomics, metabolomics, epigenomics \\
Targeted                           & Adductomics, lipidomics, immunomics                              \\
Sensor technologies           & Environmental pollutants, physical activity, stress, circadian rhythms, location \\
Imaging                            & Diet, environment, social interactions                           \\
Portable computerized devices & Behavior and experiences (EMA), stress, diet, physical activity                  \\
Improved conventional measurements & Job-exposure matrices; dietary recall                            \\ \bottomrule
\end{tabular}%
}
\caption{\textbf{Exposome tools for measurement.} Figure adapted from \cite{Wild2005ComplementingEpidemiology}.}
\label{tab:exposome-tools}
\end{table}

\textbf{Events}
Digital methods are increasingly capturing events of human life. An event is a signal capturing the occurrence of an action. These daily life actions are what make up the history of each human life and also provide an input that changes the health state. With the rise of smartphone applications capturing events, individuals can easily track daily human activities, nutrition, sleep, and location to give a few examples. Physical activities can be tracked by wearable devices that can measure activity type, location, distance, and other relevant attributes \cite{Strava2017StravaAthletes}. Photos can easily capture food consumption habits. Image recognition of these photos can help to understand nutrition behaviors. There is growing accuracy in recognition of food photos and large scale images \cite{Yanai2015FoodFine-tuning} \cite{Kawano2014FoodFeatures}. The ``personicle" concept is the most holistic application to collect lifelog events \cite{Jal2014Personicle:Events}. Life events can be at various scales of time and space. Molecular interactions occur at the atomic size and nanosecond scale, while an event such as ``attending college" can last years and span a large area of space. Building a unified data structure to capture any event is essential for a robust data collection system. There have been efforts in pulling a unified view on how to represent these events, which include six key aspects (Structural, Temporal, Causal, Spatial, Experiential, Informational) \cite{Westermann2007TowardApplications}.

Unique to the digital era, time spent with screens is a large part of awake events in the current world. There are growing mechanisms for capturing event data about an individual in this light. The ``screenome" reflects the use of digital media and screen time to analyze what people do and spend time with on their screens \cite{Reeves2020TimeProject}. Each digital interaction gives insight into a neurological event. Smartphone applications such as Apple ScreenTime and RescueTime tracks habits and time spent on particular applications or time on web \cite{Apple2020UseSupport}. Human interactions with social media capture much of an individual's time spent (i.e., interests, activities, mental well-being). The interactions give us an understanding of individual choices, behavior, culture, and habits. To further understand one's relationship with the screen, there is a proposition for a Human Screenome Project in which ``screenomics data should be sifted using a gamut of approaches — from deep-dive qualitative analyses to algorithms that mine and classify patterns and structures" \cite{Reeves2020TimeProject}. The project proposes reviewing an individual's screen time to investigate physiological and psychological health states and sociological dynamics of interpersonal relations over time. Further research continues on the ``screenome" for understanding screen time and human behaviors \cite{Yang2019UsingSmartphones}.

\textbf{Episodic sensors}
Healthcare practice relies primarily on episodic or non-continuous sensors for dependable data acquisition. Professional-grade equipment generally has a higher expense but also higher accuracy and depth, which provides greater certainty in medical decision making. Professional equipment generally needs an operator for the measurements and may be more challenging to obtain this information unless it is a medical necessity. Although useful at the time taken, data collected in this manner is generally not used for any future steps in the person's life after the clinical encounter. Professional-grade equipment provides calibrations at a current snapshot of the health state rather than continuous monitoring. Computing methods that are suitable for this include Kalman filtering \cite{Gustafsson2013StatisticalFusion}. Observing the role of accuracy and precision for each sensor is important based on the system goals. While many professional-grade types of equipment have high accuracy, it may be more relevant to hold precision (with continuous measurements), to detect a change in the health state. Episodic measurements combined with continuous measurements are therefore most ideal, increasing the accuracy of the current state with high temporal resolution to changes in the state.

Conventional professional equipment used for collecting individual data includes imaging, microscopy, and blood work. For example, magnetic resonance imaging (MRI) techniques are sensitive to changes in cerebral blood flow and blood oxygenation \cite{Kwong1992DynamicStimulation}, and cranial magnetic resonance imaging (cMRI) have been used in cardiovascular studies to determine that sulcal width, ventricular size, and white matter signal intensity change with age, sex, and race \cite{yu2005mining}. Other equipment such as dual-energy X-ray absorptiometry (DEXA) scanners, allow for one-off measurements of bone density and visceral fats \cite{Chen2012AbdominalMice}.

Several medical procedures to gather data have risks associated that may be adverse to an individual. For example, an angiogram is an invasive medical procedure that holds the risk of radiation exposure from X-rays used, major complications including injury to the catheterized artery, arrhythmia's, allergic reaction to dyes, kidney damage, and more \cite{Clinic2020CoronaryAngiogram}. Thus, in addition to being episodic, there are also non-trivial risks to these high-quality measurements. These expensive sensing mechanisms initialize, calibrate, or validate a health state, and the less expensive sensors, which are more continuous, non-invasive, and accessible, would be available to estimate the current state from the most recently calibrated state.

\subsection{External Data Sources}
There may be sources that are external to the individual that may be capturing the individual's health information. This includes domain experts who are capturing measurements or observations on the individual, or external agents that an individual interacts with, such as a business.

\textbf{Domain Experts}
Domain experts who are trained professionals can have a role in collecting data manually. For example, a physical therapist may use direct measurements and observations for determining an individual's flexibility or with goniometers. Clinicians commonly use norm-referenced, standardized measurements \cite{Horak1987ClinicalAdults}. In the field of psychology, experts question individuals to understand their psychological state and needs, and thus mental health relies on domain experts for detecting and determining psychiatric issues.  This system largely relies on domain experts to act as the sensor \cite{Treffers1990ThePsychiatry}.

\textbf{External Platforms}
Data acquisition about individuals is also common through platforms operated by external entities and businesses. The platforms designed to collect individual data are for pattern analysis to observe the individual's behavior. Loyalty programs for restaurants, credit cards, and stores collect information on individuals by observing their purchasing behavior. The programs have sophisticated tracking mechanisms between a customer and a brand that can help the program better understand the buyer's habits. Current progress in this field aims to use this information during a live interaction with a customer \cite{Davenport2013BigCompanies}. Companies may be using searches, clicks, and views on particular topics in order to understand an individual's interest and target advertisements accordingly. This process may not be at the request of the individual but based on the websites an individual visits and items they buy. External data gathering may also allow agents to track habits, behaviors, and patterns on individuals, giving the insight to develop the business \cite{Mcafee2012BigRevolution}. The world's largest consumer-facing technology companies, such as Google, Facebook, Amazon, and Netflix, critically depend on this data acquisition process for their business performance.

The data digitally collected and stored externally is vast and expanding rapidly. As a result, organizations convert this information and knowledge for their objectives, whether it be in ``astronomy (e.g., the Sloan Digital Sky Survey of telescopic information), retail sales (e.g., Walmart's expansive number of transactions), search engines (e.g., Google's customization of individual searches based on previous web data), and politics (e.g., a campaign's focus of political advertisements on people most likely to support their candidate based on web searches)" \cite{Murdoch2013TheCare}. According to Murdoch, it is inevitable that this data will have a large role in providing insight into the health state, such as data gathering mechanisms to observe broader trends to predict emergency room visits. Google Trends (a free, publicly available aggregator of relevant search terms) has been used to predict surges in flu-related emergency room visits \cite{Google2020GoogleTrends}. Similarly, Twitter updates were as accurate as official reports at tracking the spread of cholera in Haiti after the January 2010 earthquake \cite{Mcafee2012BigRevolution}.

Data collection through sensing mechanisms remains a considerable research challenge for humans due to the complex nature of understanding and capturing the health state. The above review is just the tip of the iceberg. The next step in the health state computing pipeline is to use these sensor measurements within an estimation approach.


\section{State Estimation for Individual Health}
State estimation for health allows us to understand the current health situation for an individual. We will review the various approaches for how health can be estimated: quantified approaches, classification approaches, probabilistic approaches, and subjective approaches. It is important to note that though various approaches may exist, they are highly interconnected, and each one alone may not be sufficient.

\begin{figure}[H]
  \centering \centerline{\epsfig{figure=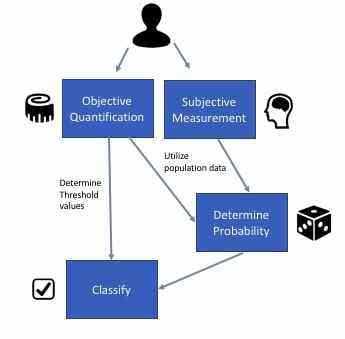,width=8cm}}
  \caption{\textbf{Estimation Approaches are Interconnected.} }
    \label{fig:estimation}
\end{figure}

\subsection{Quantified Approaches}
\begin{figure}[H]
  \centering \centerline{\epsfig{figure=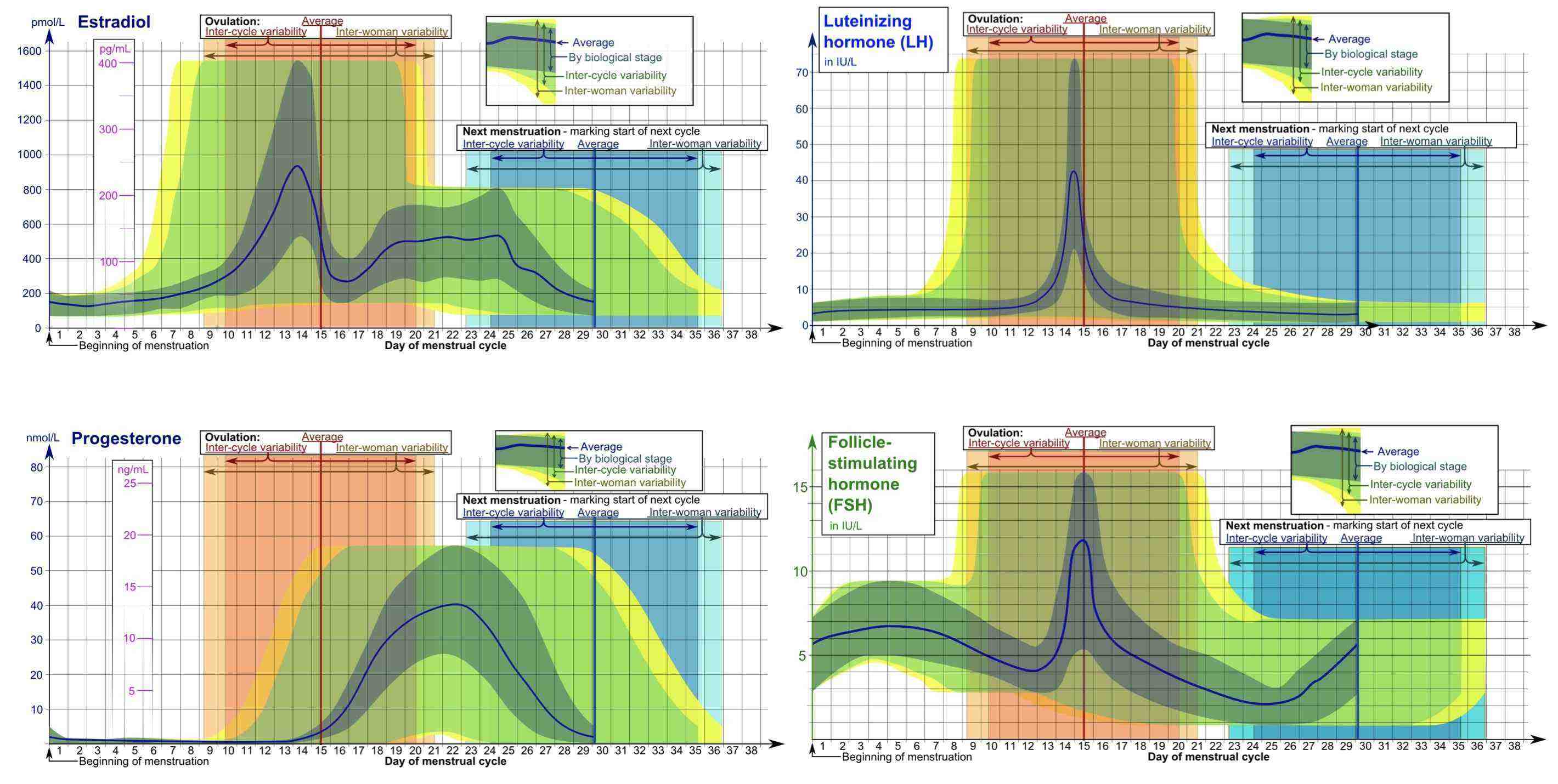,width=16cm}}
  \caption{\textbf{State Space for Quantified Health.} Here are four charts showing how hormonal fluctuation occurs during menstruation. At each moment, an ovulating female has a point on this state space they exist. Regions specific the limits to which the state can exist and what the state means \cite{UpToDate2016Evidence-BasedUpToDate}.}
    \label{fig:hormones}
\end{figure}

Quantified approaches give insight into where an individual is within a state space based on objective measurements. Individuals take blood tests, such as a basic metabolic panel, which is a group of tests that measures different chemicals in the blood, including blood glucose, calcium, electrolytes, and lipoproteins. However, the extrapolation of the information then allows the physician to come to a conclusion or learn what these levels may indicate. A physician can form a conclusion about the readings depending on the parameter of the state space. For example, if HDL cholesterol is 20, that is considered in a poor state, compared to a value of 70, which is a good state. A visualization of these quantified states in the case of hormones is in Figure \ref{fig:hormones}.

One's blood alcohol level is determined based on measurements that determine the specific amounts of alcohol in the blood. If the blood alcohol content (BAC) reaches a threshold of 0.08, it indicates legal intoxication, which is a threshold-based classification. Quantification is necessary, but may not be sufficient alone within an interpretation or added meaning of what this measurement indicates. Similarly, eye power is measured by a diopter, a unit of measurement of the optical power of a lens or curved mirror, which is equal to the reciprocal of the focal length measures in meters \cite{Dictionary.com2020DiopterDictionary.com}. The objective measurement of power can be useful when determining thresholds, such as determining whether an individual is legally blind or not. This need for classification may be determined external to the individual (i.e., safety, legal) not necessarily based on the health state of the individual. Thus placing a layer of information on top of the quantification approach provides more meaningfulness to the health state.

\subsection{Classification Approaches}

Classification methods, as we have described in Chapter 2, define health based on disease, conditions, and various measurements to define the state of one's health. Classifications may help in creating standards that may be beyond an individual's specific situation. In the example of the blood alcohol content, the standards of classifications can identify toxicity and thus concerns about safety. Disease classification is internationally recognized through the International Classification for Disease (ICD), Tenth Revision, by the World Health Organization. The ICD-10 code serves as a classifying method to label whether an individual is in a disease state. The ICD allows for systematic recording analysis, interpretation of mortability and morbidity in various locations globally, and disease classification for health problems \cite{Organization2004InternationalProblems}. 

Widely used classifications of disease include (1) topographic, by bodily region or system (i.e., vascular disease), (2) anatomic, by organ or tissue (i,.e. heart disease), (3) physiological (i.e., metabolic disease), by function or effect, (4) pathological (i.e., inflammatory disease), by the nature of the disease process, (5) etiologic (causal) (i.e., staphylococcal causing pneumonia), (6) juristic, by speed of advent of death, (7) epidemiological (i.e., typhoid), and (8) statistical (i.e., diet as a cause of atherosclerosis) \cite{Robbins2011HumanBritannica.com}. For each of these classifications, we see that there may or may not be threshold values that contribute to classification.

\begin{figure}[H]
  \centering \centerline{\epsfig{figure=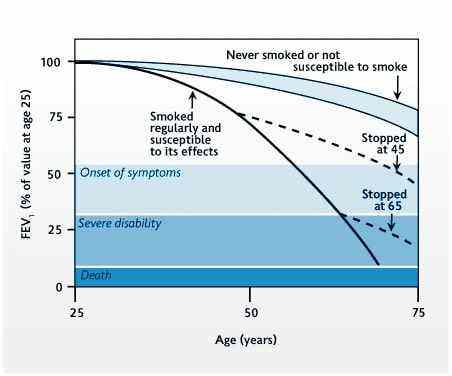,width=12cm}}
  \caption{\textbf{Fletcher Curve.} Lung function is measured through a quantitative mean of FEV1, which is the Forced Expired Volume a person can exhale in 1 second. This parameter is then placed upon the map of which regions and thresholds are relevant for understanding health state. \cite{Tantucci2012LungCOPD}.}
    \label{fig:fletcher}
\end{figure}

\textit{Threshold}
Classification approaches are defined by how an individual's health can be labeled. It may be a particular threshold met to define whether an individual has a disease or does not. Thresholds are, therefore, a region of interest within a continuum. For example, in Figure \ref{fig:fletcher} showing the Fletcher Curve for lung function decline, the classification of where one falls along the curve can be ``onset of symptoms," ``disability," or ``death" \cite{Tantucci2012LungCOPD}. The thresholds are determined generally by population data. It would not be possible to estimate the health of one individual solely and classify this individual's state unless it was a postmortem analysis. Therefore thresholds are more commonly generated given population data.

\textit{Categorical}
Classification approaches are also important in a categorical sense. For example, in many cases of cancer, a tumor can be defined as metastasized or benign without any particular threshold met. One either has cancer or does not. The same applies to infection; one can either test positive or negative for infection with no thresholds to determine the disease. We also see classifications of sex if an individual either has chromosomes XX or XY. There is no in-between threshold for these classifications.

Classification approaches alone are not sufficient for defining the state of an individual's health if used solely; there will be no observation of state change at a high resolution, and change will only be detected when crossing a threshold. The continuum is essential for recognizing alongside classifications.

\subsection{Probabilistic Approaches}
Probabilistic approaches measure the likelihood of a future event. Risk factors are needed to compute the probability of having a disease. There are methods of calculating risk that takes into consideration age, biometrics, and gender to determine chances for having a disease. The information in this approach may be based on the estimation of a population or based on an individual's unique information. Essentially this approach is giving a person a distance between an undesired region on the state space and their current state.

\textit{Population Based}
Population-based probabilistic approaches use population samples that are most similar to the individual. It is essentially using other people as a model for one's self. Population health is ``the health outcomes of a group of individuals, including the distribution of such outcomes within the group," including health outcomes, patterns of health determinants, and policies and interventions that are links to outcomes and determinants \cite{Kindig2003WhatHealth}. Health can be measured at a population level to understand the functioning and well-being of a group that is similar to an individual. Classic factors for calculating population health include geography and demographics information.

In regards to CVD, population-level health estimation is a primary way of understanding behaviors that contribute to cardiovascular disease for individuals. Multifactorial cardiovascular health risks have been investigated in many of the extensive epidemiological studies such as the Framingham Study, which lay the foundation for most modern clinical guidelines by the AHA and American College of Cardiology (ACC). Conclusions from these large studies are useful in current clinical practice through the Atherosclerotic Cardiovascular Disease (ASCVD) calculator \cite{Institute2018FraminghamNHLBI}. This pooled cohort algorithm was based on linear regression analysis for four separate cohorts of individuals female blacks, male blacks, female whites, male whites. They also take into account age, systolic and diastolic blood pressure, cholesterol, smoking history, diabetes (binary field: yes or no), medication history (hypertension, statins, aspirin only), and is only applicable for patients in the age range of 40-79. The limitations of this calculator include the requirement of invasive blood data and non-consumer based lab processing. The estimated risks and scores are specific to defined combinations of the risk factors and demonstrate how they vary over a broad spectrum of potential profiles. These determined risks are now primarily used as guidelines for understanding individual risk \cite{Whelton20182017Guidelines}. The American College of Cardiology takes the inputs of risk factors for ASCVD  such as age, gender, blood work (systolic blood pressure, diastolic blood pressure, cholesterol), medical history, and medication status. The calculation will generate the chances by percentage for the likelihood of ASCVD. More information such as genetics and behavior can further enhance the accuracy of the risk calculation \cite{Whelton20182017Guidelines,Cardiology2020ASCVD+,Lloyd-Jones2019UseCardiology}.

There are many human health risk assessments, models, and tools developed by the Environmental Protection Agency (EPA). For example, the ``All Ages Lead Model (AALM) predicts lead concentration in body tissues and organs for a hypothetical individual, based on a simulated lifetime of lead exposure" and ``statistical methods extrapolate to a population of similarly exposed individuals." This model provides risk assessors with a tool for evaluating the sources of lead in a population or potential exposure to lead \cite{EPA2020HumanDatabases}.

There have been several efforts for making long-term population health research using probabilistic approaches more expansive, include the United States Precision Medicine Initiative led by President Barack Obama. The Precision Medicine Initiative is a long-term research endeavor, involving the National Institutes of Health (NIH) and multiple other research centers, which aims to understand how a person's genetics, environment, and lifestyle can help determine the best approach to prevent or treat disease \cite{House2020PrecisionInitiative}. Other examples of long-term population health research include Mobile Sensor Data to Knowledge \cite{Kumar2017CenterMD2K}, and Alphabet's Verily division \cite{Verily2019XVerily}. These approaches seek to understand what population cohort characteristics allow for similarity comparisons between individuals to group them into sub-populations. These sub-populations allow for a more specific estimation of an individual's health state. The World Health Organization has conducted long term studies for lung cancer as rates had rapidly increased in the 20th century; sub-populations and geography were closely examined in the epidemiological studies to understand the reasons for the spike in disease \cite{Bernard2014World2014}. Long term research efforts for acquiring data may take decades before we have data available for meaningful insight, as they largely depend upon outcomes of morbidity and mortality before they become sufficiently robust. For this reason, they are inappropriate to use as the benchmark for state estimation at the individual level.

\textit{Non-population based}
Personalized models are used in probabilistic approaches to determine one's health state. Current approaches to identify patients at risk for arrhythmia are of low sensitivity and specificity. Personalized approaches to assess Sudden Cardiac Death (SCD) risk in post-infarction patients have been developed based on cardiac imaging and computational modeling. Personalized three-dimensional computer models constructed post-infarction hearts from patients' clinical magnetic resonance imaging. The propensity of each model to develop arrhythmia was assessed \cite{Arevalo2016ArrhythmiaModels}, as shown in Figure \ref{fig:arevalo} in chapter 1 (personalized heart models).

Personal experience may also account for valuable information in probabilistic approaches. For example, if an individual experiences knee pain when cycling that leads to knee tendinitis, by experience, they can determine if the probability of the same pain occurring to be associated with the tendinitis if they have data collected. These are experience-based probabilities that can help individuals estimate their health state. The personal experience and perception of pain in determining the probability of a health state may also a subjective approach.

\subsection{Subjective Approaches}
Subjective approaches for estimating health include observation and perceptual data that can be used as information to understand one's state and also future risks. If any individual holds information such as self history, changes in feelings over time (i.e., pain increase over time), this information can be used in correlation to other objective metrics to generate a conclusion about the current state. This method for gathering knowledge and information depends on a user's intent or experience. We can observe this, for example, in the dynamic nature of recreation experience. Experience patterns capture engagement over time. Seven qualities of hikers' experiences (four mood measures, two satisfaction measures, and landscape scenic beauty) were assessed at twelve moments during a hike in Hull's study. The subjective data components may be relevant for understanding the experience and how this is contributing to the state of the individual \cite{Hull1992ExperienceExperience}. 

There have been advancements in tools for data collection methods, such as facial recognition, natural language processing (NLP), and surveys that combine pieces to generate an estimation for how an individual is feeling. Facial recognition attempts to estimate emotions and track feelings over time. NLP relates thoughts, language, and patterns of behavior learned through experience to specific outcomes \cite{Manning2015TheToolkit,Fleischman2002EmotionalGeneration}. Basic survey methods for asking individuals about their lifestyle and behavior are also still common forms to gain knowledge about an individual's health trajectory. 

All of the above-mentioned estimation approaches have a unique value. The integration of the above approaches is still highly sought after in the research and clinical communities. In order to truly have the integration at an individual level for state estimation, we need to investigate modeling techniques.


\section{Personal Human Health Models}
Given the measurements described above and estimation approaches, we need to have a personal model to fulfill some essential design requirements mentioned. A model characterizes the relationship between the inputs and outputs to one's health. The most obvious benefit the ability to predict, which requires having a model by which to estimate a future state. Thus, we review various approaches in forecasting or modeling long term health. To design a robust model, we must incorporate domain knowledge, taking epidemiological data, and individual data to ultimately make a generalized personal model. We must note that the model is not always static and needs to be updated.

\begin{figure}[H]
  \centering \centerline{\epsfig{figure=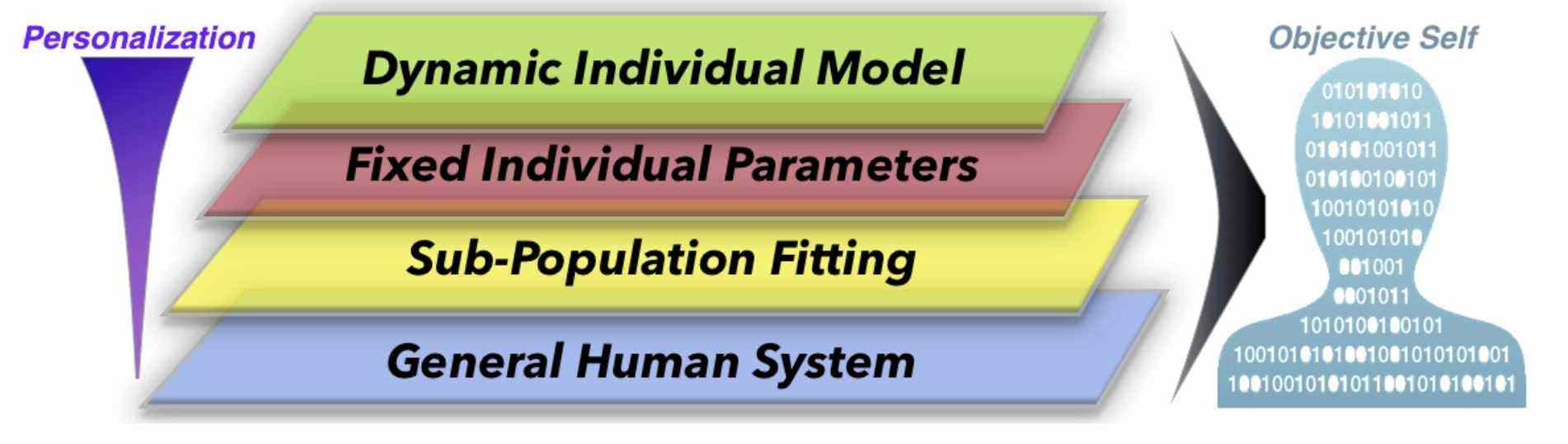,width=12cm}}
  \caption{\textbf{Layers of personal data modeling.}}
    \label{fig:personalmodel}
\end{figure}

There are several distinct levels we consider for building a personal model. In the first level, researchers have applied general rules directly related to known human domain knowledge (i.e., biology, physiology, behavior) to establish a general human system model. This model approach is most familiar in the form of textbooks. In the second, we incorporate specific knowledge that applies to sub-categories of similar people that we call cohort generalization or sub-population fitting. This layer is the function of the individual at a specific time and place. Clinicians use this method quite often when suggesting a course of action for a particular patient. Third, we consider ``static" variables such as genetics and birthdate (fixed individual parameters). Fourth, dynamic individual models use the previously mentioned three layers with constant real-time data to update the model. The visualization of these layers is shown in Figure \ref{fig:personalmodel}.

\subsection{General Human Systems}
We first review models using general rules provided by medical and biological knowledge to represent the human. These are objective models based on well-defined textbook knowledge that gives us the insight to forecast human health. Models help us understand a human's relationship to the environment, its current situation, and plans for the future. We will review models using biological and behavioral domain knowledge.

\textit{Physiological modeling} takes biological knowledge and apply this to our understanding of how a person functions. Biological information regarding the individual is captured in, for example, the Athlete Biological Passport (ABP). This model is based on the monitoring of biomarkers to catch doping with performance-enhancing drugs. The ABP gives a unique mark on an individual's hematological profile as a baseline, and any deterrence in the system, such as through doping, will trigger a physiological change \cite{Sottas2011ThePassport}. This model applies a standard set of thresholds to all athletes, such as having a hematocrit under 50 is considered normal. 

\begin{figure}[H]
  \centering \centerline{\epsfig{figure=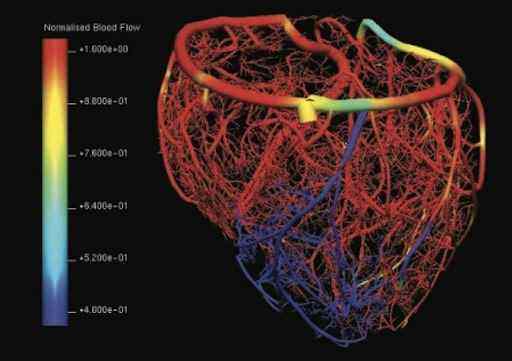,width=10cm}}
  \caption{\textbf{Modeling the Heart.} This figure shows the left coronary circulation model subjected to constriction of one of the main branches leading to blocked blood flow \cite{Noble2002ModelingOrgan}.}
    \label{fig:heart}
\end{figure}

Researchers have also built models regarding functional biological components, such as a model of the heart for better understanding the functional interactions between cells, organs, and systems, as well as how interactions between them change disease states. ``The rapid growth in biological databases; models of cells, tissues, and organs; and the development of powerful computing hardware and algorithms have made it possible to explore functionality in a quantitative manner all the way from the level of genes to the physiological function of whole organs and regulatory systems" \cite{Noble2002ModelingOrgan}. Figure \ref{fig:heart} is a model determined by several human samples, giving us a representation of the constriction of blood flow within an example heart. This type of model allows for the simulation of various physiological situations.

Computational modeling can be used for biological processes that are occurring within the mind. For example, the neurological process of associative memory has a model of processing. This model demonstrates how the brain handles pieces of information in a serial order. A study conducted sequentially arranged patterns to simulate memory serving as a digital representation of human memory \cite{Wegner1995AMemory}.

\begin{figure}[H]
  \centering \centerline{\epsfig{figure=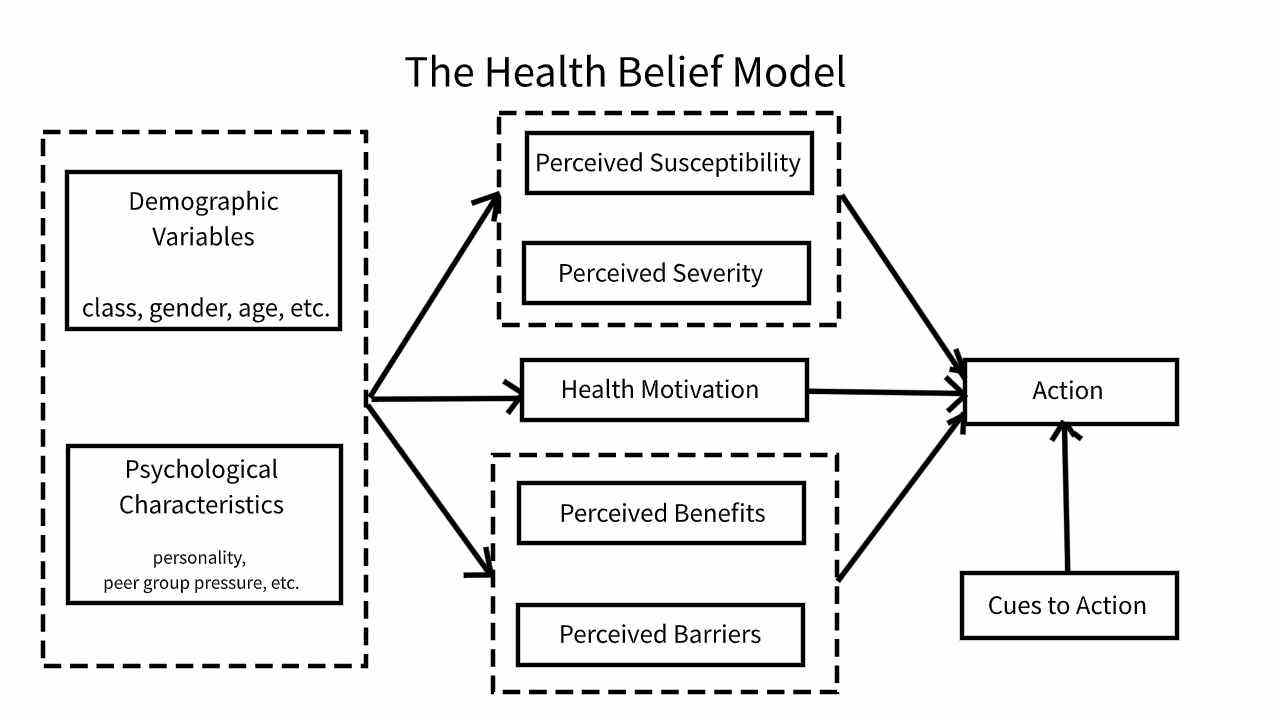,width=16cm}}
          \caption{\textbf{Health Belief Model.} This model predicts the probability of a person taking action that would benefit their health. The model focuses on understanding what people believe in and where they see benefits in certain actions. \cite{Wiley2013HealthModel}.}
    \label{fig:healthbelief}
\end{figure}

\textit{Behavioral modeling} allows for the interpretation of human behavior and preferences. Human behavior can be very complex, and models are often simplified within a system or domain. A traditional behavioral model is Pavlov's classical and operant conditioning, which pairs a neutral stimulus with an unconditioned stimulus to elicit a desired response \cite{Clark2004TheConditioning,Gormezano1975HandbookBooks}. Another early theoretical model developed for understanding health behaviors was the health belief model in Figure \ref{fig:healthbelief}, which shows how particular human attributes contribute to perceived severity, benefits, and motivations to take action on health.

\begin{figure}[H]
  \centering \centerline{\epsfig{figure=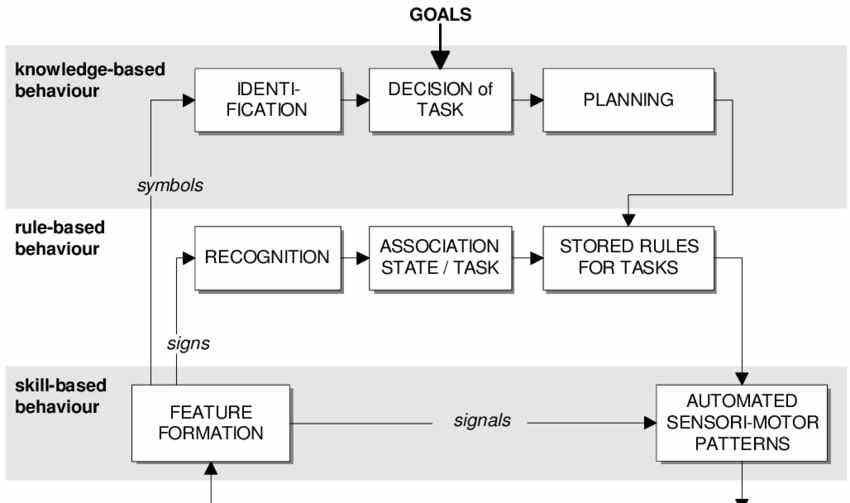,width=16cm}}
          \caption{\textbf{Behavior Models.} Rasmussen's behavior model outlines behavior in three different levels: Skill-based behaviour represents human behaviour that takes place without conscious control. Rule-based behaviour involves task execution controlled by rules. In unfamiliar situations in which there are no rules or procedures, behavior is knowledge-based \cite{Wentink2003RasmussensTraining}.}
    \label{fig:Behavior}
\end{figure}

A third example of a human behavior model shows that behavior has multiple layers that are interconnected. This research was conducted in the modeling the behavior during training laparoscopic surgeons \cite{Wentink2003RasmussensTraining}, shown in Figure \ref{fig:Behavior}.

The model of human driving gives a specific representation of behavior. ``The model of the driver includes lateral and longitudinal control tasks such as path-following, obstacle avoidance, and headway control, examples of steering and braking activities performed by the human driver" \cite{Macadam2003UnderstandingDriver}. Macdam et al. recognize that both physical limitations and other attributes that make the human driver unique, characterize human control behavior. Knowing this model predicts the performance of the human over time.

Quantification methods are needed for behavior models to be more comprehensive and precise. Human interactions have been modeled through Bayesian computer vision systems to recognize human behaviors in visual surveillance. Oliver et al. designed a system combining a top-down with bottom-up information in a closed feedback loop, comparing two different state-based learning architectures for modeling behaviors and interactions \cite{Oliver2000AInteractions}. There are also quantification methods applied to human performance modeling to analyze human function and development of systems for optimal user experience. Sebak et al. summarize this well in their review of behavior models describing how ``human performance models predict human behavior in a task, domain, or system. However, these models must be based upon and compared against empirical human-in-the-loop data to ensure that the human performance predictions are correct" \cite{Sebok2013UsingDevelopment}.

Many of these approaches use a connection between various functional abilities or biological components connected through an informal or formal graph structure. We can see this clearly in many of the figures visually. While using these objective measurements to model general health through biology and behavior may be useful, combining this with models based on demographic data gives us further precision into one's long term health state.

\subsection{Sub-Population Fitting}
Demographic, cohort, or population-based models refine general health models. Cohort generalizations consider population-level findings and apply this data to an individual, based on most similarities to the individual. For example, risk calculators predict cardiac arrest. The Revised Cardiac Risk Index is a cardiac risk stratification tool that allows us to model humans by making taking population data and making calculations based on known unique factors (gender, age, blood panel) a \cite{Gupta2011DevelopmentSurgery}. A defined cohort allows estimation of distributions of prevalence rates of relevant variables, such as risk factors. Advantages of a population-based cohort study arise from the specific health-related parameters for the reference population, which allows the estimation of population risk figures used for planning \cite{Szklo1998Population-basedStudies}. This is a typical medical approach to sub-group matching.

Sub-population data is useful for making assumptions about individuals based on others who are similar to them, known as collaborative filtering. This is a typical computer recommender system approach to sub-group matching. This method makes predictions and recommendations based on feedback information from a community of users. The method assumes that people are similar when they have similar interests and do not assume causal modeling. We see the collaborative filtering techniques commonly used for online platforms, such as Facebook, Google, Netflix, and Amazon, used in recommendations to an individual based on the preferences of others who are similar. The technique applies to healthcare in systems with limited prevalence. Personal healthcare modeling for communications between users in online forums uses this method to monitor the condition of patients and help people with similar conditions interact and exchange experiences \cite{Vlahu-Gjorgievska2011PersonalTechniques}.

\subsection{Static Individual Parameters}
\begin{figure}[H]
  \centering \centerline{\epsfig{figure=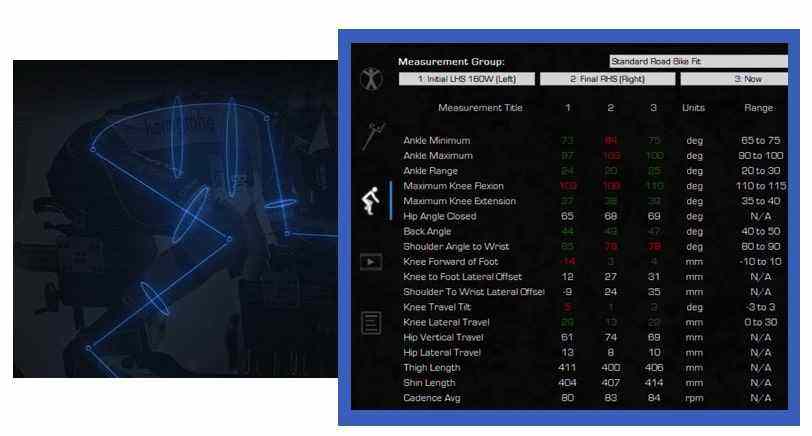,width=16cm}}
          \caption{\textbf{Retul Bike Fit Personal Model.} Individual measurements are collected and applied to a highly customized bike fit.}
    \label{fig:retul}
\end{figure}

Individual parameters that are not constantly changing over time and are ``fixed" can be used to understand an individual's model at a given point in time. These parameters may include genetics data, birth information, and their unique medical history, or developmental disorders. ``Fixed" is in quotations because all components are subject to change, including genetics, thorough DNA damage, and technologies like CRISPR \cite{Cong2013MultiplexSystems}.  Recognizing fixed parameters allows for modeling an individual upon these constraints. For example, the bone length is usually a fixed parameter when an individual reaches adulthood. As the growth plates close, the bone will remain a particular length going forward, providing information on one's body size and enabling customization for health or shopping for clothing. We have seen these models developed in Retul's bike fit testing for helping users select appropriate bike models, improving pedaling technique and selecting bike components for one's specific biomechanics and measurements, as shown in Figure \ref{fig:retul} \cite{Retul2020RetulFitters}. Orthotic specialists use this approach as well. Many multimedia efforts focus on understanding which clothing best suits the fit of an individual's parameters \cite{Ashdown2006AIndustry,Fan2004ClothingTechnology}.

Other static parameters, including genetics, can be useful to model an individual's exposures and future health risk. Companies such as 23andMe are using genetics data to model an individual and give them information about their objective genetic parameters. With this, we can make predictions about a person's future health state, such as genetic predispositions for developing heart disease, cancers, relationship to various foods and activities, and behavioral tendencies \cite{23andMe2020DNA23andMe}.

\subsection{Towards Dynamic Personal Models}
The modeling techniques mentioned above so far assume that the person is an entity that does not change over time. These models assume that the individual model is static. In reality, this is a false assumption. A human model must continuously change due to circumstance, aging, and events that are impacting the person. There have been further efforts in building adaptive modeling techniques.

Contemporary digital technologies measuring constant change, such as on-engine monitors in airplanes, are used to maintain the well-being of jet engines and also similarly could be applied to humans \cite{Tarassenko2018MonitoringPeople}. Tarrasenko's ``digital twin" consists of a particular engine and a unique computer model that accurately adapts to estimate the state of that particular engine. For example, if an engine flies from Los Angeles to Houston where the weather is dry, compared to an engine that flies a rainy route from Vancouver to Montreal, the model of how the engine state changes is different, even if the engine is from the same manufacturer. When applied to people, the digital twin is a computer simulation of an individual that dynamically reflects the individual's molecular status, physiology, and lifestyle over time. An alternative concept of digital human twins relies on a large information infrastructure that characterizes each individual for demographics, biological data, physiology, anatomy, and environment, along with treatment and outcomes for medical conditions. Tarrasenko et al. describe ``the twin represent(ing) the nearest-neighbor patient, derived from the algorithmic matching of the maximal proportion of data points using a sub-type of AI known as nearest-neighbor analysis." This example harks back to the collaborative filtering methods mentioned above.

\begin{figure}[H]
  \centering \centerline{\epsfig{figure=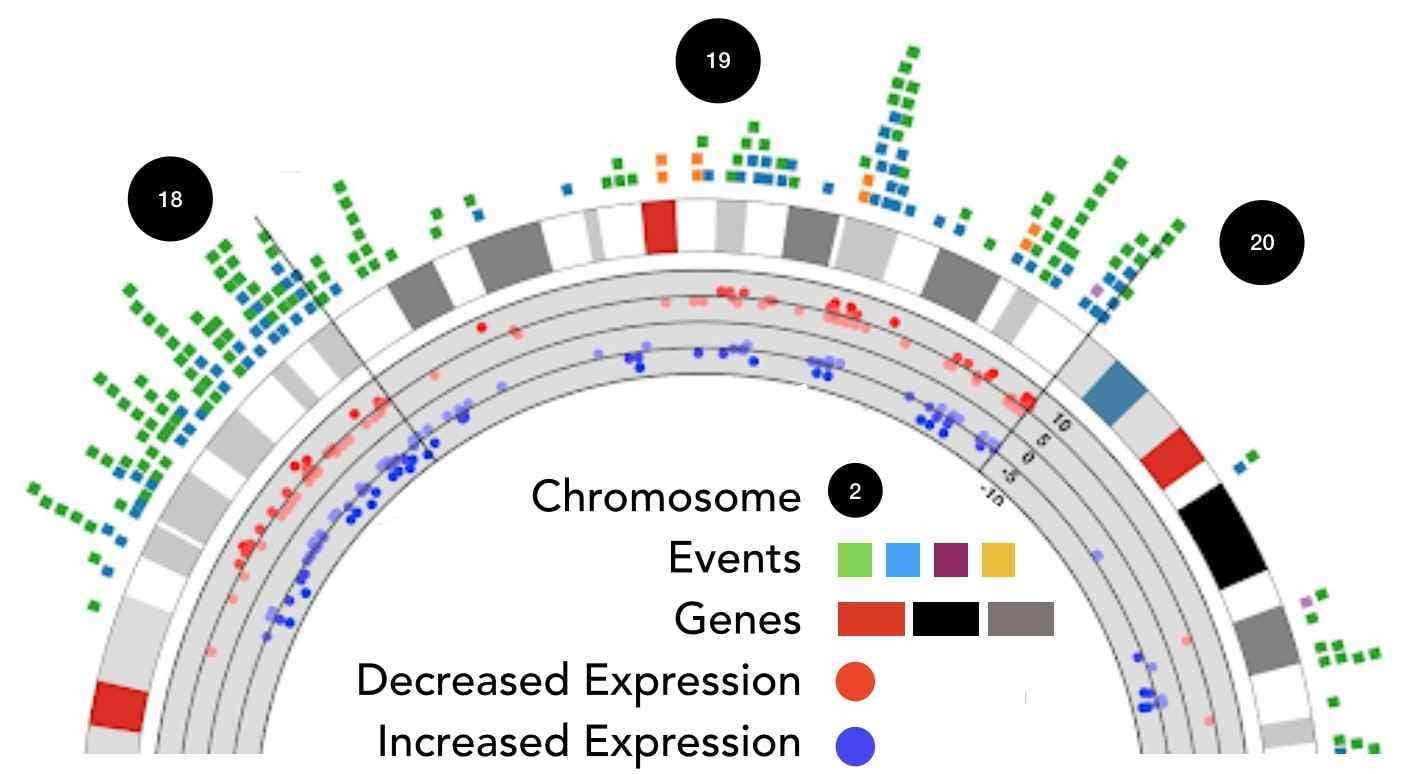,width=14cm}}
          \caption{\textbf{Personal Regulome.} This consists of the complete set of switches for all genes within an individual. Here is a rendition of chromosomes 18-20 of an individual whereby certain events either turn on or turn off the gene expression of specific genes within the individual. This regulome has specificity down to the cell level and is a key pillar in developing dynamic individual models.}
    \label{fig:regulome}
\end{figure}

The Howard Chang Personal Dynamic Regulome Lab works on understanding what inputs change the state of an individual. Their concept is to ``create the GPS system for navigating the regulomic landscape of human health and disease. Each gene has a system of switches that controls when and where a gene will turn on. The regulome is the complete set of switches for all genes. Regulated gene expression plays key roles in nearly every developmental program and disease state, and dozens of human therapeutics act through this regulation to alter gene expression. We aim for technologies to determine rapidly how the regulomes of individual patients and disease tissues maps into a global landscape, and how each intervention changes their trajectories in real-time. This real-time information feedback between perturbation and outcome is essential to tailor precise medical treatments for individual patients. They face challenges of how these new technologies must be compatible with the small sample sizes of human biopsies and clinical workflows if they are to be widely adopted" \cite{Qu2015IndividualityCells}. Essentially, the regulome integrates events with specific changes in biological gene expression. This concept is visualized in Figure \ref{fig:regulome}.

Dynamic modeling can be crudely done currently in systems like STELLA, a visual programming language created to run models through a graphical interface. This model is represented by stocks, flows, and variables to represent a single entity moving in the system. Each of the interconnected icons is defined mathematically. STELLA can be used to represent a biological model for human cholesterol dynamics, circadian rhythm control, and more \cite{Hargrove1998DynamicSciences}. STELLA can be used for modeling cardiovascular function \cite{Romstedt2003ModelingBiology}. Other examples of systems dynamics programs to see entities changing in a system is through Insight Maker, an open-source program to use loop diagrams to create simulation models \cite{Maker2019InsightBrowser}. Important to note, the ``dynamic" word in these approaches refers to the system state is dynamic, while the relationships between variables are generally static.

In order to have a successful dynamic model, the constantly changing health measures are recorded using real-time sensor data. Firstbeat Technologies collects various data sources through wearables and makes sense of the information together and its relevance to an individual's health state. The dynamic model of a person's aerobic fitness level uses the calculation of the maximum volume of oxygen (VO2max). The method uses the heart rate and speed relationship and detects the reliable data period for VO2max estimate during the exercise. Firstbeat has created a method to calculate the VO2max in an uncontrolled real-life exercise. This includes a physiological basis, the linear relationship between oxygen consumption and running speed. The ``technology (such as GPS sensors and foot pods in wrist devices or mobile phones) enables reliable measurement of running speed along with heart rate. These parameters can be monitored continuously and automatically during each workout" \cite{Firstbeat2020WhiteFirstbeat}. They capture the personal information of the individual, such as age. They use a device that calculates heart rate segmentation to estimate the fitness data through the person's heart rate and speed data \cite{Firstbeat2020WhiteFirstbeat}. These models are a good start, but still remain quite primitive in that they only use a handful of variables to build a dynamic model about the individual.


Once again, there is a common theme of needing to use continuous data streams to build models about people that improve over time—connecting the data to the person's model has the most promise in the simple graphical approach. Modern artificial intelligence modeling techniques also have been utilizing graphs as a primary method of connecting data and meaning \cite{RussellStuartNorvig2003NoApproach}. For this reason, we move into reviewing the various techniques for representing the individual dynamically through graph modeling below.


\section{Artificial Intelligence and Graphs}
Humans have spent a greater part of the last century developing machines that can improve the quality of life. Computing, with machine processing information towards a goal, contributed to a large portion of this progress. Machines that can execute complex thinking tasks such as reasoning, planning, and learning is artificial intelligence (AI) \cite{RussellStuartNorvig2003NoApproach}. Modern research efforts continue to investigate how these artificial intelligence principles can apply to health. Health is a product of functional interactions of many objects such as molecules, cells, organisms, and the environment. Based on this premise of connectedness in the world, graphs are one of the most potent ways to represent the real world computationally. We review key relevant principles in AI for HSE.

\subsection{Graph Networks}
\begin{figure}[H]
  \centering \centerline{\epsfig{figure=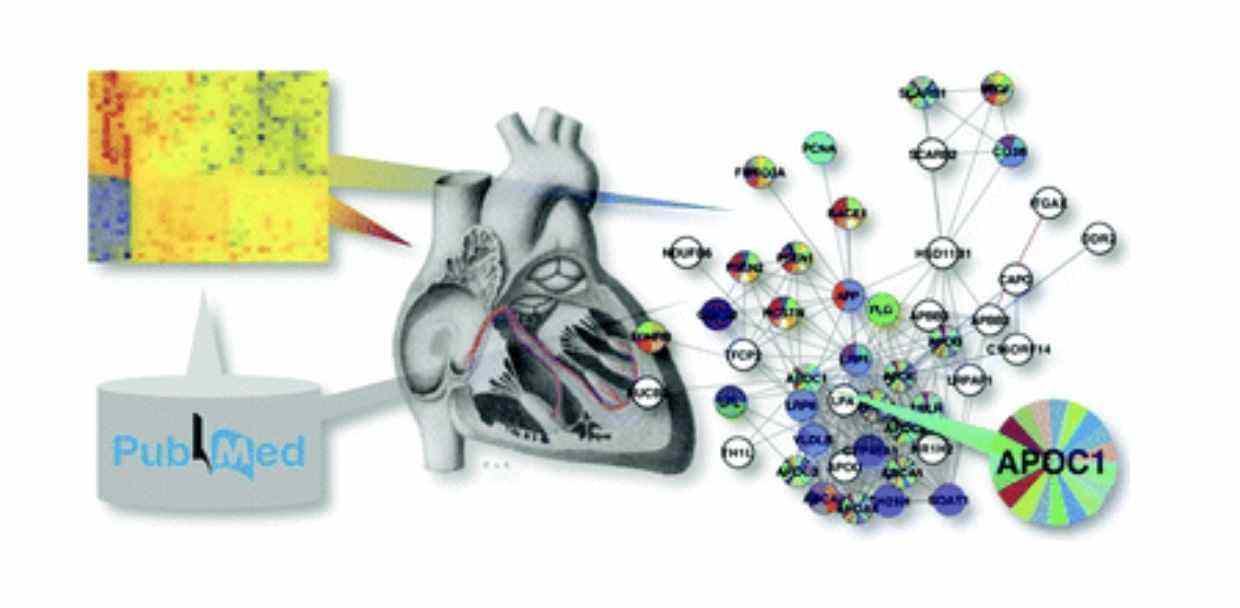,width=15cm}}
          \caption{\textbf{The Heart Network.} The merging of omics data and biological domain knowledge is used to create graph networks, highlighting causal relationships.}
    \label{fig:heartnetwork}
\end{figure}

Established by Leonhard Euler in 1736, graph theory is the study of modeling relationships between objects \cite{Guichard2016IntroductionTheory}. Undirected graphs are a tool used to capture symmetrical relationships between objects. Directed graphs contain connections that have a source object and a target object, thus establishing a direction by which relationships flow. Graphs are an excellent computing methodology for capturing the connectedness of information about our health. Graph networks connect how genes are expressed in cancer and healthy tissue to identify topology that distinguishes them apart \cite{Sandhu2015GraphNetworks}. Gene identification methods can also determine pathways for CVD, merging omics data with biological knowledge, as shown in Figure \ref{fig:heartnetwork}. By identifying the biological networks for a system, we can see transitions from a healthy to a diseased state \cite{Diez2010TheDisease}.Graph networks can construct the human brain networks (brain regions as nodes and pairwise associations as edges) by combing functional and material causes (shown in Figure \label{brainnetwork}) \cite{Bassett2009HumanDisease}. There may be a clustering of physiologic patterns related to the nodes, as seen by measuring REM activity as a risk for CVD (shown in Figure \label{heartnetwork}) \cite{Hou2016VisibilitySleep}. Lastly, by using graph structures, we can start observing causal relationships.


\begin{figure}[H]
  \centering \centerline{\epsfig{figure=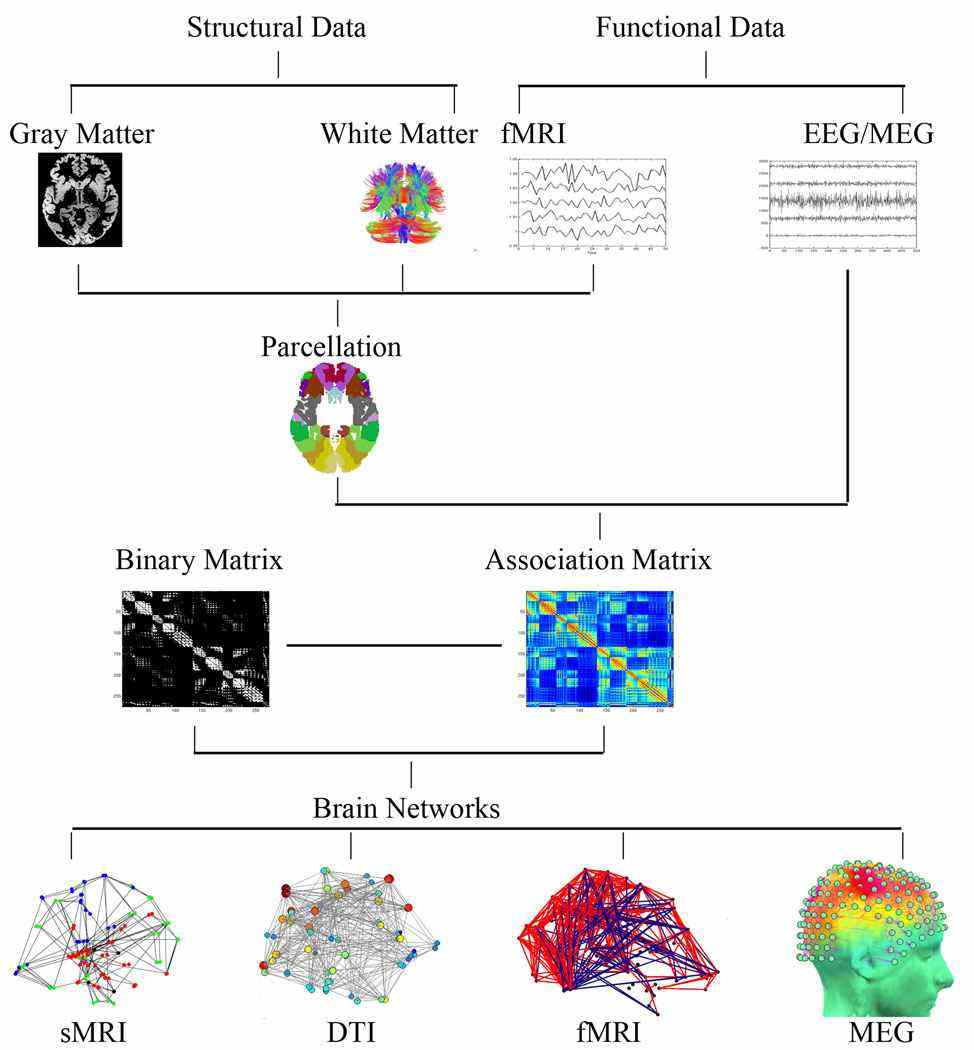,width=12cm}}
          \caption{\textbf{Workflow of Human Brain Network.} Connections can be made using both functional and material causes for establishing the graph network.}
    \label{fig:brainnetwork}
\end{figure}

\subsection{Causal Reasoning}
Causal reasoning is the process of understanding the relationship between a cause and its effect. Causal events usually have a temporal aspect in which the cause is antecedent to the effect. Graph structures are typically employed to represent causal pathways. Computing methods that understand and can discover causal reasoning is of great interest to the AI and statistical fields. Turing award winner, Judea Pearl, describes causal calculus as ``the ability to infer intervention probabilities from conditional probabilities in causal Bayesian networks with unmeasured variables" \cite{Pearl2013AActions}. A benchmark of this work is the detection and description of confounding variables within data. Pearl's criterion, which he calls ``backdoor," provides a robust definition of confounding variables, enhancing the identification of variables worthy of measurement \cite{Pearl2018TheWhy}. An investigation into causal reasoning in early AI research conducted, uses pathophysiological knowledge in medical diagnoses \cite{Patil1986REVIEWDIAGNOSIS.}. The medical community continues to seek a causal understanding. Goodman et al. use the Framingham model as an example to identify causal factors (such as smoking cholesterol, blood pressure, i.e.) through prior knowledge and changes in risk factor \cite{Goodman2018MachineReasoning}. Causal relationships also conduct simulations using demographic information \cite{Chen2013CanSimulations}.

\subsection{Knowledge Structures}
Similar to a database, a knowledge base stores information about the real world. The first knowledge bases were ``expert systems" \cite{Chandrasekaran1986GenericDesign}. Knowledge in health largely consists of understanding the output of a system after it is perturbed given an input. This knowledge arrives from established causality from scientific discovery. A graph structure can store the causal relationship of various parts of health through directed edges which store the cause as a source node, and the effect as the target node. From a molecular level to a societal level, these relationships are routinely in the scientific literature. In light of this, we review knowledge structures in graph format, also known as knowledge graphs.

\begin{figure}[H] 
  \centering
  \includegraphics[width=0.8\textwidth]
  {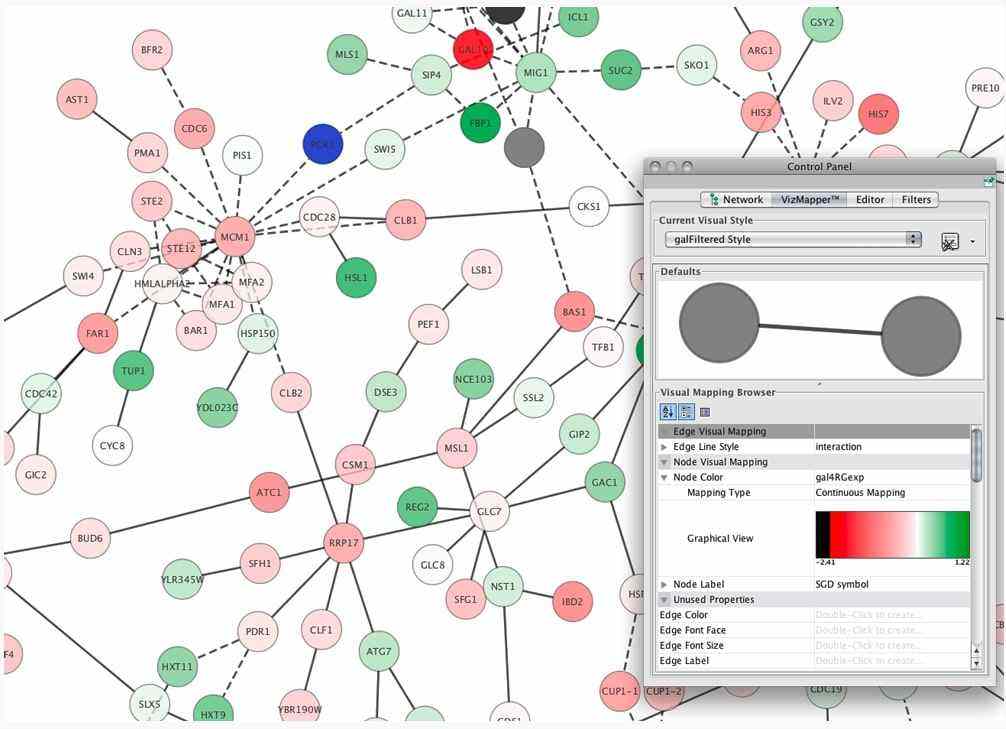}
  \caption{\textbf{Cytoscape.} Cytoscape is an open source bioinformatics software platform for visualizing molecular interaction networks and integrating with gene expression profiles and other state data. \cite{Otasek2019CytoscapeAnalysis}.}
  \label{fig:cytoscape}
\end{figure}

\begin{figure}[H]
  \centering
  \includegraphics[width=0.8\textwidth]
  {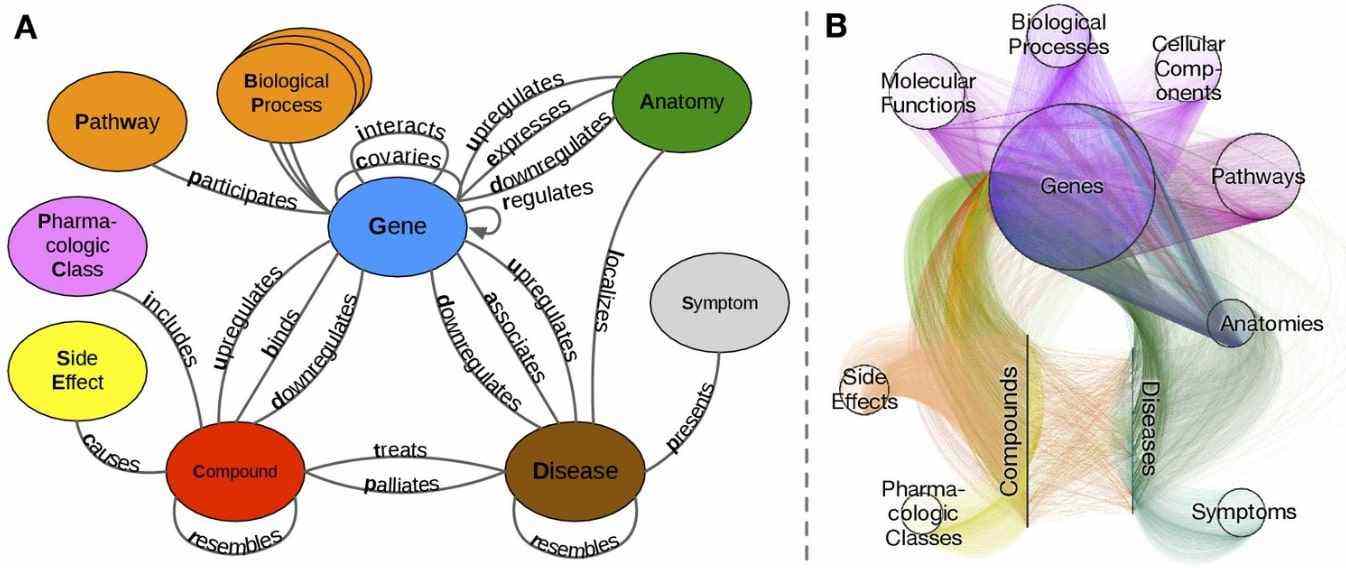}
  \caption{\textbf{Heterogeneous Networks in Biology.}  Hetnets contain edges and nodes that have type specificity as part of their properties. Using these types as distinctions, various visualizations, queries, and algorithms can be applied. This particular HetNet maps relationships between compounds and diseases \cite{Himmelstein2017SystematicRepurposing}.}
  \label{fig:hetnet}
\end{figure}

Google harbors one of the largest knowledge graphs at the current time, with over 70 billion facts \cite{Vincent2016AppleVerge}. Knowledge graphs have been described in various forms \cite{Ehrlinger2016TowardsGraphs}. Collectively, the aim of knowledge captured in this format is for use in further sharing and computing. Within the domain of health, there have been various efforts to capture this knowledge. Cytoscape is an open-source software platform to capture the relationships between molecular entities for biological research \cite{Shannon2003Cytoscape:Networks}. An example of Cytoscape software is given in Figure \ref{fig:cytoscape}. Non-profit databases such as Pathway Commons store the molecular pathways used in Cytoscape \cite{PathwayCommons2020PathwayAnalysis}. Heterogeneous networks, or HetNets in short form, have also been described in biology through the work of Himmelstein et al. \cite{Himmelstein2017SystematicRepurposing}. By gathering public domain literature into a connected knowledge base, queries are run to determine distances between entities to discover pathways for re-purposing drugs. A data-model of how these HetNets are shown in Figure \ref{fig:hetnet}. These efforts in accumulating knowledge into data structures can be extended for use in further applications, yet have mostly been limited to specialized domains of research.


\begin{figure}[H]
  \centering
  \includegraphics[width=0.8\textwidth]
  {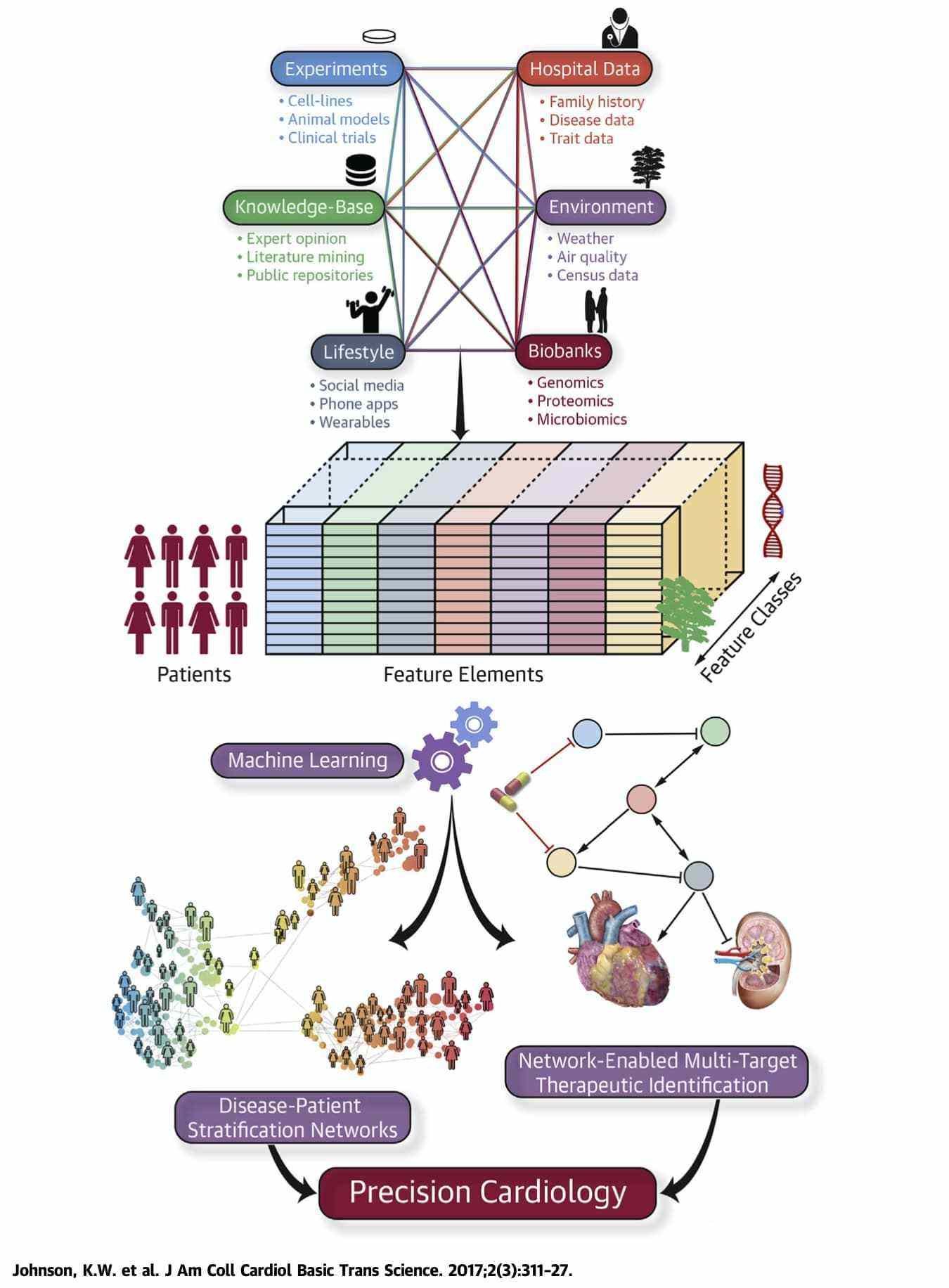}
  \caption{\textbf{Multiscale Cardiology.} There are multiple layers of consideration for practicing precision cardiology including integration of knowledge, experiments, data, environment, and person's lifestyle. Computing techniques to extract understanding of how actions can be best taken to help the cardiovascular state are still needed \cite{Johnson2017EnablingMedicine}.}
  \label{fig:multicardio}
\end{figure}

In the field of cardiovascular research, there has been a call for the integration of diverse knowledge graphs to work together. This approach has been called multi-scale cardiology by Johnson et al. Current clinical knowledge in cardiovascular health derives insight broad clinical trials and studies. This knowledge is placed into sweeping standardized guidelines. Johnson et al. describe future research needs in applying AI and knowledge network models to enable a more precise understanding of patients and matching the best therapies for a particular person \cite{Johnson2017EnablingMedicine}, shown in Figure \ref{fig:multicardio}. This work highlights the issue of integrating knowledge even within a specific sub-field of research as a formidable challenge.

As we think forward about establishing a state estimation model for health, knowledge graphs can help instantiate a model of an individual at various abstraction scales. Knowledge graphs are essential to overcome the cold start problem of personal HSE.

\subsection{Systems Thinking}
Systems thinking approaches problems through an eagle-eye view of interrelated components, contrasting reductionist approaches that try to isolate the view of study to a highly controlled environment within which we understand perturbations. A system may express emergent behavior when the parts accumulate, giving homage to the famous term ``greater than the sum of its parts." As described by Fang et al., ``a fundamental tenet of systems biology is that cellular and organismal constituents are interconnected, so that their structure and dynamics must be examined in intact cells and organisms rather than as isolated parts" \cite{Fang2011ReductionisticScience}. Inputs to one part of the system can propagate to other parts of the system. Systems approaches can predict how these effects will propagate.

\begin{figure}[H]
  \centering \centerline{\epsfig{figure=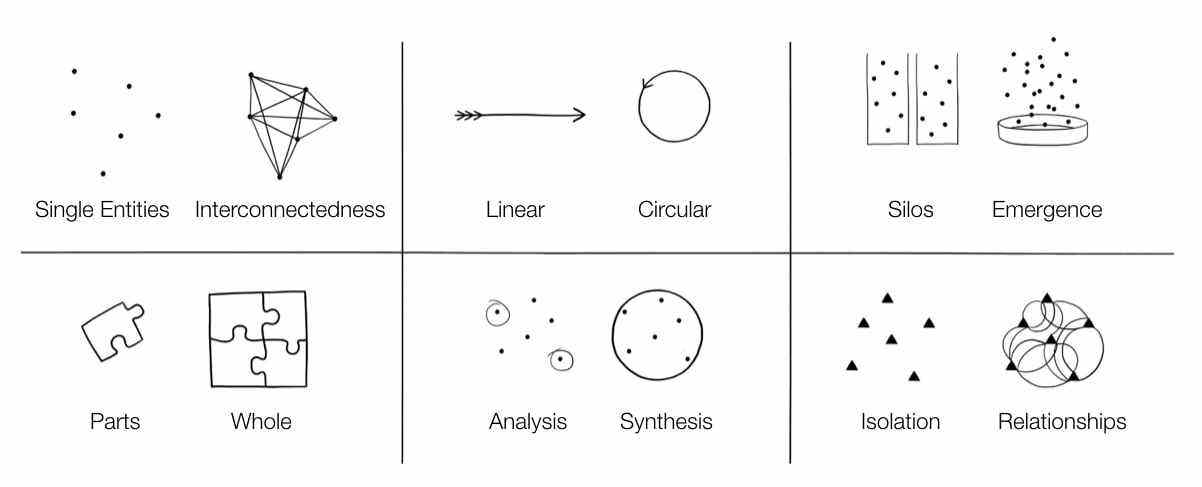,width=16cm}}
  \caption{\textbf{Systems Thinking.} By considering an entire ecosystem of components, system thinking aims to understand the behavior of complex entities. This approach includes understanding how sub-entities are connected and related, how effects can circularly impact the system, what emergent properties arise from the system as a whole through synthesizing the system into a single architecture. \textit{Adapted from Leyla Acaroglu.}}
    \label{fig:systemsthinking}
\end{figure}

\begin{figure}[H]
  \centering \centerline{\epsfig{figure=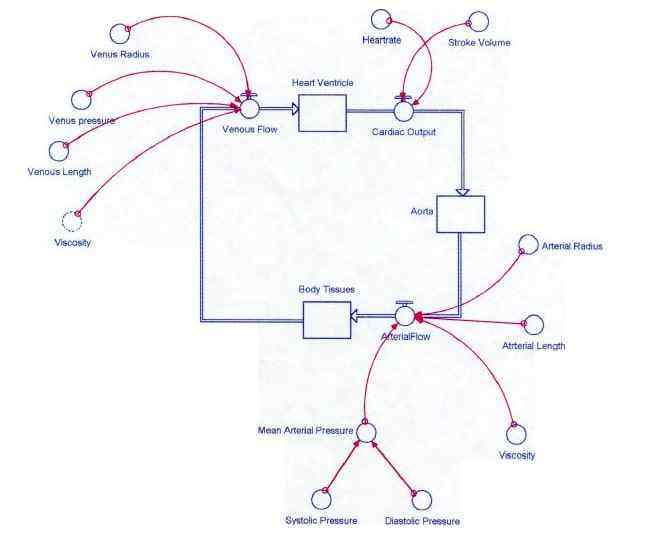,width=16cm}}
  \caption{\textbf{Cardiovascular Function.} Blood flow is shown here moving through stocks and flows through the STELLA interface. This model can only focus on a single dynamic variable within these containers \cite{Romstedt2003ModelingBiology}.}
    \label{fig:cvdstella}
\end{figure}

STELLA and InsightMaker are examples of modeling systems with stocks and flows \cite{Maker2019InsightBrowser, Doerr2004StellaLiterature}. The user places interconnected icons and defines functions mathematically of how a particular value is flowing into and out of stocks. The most classic example of this method is a bathtub with an in-flow from the faucet and an out-flow through the drain. This approach has been used in modeling cardiovascular function, as shown in Figure \ref{fig:cvdstella} \cite{Romstedt2003ModelingBiology}. Systems models that are intelligent that change their structure remain a challenge in this field.

A robust HSE requires a systems approach due to the complexity of biology. At the same time, we must incorporate reductionist science discovery to have the best performance. To address this computationally, we look towards microservices.

\subsection{Microservice Architecture}
Microservice architecture is a software development technique whereby an application is made from a compilation of loosely coupled services. These services provide a specified interaction with other services. Reductionist science is a microservice reporting on the behavior of an isolated system. An extensive collection of these systems that have knowledge-driven connections between them can start to build a systems approach to using these smaller nuggets of knowledge. The modular nature of microservices also means that a system can easily swap out, add, or delete a service if there is an updated or out-dated version without breaking the whole system structure. Hence microservices are readily understood as a graph network of interacting Application Programming Interfaces (API).

This approach is used in primarily identifying device or software performance health, and security. As examples, microservices are used in modeling IoT devices for health \cite{Hill2017EnablingMicroservices}, monitoring the health of the software itself \cite{Toffetti2015AnMicroservices}, and implementing security measures \cite{Esposito2017SecurityApproach}. No one known to date has used the microservice approach to represent the health of an individual directly. 




\subsection{Data-Driven Learning}
Uncovering complex patterns specific to an individual requires harmony between top-down approaches through knowledge, as mentioned above, alongside data-driven bottom-up approaches. These data-driven approaches can discover patterns that human analysis may fail to see. We can use data in learning how to adapt models to best fit reality constantly. A large part of these patterns in health and life is temporal. This approach is the particular aim of event-mining research.

\textit{Event mining} encompasses techniques for automatically and efficiently extracting valuable knowledge from the historical event and log data \cite{Li2015EventApplications}. Event Mining: Algorithms and Applications presents state-of-the-art event mining approaches and applications with a focus on computing system management \cite{Li2015EventApplications}. Event mining can be challenging as it requires familiarity with domain literature for meaningful pattern discovery.

In general, temporal data is considered as a sequence of a primary data type, most commonly numerical or categorical values and sometimes multivariate information. Temporal data is presented in a numerical time series of values or events (e.g., EEG, stock price), click streams). Some primary goals of temporal data mining include pattern analysis, search, prediction, and clustering. Translating knowledge maps for events into a temporal visual representation \cite{Wang2013AData}. Different variants of sequential patterns are described \cite{agrawal1995mining,chaochen1991calculus,li1996hierarchyscan}. Multidimensional patterns are particularly interesting for health since there are usually many parameters to compare at once \cite{yu2005mining,han2000freespan,ayres2002sequential}. There have been extractions of knowledge in literature for event mining in the context of systems biology \cite{Ananiadou2010EventLiterature}. By using the knowledge above as a basis for constraints, the search space for mining these patterns can be reduced in computational complexity \cite{pei2001prefixspan,morzy2002efficient,pei2002mining,wojciechowski2001interactive}. Machine learning approach based on Bayesian networks is trained on medical health records to predict the probability of having a cardiovascular event. Unfortunately, their model fails to use high temporal resolution data \cite{Bandyopadhyay2015DataData}. Temporal patterns are just one avenue by which to use data-driven approaches. A breakthrough in machine learning and AI systems in the last decade has been through the advancement of neural networks.


\textit{Neural Networks} represent a broad class of algorithms modeled after the human neuronal connections for pattern recognition. A network starts with input as stimuli and propagates a signal to neighboring nodes by an activation function, such as sigmoid or tanh. Each pair of nodes have particular edge weights. Below are these activation functions and weights.

${\displaystyle Activation(W(eights)*X+b(ias))}{\displaystyle Activation(W(eights)*X+b(ias))}$

\begin{figure}[H]
  \centering \centerline{\epsfig{figure=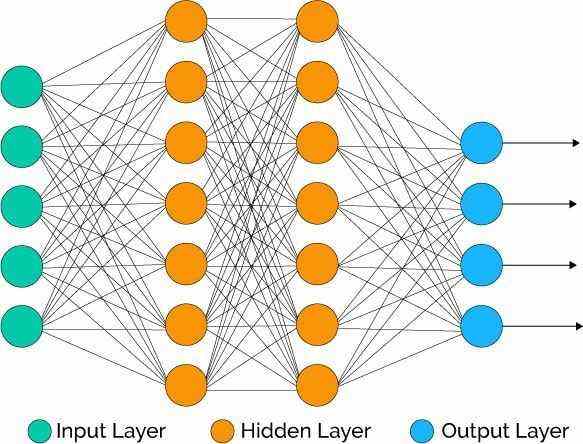,width=9cm}}
  \caption{\textbf{Neural Networks.} Given an input layer of data, the neural network learns patterns and features to match an output layer.}
    \label{fig:neuralnet}
\end{figure}

Training the network is based upon defining a cost function that is optimized. This concept is shown visually in Figure \ref{fig:neuralnet}. As of 2020, neural networks have provided a valuable computing method to recognize patterns when given large amounts of data. This data-hungry need for neural networks is a double-edged sword. For systems that continuously produce large amounts of homogeneous data, it is advantageous. Unfortunately, large amounts of data about an are rarely collected, and thus there has not yet been much progress in using this method for personal HSE. Regardless, there has been some research into using neural networks for HSE.

We explore two avenues within neural networks for HSE: 1) temporal data streams and 2) deep learning. Temporal data streams such as signal based learning for EEG signals \cite{Oh2018ASignals}. We also see temporal-based data streams in health condition management and predictions using neural networks \cite{Wu2013Condition-basedPrediction,Guo2017ABearings}. There are several examples of using deep learning for better classifying diseases. Image recognition in a clinical setting is prevalent. We see this in radiology, dermatology, pathology, as these fields are dependent on images \cite{Ravi2017DeepInformatics, DeFauw2018ClinicallyDisease,Esteva2017Dermatologist-levelNetworks}. Deep imaging continues to grow in in biomedicine, such as for predicting genetic variations and altercations \cite{Wainberg2018DeepBiomedicine}. Image Recognition applied to health state has a broad scope of research conducted to date and will be beyond the scope of this thesis to cover.

\subsection{Robust Artificial Intelligence}
Recent AI progress in domains such as vision, language, and decision-making fit the strengths of neural networks. Many characteristics desired from intelligent systems are still sought after and remain out of reach. In particular, generalizing beyond the story of observable data remains a formidable challenge for modern AI, requiring approaches that structure meaningful representations and integrate data coming from the real world. We will explore approaches that benefit from the complementary strengths of top-down and bottom-up thinking below. 

\textit{Probably Approximately Correct} approaches use a combination of estimation and probability as a learning approach. Turing award winner, Leslie Valiant, describes how probably approximately correct algorithms can explain how effective models are learned without direct, explicit instructions. Satisfactory solutions are generated in the absence of any theory about the problem. This approach rests on choosing the appropriate information gathering mechanism and a reasonable learning protocol \cite{Valiant1984ALearnable,Valiant2013ProbablyWorld}.

\textit{Knowledge Infusion} aims to endow computers with the ability to apply commonsense knowledge with human-level performance. Here the computer represents pieces of knowledge about the world with uncertainty. Knowledge infusion rules are learned from the world to reduce this uncertainty \cite{Valiant2008KnowledgeIntelligence}.

\textit{Combinatorial generalization} is the ability to understand and produce a new understanding of familiar components through new inferences, predictions, and behaviors from known building blocks. This signature of human intelligence to extend data from life experiences with knowledge is ``infinite use of finite means" \cite{Joseph1991OnMankind,Chomsky1965MethodologicalAspects}. Battaglia et al. have explored implementing a combinatorial generalization approach by biasing learning towards structured representations and computations that operate on graphs. They claim, ``Just as biology uses nature and nurture cooperatively, we reject the false choice between hand-engineering and end-to-end learning, and instead advocate for an approach which benefits from their complementary strengths" \cite{Battaglia2018RelationalNetworks}. They explored how to capitalize on the structured representations through nested layers of graphs called graph network blocks (GNB). Combinatorics can help answer what combination of blocks is optimal in modeling the system \cite{Guichard2016IntroductionTheory}.

Other researchers have attempted to build frameworks for combinatorial generalization. A hybrid model approach of mechanically producing real-world data to feed into a neural network has been attempted by combining neural networks with a physical audio engine \cite{Wang2019ASound}. This approach is mostly data-driven and doesn't take advantage of knowledge, even though the authors claim to have achieved combinatorial generalization. One explored approach to using both data and knowledge has been to use combinatorial building blocks into neural networks \cite{Vlastelica2019DifferentiationSolvers}. Representing symbols through ``vectors approach to representing symbols" (VARS) allows training standard neural architectures to encode symbolic knowledge explicitly at their output layers \cite{Vankov2020TrainingGeneralization}. In these cases, the knowledge put into an overarching data-driven approach is the primary driver. At the time of publication, there are no known HSE approaches that use graph network blocks or combinatorial generalization techniques.


\section{Unmet Challenges and Scope of Dissertation}

Especially in chronic diseases, the current medical practice continues to use sparse hospital-based biological metrics (i.e., blood tests, expensive imaging) to understand the health status of an individual. The apparent research direction ahead points to utilizing a combination of the approaches above to shift towards a continuous and abundant HSE paradigm. To achieve this requires the harmony of using personal data, estimation techniques, and modeling. Enabling data assimilation will require an intelligible understanding of how sensor data relates to health status. We must use existing medical and biological scientific domain knowledge to guide the data assimilation and conversion matrices from signal to state, as shown in Equation \ref{truecyberneticstate} and Equation \ref{measuredcyberneticstate} with causal intelligibility. Knowing \textit{why} an individual has a particular health status will be paramount to explanation, recommendation, and treatment. Purely data-driven methods can be components in an extensive system but will have difficulty in leveraging existing knowledge. Purely knowledge-based systems do not take advantage of data giving insight into the dynamic state of the individual.
For this reason, complementary methods needed to take advantage of both approaches. Combinatorial generalization approaches allow for the pooling of understanding at different levels of abstraction, the encoding of domain knowledge, and learning from data. In so far, at the time of writing, these approaches seem the most promising for leveraging computing power to aid in the HSE problem.

We will discuss the unmet challenges in this work by focusing on bridging together several components of the literature mentioned uniquely. Foremost, continuous sensing through the self, sensors, and external agents is an absolute must for in-depth high temporal resolution HSE. A graph-based approach aims to fuse the connectedness of causal reasoning with domain knowledge. By nesting these graphs into microservice blocks, we gain the ability to zoom between abstraction layers fluidly. This ease in movement allows for a flexible, holistic, and reductionist approach to work symbiotically. These blocks can contain their unique models and estimation techniques that suit their level of abstraction optimally. We will describe these efforts in the context of bioenergetics and cardiovascular health at an individual level.
\chapter{Health State Estimation Framework}

\epigraph{Knowledge is power.}{\textit{Francis Bacon}}
\vspace{12pt}


In the era of information technology, we harness the power to know the state of key aspects of our life. In an instant, we can know our physical location, financial situation, or the quality of our batteries. Based on this high-quality state estimation, we have a host of rich computing services that serve our needs, such as mapping directions, trading assets, and energy optimization. To date, we cannot see the details of our evolving health state in this way: on-demand and at every moment. The host of problems this issue poses are described in Chapters 1, 2, and 3. As defined in Chapter 2, health must go beyond just biological components by including elements of function. In order to know the state of our health in each moment, we need to combine all possible sources of data and knowledge into a seamlessly connected framework in the context of both biology and real-life needs. These two factors are the core of how we build a useful HSE paradigm. This chapter describes the principal concept of this dissertation by which to build a state estimation model for each person connecting these two factors.

\section{Health State Estimation General Overview}
The value of good health to an individual is largely based upon the wishes of how they wish to live. In order to practice a well-lived life, certain abilities determine the possibility of these wishes coming true. Take the example of an individual who is keen to play the guitar, ride a bicycle, travel to hiking destinations, or solve interesting computing problems. Each of these endeavors requires abilities for the individual to embark on the respective goal. For example, an individual who wishes to play the guitar and enjoy music must have the ability to use their fingers with dexterity, have the endurance to hit the strings repeatedly, and a heightened sense of hearing to engage satisfactorily in this activity. All these abilities depend upon harmoniously operating biology of neuro-muscular function, cellular metabolism, and auditory sensory cells. These biological mechanisms are in constant flux. So in order to determine how the state is continuously changing, we must understand what inputs cause change to the system state as given in Equation \ref{truecyberneticstate} and Equation \ref{measuredcyberneticstate}. 

\begin{figure}[H]
  \centering
  \includegraphics[width=0.8\textwidth] {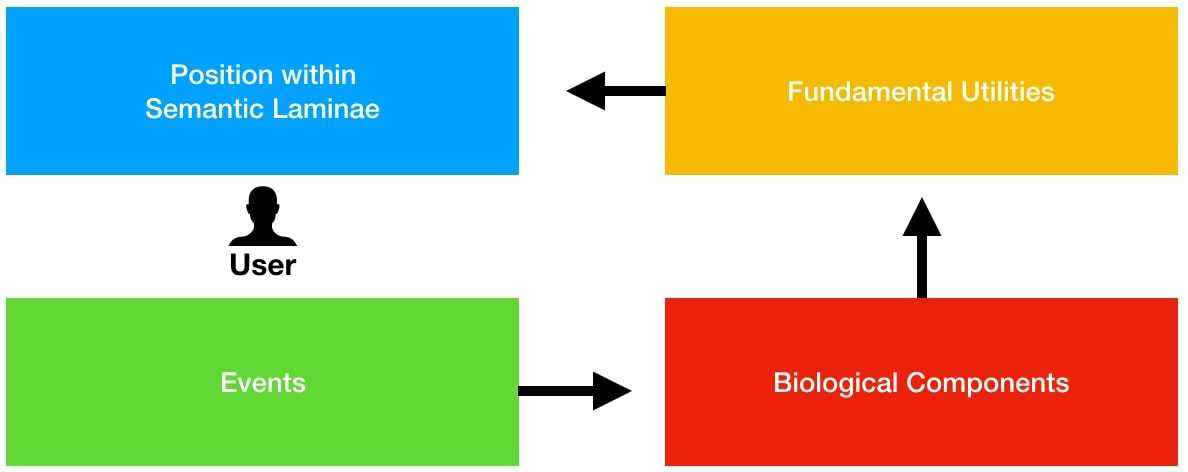}
  \caption{\textbf{Health State Estimation Flow.} Events drive biological changes in the user's body. These biological components determine utility state dimension values. The combination of these utility dimensions provide the health state, akin to the GPS position. The user's position is visualized to view the current state within a certain semantic context defined by the user intent.}
  \label{fig:hsebasic}

\end{figure}

These inputs are the events of our life in every moment aggregated in sequence. Each event interacts with our biology and causes changes in the system. The basis of this change rests in the central dogma of biology, whereby these events change our gene expression, metabolism, and other molecular processes, which then cause a change in the biological state \cite{CRICK1958OnSynthesis.}. Our health state is rooted in the flow of events shaping our biology, which in turn determine our abilities to do the things we want to do. This flow is shown in Figure \ref{fig:hsebasic}. This HSE and modeling distinctly capture these flowing elements. The proposed framework thus contains four key abstraction blocks that serve these unique needs. The first is the semantic laminae, which are the layers of information by which humans can easily interact with natural language and visualizations. In the above example, this is at the abstraction level of wanting to play the guitar. The second is the utility state block, which captures all the fundamental functionality a human has. In the above example, this refers to the abilities to capture audio and the mechanics of playing the strings. The third block is the biological block, composed of all the natural components that give rise to the utility functions, exemplified through the muscular, neuronal, and sensory biological systems that allow us to play and enjoy music from a guitar. The fourth block contains all the events that are inputs into the human system. In the case of a guitar player, they could be practice or jam sessions with other musicians. The four blocks are described below. The remainder of this chapter outlines the intricacies and interactions of these blocks. 

These four blocks interact in a specific flow. Events in a person's life change their biology, and these biological changes are reflected in the change of our utilities. The set of utilities relevant for a particular domain provides the dimensions by which to position the user state within the health state space for that domain. The flow of these four blocks is demonstrated in Figure \ref{fig:hsebasic}. We go over the atomic construction material of this system next, the GNB.

\section{Graph Network Block Structure}
\begin{figure}[H]
  \centering
  \includegraphics[width=0.8\textwidth] {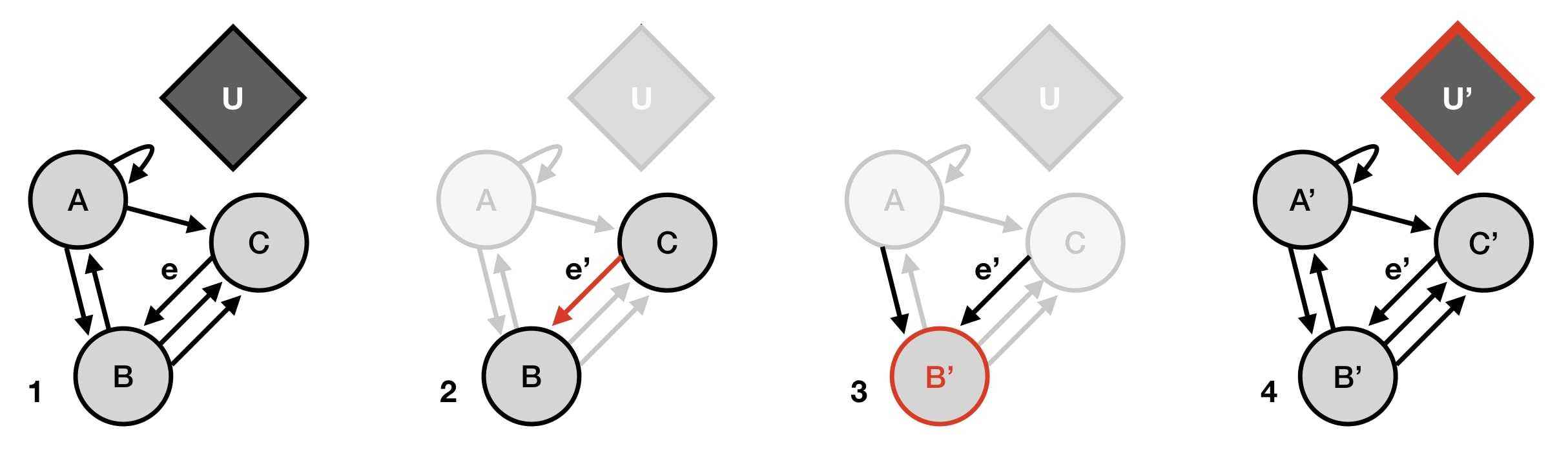}
  \caption{\textbf{Graph Network Block Unit and Update Process.} Each GNB describes a directed multi-graph, including multiple edges in the same direction or opposite directions, as shown between node B and C. Edges can also be to a self node as shown for Node A. The properties characterizing the entire multi-graph are summarized as global attributes within U. Step 1 shows the initialization of a GNB. In the following steps, red highlights the updates. Step 2 shows the edge update. Step 3 shows the node update. Step 4 shows the global update, a visualization of algorithm \ref{alg:gnbupdate}.}
  \label{fig:gnb}
\end{figure}

To begin describing the framework details, we start with the most basic component, a GNB. This approach takes inspiration from the work of Battaglia et al. to use relational induction within a system to elucidate a generalized understanding of a system \cite{Battaglia2018RelationalNetworks}. Because all the parts of life are related together, a graph concept suits our needs of connectedness the best. The GNB structure is defined by a directed multi-graph, summarized into a global set of attributes. A visual description is given in Figure \ref{fig:gnb}. This GNB concept represents the utility and biological blocks in this estimation framework.

\begin{description}[noitemsep]
\item[Directed]: Edges travel from a ``source'' node to a ``target'' node.
\item[Attribute]: properties of each node or edge are stored as a set within the respective edge or node.
\item[Global attributes]: This is a set of attributes about the entire graph structure, which may be computed from the properties of the nodes and edges.
\item[Multi-graph]: Between two nodes, there can be edges in both directions and multiple edges in any given direction. A node may also have a self edge.
\end{description}

We represent the relationship between two parameters as a directed edge from the cause to its effect. Every node is a GNB with its global update. The update cycle for each block is as follows:

1. Update incoming edges: We call the update method associated with each edge coming in to block and defined within the block. For most edges, this method reads the global parameters of the source node and passes those to the target node. If needed, we can also perform more sophisticated transformations on the incoming data. An edge can be updated when there is a new update in an associated source node or an update in the relationship itself.

2. Update the nodes in the block: Every node in the block updates by applying the node's global update function on the incoming edge parameters.

3. Global update: Once all the edges and nodes in the block updates, we can aggregate the updated values of nodes and edges to update the global parameters of the block.

\subsection{Nested Microservices}
\begin{figure}[H]
  \centering
  \includegraphics[width=0.95\textwidth] {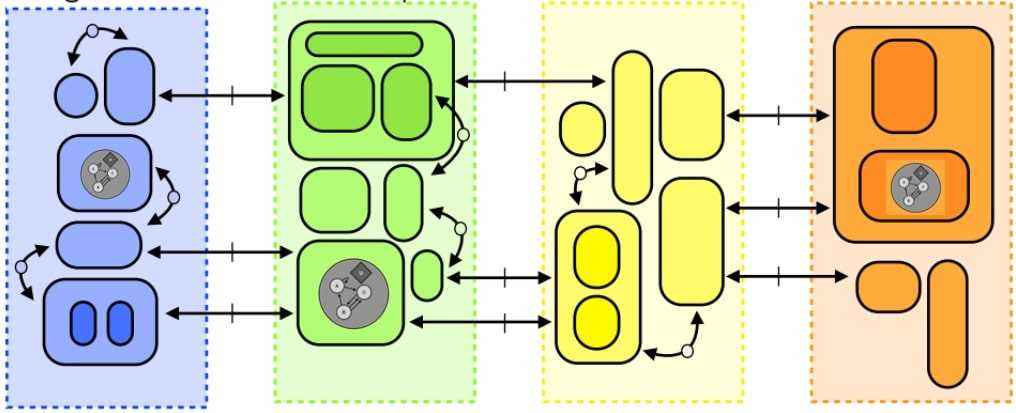}
  \caption{\textbf{Nested Microservices.} The GNB structure global attributes give rise to a microservice. These microservices are structured in a nested manner that can interact with each other in complex networks. This allows for the relevant level of abstraction to communicate in any system that wishes use the GNB. \textit{Figure adapted from Meyer et al.} \cite{Meyer2015TheModel}.}
  \label{fig:micro}
\end{figure}

Each of these GNBs can represent a set of microservices working together to provide a higher level microservice. The output of the higher level service is within the global attributes of a GNB. The purpose of using this microservice approach is to ensure the framework is able to allow for modular exchange with any biological (i.e., organ transplant, chimera) or non-biological system (i.e., metal bones). It also allows for systems and focused reductionism to work together by zooming to the appropriate block. This structure is shown in Figure \ref{fig:micro}. This design allows for incorporation of the nested service layers that biological systems provide to the human body. This hierarchy of services is shown in Figure \ref{fig:metabolism}.

\subsection{Causal Knowledge Infusion}
\begin{figure}[H]
  \centering \centerline{\epsfig{figure=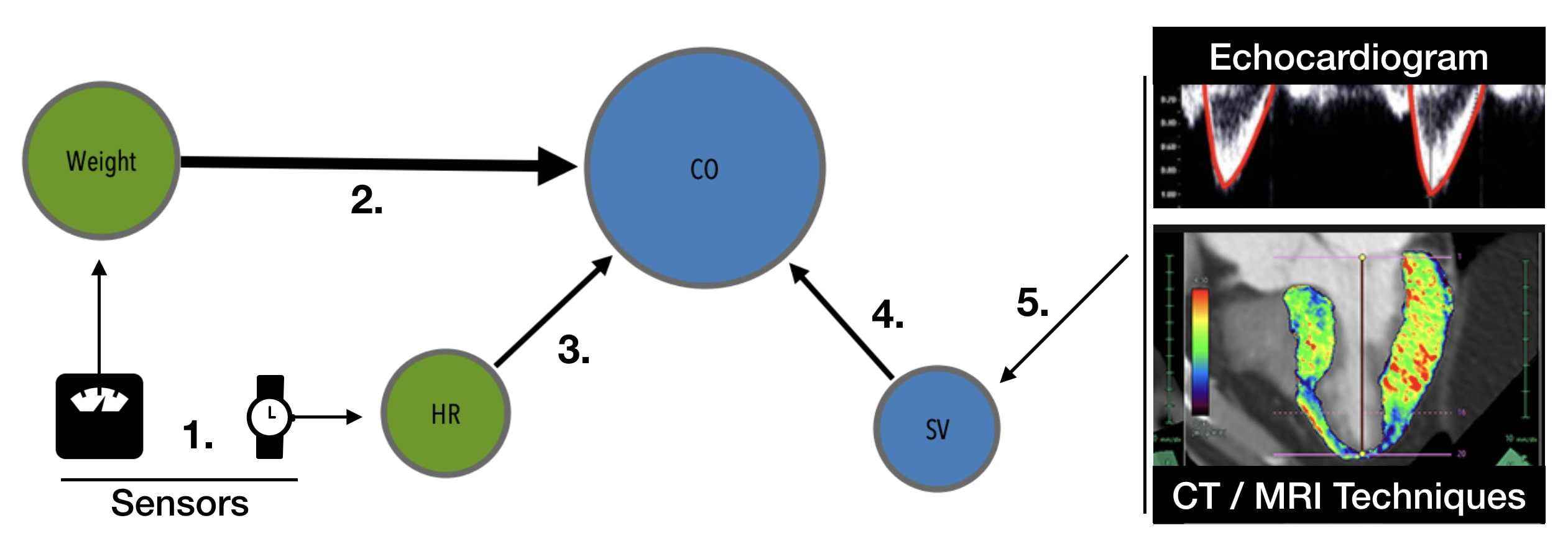,width=14cm}}
  \caption{\textbf{Knowledge to Graph.} Here, we translate cardiac output equations to the graph structure. Cardiac output refers to the volume of blood that the heart can pump per minute. Once we have the relationship in graph format in conjunction with data streams from either cheap continuous sensors like a weight scale or watch, or a more costly sensor like an ultrasound, we can update the relationship to more accurately reflect the model of the individual. } 
  \label{fig:k2g}
\end{figure}

In the case of the directed edges within a GNB, each edge represents a relationship between two real-world entities. These entities and relationships are built initially on domain knowledge. Thus, each edge is a causal relationship that extends for causal calculus or backdoor confounder analysis (beyond the scope of this dissertation). By using the nested microservice approach, we can insert knowledge at the correct abstraction level and can scale between depth and breadth through the blocks. The knowledge sources for each edge are also stored within the edge for reference. An example of the knowledge transformation process is shown in Equation \ref{fig:fick} and Figure \ref{fig:k2g} as an example for cardiovascular physiology. The equation dictating cardiac output \textit{CO} is below, where \textit{SV} is the stroke volume of blood each heartbeat ejects and \textit{HR} is heart rate. \textit{CO} at rest is determined by body weight. The knowledge source is from \textit{Exercise Physiology: Human Bioenergetics and Its Applications} \cite{Brooks1996Exercise2}.

 \begin{equation}   \label{fig:fick}
 CO = SV * HR 
 \end{equation}

\subsection{Data-Driven Model Updates}
The relationships encoded in the edges can be modified in the future when we identify two nodes that have incoming data streams of both the source and the target node. This approach allows for both domain knowledge and data-driven structures to co-exist and help each other. In Figure \ref{fig:k2g}, this is shown as either low-cost or high-cost sensor data validating the value of specific nodes. If we can measure the heart rate and stroke volume, then the value of cardiac output can be known with high accuracy. We modify the base relationship of weight to cardiac output to fit the data.

\subsection{Combinatorial Generalization}
In any situation where we update a value of a node or edge through either a learning process or through new domain knowledge, we can recompute the global attributes of all edges and nodes within the total graph network system. These updates result in new values for all global attributes, including the utility states, resulting in a combined new generalized state of the system used in the next time iteration. Hence, this updates the value along each dimension of the user health state.

\begin{algorithm}
\caption{Graph Network Block Update}
\label{alg:gnbupdate}
\begin{center}
    \begin{tabular}{  p{16cm}  }
\textbf{function:} Update ($E$, $N$, $\mathbf{U}$)

\textbf{Input: } All blocks ($E$, $N$, $\mathbf{U}$), Updated Knowledge, Attributes, Values

\textbf{Output: } Updated Values of ($E$, $N$, $\mathbf{U}$)'

\textbf{Begin}

1: for each block 

2: \qquad for each edge ($E$)

3: \qquad \qquad Update all edge relationship attributes with domain knowledge.

4: \qquad \qquad Update all edge relationship attributes with learned knowledge.

4: \qquad \qquad Return ($E$)'

5: \qquad endfor

6: \qquad for each node ($N$)

7: \qquad \qquad  Updated node attributes with new values from data.

8: \qquad \qquad  Propagate updates through outgoing edges from updated nodes.

4: \qquad \qquad Return ($N$)'

9: \qquad endfor

10: \qquad for each set of global attributes within the block ($U$)

11: \qquad\qquad Using ($N$)' and ($E$)', compute ($U$)'

12: \qquad endfor

13: endfor

14: Return ($E$', $N$', $\mathbf{U}$)'

\textbf{End}
    \end{tabular}
\end{center}
\end{algorithm}

Algorithm \ref{alg:gnbupdate} is the method by which we compute the estimated global attributes of each block microservice within the system for both biological and utility blocks at large. Each update cycle refreshes the utility dimensions to reflect the current state.

\section{Semantic Laminae: Interactive Layers}
Lamina (plural: Laminae) refers to a thin sheet. These layered sheets represent themed data in a particular state space, such as in the case of bicycle riding or guitar playing. The laminae integrate perceptual elements so that a user can interact at the abstraction level that is relevant for their needs. Parallel to Geographic Information Systems (GIS), the map layer selected can be a street map, satellite imagery, or weather information depending on if the user is looking for directions, planning a camping spot, or choosing to go to a pleasant vacation destination. The underlying dimensions that support these map layers are all tied to latitude, longitude, and altitude. These are the three dimensions of space that GIS systems use. For a person to exist in Irvine, they must ensure that these three dimensions are within the appropriate range.
Similarly, people generally think about their life in the semantic space of natural language. Engaging in any particular goal in life requires having the person exist in a specific range of health state dimensions. These interaction layers need a foundation concept of dimension. In the case of health, we call the dimensions, utilities.

\begin{figure}[H]
  \centering
  \includegraphics[width=0.75\textwidth] {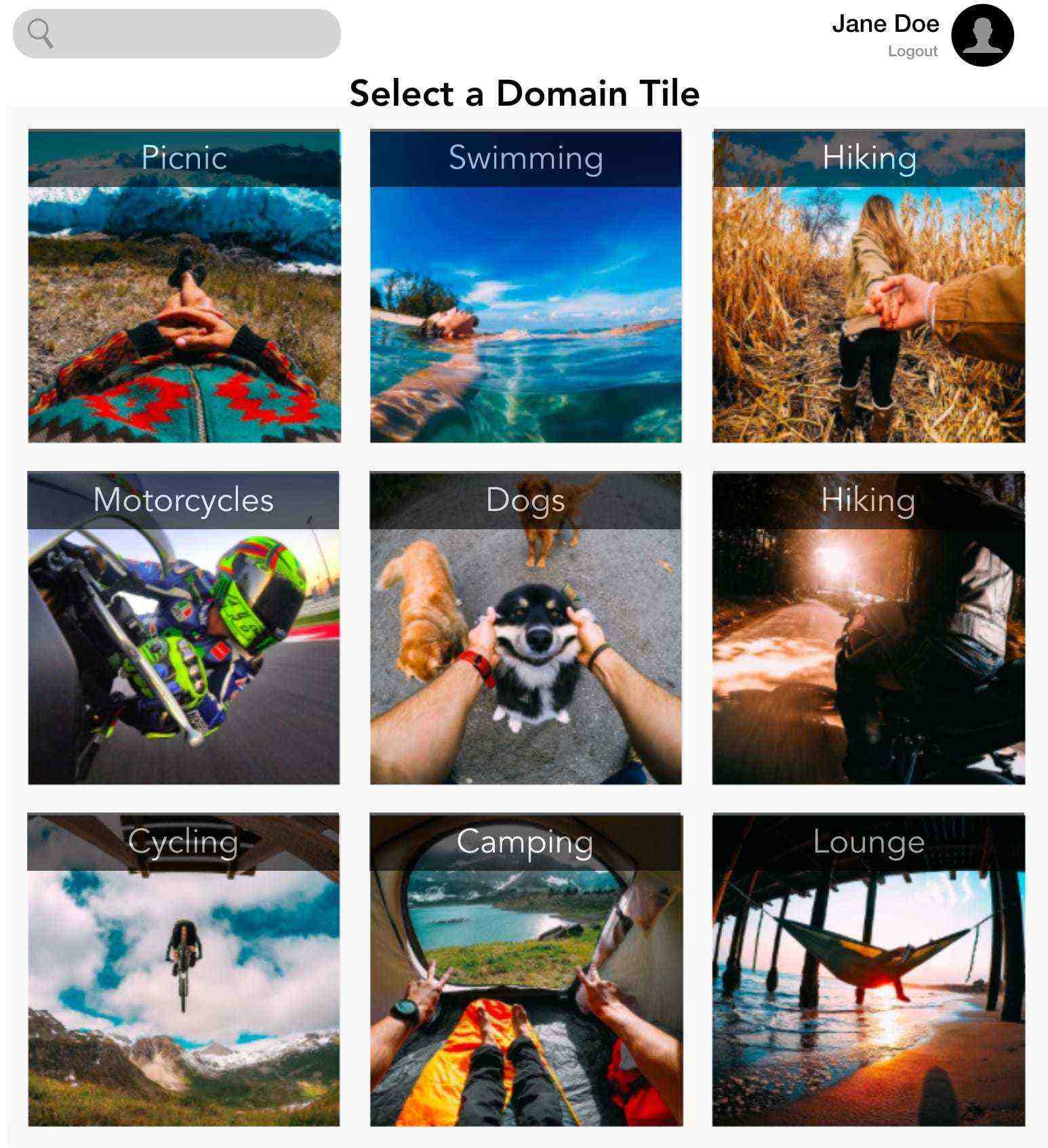}
  \caption{\textbf{User Interface.} A user can browse through various domains of interest, or search for a specific domain of interest in order to populate the a lamina.}
  \label{fig:ui}
\end{figure}

\begin{figure}[H]
  \centering
  \includegraphics[width=0.95\textwidth] {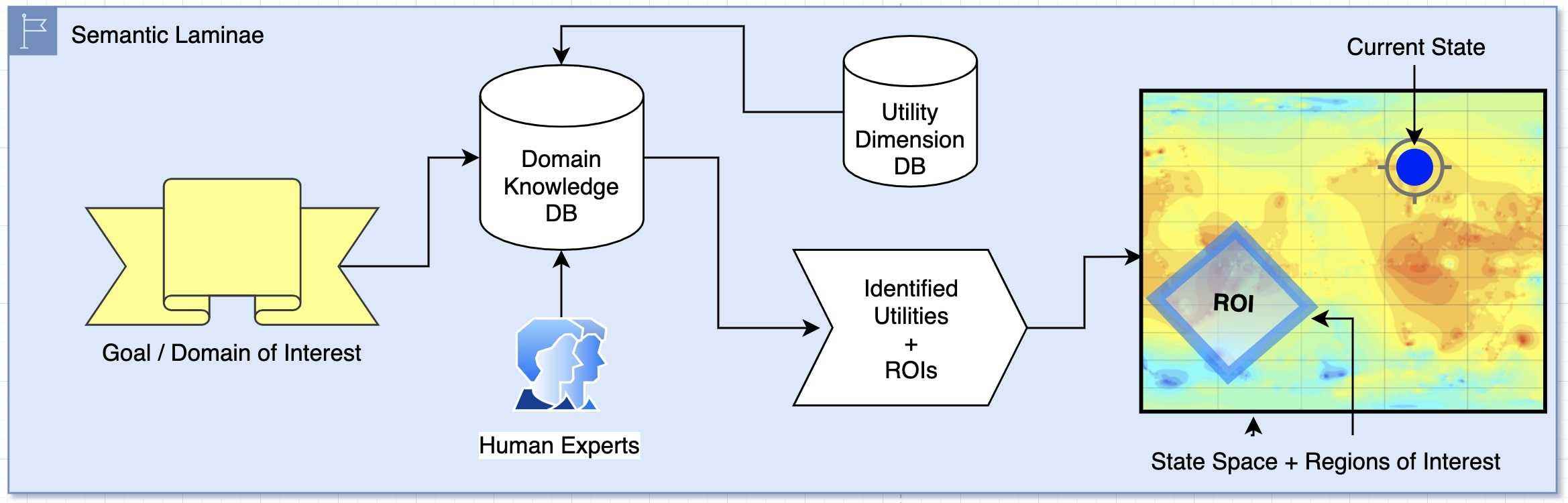}
  \caption{\textbf{Semantic Laminae.} For the system to initialize, the user must specify what goals or domain of interest they have, which is referenced against a domain knowledge database (DB) curated by experts of various fields. These experts outline the relevant utility dimensions and have map layers that contain ROIs. These map layers may also contain the connections between regions and the layers of information on how to move between states in the region. These connections will be expanded upon further in Chapter 6.}
  \label{fig:laminae}
\end{figure}

User intent can be captured through various means. A sample browser interaction to capture intent is given in Figure \ref{fig:ui}. In this case we use natural language to query the user for a general topic of interest or regions-of-interest (ROI). If the user chooses a ROI, the associated topic is retrieved. This topic then is used to retrieve the relevant coordinates or dimensions. Mapping the coordinates of a ``healthy pregnancy" ROI or ``marathon running" topic in the state space requires matching relevant laminae and dimensions that are of interest to the user intent. Each lamina has a name, a set of dimensions, and a set of ROIs. ROIs each contain a label, an enclosed boundary, and a set of attributes. The operation of the laminae within the developed system occurs through four components, which are explained in Figure \ref{fig:laminae}. The first is a user interface where the goal or domain of interest is entered. The second component uses the user input to retrieve all laminae that relevant to the user. The user can select a lamina to view from this list. When selected, the visualization loads the utility dimensions and plots the regions of interest on the user interface. The user health state location may or may not be shown on this visualization. The state space shown on the screen can be selected to be the general health state space of all possibilities within the domain, or confined to just the user's own health state space. This whole system relies upon the set of utility dimensions, which we discuss next.

\section{Utilities: The Dimensions of Health}

Utilities describe the fundamental dimensions of health state, akin to the latitude, longitude, and altitude of GIS systems. Utilities are the fundamental abilities of a living human. In this work, we define five parent nodes for the utility taxonomy based upon consensus literature in the life sciences. These major categories include movement, sensing, thinking, managing homeostasis, and reproduction. These primary groups have many nested sub-dimensions within them. The importance of utilities rests in abstraction away from the material cause of existence and instead focuses on the final cause of use, as described in Figure \ref{fig:Aristotle}. The utilities generally arise from biological underpinnings but are not dependent on biological function. A person may have a biological joint or a metal joint. The end utility is the same: to move the body in a specific manner.

\begin{figure}[H]
  \centering
  \includegraphics[width=0.95\textwidth] {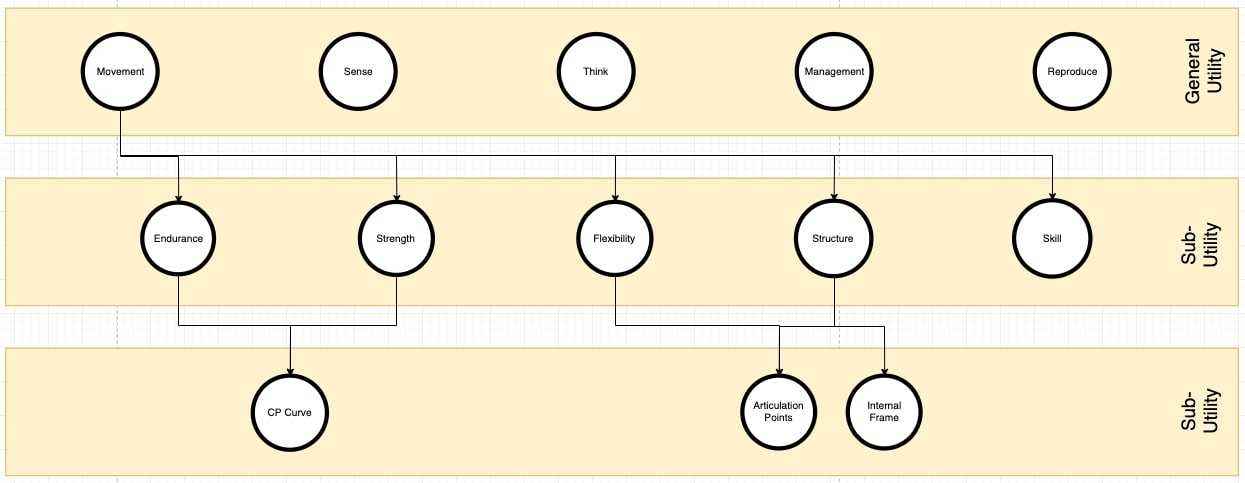}
  \caption{\textbf{Basic Utility Taxonomy.} These nodes show the nested microservice architecture of the utility dimensions. The movement node is expanded to reveal the next layer of microservices. This layer can be further expanded to more detailed child nodes. This is a simple overview of the utility GNB at large.}
  \label{fig:utility4}
\end{figure}

As mentioned above, the dimensions of health state space are defined through the utilities they provide. Utilities serve as a parameter with a particular value or categorical assignment. Structured taxonomy of human utility state space in this work is given in a simplified version to demonstrate the concept of utility space. These five categories within this block are shown in Figure \ref{fig:utility4}. If ``movement" is the parent node, then there are many child nodes that are part of movement. One of the child nodes in Figure \ref{fig:utility4} is the flexibility of the skeletal system, which encapsulates the entire summary of joint function in the body. A list of joint articulation utilities is in Table \ref{tab:joint-table} to show how there can be categorical strings, Booleans, floats, or integers as attributes for each utility. Because the majority of a person is generally composed of biology, we need to connect how utilities arise from biological systems.

\begin{table}[H]
\centering
\scalebox{.8}{
\resizebox{\textwidth}{!}{%
\begin{tabular}{@{}lll@{}}
\toprule
Joints                          & Joint Classification  & Range of Motion \\ \midrule
proximal radioulnar joint       & Pivot joint           & 190             \\
distal radioulnar joint         & Pivot joint           & 210             \\
median atlanto-axial joint      & Pivot joint           & 50              \\
elbow                           & Hinge joint           & 150             \\
jaw                             & Hinge joint           & 30              \\
proximal interphalangeal joints & Hinge joint           & 90              \\
distal interphalangeal joints   & Hinge joint           & 100             \\
foot                            & Hinge joint           & 100             \\
knee                            & Hinge joint           & 150             \\
talocrural joint                & Hinge joint           & 50              \\
subtalar joint                  & Hinge joint           & 60              \\
distal tibiofibular joint       & Hinge joint           & 45              \\
carpometacarpal joint           & Saddle joint          & 53              \\
sternoclavicular joint          & Saddle joint          & 35              \\
incudomalleolar joint           & Saddle joint          & 20              \\
acromioclavicular               & Plane joint           & 30              \\
metacarpophalangeal joints      & Condyloid joint       & 40              \\
metatarsophalangeal joints      & Condyloid joint       & 45              \\
hip joint                       & Ball and socket joint & 300             \\
shoulder joint                  & Ball and socket joint & 320             \\ \bottomrule
\end{tabular}%
}}
\caption{\textbf{Joint Utility Table.}This table is a sample list of human joint utilities. For each utility, there is a set of output attributes that determine the utility value, in this case ``Range of Motion."}
\label{tab:joint-table}
\end{table}

\section{Biological Blocks: Natural Micro-Services}
\begin{figure}[H]
  \centering
  \includegraphics[width=0.98\textwidth] {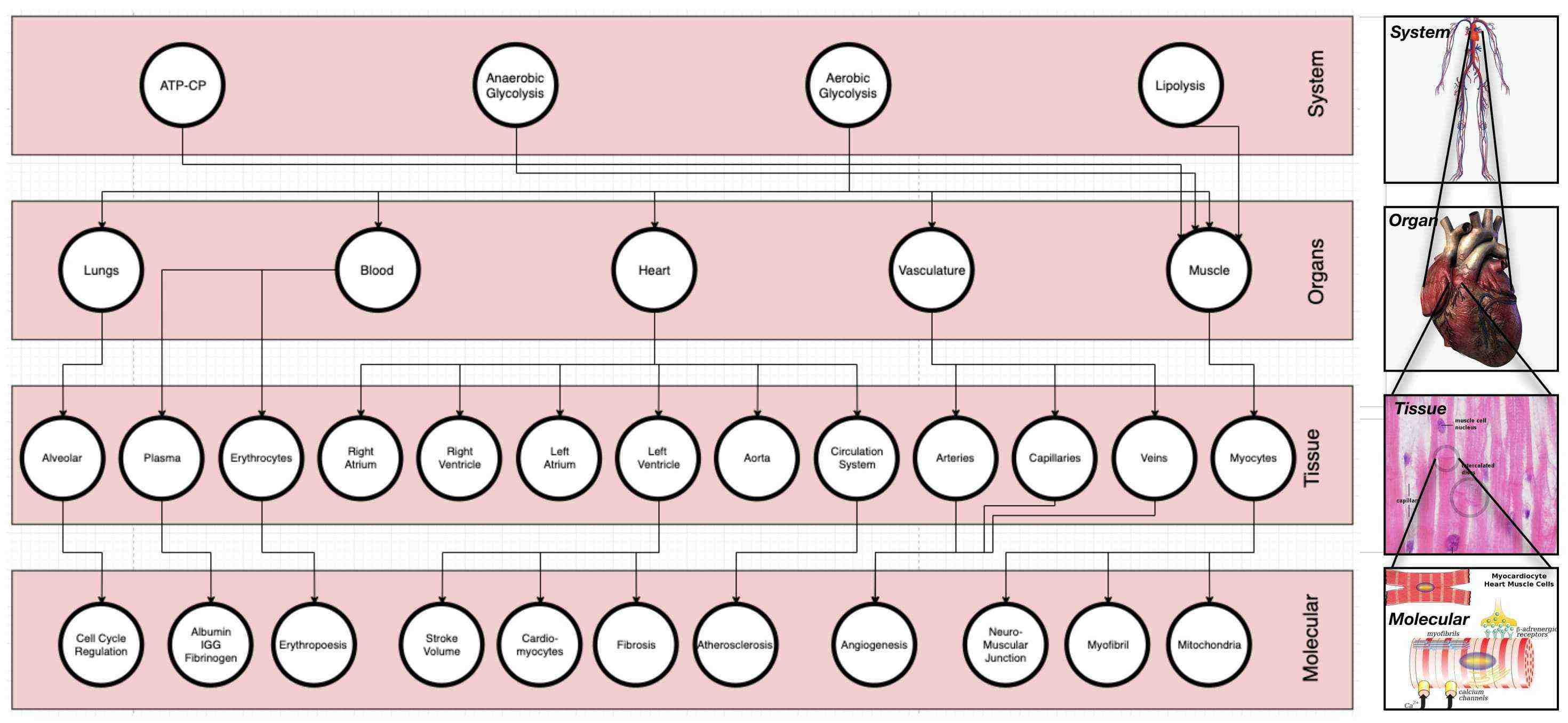}
  \caption{\textbf{Nested Biological Microservices.} Human biology is represented as GNB nodes at various layers of abstraction within the framework. This figure translates the cardiovascular system knowledge layers in Figure \ref{fig:metabolism} into the GNB structure. The four chosen abstraction layers for this work are the system level, organ level, tissue level, and molecular level.}
  \label{fig:bio4}
\end{figure}

This block contains the material cause of what humans are composed of \cite{Falcon2006AristotleCausality}. The biological components all serve certain utilities. The isolated state of biology has been traditionally the study of bioscience and is the target of health state study in most medical fields. These components nest in various levels of scientific and spatial abstraction, which can be molecular to an organ system. The microservice architecture for the GNBs allows for representing these levels of abstraction. The nesting is represented in GNB format in Figure \ref{fig:bio4}.

Personalized information that is in this structure includes anatomical, physiological, cellular, and signaling interactions. Demographic information about an individual such as age and gender may also be used to personalize the biological blocks. Genetic information is stored as relationship or node attributes that modify a generic model. Biological systems are highly complex and always in flux. A changing system inherently is temporally dependent on its inputs. Temporal data about an individual is captured in the details of their life events.
 
\section{Events: Inputs to Life}
\begin{figure}[H]
  \centering
  \includegraphics[width=0.98\textwidth] {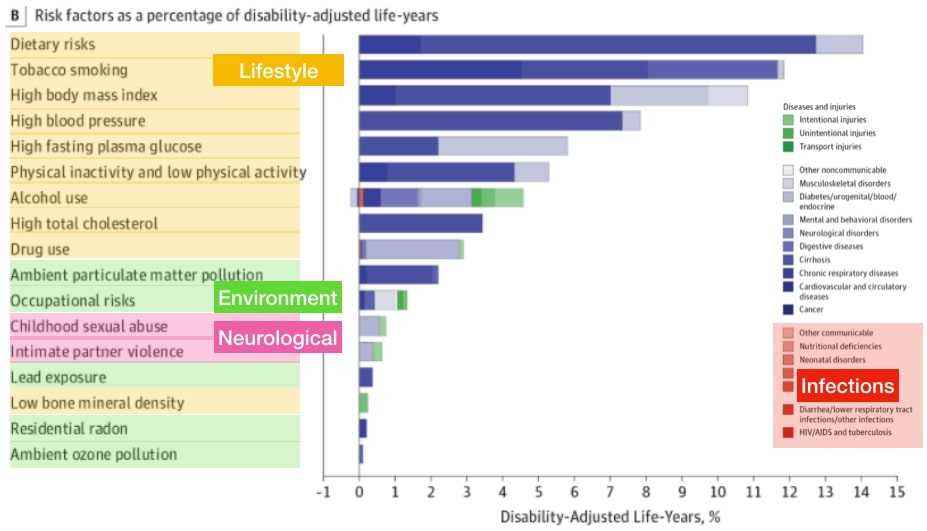}
  \caption{\textbf{Major Factors that Change Health.}  This figure displays the burden of health factors in the USA for the 20 years of 1990 to 2010 \cite{Murray2013TheFactors}. The top three items that change the health state of an individual in the United States are lifestyle events, followed by environmental and neurological events, validating the necessity of tracking events for estimating health state.}
  \label{fig:jamariskfactors}
\end{figure}

What changes our health state, and when does it happen? Every event at each moment of the day affects our health. Investigating the significant factors that impact health, we see that life events are the central premise behind changes in health state. As shown in Figure \ref{fig:jamariskfactors}, these factors are all part of daily life events for an individual.

\begin{figure}[H]
  \centering
  \includegraphics[width=0.98\textwidth] {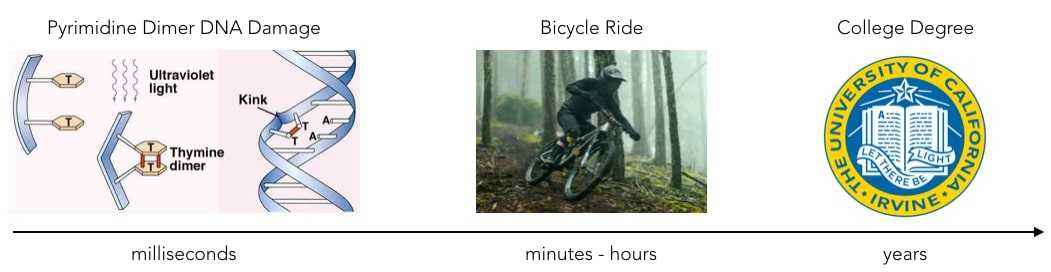}
  \caption{\textbf{Personal Events: Time Scale of Intervals.} The temporal scale of events that happen to a person have a large range that can span over 15 orders of magnitude.}
  \label{fig:dimer}
\end{figure}

A person's life history is traditionally documented with a biography. A biography consists of a series of events that flow over time. These events of life happen at varying time-scales. Take, for example, nanosecond molecular interactions of cell receptor binding compared to a four-year college degree. The difference between these time scales is 16 orders of magnitude. The temporal scope of life events is shown visually in Figure \ref{fig:dimer}. Event architecture allows for a flexible design where various event types are recorded. Within any particular time window, a set of events will contain inputs that the change in biology, a phenomenon constant from conception until death.

The HSE framework classifies events into three main categories. The first category, the exposome events, are all events happening outside the body of the person, including all environmental exposures of light, sound, weather, pollution, and any other physical attribute outside the skin. This environment can also include social events or interactions with external people or agents. The second category of events, the life events, includes all events the person's body engages in from a molecular level to the whole body. At a coarse level, these events include any activity, food intake, sleep, or any general life events as described by the Personicle system or Daniel Kahneman's day reconstruction method \cite{Oh2017FromChronicles,Kahneman2004AMethod}. At a granular level, this could include events of organ systems such as the heart (such as an arrhythmia event, volume overload event, pressure overload event), pancreas, or gut. Even more granular are events such as reactions of molecules such as ultraviolet light damaging DNA. The third category of events is called neurological events, including all events that are perceived by the mind. These events are generally subjective in nature and include events such as joy, stresses, or dreams. In summary, the three classifications of stored events are everything outside the skin, inside the skin, inside the mind.

\begin{figure}[H]
  \centering
\scalebox{0.95}{
  \includegraphics[width=0.95\textwidth] {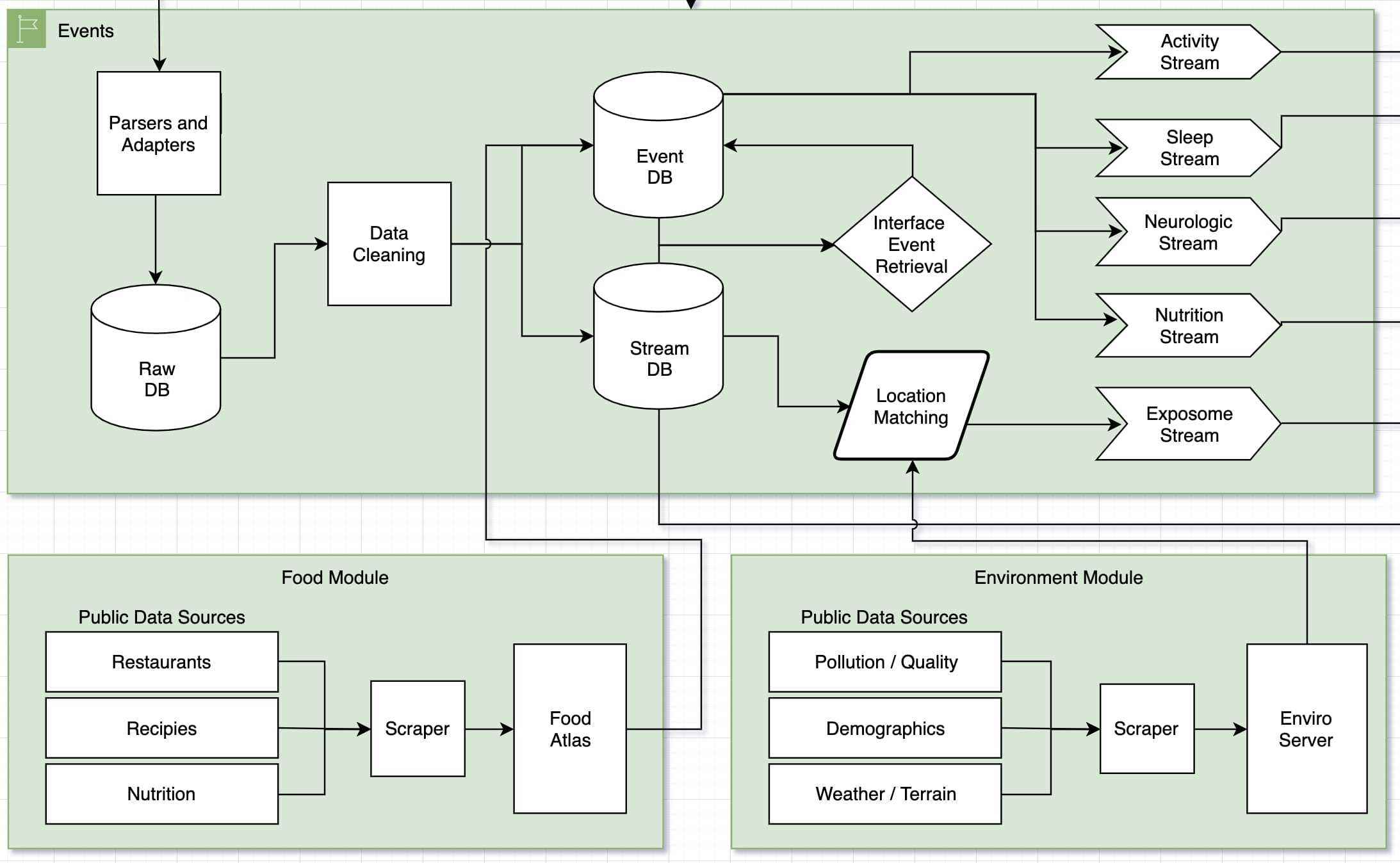}
  }
  \caption{\textbf{Event Processing Framework.}}
  \label{fig:eventframework}
\end{figure}

Events happening to a person are captured and then stored in a DB system with three containers. The first container is an event stream, where interval-based events can be stored and retrieved. The second container is the stream database, where continuous data streams can be stored and retrieved. The third container is a specialized version of the stream database, where the location data stream can be stored and retrieved. A table characterizing this storage schema is in Table \ref{tab:sample-schema}. Operationally, the event module connects to incoming data streams through parsers and adapters. These data are stored in a Raw file DB and redundantly backed up as the most original form of the user's data. After cleaning the data of erroneous data through predetermined filters, the data is translated to either the event format or the stream format and copied into the Event DB or Stream DB appropriately. The location stream in the Stream DB then fetches environmental data at the particular location and time stamps the user has data to gather weather, pollution, demographics and any other requested geospatial data. Both the Event DB and Stream DB are used in the Interface Event Retrieval process, which we will go over in the following section. A visualization of this framework is given in Figure \ref{fig:eventframework}.

\begin{table}[H]
\centering
  \begin{tabular}{cll}
    \toprule
    Stream & Field name & Data type\\
    \midrule
    Event & Event type & String\\
          & Event name & String\\
          & Start time & Timestamp\\
          & End time & Timestamp\\
          & Parameters & Key-Value pairs\\
          & Data streams & Set\\
          \\
    Real valued Data Stream & Timestamp & Timestamp\\
          & Value & Numeric\\
          & Unit & String\\
          & Source & String\\
          \\
    Location Data Stream & Timestamp & Timestamp\\
          & Value & Location\\
          & Unit & String\\
          & Source & String\\
  \bottomrule
\end{tabular}
\caption{\textbf{Event Schema.}Event schema uses an interval approach to define the event and associated parameters. Data streams are split into two schemas, one for general data streams, and one specifically for a location that includes a geospatial value of latitude, longitude, and altitude stored as ``LOCATION."}  
\label{tab:sample-schema}
\end{table}

We conclude the introduction to the four major components of the HSE framework. To set up the system, we jump into initializing the system in the next section.

\section{Initializing the System}

\begin{figure}[H]
  \centering
  \includegraphics[width=0.99\textwidth] {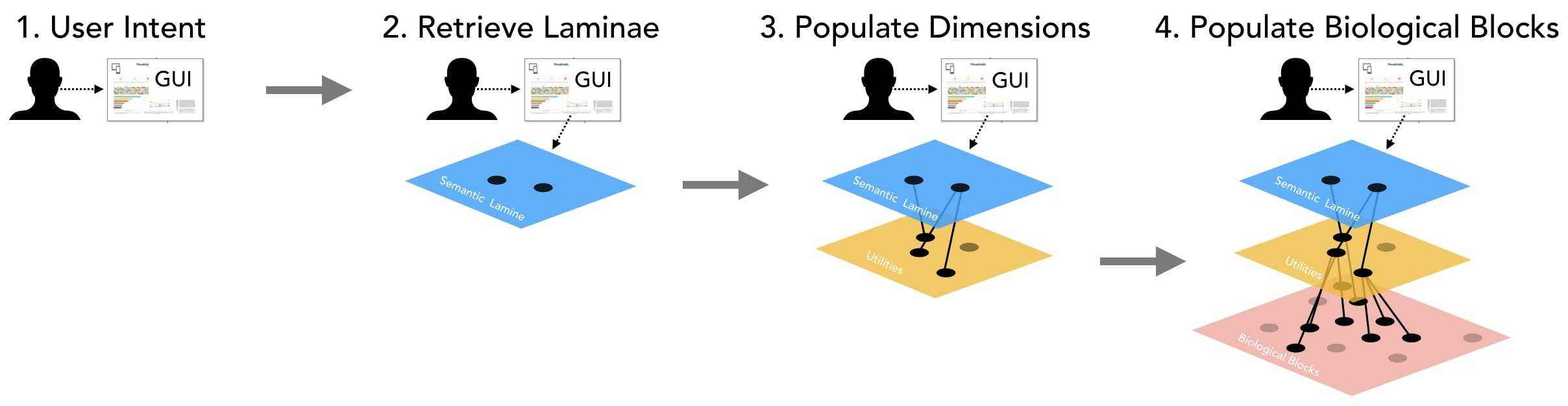}
  \caption{\textbf{Initialization Steps.} These are the four steps in initializing the graph structure for the individual. \textbf{1.} First, we query the user for their intent. \textbf{2.} Using this query, we retrieve the appropriate laminae they can visualize. \textbf{3.} These laminae require a certain set of utility dimensions to be displayed. These are populated next. In order to know the user position on these dimensions, the appropriate biological blocks need assessment. \textbf{4.} The fourth step populates these appropriate blocks to monitor.}
  \label{fig:initial}
\end{figure}

In order to begin the state estimation process for the individual, we need to populate the four main components concerning the person. The state space of all possible dimensions of health is vast, and to narrow the state space to an appropriate level, we query the user for their particular interest. Given this intent, we can give a list of laminae that match the intent. The user can then select to visualize a particular lamina. Once this lamina is chosen, the relevant dimensions must be populated to generate the state space. This process requires retrieving the utility dimensions and the supporting biological blocks, as visualized in Figure \ref{fig:initial}.

\begin{enumerate}
    \item \textbf{Obtain User Intent.} By providing a certain intent, the user can request to view their health state on a lamina that is their interest. Similarly, if a computing agent wants to understand a particular aspect of the health state, it may also have an intent.
    \item \textbf{Retrieve Associated Semantic Laminae.} Given this intent, relevant layers are retrieved from a database with domain knowledge of the particular query intent.
    \item \textbf{Populate Utility State Dimensions.} These layers are placed upon their respective dimensions of utility.
    \item \textbf{Populate Biological Blocks.} Given this set of utilities, the system then populates all relevant biological components that determine the utility state.
\end{enumerate}

\section{State Update Method}
\begin{figure}[H]
  \centering
  \includegraphics[width=0.99\textwidth] {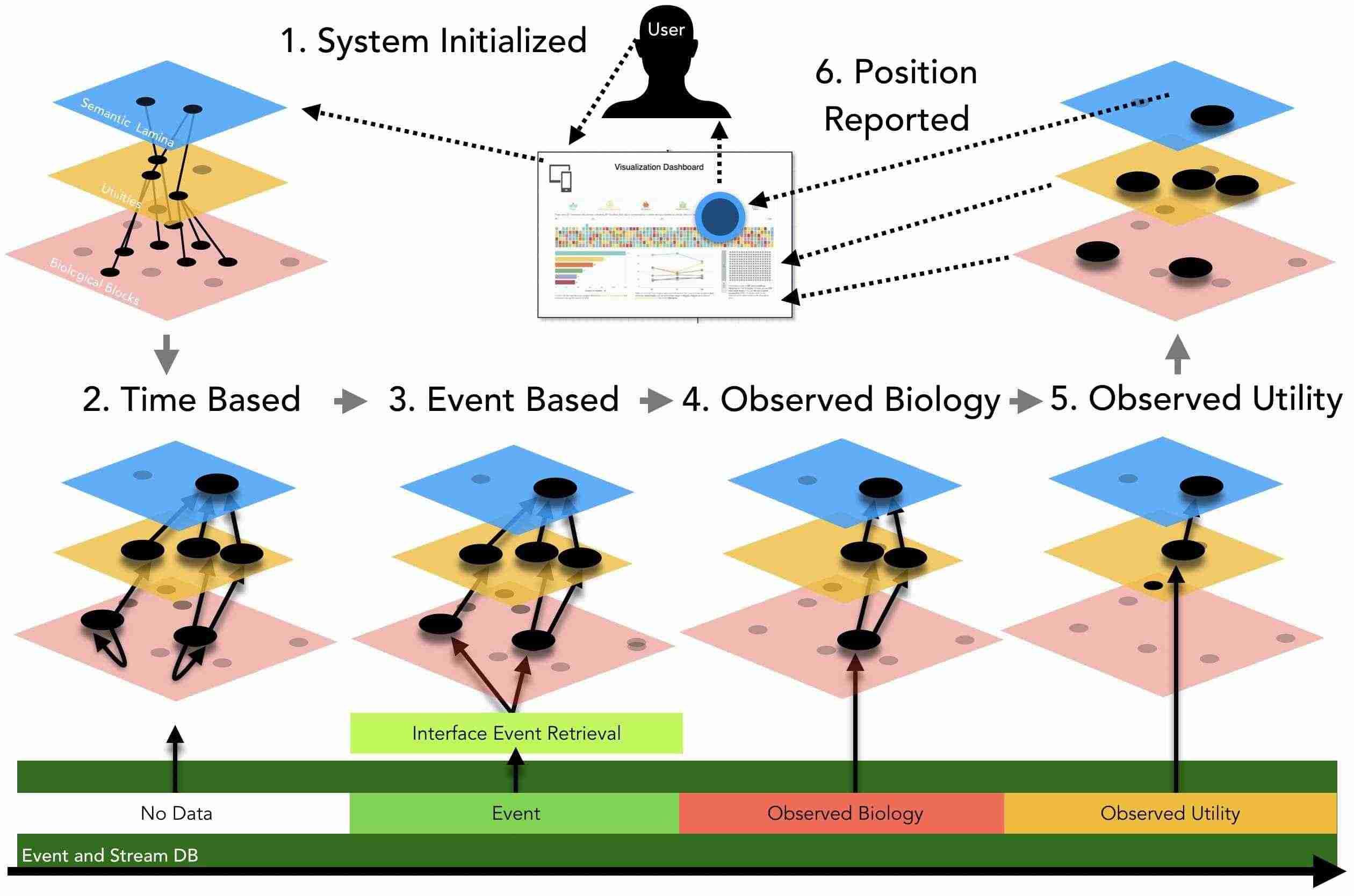}
  \caption{\textbf{State Update Methods.} \textbf{1.} The system initiates from the user intent. \textbf{2.} If there is no data from the event or stream DB reported, self-edges update nodes which then propagate along the directed edges to give new utility values. \textbf{3.} If interface events are detected from the event stream, they update the relevant biological states, which then propagate. \textbf{4.} If there are observed biological attributes directly from sensors, they update the respective biological nodes. \textbf{5.} If there are observed utility attributes directly from sensors, they update the respective utility nodes. \textbf{6.} The health state position is reported on the lamina of interest, and/or can be viewed as raw utility dimensions, and/or can be viewed as biological attributes.}
  \label{fig:stateupdate}
\end{figure}

There is constant flux within an individual's personal health state space (PHSS). We must be able to assign an accurate location within the PHSS for state estimation. Understanding this location in real-time is best powered by multi-modal data sources that update the health estimation. For example, monitoring a cardiovascular health state is useful to both endurance athletes and heart disease patients. Estimation techniques have been of great interest in designing many applications, but health applications will require increasingly deep biological knowledge layers to define and estimate health states. This process uses the knowledge generated graph structure model from the system initialization of an individual to deliver updated states. To compute the state of health using this flow, we need to continually compute the individual's position along each dimension of utility from any new source of data or knowledge. Below is outlined the step-by-step overview of how this occurs. Figure \ref{fig:stateupdate} shows a visual flow of how these steps carry out.

\begin{enumerate}
    \item \textbf{Initialize System and Connect to Personicle.} As given above, the set of utilities and relevant biological components are populated. The person's event and data history are then loaded from a personicle.
    \item \textbf{Time Based Updates} Any nodes with self-edges update within the biological blocks. This will be the only update within the whole cycle if there is no data in the personicle. The only piece of information used here is the time since the last update.
    \item \textbf{Event Based Updates.} If there is an event that has occurred within the personicle, the interface event retrieval system gathers any events that connect to the populated biological states. The attributes of these events determine the change in a particular biological block. Interface event retrieval is in more detail below.
    \item \textbf{Observed Biology} If there are direct observations in the personicle of any biological components (nodes or edges), they are updated.
    \item \textbf{Observed Utilities} If there are direct observations in the personicle of any utility dimensions (nodes or edges), they are updated. 
    \item \textbf{Update Utility dimensions.} All estimated values are propagated forward along the directed edges to update the utility nodes required to show the position of the user in the health state space.
    \item \textbf{Visualization of Current Health State Position.} The updated values for the utility states are then shown as an updated position on the selected visual lamina to the user.
\end{enumerate}

\subsection{Interface Event Retrieval}
Interface events are particular sub-events with specified attributes that are responsible for a particular change in health state, interfacing between the actions in our lives and the state of our health. After combining general events and data streams from different sources, we specify the interface events as logical statements. Each interface event has direct causal edges on particular biological nodes. These events and relationships are well studied in biomedical literature and can be defined as a combination of data and event streams using an event operator language. When we detect an interface event, the relationship with the biological variables contributes to the update of the health state. This work will retrieve interface events from five main event categories in chapter 5. These categories include nutrition, exercise, neurological, environment, and sleep events. Logical statement examples are given below describing left ventricular volume overload cardiac events (\textit{VolOverload}), and pressure overload left ventricle cardiac events (\textit{PressOverload}).

Representations of these events are given below:

    \begin{displaymath}
      VolOverload := (HR > 140) \vee (Cycling \wedge detect-climb(Altitude))
          \end{displaymath}
   \begin{displaymath}
      PressOverload := detect-spike(HR) \vee (Power > 400 W)
    \end{displaymath}

The state update mechanism uses the same node and edge structure as new data arrives. In order to change the graph structure itself, we need to update the GNB model itself.

\section{Individual Dynamic Model Updates}
To improve the performance of the GNB model, we use both data and domain knowledge approaches to edit the GNB structure itself. We update the node attributes, edge relationships between nodes, adding new nodes, adding new edges, or removing edges and nodes. There are four key methods by which these updates can occur, one based in knowledge-driven approaches, and the other three based on data-driven approaches. 

\begin{figure}[H]
  \centering
  \includegraphics[width=0.99\textwidth] {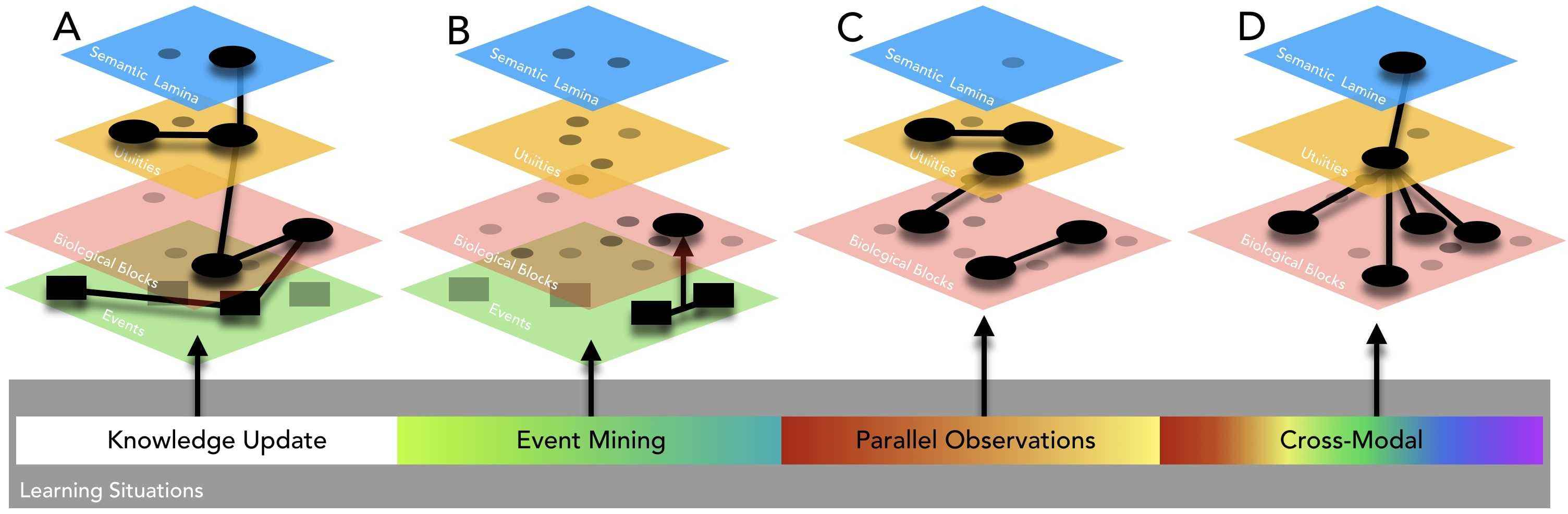}
  \caption{\textbf{Learning Situations.} Relationships between nodes update in both inter-layer edges or intra-layer edges. \textbf{Situation A.} When new knowledge is given to the system, the appropriate edges or nodes are updated. \textbf{Situation B.} Event-mining can discover relationships between temporal events within the user personicle. These events can reflect an update to a biological node that models behavior or any other biological component. \textbf{Situation C.} When two nodes that are connected are both measured in parallel, the relationship between the two nodes can be characterized to match the incoming data. \textbf{Situation D.} When many source nodes are related to a target node, we can use all the nodes to learn how they impact the target node.}
  \label{fig:learning}
\end{figure}

Knowledge-based approaches reference a knowledge DB in order to establish the graph structure. When a scientist discovers new knowledge or a new genetic association, the system can access these updates to update the individual GNB. An overview is given below in a list format. Figure \ref{fig:learning} describes these four approaches visually.

\begin{enumerate}
    \item \textbf{Knowledge Update.} If there are new static knowledge that is associated with the GNB or individual (such as their genetic information), they apply to the relationships within the graph.
    \item \textbf{Event Mining.} Event mining establishes relationships on a temporal scale between events. This work does not directly address event mining. It provides a framework that allows for this type of approach in future research.
    \item \textbf{Parallel Observations.} When data is observed regarding two nodes that are connected, the relationship can be updated using the data relationships. If the predictive quality is suitable, the respective edges are updated with this relationship.
    \item \textbf{Cross-Modal Estimation.} If there are multiple relationships with a certain node, they can be used in combination to learn their relationship to the target node. If the predictive quality is suitable, the node updates with this relationship to its sources.
\end{enumerate}

The above two sections cover the process of how we estimate the health state and continuously modify the graph structure model of an individual to reflect what we learn from data. In order to verify if these approaches satisfies the needs of HSE, we compare this framework overview against the design requirements.

\section{Revisiting the Design Requirements}
Based on the discussions so far, we have defined the following design principles essential to the HSE framework in chapter 2. Below describes how we meet these particular demands with the above framework.

\begin{enumerate}
    \item \textbf{Participation with User.} We begin addressing the design requirements by directly engaging with the user's intent in the semantic laminae. This is the root of instantiating the system and ensuring that it works towards the user's goals.

    \item \textbf{Encompass Holistic Definition of Health.} Because a holistic definition of health goes beyond biology, we abstract health needs into utility states.

    \item \textbf{Capture Connectedness.} In order to capture how aspects of health are constantly evolving as a system, we choose to use a graph-based approach.
    
     \item \textbf{Flexible and Modular System.} By nesting the graphs into blocks, we can take advantage of microservice architecture to represent and model the components of interest.
    
    \item \textbf{Zoom Fluidly Between Abstraction Scales.} These nested blocks allow the user or system to interact at an abstraction level that is appropriate for the situation. Scales can range from molecular to person level health.

    \item \textbf{Causal and Explainable.} To have a causal and explainable system, we choose to have directed edges in the graph that represent causal relationships. Understanding the root of how something is changing can be traced in reverse on this causal pathway.
    
    \item \textbf{Domain Knowledge Integration.} To determine what directed edges to add to the system, we reference domain knowledge to establish the relationships as a cold start instantiated personal model.
        
    \item \textbf{Diverse Data Fusion and Transformation.} Using the high-level abstraction of events, biology, and utility, we can transform various types of data into these three layers.
    
    \item \textbf{Perpetually Updating.} When new data arrives into the system, we can propagate updates to other nodes via the directed edges.
    
    \item \textbf{Personalized Modeling.} Physiologic function, psychological function, risk, and any other attributes that are related to genetics, life history, age, or other demographics data can modify the relationships between nodes to personalize the graph network.
        
    \item \textbf{Precision through Understanding and Learning.} When data about two related nodes arrive in parallel, we can use learning approaches to characterize the relationship better. The general domain knowledge relationship can then be updated to reflect a precise model for the individual.
    
    \item \textbf{Predictive and Preventive Capability.} By using this constantly adapting network, we can simulate inputs to the graph and compute future outputs. If a future output is undesirable, we can prevent it by testing inputs that lead to a desirable future state.
\end{enumerate}

In summary, the structure of this framework satisfies the design requirements we set out to address. In the next chapter, we dive into the details of how we can operationally implement such a framework for a real person.
\chapter{Instantiating Personal Health State Estimation}

\epigraph{A healthy vision of the future is not possible without an accurate knowledge of the past.}{\textit{Daisaku Ikeda}}
\vspace{12pt}

\begin{figure}[H]
  \centering \centerline{\epsfig{figure=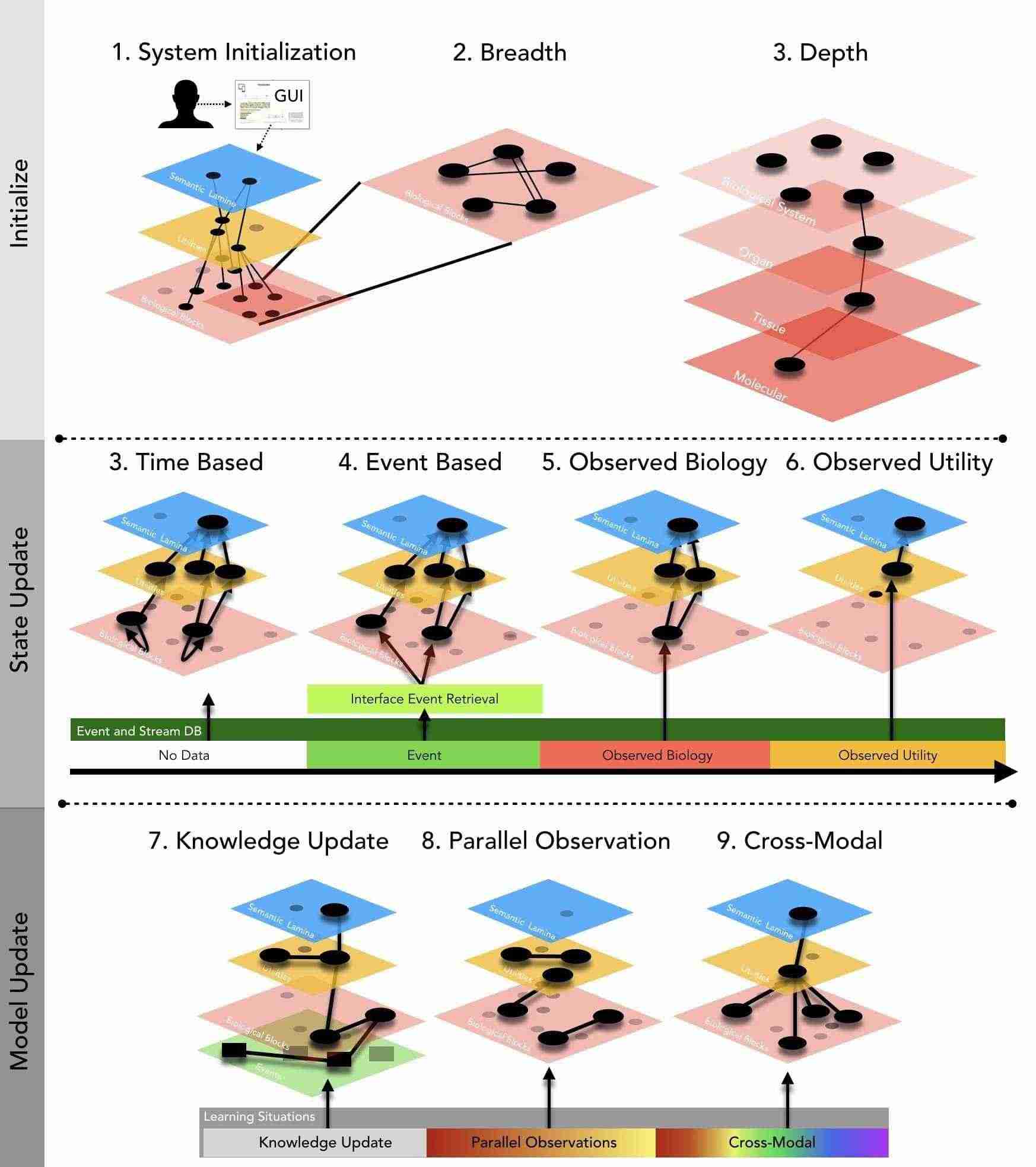,width=16cm}}
  \caption{\textbf{Chapter 5 Experimental Overview.} The first part of Chapter 5 is initializing the system for the test subject in both breadth and depth. The second phase demonstrates the state estimation in action. The final set of experiments demonstrate three approaches to update the personal GNB model.}
    \label{fig:overview}
\end{figure}

Chapter 5 describes using the framework described in chapter 4 within a real-world implementation of personal HSE. The chapter begins with an overview of the methods and tools used to initialize the HSE GNB at an individual scale. Initialization is for both biological breadth and depth with a focus on CRF. We then cover the types of data collected, and how they are processed towards the state estimation given various data situations. Insights from the state estimation are discussed. We conclude this chapter by showing how we dynamically edit the model with new knowledge and data-streams to fit the individual better. A visual experimental set up is shown in Figure \ref{fig:overview}.

\section{Experimental Overview}

\section{Methods}
Applying the concept described in Chapter 4 within a real experiment set-up provided a few challenges. First, there is no dataset in existence at the level of detail required for this research, posing the most significant barrier to implementing this research. Although a great task to build such a comprehensive data collection pipeline, we decided to create the dataset organically to specification around CRF. We chose to focus on a single subject, with a personalized approach. We also needed data over a long temporal span to show how the health state is changing. Data collection is also necessary to be non-intrusive to cause minimal observer effect. We also want these experiments replicated within the scientific and clinical settings. Hence we chose mostly commercially available hardware and software. These decisions resulted in exemplifying the state estimation framework in chapter 4 within real-world settings and allow for using complex personal data. The next four subsections cover these experimental methods in detail.

\subsection{Longitudinal}
As we want the state estimation to provide continuous estimates over time, the most appropriate experimental approach would be a longitudinal one on a specific subject, rather than a cross-sectional study of many individuals. Longitudinal studies consist of repeated observations and measurements of the same variables, such as people, over periods (could be decades at a time). The data collected is a combination of quantitative and qualitative exposures and outcomes data of the controlled variables. This type of study is useful in determining relationships between risk factors and disease development \cite{Caruana2015LongitudinalStudies}.

A cross-functional observational study, on the other hand, analyzes data from a population with multiple variables at a specific point in time, often used in cohort studies \cite{Lee1994OddsData}. These studies can be useful to understand the prevalence of disease in a clinic-based sample \cite{Setia2016MethodologyStudies}. However, with no information with the influence of time, the difficulty is to derive causal relationships \cite{Caruana2015LongitudinalStudies}.

\subsection{N=1}
Because we are trying to build personal estimation models, the incoming data must only be from the individual. Population-level data is helpful as a part of the knowledge-based edges, but cannot be used to personalize the graph network for the individual. For this reason, we choose to have an N=1 approach. N=1 approaches are also gaining traction in clinical settings. An N of 1 trial is a clinical trial in which one patient is the entire scope of the case study. These studies include experimental and control interventions in the design. The goal of this type of trial is to determine the optimal intervention for a patient, given data-driven criteria \cite{Lillie2011TheMedicine}. While N=1 studies are not highly common in the medical community, Lillie proposes further use of this method given the ``contemporary focus on individualized medicine." There has been further research conducted supporting the feasibility of N=1 clinical trials \cite{Guyatt1990TheExperience,McQuay1994DextromethorphanDesign,Odineal2020EffectTrial}. Table \ref{tab:n1-table} lists several simple experimental designs to conduct these N=1 trials. When updating the state, we use an A-B method using domain knowledge to establish a causal relationship. When we update the model using data-driven approaches, we use the A-B-A-B experimental method to capture many instances of A and B and then derive the relationship.

\begin{table}[hbt!]
\centering
\begin{tabular}{@{}lll@{}}
\toprule
Design & Causality        & Use                            \\ \midrule
A-B    & Quasi experiment & Often the only possible method \\
A-A1-A    & Experiment       & Placebo test design where A1 is a placebo     \\
A-B-A  &  Experiment   & Withdrawal Study where effect of B can be established \\
A-B-A-B       &  Experiment   & Further Validation of Above Study Design   \\
\bottomrule
\end{tabular}
\caption{\textbf{Quasi Experiments.} Quasi experiments cannot definitively establish causality, but provide evidence towards a related mechanism (correlation or causal). Experiments establish proof through demonstration \cite{Chapple2011FindingTrials}.}
\label{tab:n1-table}
\end{table}
    
\subsection{Multi-Modal Fusion}
We explored the vast multitude of data acquisition approaches in Chapter 3. To have the most holistic and high-resolution health state estimate, we need data of all types about the individual. This data range includes low-cost sensors that have a high temporal resolution or low user burden and high-cost sensors that may only have sparse readings. This approach takes inspiration from sensor fusion approaches that aim to use all possible information to get better situational awareness, precision, and accuracy. Sensor fusion combines sensory data from different sources of uncertainty but provides greater insight when combined. There may be ``uncertainty reduction" as the aggregation of the data can be more dependable than each piece alone \cite{Gustafsson2013StatisticalFusion,Elmenreich2002SensorSystems}. Distributed sensors are represented in a network with dynamically changing topology between each data source \cite{Xiao2005AConsensus}. Numerical representations of sensor fusion, such as the ``certainty grid," create probabilistic geographic mapping given the assimilation of sensor data \cite{Moravec1989SensorRobots}. Our approach uses a centralized sensor fusion approach, where all sensors are uploaded together into a single system for analysis. When appropriate, we use weighted averages of various sensors depending on their data quality to derive the fusion.

\subsection{Living-Labs Implementation in the Real World}
The final choice involved in choosing methods for the following experiments was to use devices and software that is readily available to the public. Other than one of the exposome sensors, the entire suite of products used for the experiments is available in the consumer goods market without any special restrictions. We chose this approach to reduce the burden of developing proprietary components, demonstrate how the fusion of various sources is key for novel health state insight, and scale and implement these types of experiments easily for future researchers or for those who wish to implement these experiments for themselves. Furthermore, the available sensors are minimally intrusive to collect the best continuous data streams in a natural setting. Nobel Laureate Richard Thaler's work inspired the thought process in behavioral economics regarding nudges that unconsciously incentive our behaviors \cite{Thaler2016BehavioralFuture}. By designing the data collection to be mostly unconscious, we can help ensure that observer effects are at a minimum, and the data represents real life most accurately. Observer effects are also known as the Hawthorne effect, a type of response in which people modify their behavior when they are observed. This response can change the integrity of the research and relationships between variables \cite{Salkind2010EncyclopediaDesign}. Aims for quantifying the Hawthorne effect have been conducted \cite{Adair1984TheArtifact} and specifically in clinical trials \cite{McCarney2007TheTrial,DeAmici2000ImpactAnesthesia}. By having measurements as a background part of normal daily life, these observer effects are minimized.

The first step in starting these experiments is to collect data about the subject. We will review the tools for data compilation in the next section.


\section{Data}
To fuel the HSE framework, we need to collect various types of data. The data needs are dictated by the user's intent and any biological or utility parameters related to this intent. We use a combination of hardware and software alongside public data resources and open source-based tools to collect data about the user within our scope of CRF. 

\subsection{Collection Tools}
\begin{figure}[H]
  \centering \centerline{\epsfig{figure=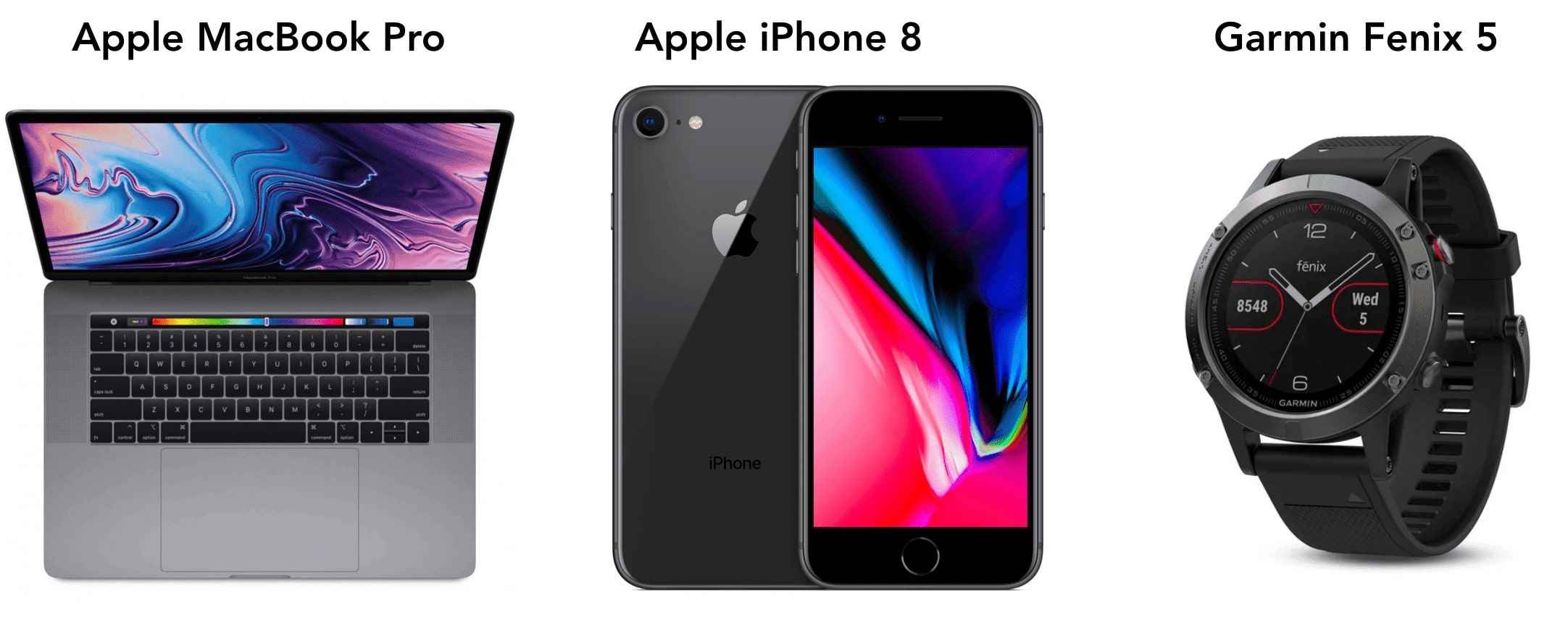,width=8cm}}
  \caption{\textbf{Central Hardware Devices.} These are the three main central devices to which other sensors link via local wireless connections of either Bluetooth, WiFi, or ANT+ protocols. The devices themselves include their sensors. The notebook computer contains a microphone, front camera, and HCI capture software. The mobile phone includes a light sensor, microphone, accelerometer, gyroscope, barometer, front, and rear camera. The wearable watch includes an optical sensor, accelerometer, gyroscope, barometer, and GPS.}
    \label{fig:centralhardware}
\end{figure}

\begin{figure}[H]
  \centering \centerline{\epsfig{figure=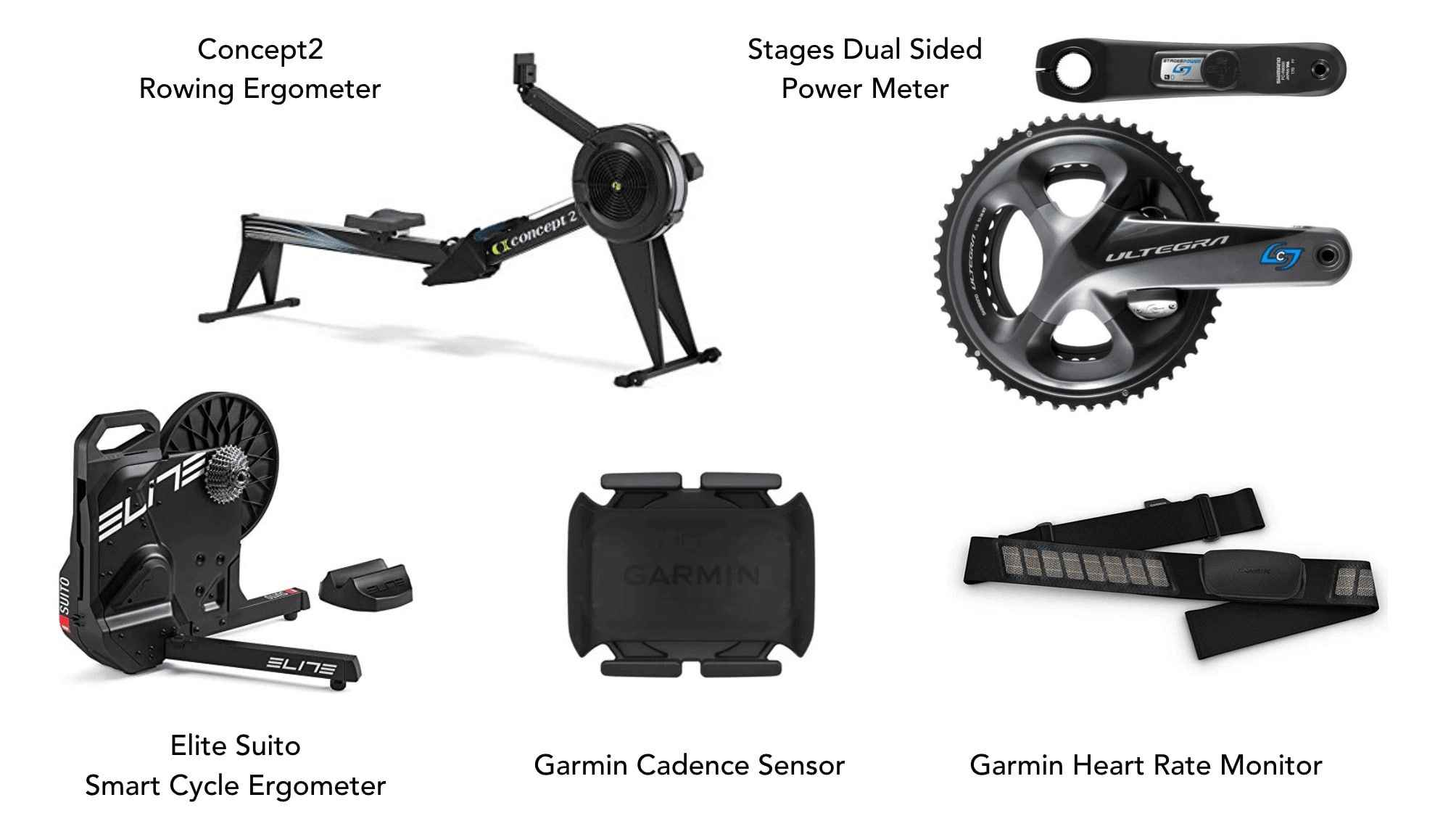,width=10cm}}
  \caption{\textbf{Activity Sensor Hardware.}}
    \label{fig:activityhardware}
\end{figure}

\begin{figure}[H]
  \centering \centerline{\epsfig{figure=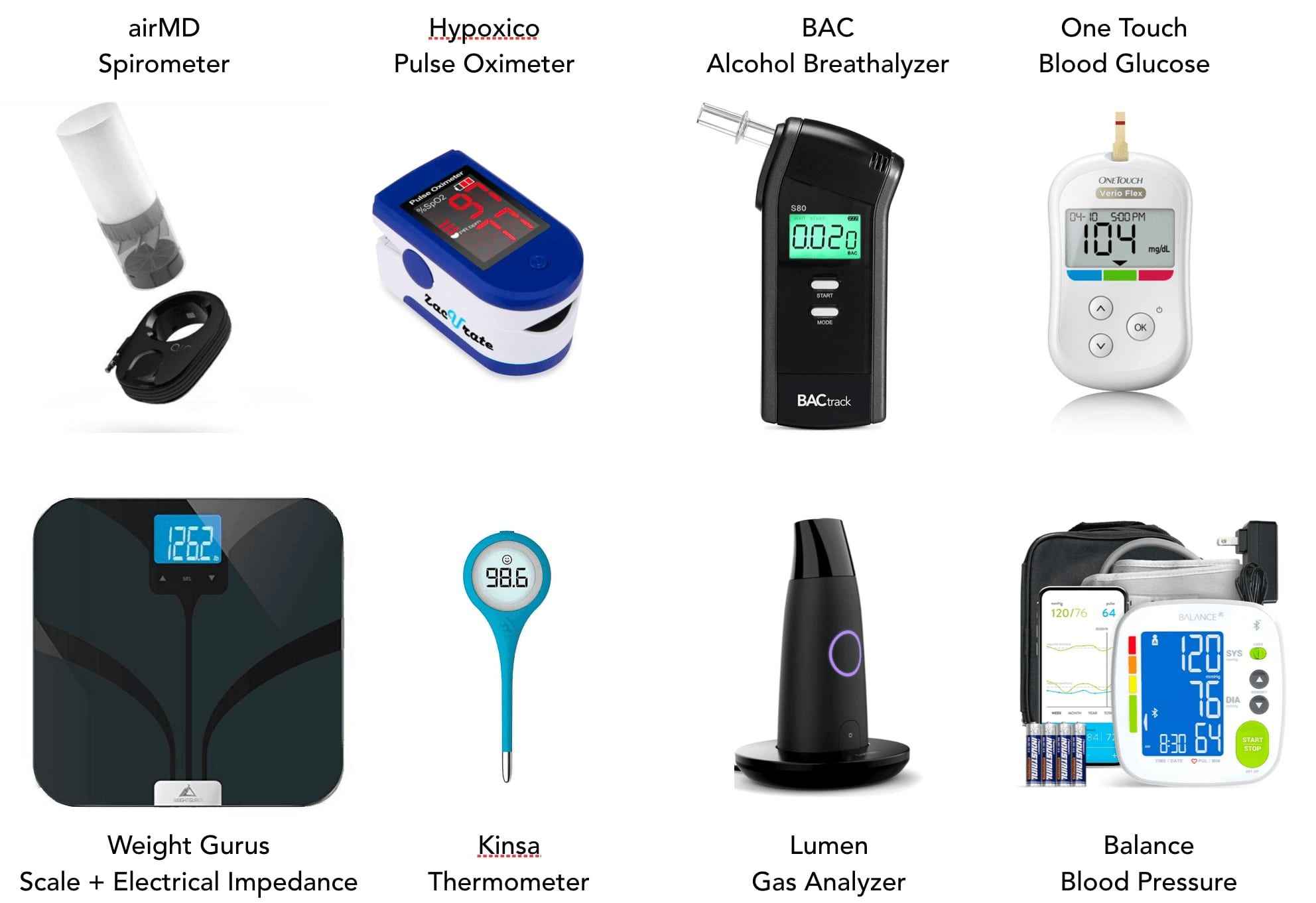,width=10cm}}
  \caption{\textbf{Biological Measurement Hardware.}}
    \label{fig:bodyhardware}
\end{figure}

There are three main central hardware components: a computer, mobile phone, and wearable described in Figure \ref{fig:centralhardware}, which tether together all the sensors and data collection through local or cloud network connections. Because the user participates in many activities, the suite of activity sensors includes rowing machines, cycle ergometers, heart rate, and cadence accelerometers, as shown in Figure \ref{fig:activityhardware}. The data streams resulting as an output from these devices are explained further in Figure \ref{fig:bikestream} and Figure \ref{fig:bikestream2}. We capture biological measurements about the user through a host of smart-connected devices, as shown in Figure \ref{fig:bodyhardware}. The ability of these devices to track and record metadata alongside the primary sensor without any extra burden to the user is critical towards encouraging data collection and reducing friction to use. This reduction in friction is part of the real-world implementation concept as people need to have a seamless experience to be nudged towards engaging with these tools. Some measurements are still challenging to capture through smart devices due to this need for user engagement.

\begin{figure}[H]
  \centering \centerline{\epsfig{figure=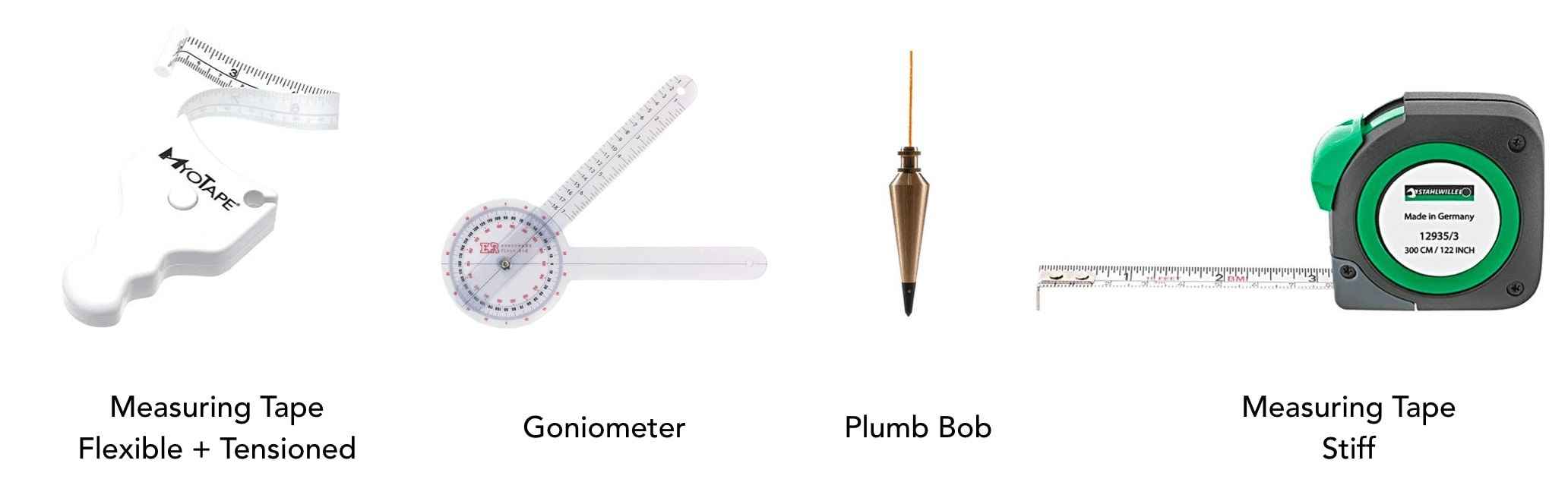,width=10cm}}
  \caption{\textbf{Manual Measurement Tools.} These tools were used to capture basic physical measurements.}
    \label{fig:manualhardware}
\end{figure}

\begin{figure}[H]
  \centering \centerline{\epsfig{figure=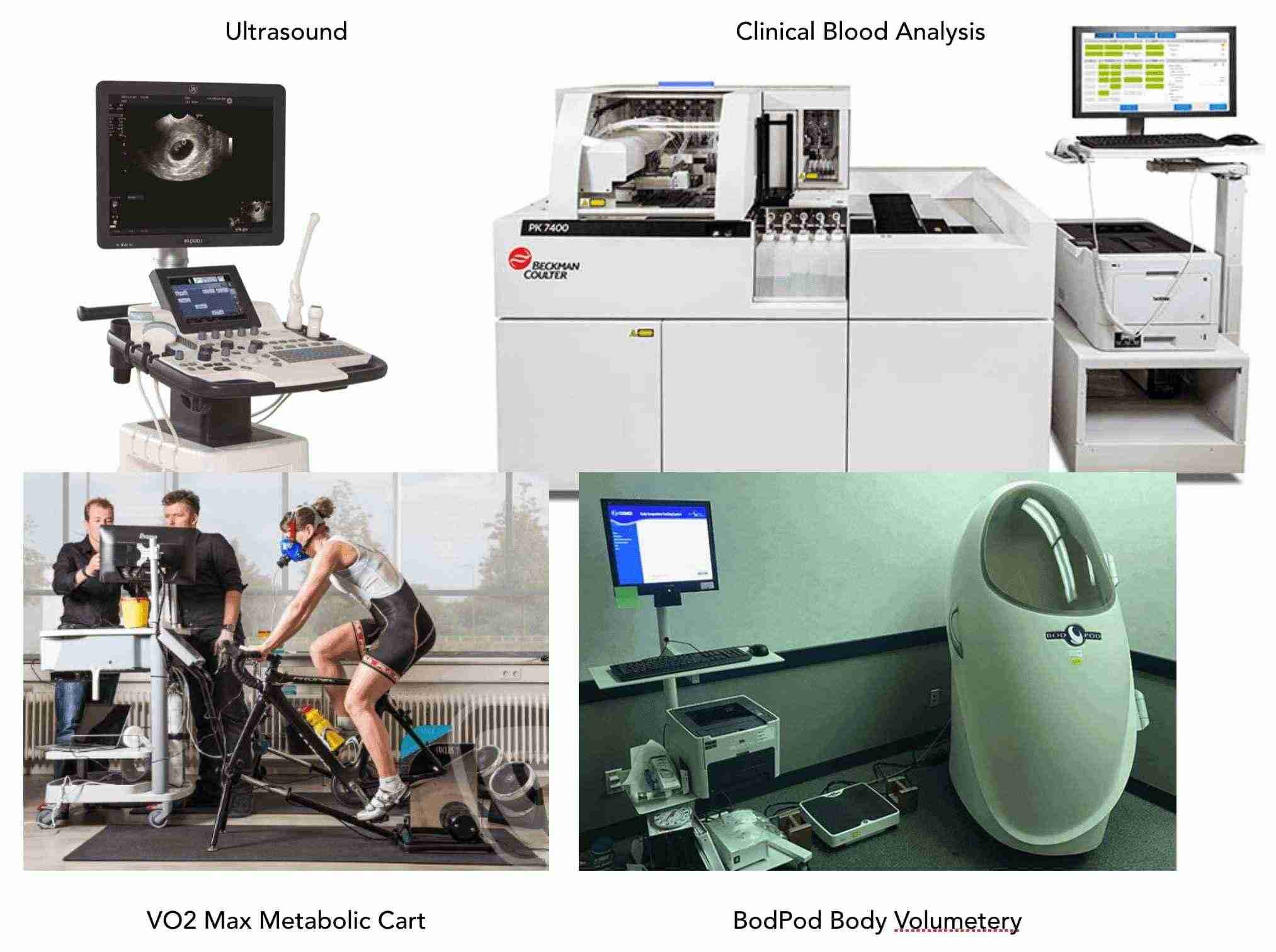,width=10cm}}
  \caption{\textbf{Clinical Hardware.} These machines are used as examples of reference values to compare with the accuracy of lower cost sensors.}
    \label{fig:clinicalhardware}
\end{figure}

\begin{figure}[H]
  \centering \centerline{\epsfig{figure=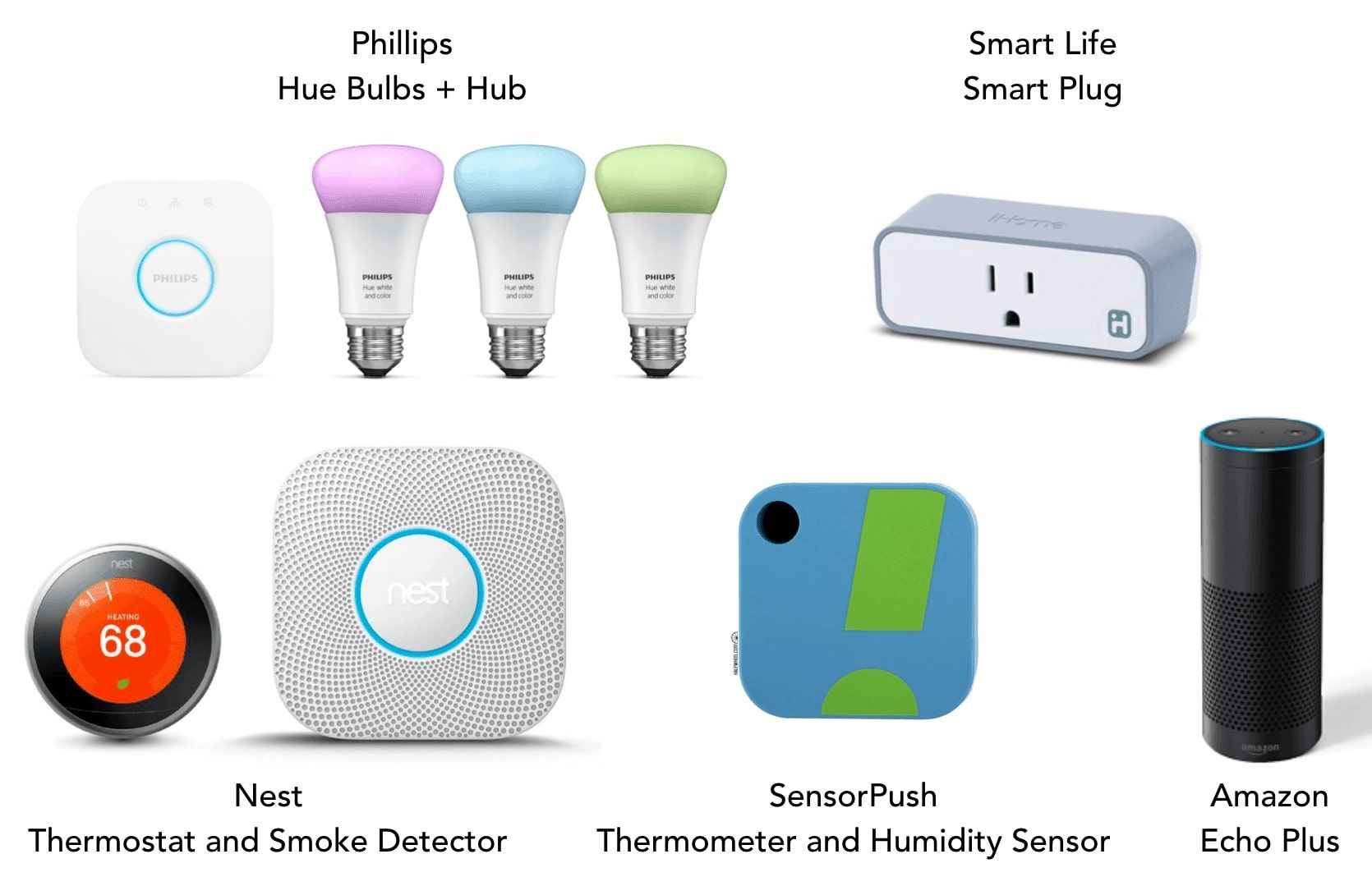,width=10cm}}
  \caption{\textbf{IoT Devices.} These sensors are used to monitor the local exposome of the individual including light, sound, temperature, humidity, and smoke.}
    \label{fig:iothardware}
\end{figure}

\begin{figure}[H]
  \centering \centerline{\epsfig{figure=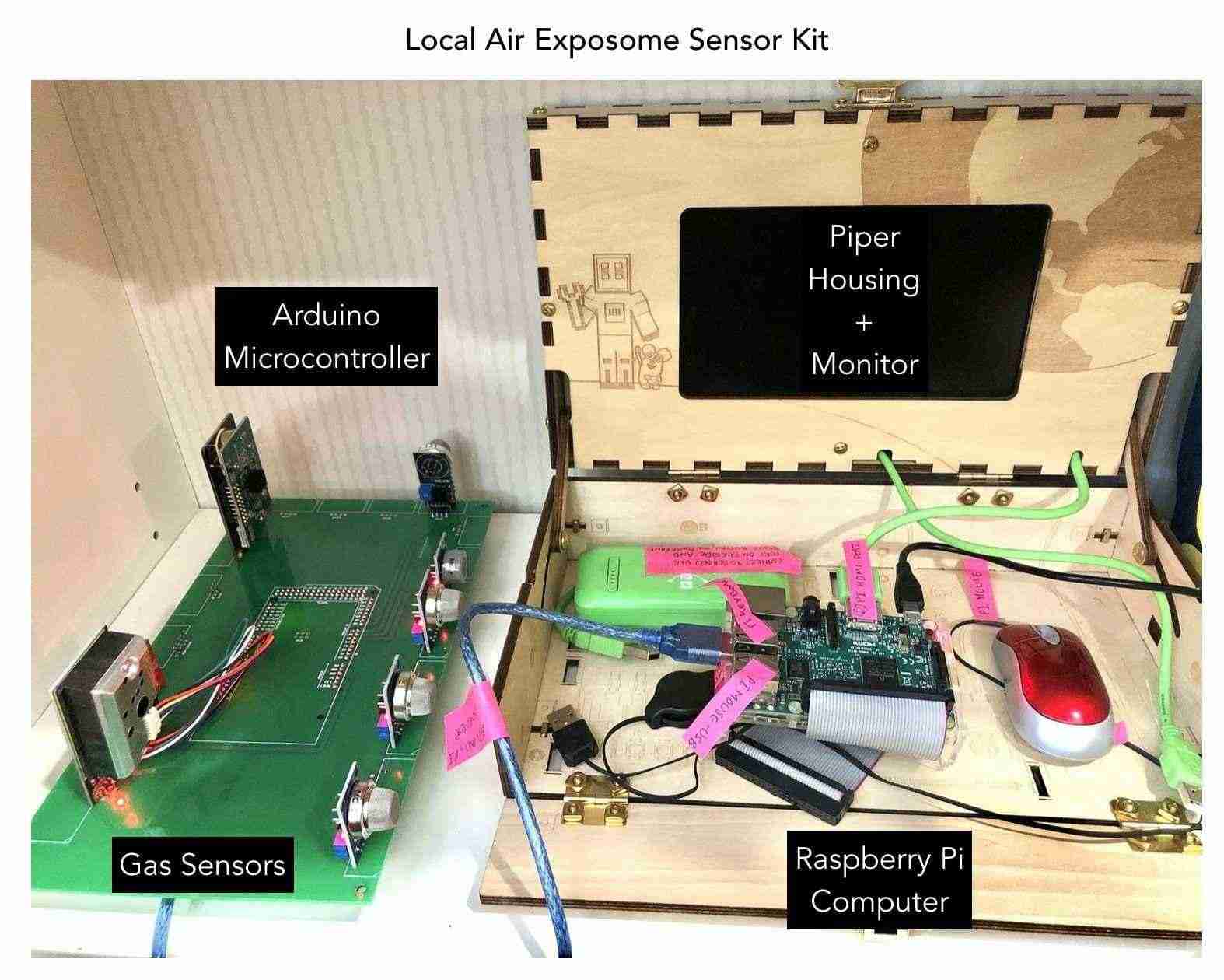,width=14cm}}
  \caption{\textbf{Custom Gas Sensor IoT Device.} This sensor IoT system was built with multiple gas sensors connected to an Arduino micro-controller that links to a Raspberry Pi computer. The computer sends the signal via a web Rest-API that allows for a minute by minute monitoring of the gases in the local environment.}
    \label{fig:pi}
\end{figure}

In this experiment, all these measurements are related to body surface measurements are captured with traditional analog tools, as shown in Figure \ref{fig:manualhardware}. Integration with costly clinical tools is also an important factor in using data for HSE. These types of tools can aid in ensuring the accuracy of lower-cost sensors and validate their use in the implementation. They can also serve as a ground truth reference from which low-cost sensors can track the change in the state. The four clinical tools that are used in this work to demonstrate this integration (Figure \ref{fig:clinicalhardware}). Three sets of hardware are used in sensing the environment of the individual. The first includes synchronizing the GPS location from the mobile phone and wearable with the nearest public sensors and data. The second involved a host of readily available IoT devices that monitor the home setting (Figure \ref{fig:iothardware}). The third includes a low-cost constructed gas sensor kit linked to a Raspberry Pi (Figure \ref{fig:pi}). An overview of these hardware components and the outputs they produce is in Table \ref{tab:data-streams}.

\begin{table}[hbt!]
\centering
\resizebox{\columnwidth}{!}{%
\begin{tabular}{@{}llll@{}}
\toprule
\multicolumn{1}{l|}{\textbf{Hardware}} & \multicolumn{1}{l|}{\textbf{Manufacturer}} & \textbf{Sensors} & \textbf{Datastreams} \\ \midrule
Cadence                & Garmin       & Accelerometer          & Cadence                    \\
Crankset               & Stages       & Accelerometer          & Torque                     \\
                       &              & Gyroscope              & Force                      \\
                       &              & Strain Gauge           & Power                      \\
                       &              & Thermometer            & Cadence                    \\
Chest Strap            & Garmin       & Electrical conductance & Heart Rate                 \\
Rowing Ergometer       & Concept2     & Strain Gauge           & Power                      \\
                       &              & Magnet                 & Cadence                    \\
Cycle Ergometer        & Elite        & Accelerometer          & Torque                     \\
                       &              & Gyroscope              & Force                      \\
                       &              & Strain Gauge           & Power                      \\
                       &              & Thermometer            & Cadence                    \\
Smart Watch (Fenix 5)  & Garmin       & Optical                & Heart Rate                 \\
                       &              & Accelerometer          & Cadence                    \\
                       &              & Gyroscope              & Movement                   \\
                       &              & Barometer              & Altitude                   \\
                       &              & Thermometer            & Temperature                \\
                       &              & GPS                    & Location                   \\
Measuring              & MyoTape      & Distance               & Body Measurements          \\
Goiniometer            & EMI          & Angle                  & Joint Measurements         \\
Plumb Bob              & Generic      & Gravity                & Body Alignment             \\
Spirometer             & AirMD        & Air Pitot              & FEV1 / FVC                 \\
Pulse Oximeter         & Hypoxico     & Optical                & O2 Saturation              \\
Breathalyzer           & BAC Track    & EtOH                   & Blood Alcohol              \\
Blood Glucose Monitor  & One Touch    & Glucose                & Glucose                    \\
Weight Scale           & Weight Gurus & Strain Gauge           & Weight                     \\
                       &              & Electrical Impedance   & Body Fat Percentage        \\
Thermometer            & Kinsa        & Thermometer            & Body Temperature           \\
Gas Analyzer           & Lumen        & CO2 / O2               & Respiratory Exchange       \\
Blood Pressure Machine & Balance      & Pressure               & Blood Pressure             \\
Thermostat             & Nest         & Thermometer            & Local Temperature          \\
Smoke Detector         & Nest         & CO2 / CO               & Gas / Smoke                \\
Thermometer            & SensorPush   & Thermometer            & Local Temperature          \\
                       &              & Humidity               & Humidity                   \\
Echo Plus              & Amazon       & Microphone             & Audio                      \\
Smart Plug             & SmartLife    & Electrical Usage       & Electrical Usage           \\ \bottomrule
\end{tabular}
}
\caption{\textbf{Hardware Sensors.}}
\label{tab:data-streams}
\end{table}

Software that was used as a sensor to track the user includes two varieties. The first variety of software sensors directly query the user for input. For the mental state of the individual, the open-source tracking mobile application (app) Nomie was used. The Sleep Cycle app was used to gather information about the mental state before falling asleep and asses the mood and perceived quality of sleep upon waking up. The second variety of software used as a sensor operates through background tracking that does not interact with the user. Google Timeline captures location history and the personicle of the individual. LifeCycle app is a secondary personicle capture. Garmin MoveIQ is a third source that captures background activity personicle. The RescueTime mobile app and desktop applications track screen time. Time Out software tracks desktop computer visual strain by engaging with the user for screen breaks. The software used to track the individual is summarized in Table \ref{tab:software}. Public datasets are the third source of ``tracking" we use about an individual. Based on the GPS location and timestamp, we match the location to the current nearest sensors for various environmental streams. These environmental streams are in Table \ref{tab:software}. We will discuss the details of any relevant sensing in the sections below. The Appendix contains further details on hardware, software, and public data sources used in this work.

\begin{table}[hbt!]
\centering
\begin{tabular}{@{}lllll@{}}
\toprule
\textbf{Category} & \textbf{Interaction} & \textbf{Method} & \textbf{Source} & \textbf{Data-Streams} \\ \midrule
Neurological & Active     & Mobile App     & Nomie           & Mood            \\
             &            &                &                 & Pain            \\
             &            &                &                 & Symptoms        \\
Neurological & Active     & Mobile App     & Sleep Cycle     & Mood            \\
             &            &                &                 & Rest Quality    \\
Personicle   & Background & Desktop+Mobile & Google Timeline & Location        \\
             &            &                &                 & Activities      \\
Personicle   & Background & Desktop+Mobile & LifeCycle       & Location        \\
             &            &                &                 & Activities      \\
Personicle   & Background & Wearable       & Garmin Move IQ  & Activity Type   \\
             &            &                &                 & Intensity       \\
Screenome    & Background & Desktop+Mobile & RescueTime      & Focus Level     \\
             &            &                &                 & Screen Timeline \\
Screenome    & Active     & Desktop        & Time Out        & Eye Strain      \\ \bottomrule
\end{tabular}
\caption{\textbf{Software Sensors.}Software sensor systems used to track the user's neurological state and lifestyle events.}
\label{tab:software}
\end{table}


\subsection{Events}
The scope of implementation around CRF determined the tools we choose for data collection regarding events that can power state estimation. The data collection methods focus on capturing the details of five categories regarding the subject's life events. These events were chosen because they have a well-understood impact on health state in large bodies of literature. They also compose a large portion of the daily life of individuals. We will describe these five categories in the next sections.

\subsubsection{Activity}
\begin{figure}[H]
  \centering \centerline{\epsfig{figure=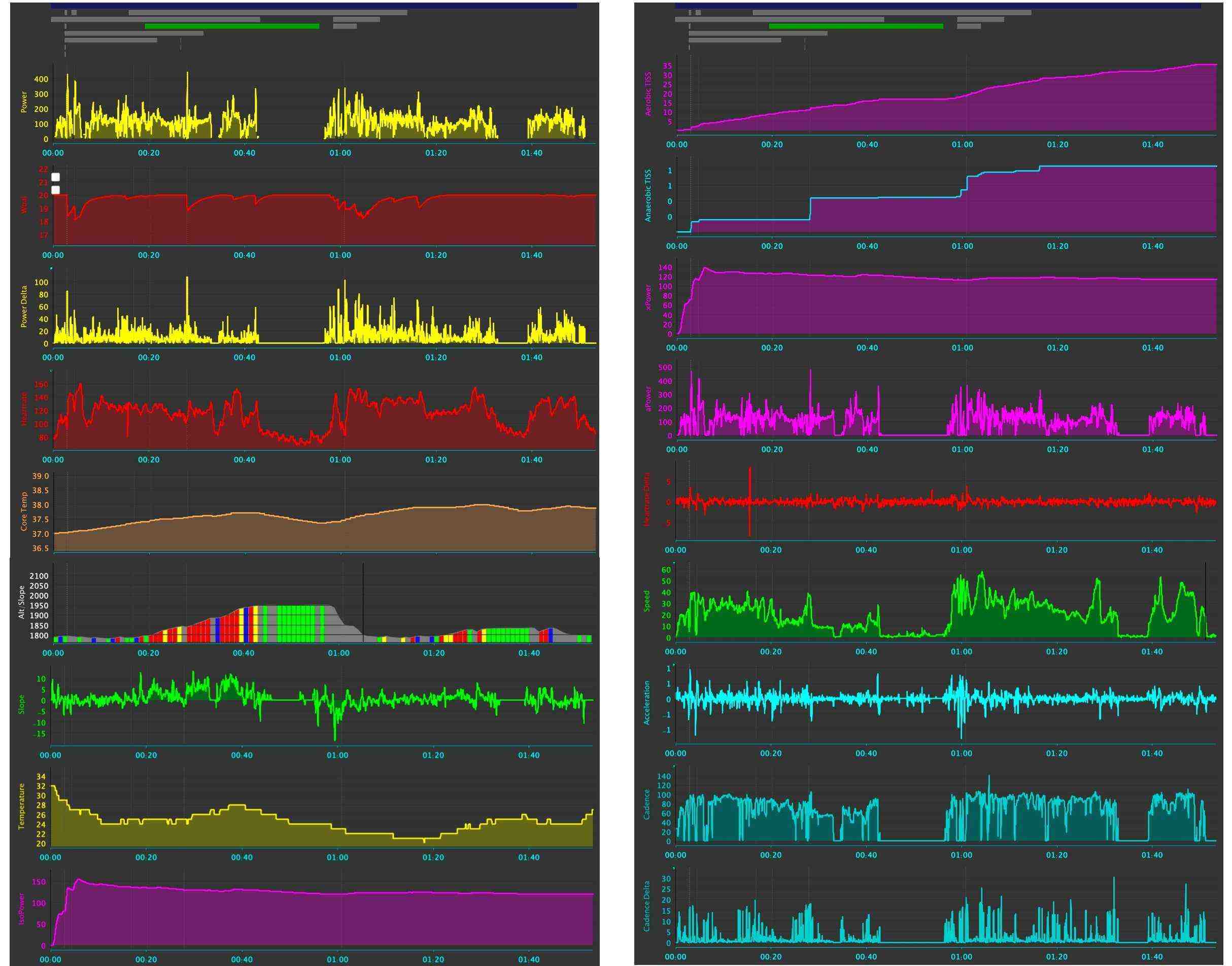,width=16cm}}
  \caption{\textbf{Exercise Event Understanding.} Single exercise event streams.}
    \label{fig:bikestream}
\end{figure}

\begin{figure}[H]
  \centering \centerline{\epsfig{figure=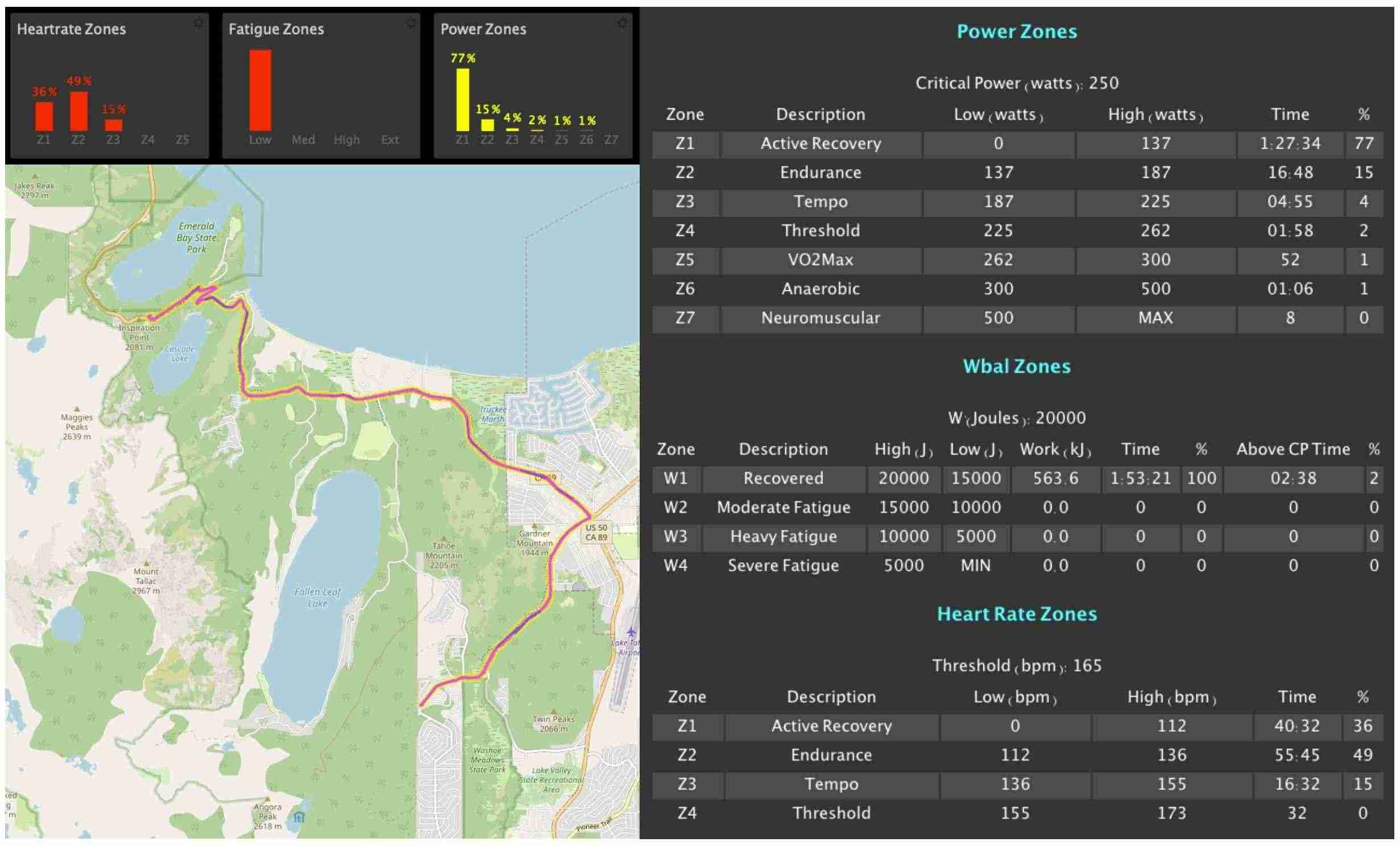,width=16cm}}
  \caption{\textbf{Exercise Event Understanding.} Single exercise event summary with location data, heart rate zone charts, and power data zone charts.}
    \label{fig:bikestream2}
\end{figure}

All instances where the user is conducting any movement, physical activity, and exercise are captured in detail. Movement and physical activity are defined as energy expended to move, occuring through any physical action taken. Exercise, in particular, is defined as an intentional activity designed around exerting effort to improve physical fitness. These terms have been well described in the literature \cite{Caspersen1985DearI}. Both activities at large and exercise events are logged in the user's personicle. The activities at large are recorded as events with time intervals of 2 minutes. Because exercise events provide a higher insight into the operational status of both the utility and biological states, they are an excellent event source to capture in detail for state estimation. Therefore exercise events are captured at 1-second intervals. Figure \ref{fig:bikestream} and Figure \ref{fig:bikestream2} show sample exercise event data-streams. Table \ref{tab:exercise-sensors} lists all the sensor streams collected during the activity.
 
\begin{table}[hbt!]
\centering
\begin{tabular}{@{}llll@{}}
\toprule
\textbf{Category} & \textbf{Source} & \textbf{Data-Stream} & \textbf{Units}         \\ \midrule
Utility           & Stages          & Power                & Watts                  \\
Biology           & Stages          & Energy Balance       & Wbal                   \\
Utility           & Stages          & Power Delta          & Watts/s                \\
Biology           & Garmin          & Heart Rate           & bpm                    \\
Biology           & Garmin          & Temperature          & Celsius                \\
Environment       & Garmin          & Altitude             & meters                 \\
Environment       & Garmin          & Slope                & percent                \\
Environment       & Garmin          & Temperature          & Celsius                \\
Utility           & Stages          & IsoPower             & Watts                  \\
Biology           & Stages + Garmin & Aerobic TSS          & TSS                    \\
Biology           & Stages + Garmin & Anaerobic TSS        & TSS                    \\
Utility           & Stages          & xPower               & Watts                  \\
Utility           & Stages          & aPower               & Watts                  \\
Biology           & Garmin          & Heart Rate Delta     & bpm                    \\
Utility           & Garmin          & Speed                & kph                    \\
Utility           & Garmin          & Acceleration         & m/s\textasciicircum{}2 \\
Biology           & Garmin          & Cadence              & rpm                    \\
Biology           & Garmin          & Cadence Delta        & rpm/s                  \\ \bottomrule
\end{tabular}
\caption{\textbf{Sensor Streams.} Sensor streams captured at 1 second intervals during exercise activity.}
\label{tab:exercise-sensors}
\end{table}

\subsubsection{Nutrition}
Epidemiological data suggest that food is the most influential event on human health, according to the Journal of the American Medical Association \cite{Murray2013TheFactors}. Nutrition includes all energy or nourishment obtained from food intake. All nutrition events include any food consumption that occurs. Nutrition events are tracked at each instance of food or beverage ingestion through images. Images are run through the Calorie Mama platform for object detection. Objects are then verified for labels and weights by the user. The app references the United States Department of Agriculture DB, extracting nutritional content of each item and then can aggregate the nutrition intake per meal. The nutrition data is sent to HealthKit on the user's mobile phone as a central data aggregator. Metadata of timestamps and location are also captured.

\subsubsection{Neurological}
Understanding the mental perception of each individual's state is a subjective component of a health state. Neurological events include stress, mood, and even one's screen time. \textit{Stress}, in the medical context, is a physical, mental, or emotional factor that causes psychological or physical strain to an individual. Stress also includes an individual's response to a change. One's \textit{mood} is a temporary state of feeling that the individual experiences, reflecting their current state of mind. Each digital interaction also gives insight into neurological events. The \textit{screenome} consists of all interactions with one's screen. The screenome reflects the use of digital media and screen time to give insight into how individuals are choosing to spend their time \cite{Reeves2020TimeProject}. Although viewed as subjective, there are clear physical impacts on human physiology, as described in Chapter 2. Therefore, the mental state of the individual is in both subjective and objective formats. Stress and mood are captured subjectively through a direct query to the user, as given in Table \ref{tab:software}. Stress is captured objectively with heart rate variability (HRV) through the Garmin Fenix 5 smartwatch, as HRV is a strong indicator of stress. Lastly, the screenome is collected through background tracking software RescueTime and Time Out, also shown in Table \ref{tab:software}.

\subsubsection{Sleep}
This portion of data collection includes time spent in the prone position. Prone position activities refer to the body in a horizontal position. The individual may be awake or asleep while tracking this activity. Sleep is a condition in which the nervous system is relatively inactive, eyes are closed, and postural muscles relaxed. During sleep, consciousness is typically suspended. Sleep data is collected through objective and subjective sensors. The objective sensors include ambient noise, accelerometer-based movement, heart rate, HRV, temperature, humidity, and timestamps. Subjective measures include mood before sleep and perceived quality of sleep upon waking up.

\subsubsection{Environment}
\begin{figure}[H]
  \centering \centerline{\epsfig{figure=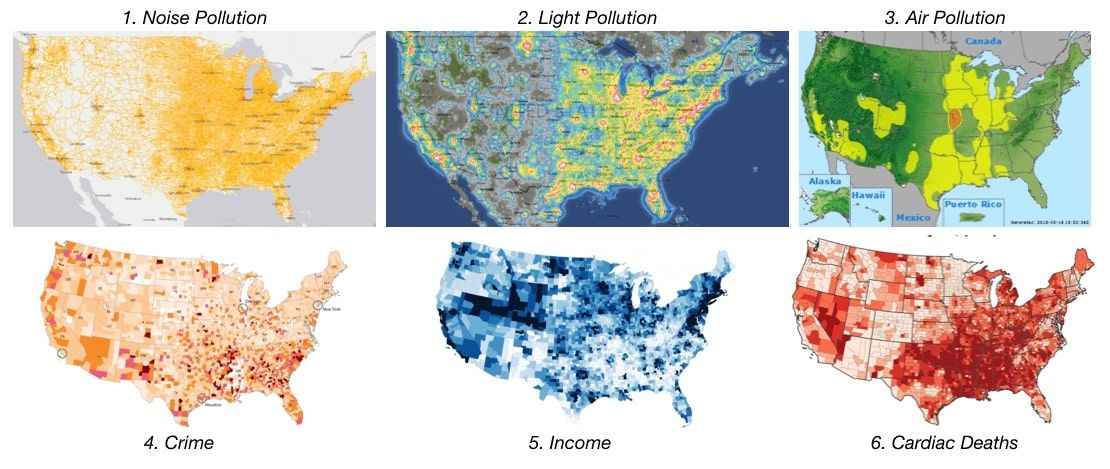,width=16cm}}
  \caption{\textbf{Public Geospatial Dataset Integration.} Many aspects of the environment, including ones that induce stress, can be elucidated from domain literature and publicly available GIS datasets. Related to pollution, we gather air, audio, and light as factors that impact the health state. Related to demographic information, we gather crime, income, and cardiac deaths to estimate socioeconomic stress \cite{Nag2018Cross-modalEstimation}.}
    \label{fig:maps}
\end{figure}

The external environment directly impacts various components of health, as described in Chapter 2. The external environment includes all factors outside the body that affect the individual; these can be local factors in one's home like lighting or broader environmental exposures such as the air. We use both public GIS data and devices that are precisely placed within the immediate vicinity of the user to collect environmental information. GIS data is collected through public sensors near the user's time and location. Public data about the user covers a large variety of data types, as shown in Figure \ref{fig:maps}. Hyperlocal systems gather data about the user's in-room environment. These streams come from the IoT devices and the Raspberry Pi. Table \ref{tab:exposome-sensors-table} lists the sensors we use to build the exposome stream for the individual.

\begin{table}[hbt!]
\centering
\begin{tabular}{@{}lll@{}}
\toprule
\textbf{Type} & \textbf{Source} & \textbf{Data-Stream}      \\ \midrule
GIS Matched   & EPA             & PM2.5                     \\
              & EPA             & PM10                      \\
              & EPA             & NO                        \\
              & EPA             & NO2                       \\
              & EPA             & NO3                       \\
              & EPA             & Ozone                     \\
              & EPA             & UV                        \\
              & Open Street Maps  & Altitude                  \\
              & In-the-Sky      & Light                     \\
Hyper Local   & Sensor Push      & Temperature               \\
              &                 & Humidity                  \\
              & Decibel-X        & Sound Level (dB)          \\
              & Sleep Cycle      & Sound Level (dB)          \\
              & Nest Protect    & Smoke                     \\
              & Nest Thermostat & Temperature               \\
              &                 & Humidity                  \\
              & Raspberry Pi    & Temperature               \\
              &                 & Pressure                  \\
              &                 & Altitude                  \\
              &                 & Humidity                   \\
              &                 & Carbon Monoxide (CO)      \\
              &                 & Smoke                     \\
              &                 & PM10                      \\
              &                 & PM2.5                     \\
              &                 & Large Particulate (LPG)   \\
              &                 & Volatile Compounds (tVOC) \\
              &                 & Ethanol / Fuel            \\
              &                 & Carbon Dioxide (CO2)      \\
              &                 & Ozone (O3)                \\
              &                 & Dust                      \\ \bottomrule
\end{tabular}
\caption{\textbf{Exposome Sensors.} There are two forms of exposome data collection within this experiment. The first is GIS matched data, which uses the user's location to retrieve the nearest public sensors as an estimate of their local environment. The second set of data uses hyper-local in-room sensors to capture data about the immediate vicinity of the user's environment.}
\label{tab:exposome-sensors-table}
\end{table}

We use this detailed collection of data for the event and stream DB system to power the HSE.

\subsection{Biology and Utility Direct Measurements}

\begin{figure}[H]
  \centering \centerline{\epsfig{figure=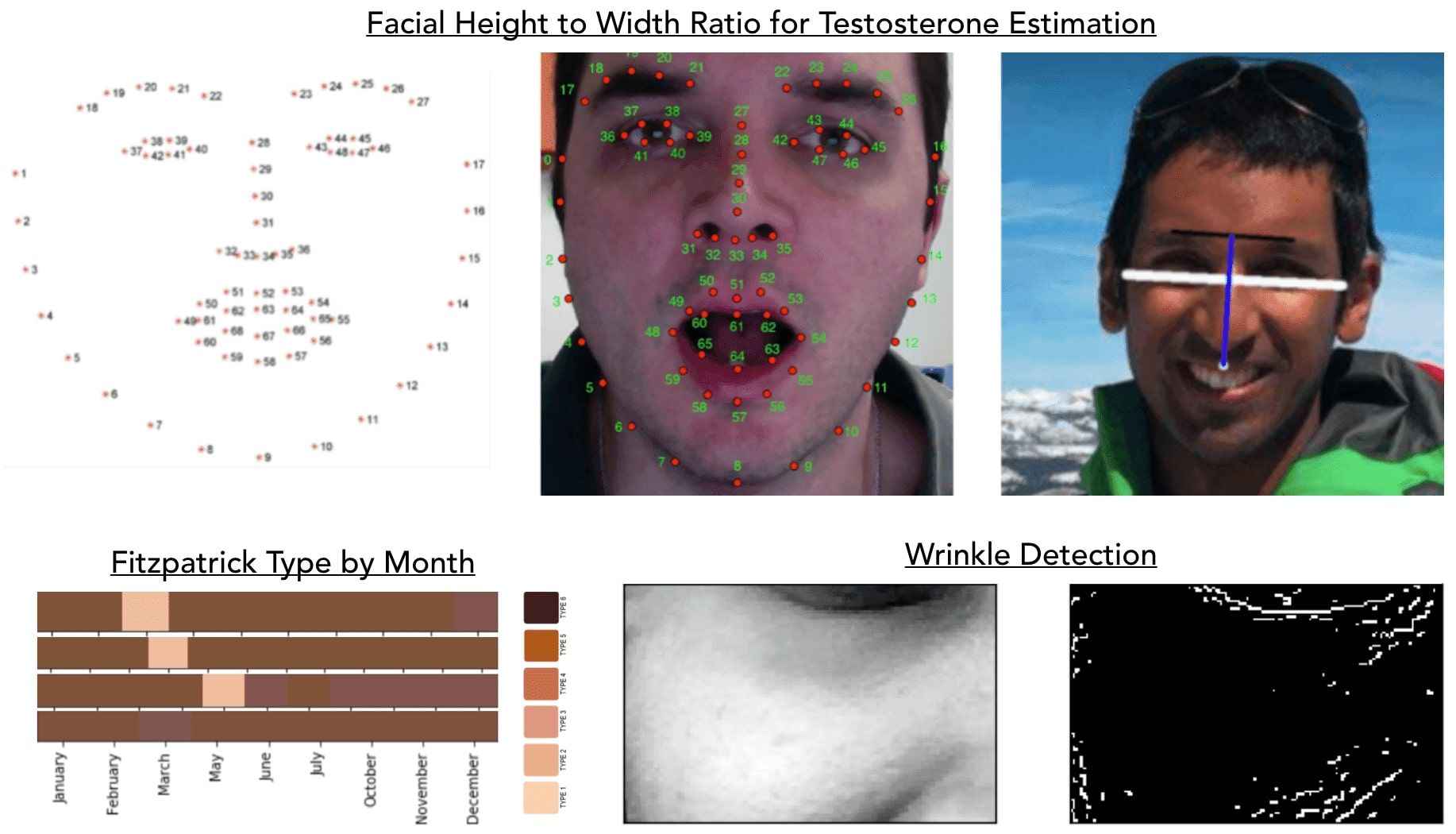,width=16cm}}
  \caption{\textbf{Computer Vision for Biological Measurements.} Images can give insight into a person's health state. In this figure, we use facial images for three purposes. First, we detect landmarks on the person's face to identify their height-to-width ratio, lending insight into their baseline testosterone production. Second, we use color detection to see how the user's skin color is changing based on each month. Skin color is classified using the Fitzpatrick type used routinely by dermatology physicians \cite{Roberts2009SkinNew,Sachdeva2009FitzpatrickDermatology}. We see on average that spring time, the user has lighter skin than the rest of the year. Third, we use edge detectors to characterize the wrinkling of the skin to understand collagen health.}
    \label{fig:face}
\end{figure}

Data capture about the biology of an individual is done concurrently with many of the events, as shown above. Data from smart devices and manual tools are shown in Figure \ref{fig:bodyhardware} and Figure \ref{fig:manualhardware}. The data capture of utility in the case of this experiment is the power output capability of the individual, also captured concurrently with activities. One additional method of biological data capture is with images of the user. Jaw shape and facial proportions are associated with testosterone levels. Here we use social media images of the individual to estimate the facial proportions and shape and derive a relative score of testosterone for the individual \cite{Nag2018Cross-modalEstimation}. Testosterone levels affect the cardiovascular health of individuals and the ability to recover from fatigue \cite{Barrett-Connor1995TestosteroneMen,Shores2014TestosteroneStudy,Axell2006ContinuousMice,Budgett1998FatigueSyndrome}. From a user's full social media data, images were filtered that had a high-quality resolution and direct frontal view of the subject. Open Computer Vision resources were used to pinpoint the face height-to-width ratio. An example of a photo analysis is shown in Figure \ref{fig:face}. The theme of this biological measurement technique from a computing perspective is that we can garner a plethora of information regarding the subject's health if we use modern image recognition techniques. We also investigated these images for skin health through pigmentation tracking and wrinkle detection. This investigation is shown here as part of the image-based health tracking theme, but it will not be used in further CRF estimations. The appendix contains resources used in data collection, analysis, and visualization in Table \ref{tab:hardware_software}.

\newpage
\section{HSE System Initialization}
\begin{wrapfigure}{l}{0.5\textwidth}
    \includegraphics[width=0.48\textwidth]{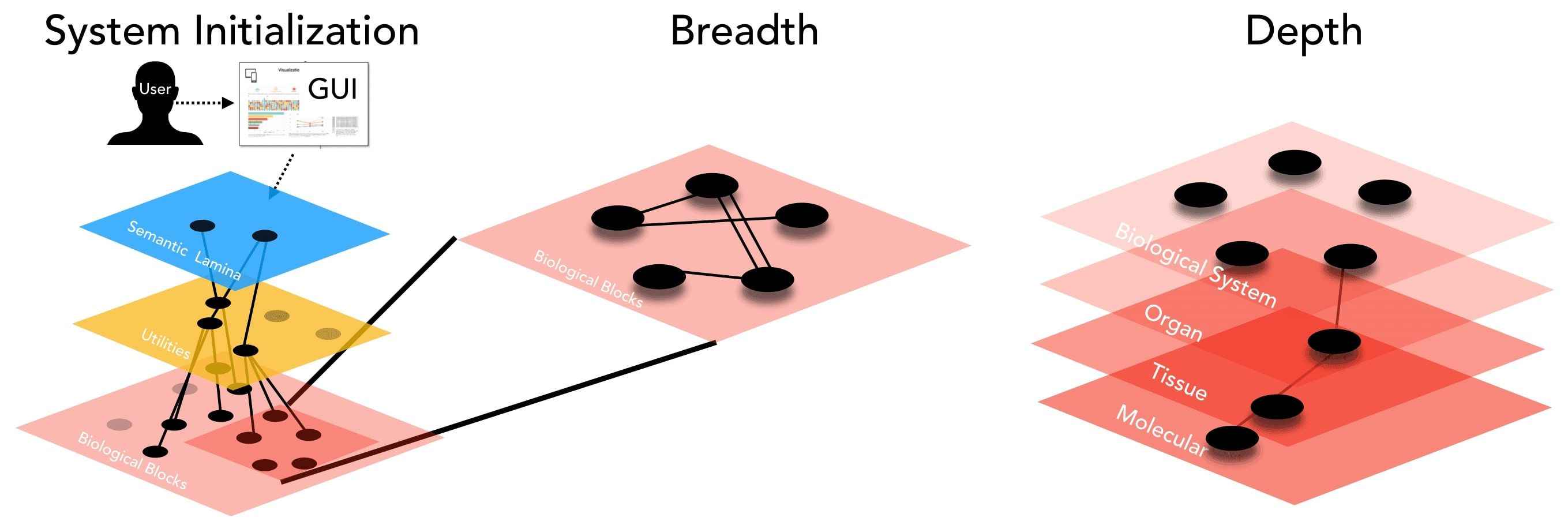}
      \label{fig:iHSEinit}
\end{wrapfigure} 

The precursors to the state estimation system require having the above data streams and background work. Next, we begin the process of implementing the system for the test subject. The first sequence of steps is initializing the system, as explained in Chapter 4. We run through this process to set up the GNB structure, created once the user specifies a domain or intent. Following this, we can populate the relevant layers of utility dimensions and biological layers. This section walks through all these steps for the test subject.

\subsection{Translating User Intent}
\begin{wrapfigure}{l}{0.35\textwidth}
    \includegraphics[width=0.3\textwidth]{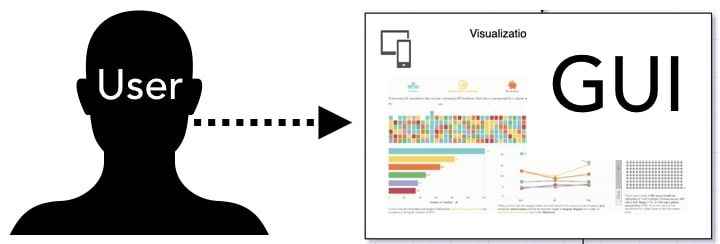}
      \label{fig:intent}
\end{wrapfigure}
\begin{figure}[H]
  \centering \centerline{\epsfig{figure=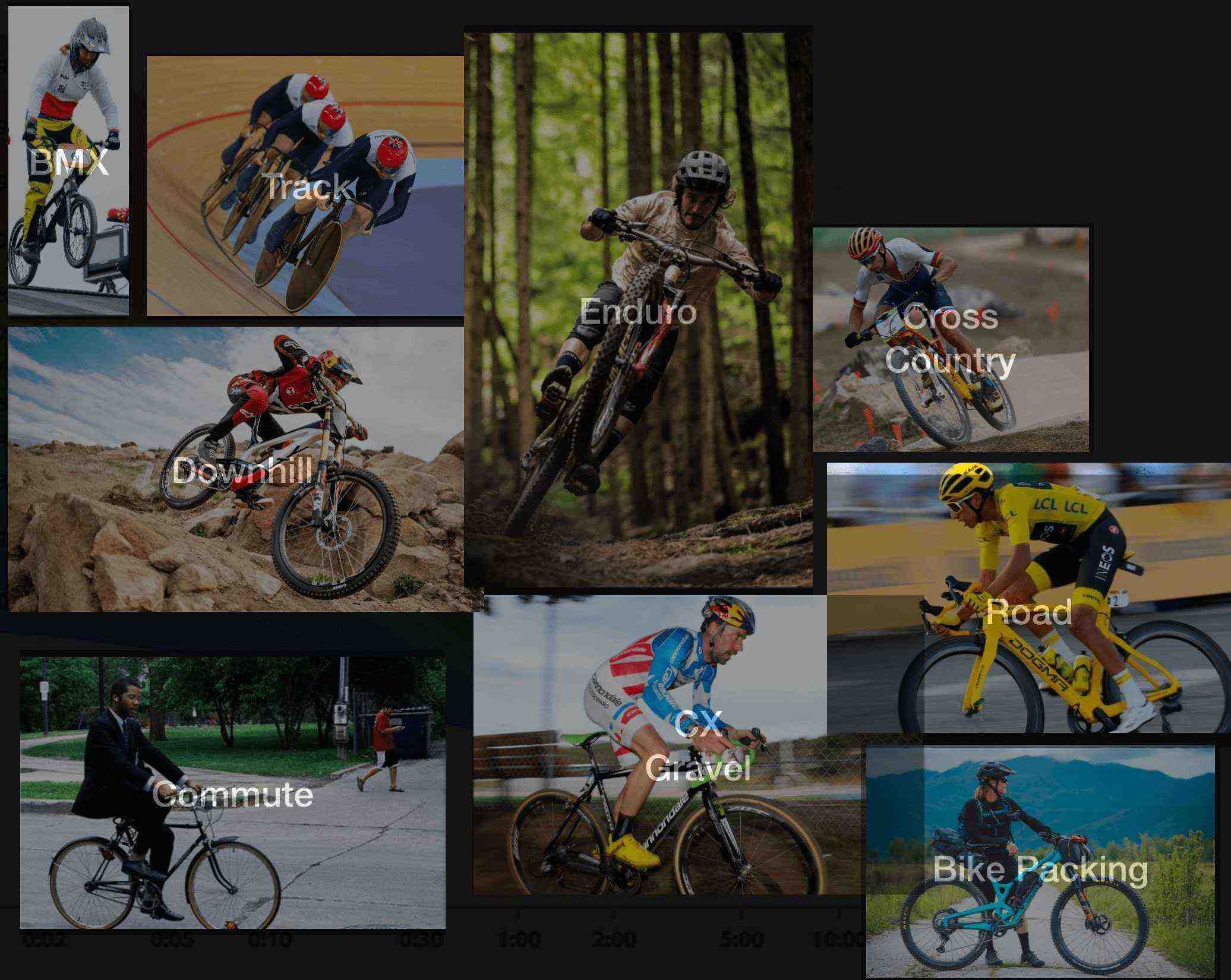,width=16cm}}
  \caption{\textbf{Cycling World Lamina.} This interface displays many different types of cycling of interest to the user. The user can select a specific type, zoom in and out, or view the whole lamina at once.}
    \label{fig:cycling}
\end{figure}

 
To start the HSE process for the subject, we begin by inquiring the user for their intent or domain of interest. The test subject's chosen interest in the experiment is in the realm of cycling. We display a lamina of cycling types to the user for visualization and interaction, as shown in Figure \ref{fig:cycling}. The user can see the relationships of different types of cycling on a particular health state space. The proximity of the various types of cycling shows the similarity of the health state required to participate. For example, BMX, track, and downhill styles of riding are all quite comparable on the upper left-hand side of Figure \ref{fig:cycling}.
Similarly, cross country, road, and gravel are also close together. Enduro riding sits in between these two clusters. The subject selects interest in the disciplines of enduro and gravel riding and wishes to see the health state in the context of this region-of-interest in the lamina. The next step for the system is to generate the utility dimensions for this lamina.

\subsection{Retrieval of Relevant Utility State Space}
\begin{wrapfigure}{l}{0.35\textwidth}
    \includegraphics[width=0.3\textwidth]{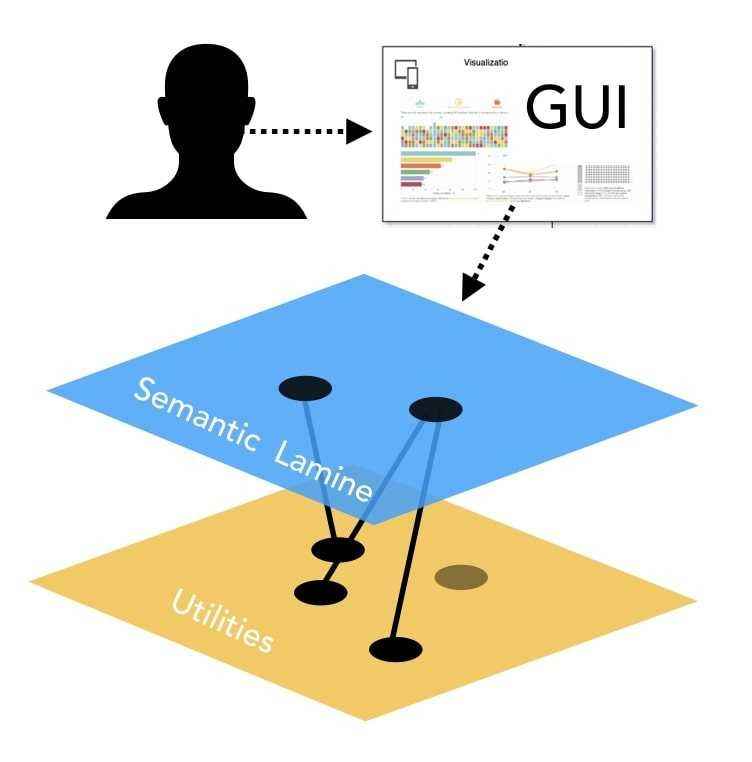}
      \label{fig:r-utility}
\end{wrapfigure}

\begin{figure}[H]
  \centering \centerline{\epsfig{figure=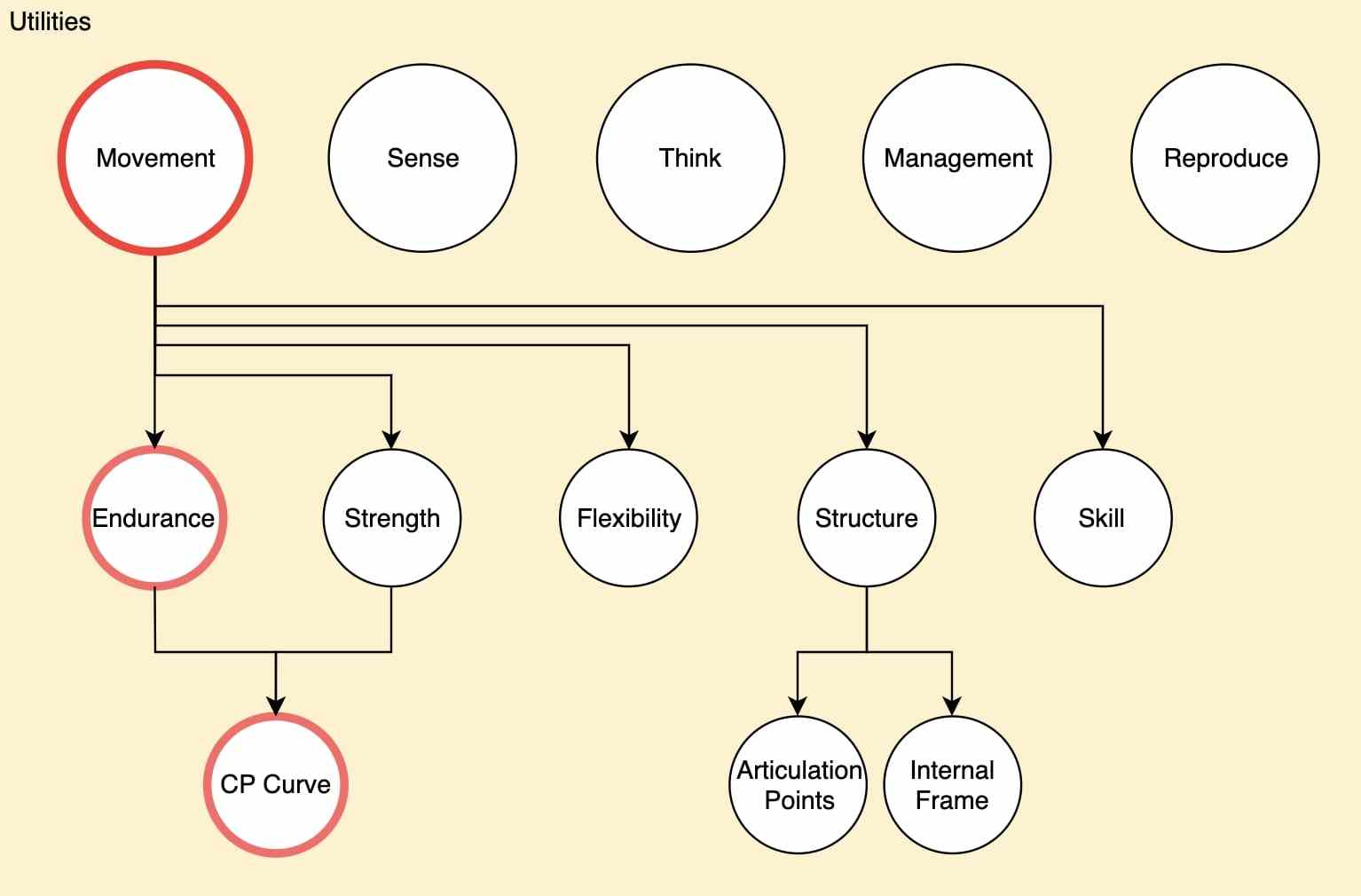,width=14cm}}
  \caption{\textbf{Utilities of Human Health.} In this work, we have defined five basic clusters of human utility. Given the selection of the user, we focus on the utility blocks that are connected to power production. In this work, we focus on the endurance capacity of critical power production. These nodes are highlighted in red.} 
    \label{fig:utilities}
\end{figure}

Given the intent of cycling by the user, the knowledge DB retrieves the dimensions of endurance, specifically the Critical Power (CP) curve, to display this specific lamina. The CP is the amount of power that can be applied maximally in a given period. It serves as the tolerance or biological threshold for the whole body in exercise (such as in cycling) for an individual \cite{Jones2017TheExercise,Poole2016CriticalPhysiology}. For example, a 5-minute CP is the maximum amount of power an individual can produce for 5 minutes. An untrained person can generally produce about 2 Watts per kilogram of body weight. For a 70kg person, this would be 140 Watts of power. A world-class athlete can produce around 7 Watts/kg, sustaining 490 Watts of power for a 70kg person. CP is usually shown as maximum power for each one-second interval of time from 1 second to several hours. Every living individual has a CP utility dimension value for producing power for 1 second, 2 seconds, 3 seconds...60 seconds (1 minute)...3600 seconds (1 hour). For measurement of these dimensions up to 5 hours, there are  18,000 utility state dimensions (5 hours x 60 minutes x 60 seconds). There are many other laminae related to cycling using utility dimensions of skill, sensing, or thinking abilities. We do not cover these in the dissertation for the sake of brevity. This selection of CP as part of the dimensions within the GNB of movement is shown in the utility state space in Figure \ref{fig:utilities}.

\begin{figure}[H]
  \centering \centerline{\epsfig{figure=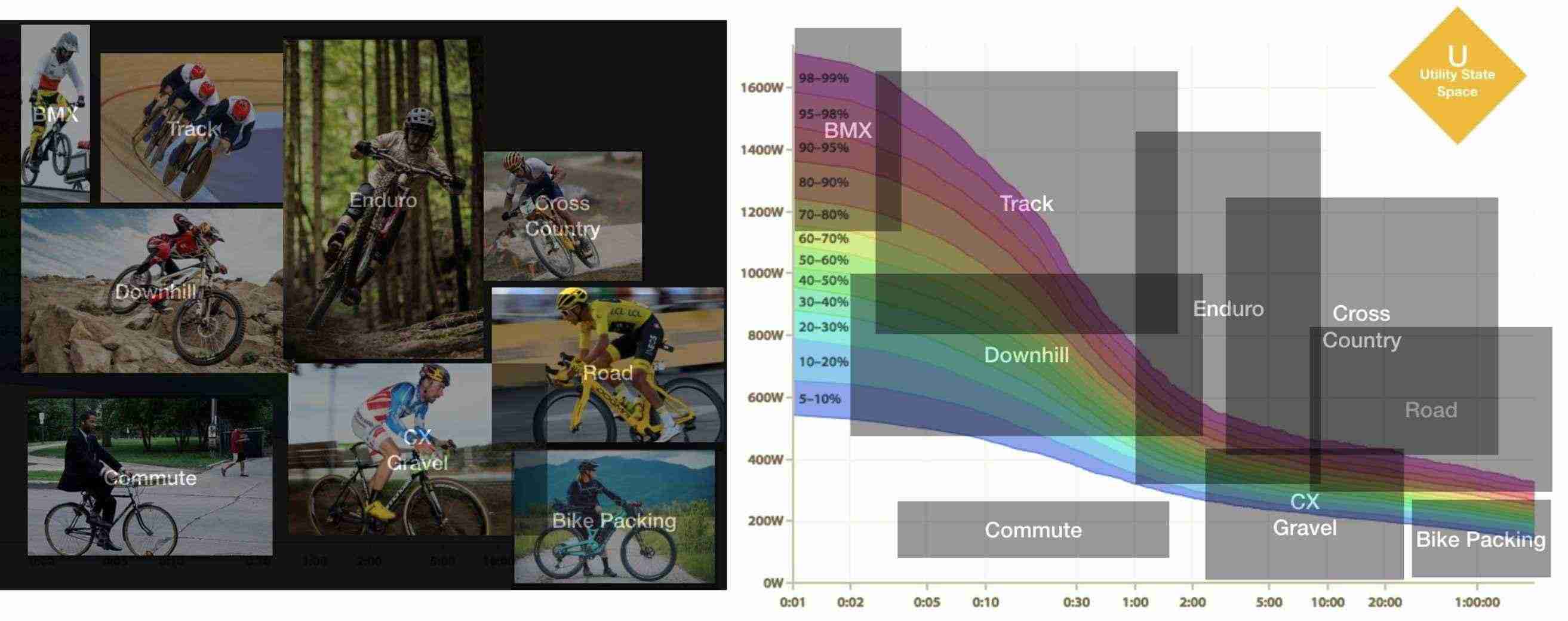,width=16cm}}
  \caption{\textbf{Translating Regions-of-Interest to Utility State Space.} On the left, we see all different types of cycling that a user may be interested in participating in a cycling lamina. On the right, these types of cycling are overlaid on the utility dimensions of power production versus duration. This state space of power given duration is the CP curve.}
    \label{fig:semantic}
\end{figure}

\subsection{Populating Biological Blocks}
\begin{wrapfigure}{l}{0.35\textwidth}
    \includegraphics[width=0.3\textwidth]{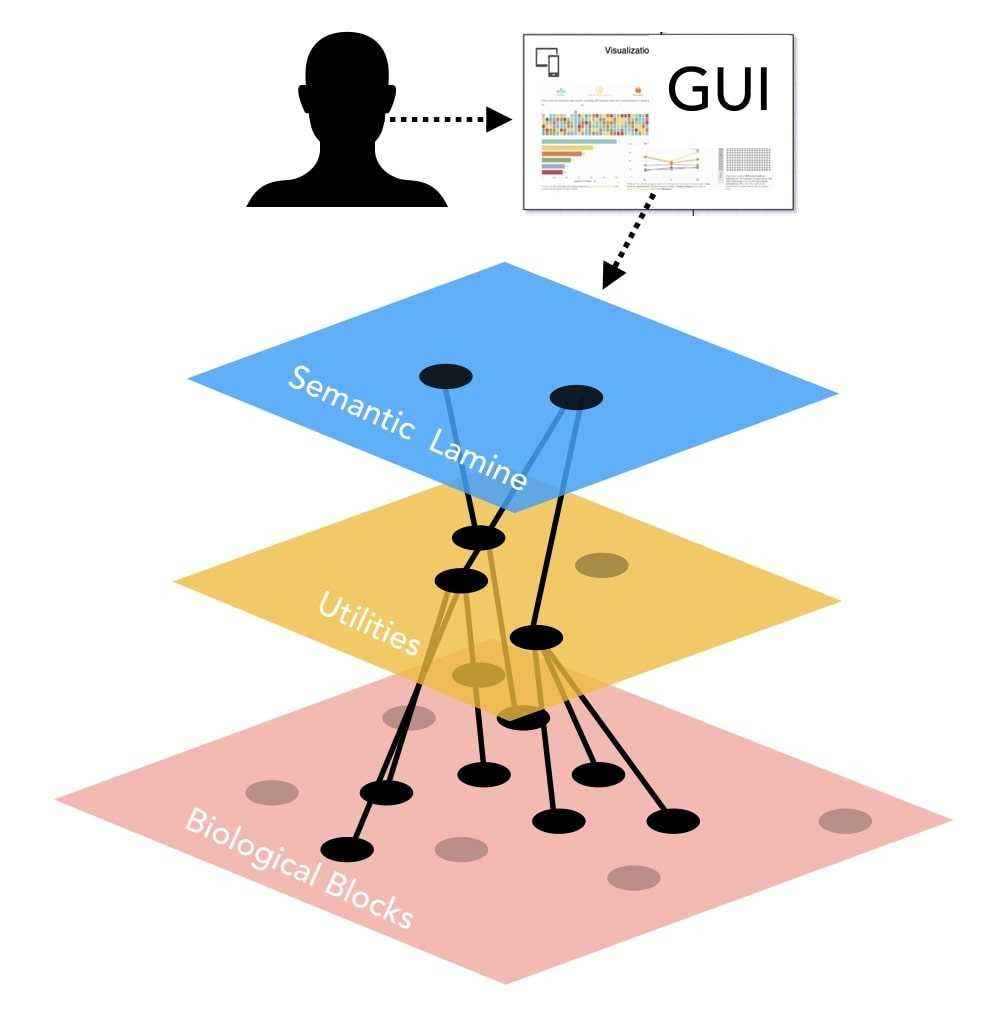}
      \label{fig:popbio}
\end{wrapfigure} 
Once we have retrieved the relevant utility dimensions, such as CP, we need to retrieve next what nodes in the biological block determine the value of these dimensions. There are two challenges to implement this. First, there is a breadth of biology that is needed to explain a particular utility dimension. For example, if the utility is walking ability for movement, it is insufficient only to understand the isolated biology of the toes or knees. The whole biomechanics and bioenergetics of walking must be populated into the system for a proper understanding of walking. A superficial understanding is a good start, but understanding the depth of details about the system will give a better picture of the factors that are changing the utility state. The next two sections cover these two challenges of breadth and depth in the biological GNBs.

\subsubsection{Breadth of Biology}
\begin{wrapfigure}{l}{0.35\textwidth}
    \includegraphics[width=0.3\textwidth]{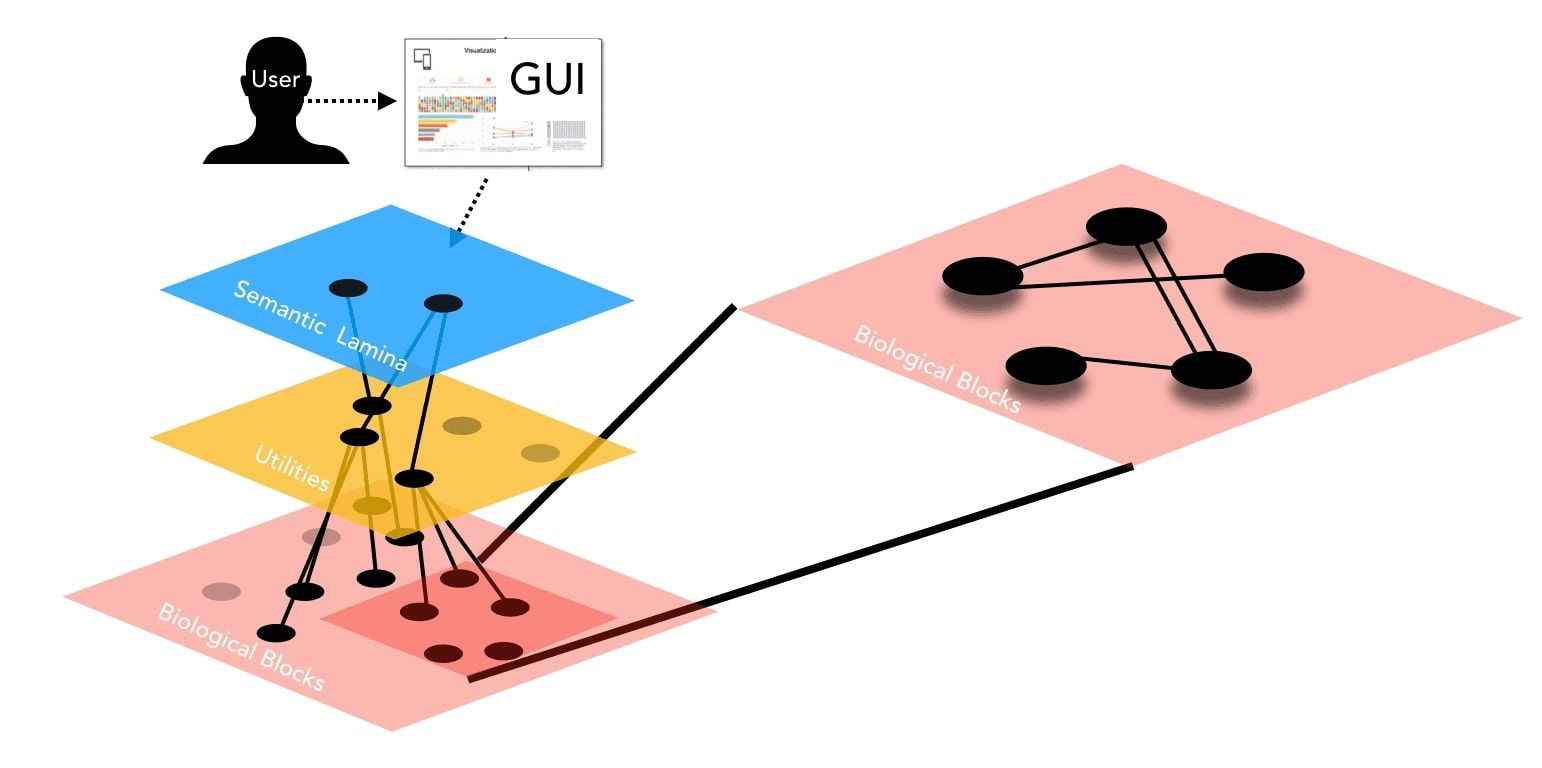}
      \label{fig:breadth}
\end{wrapfigure}
\begin{figure}[H]
  \centering \centerline{\epsfig{figure=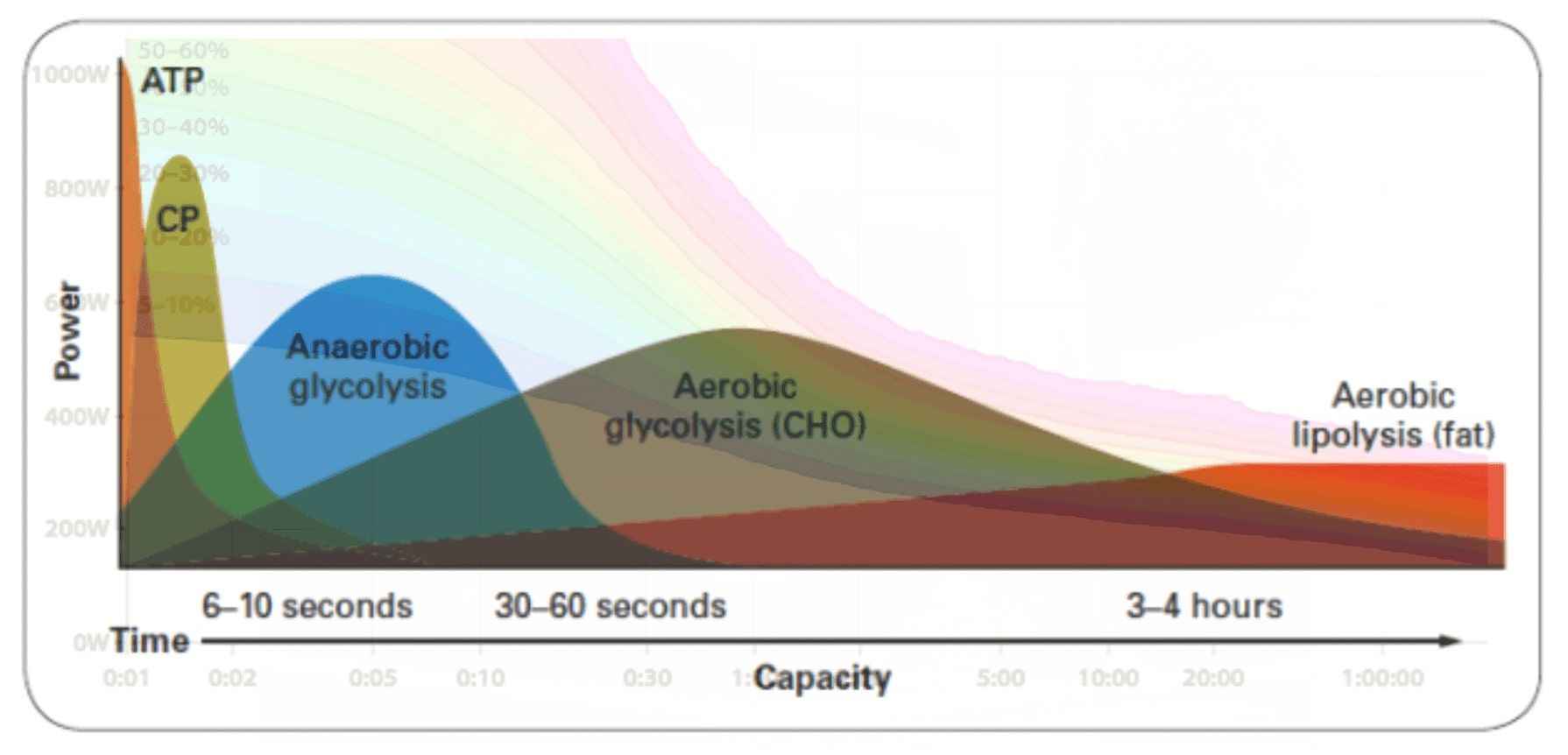,width=16cm}}
  \caption{\textbf{Populating Biological Blocks from Utility Blocks.} The CP utility state space (background) exists through 4 main biological systems (foreground). The first is the ATP-CP system (orange and yellow), the second is the anaerobic glycolysis (blue), the third is aerobic glycolysis (gray), and the fourth is lipolysis (red).}
    \label{fig:bioenergetics}
\end{figure}

 
To capture the parent nodes that are directly related to the utility dimensions of CP, we reference the pertinent domain knowledge of human bioenergetics \cite{Brooks1996Exercise2}. Bioenergetics focuses on how cells transform energy in cells, tissues, and organisms, which may be done by storing, transferring or consuming energy \cite{Nature.com2020BioenergeticsNature}. Depending on the duration of the energy need, there are four significant systems the human body uses to move. These four systems of interest are in Figure \ref{fig:bioenergetics} overlaid on top of the utility dimensions. The dimensional boundaries of enduro and gravel riding (the user's selected region-of-interest), the utility dimensions to focus on are from the 100 second CP value to the 1,200 second CP value (20 minutes). The biological energy system connected to these utility dimensions is aerobic glycolysis, as shown in Figure \ref{fig:bioenergetics}. Aerobic glycolysis ability is synchronous with CRF and measured in units of VO2max (mL O2 / minute / kg).

\begin{figure}[H]
  \centering \centerline{\epsfig{figure=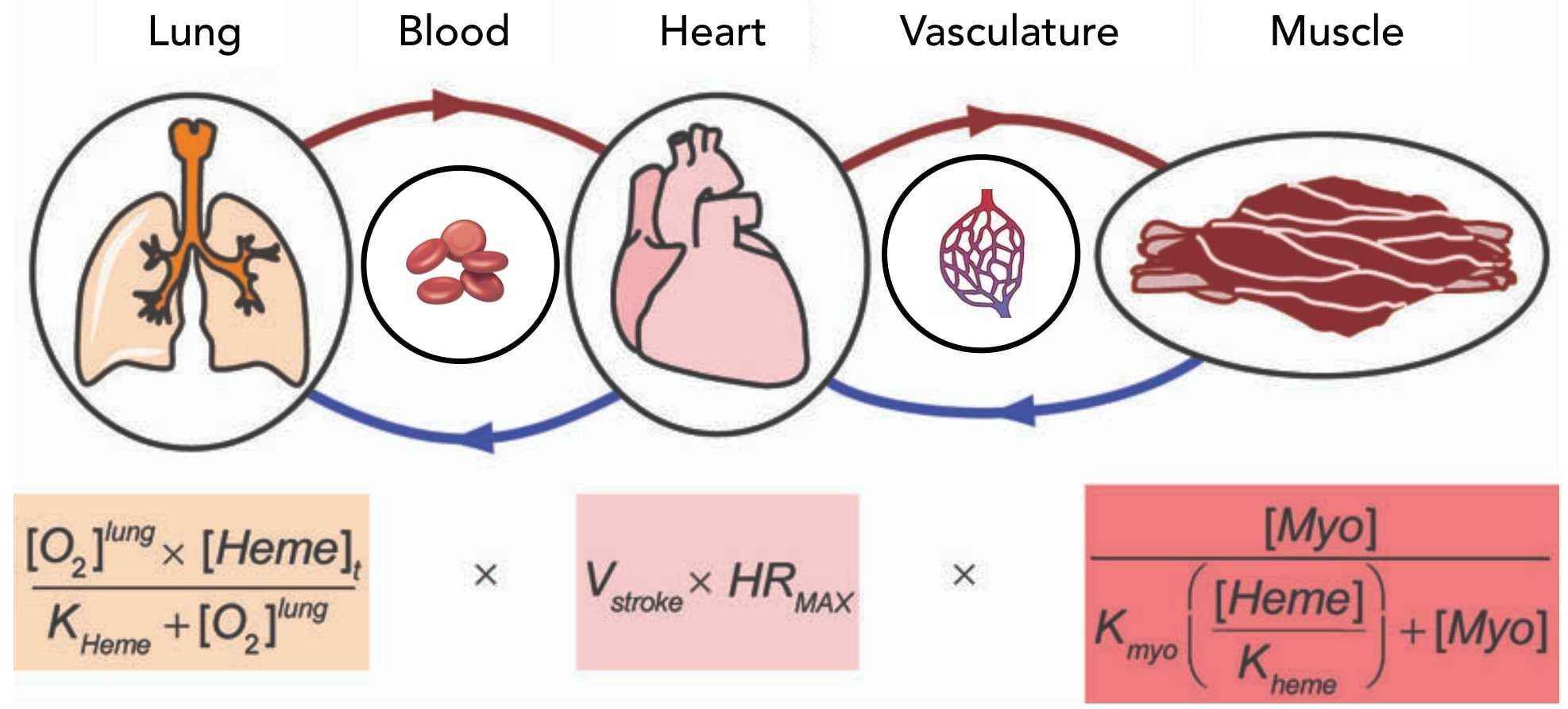,width=14cm}}
  \caption{\textbf{Aerobic Glycolysis Domain Knowledge.} This figure shows the flow of oxygen (O2) from the air to the muscles used in glucose oxidation, also illustrating how the nodes are connected in the GNB. The processes are well-characterized through the scientific literature, and in some instances, can be described mathematically. These processes are captured in the edges of the GNB.}
    \label{fig:o2flow}
\end{figure}

Given this target system, we populate the range of involved biological nodes at the gross organ level. The organ systems involved in this process and the edges between them are given in Figure \ref{fig:o2flow}.
Each organ has a set of major global attributes describing its microservice output to the system. The microservice edge relationships are captured in the equations in Figure \ref{fig:o2flow}. The global attributes of these organ nodes are in Table \ref{tab:node-attribute}. Next, we dive into deeper layers of biology by expanding these organ level nodes.

\begin{table}[hbt!]
\centering
\resizebox{\textwidth}{!}{%
\begin{tabular}{@{}ll@{}}
\toprule
Organ Node  & Global Attribute                                           \\ \midrule
Lungs       & O2 diffusion, CO2 diffusion                                \\
Blood       & O2 carrying capacity, heat capacity (through fluid volume) \\
Heart       & Cardiac Output                                             \\
Vasculature & O2 diffusion, CO2 diffusion, density                       \\
Muscles     & O2 extraction, rate of ATP synthesis, myofibril force      \\ \bottomrule
\end{tabular}%
}
\caption{\textbf{Organ Global Attributes.} Global attributes of major organ system GNBs}
\label{tab:node-attribute}
\end{table}

\newpage
\subsubsection{Depth of Biology}
\begin{wrapfigure}{l}{0.35\textwidth}
    \includegraphics[width=0.3\textwidth]{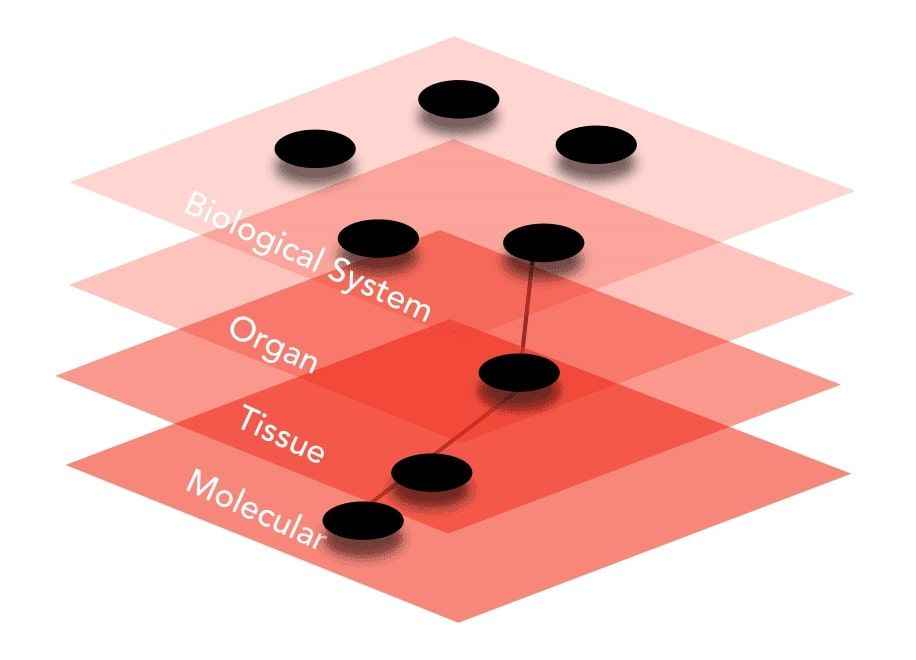}
      \label{fig:depth}
\end{wrapfigure}
\begin{figure}[H]
  \centering \centerline{\epsfig{figure=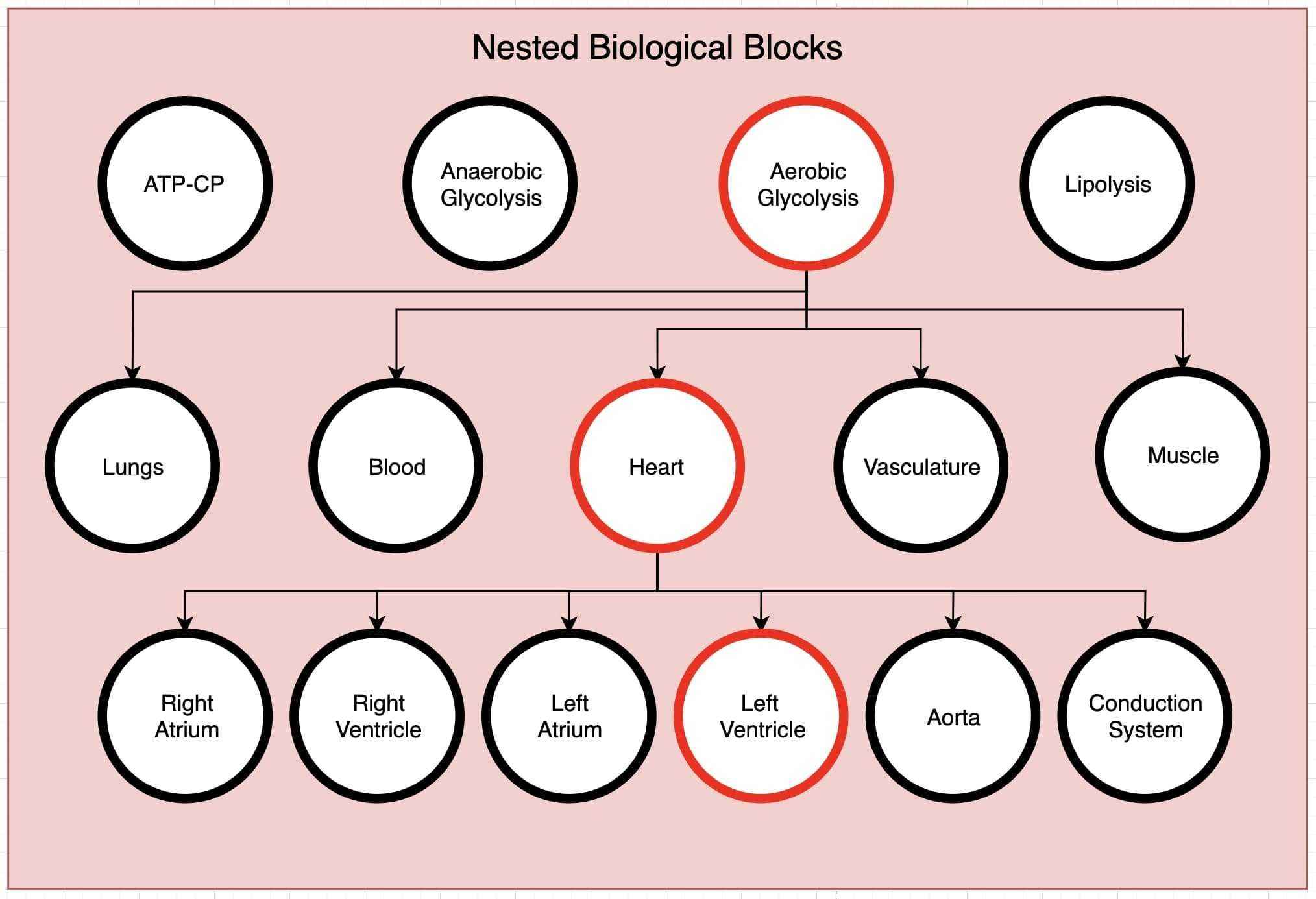,width=14cm}}
  \caption{\textbf{Nested Graph Network Blocks within Aerobic Glycolysis.} There are many biological components nested within aerobic glycolysis. We focus on the details of gene expression inside the left ventricle in this experiment set. These nodes are highlighted in red.}
    \label{fig:nestedcardioblocks}
\end{figure}

 
To face the challenge of biological depth, we build the GNB with deeper layers of biology. We will address four layers of depth in the experimental model. The first and second layers are the system layer and organ layer, as described in the broad sense above. We then dive into the tissue level and the molecular cell biology level in further depth. For the cardiac (heart) system, we see there are several key tissues that comprise the heart. The left ventricle is the tissue chamber that is responsible for cardiac output, which is the key attribute of interest for the utility dimensions at hand. This nesting of microservices in the GNB for biological depth is shown in Figure \ref{fig:nestedcardioblocks}.

\begin{figure}[H]
  \centering \centerline{\epsfig{figure=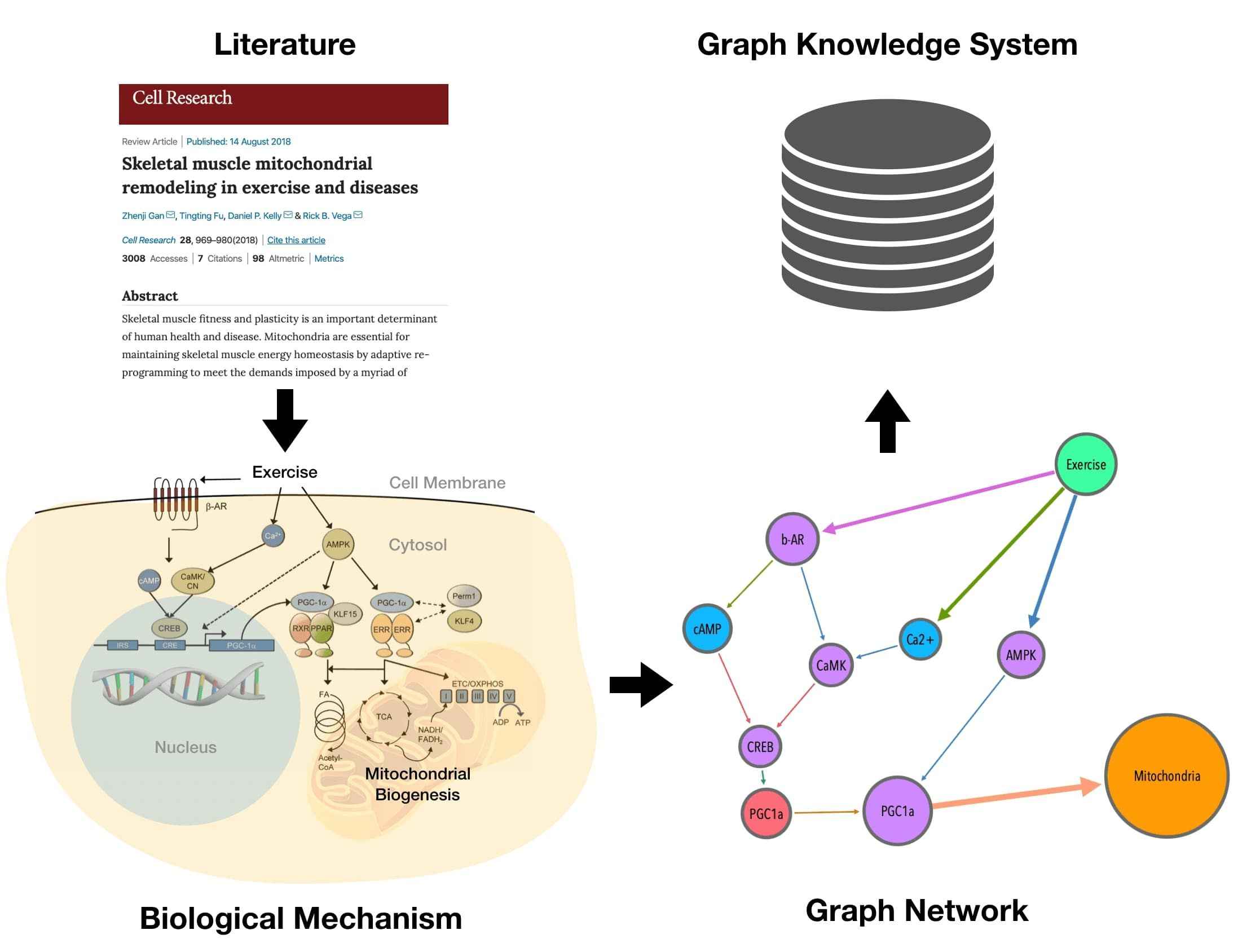,width=16cm}}
  \caption{\textbf{Translating Domain Knowledge to Graphs.} Mitochondrial biogenesis pathways are taken from literature to populate the molecular GNB layers of the HSE system for our test subject. This GNB is stored as a node within the muscle tissue nodes.}
    \label{fig:k2gmito}
\end{figure}

\begin{figure}[H]
  \centering \centerline{\epsfig{figure=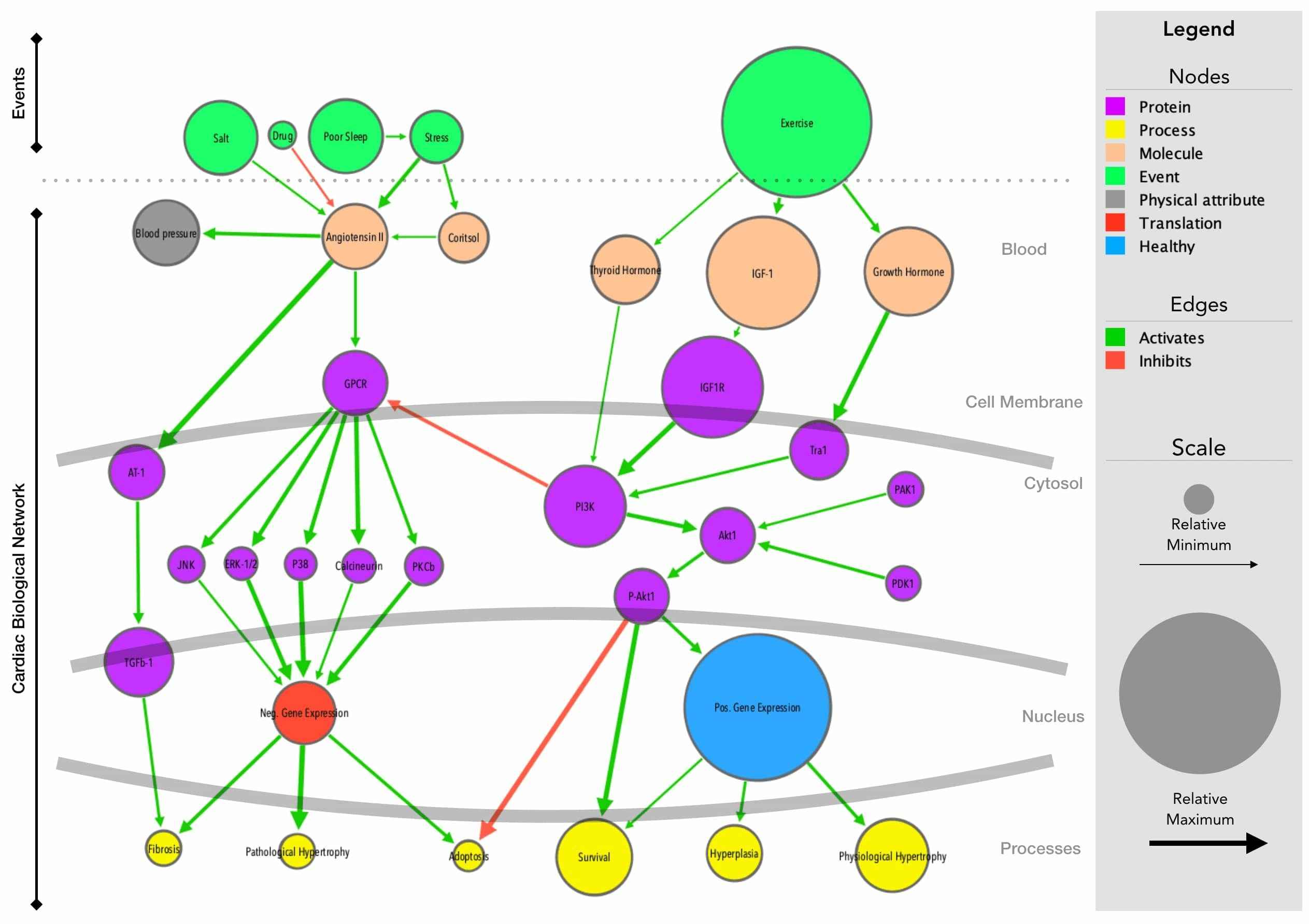,width=16cm}}
  \caption{\textbf{Cardiovascular Graph Network Block.} This network shows how events impact the cardiovascular state down to the gene expression level. The graph network of biology is taken from compiling open-source data from graph knowledge database Cytoscape and other literature. Consolidation of this into a unified graph was done by the author. Green arrows denote an activating effect on the target node. Red arrows denote an inhibitory effect of the source node to the target node.}
    \label{fig:cardiacnetwork}
\end{figure}

Within the left ventricle, there is a highly complex network of molecules that signal changes in the tissue. These changes include growth or degradation of tissue and cellular proteins through changing gene expression. Events are the primary input that alters gene expression. The understanding of these networks is usually stored in natural language within the body of traditional literature. Recently, graph DB systems that store domain knowledge in cellular signaling have also started flourishing. Cytoscape is the most popular of this graph-based knowledge and research systems \cite{Otasek2019CytoscapeAnalysis}. Harnessing the knowledge in these systems, along with traditional literature, helps to populate the molecular networks for the tissue of interest. Figure \ref{fig:k2gmito} describes the process by which mitochondrial biogenesis networks for this experimental GNB is populated from public domain knowledge. Domain knowledge of how events impact this network are also curated from traditional literature to build this knowledge graph. The process of using public graph networks of biology and traditional literature is also shown for cardiac tissue in Figure \ref{fig:cardiacnetwork}. The biological network for cardiac tissue is sourced from Cytoscape, and the event relationships to these nodes are sourced from traditional literature.

These knowledge graphs at the molecular level populate the deepest level of the personal GNB structure in this experiment. Now that we have the HSE GNB system built at all the desired levels of abstraction, we can plug incoming data streams to see how they are changing the state.

\begin{figure}[H]
  \centering \centerline{\epsfig{figure=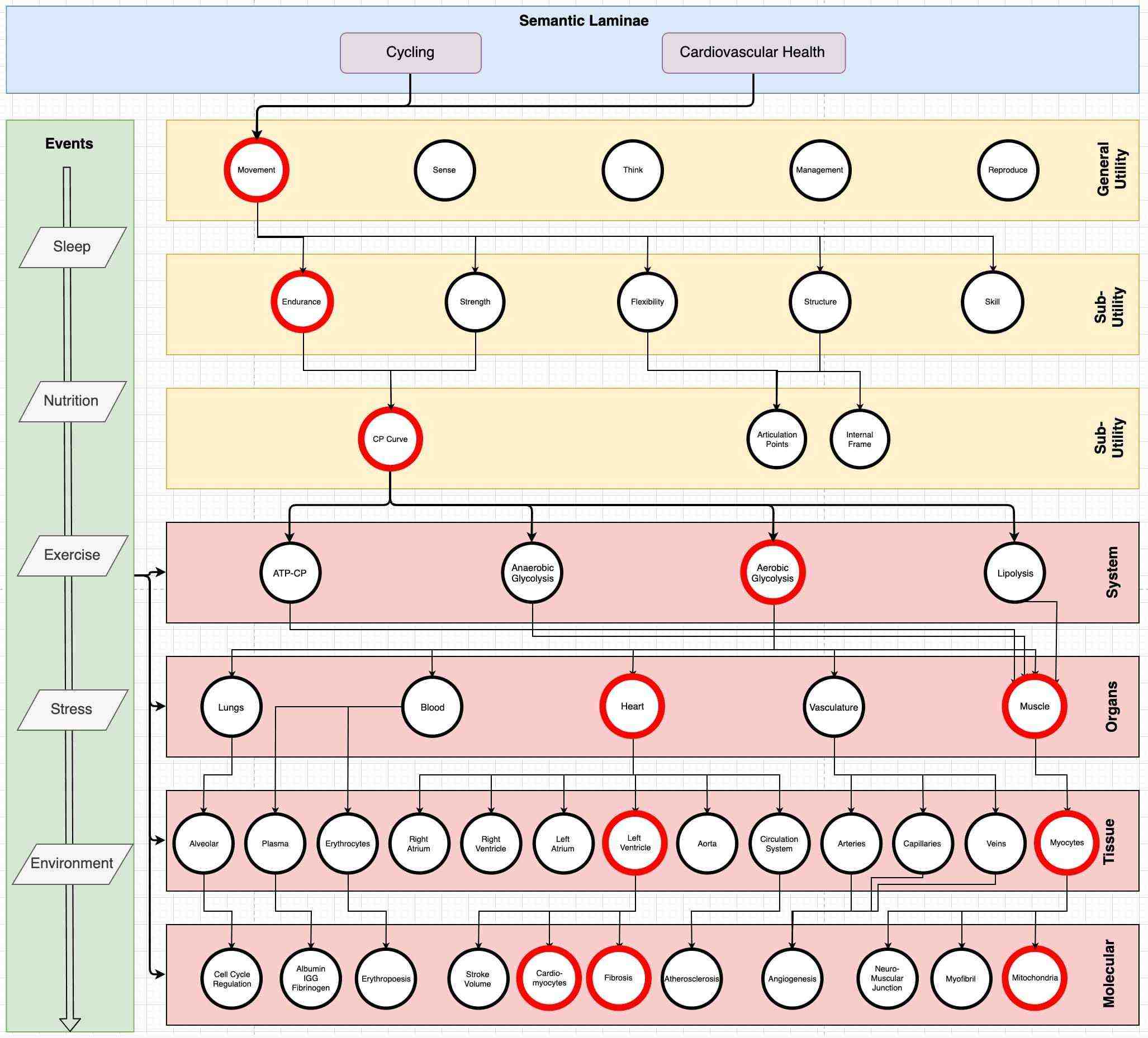,width=16cm}}
  \caption{\textbf{Graph Network Block Structure Summary.} The total overview of the 4 main system components instantiated for the test user. Red circles denote areas of focus for the experiments carried out in the next chapter.}
    \label{fig:gnbstructure}
\end{figure}

\newpage
\section{Health State Update}
\begin{wrapfigure}{l}{0.5\textwidth}
    \includegraphics[width=0.48\textwidth]{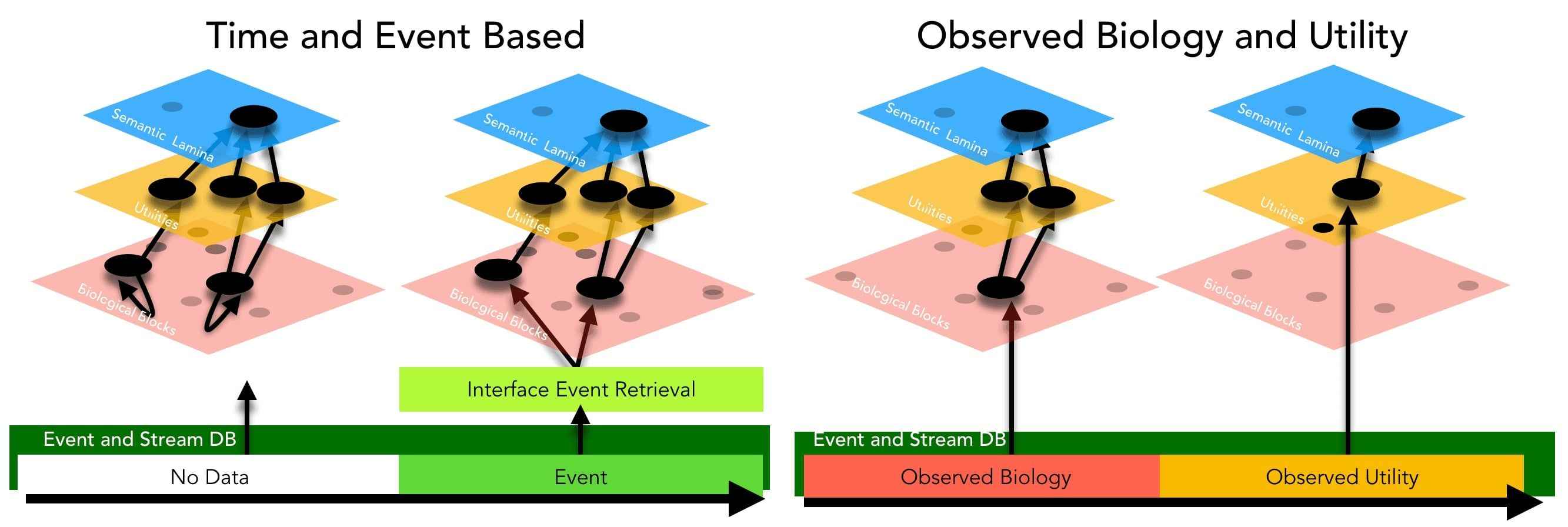}
      \label{fig:iHSEupdate}
\end{wrapfigure}

After setting up the GNB structure for the individual's intent, the next step is to use the network to show how the state is changing. Time intervals at which to update the state were chosen to be one day. There are two overarching approaches to update the state. The first approach uses event data streams about the individual as inputs. Events can include time intervals where there is no other data stream other than time itself. The second approach is measuring a biological or utility attribute directly through a sensor, and we use that information to arrive at an updated estimation of the health state. This section addresses these two approaches in separate sub-sections below. 

\subsection{Time and Event Based HSE Updates}
\begin{wrapfigure}{l}{0.35\textwidth}
    \includegraphics[width=0.3\textwidth]{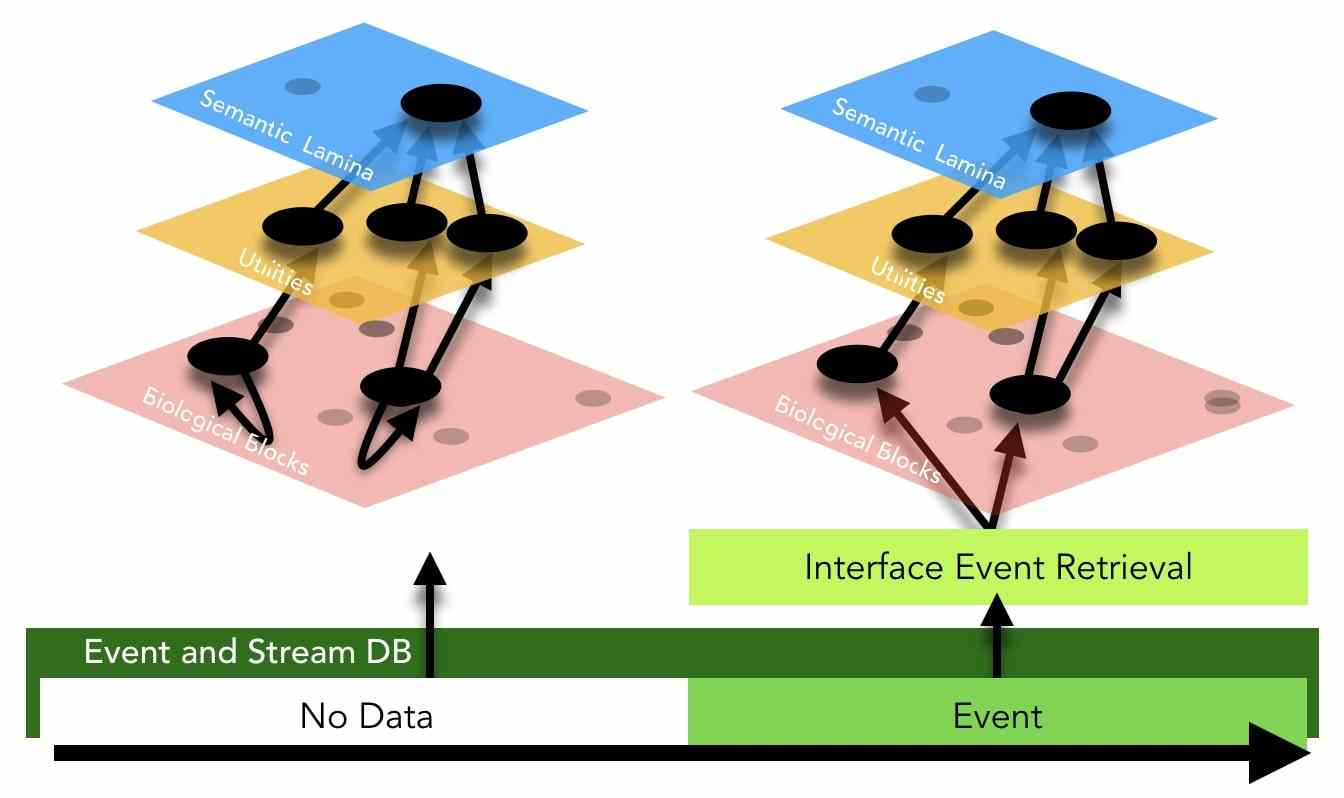}
      \label{fig:time-event}
\end{wrapfigure}
The first update approach uses temporal streams as events that perturb the GNB. Temporal streams are understood as events in this work. Not all events impact the health state regarding the utility dimensions of interest. To detect events that are relevant in changing the health state we are focused on, we need to know what types of features to look for and characterize the impact of that event on the health state. This process is called \textit{interface event retrieval}. 

\subsubsection{Interface Event Retrieval}
\begin{figure}[H]
  \centering \centerline{\epsfig{figure=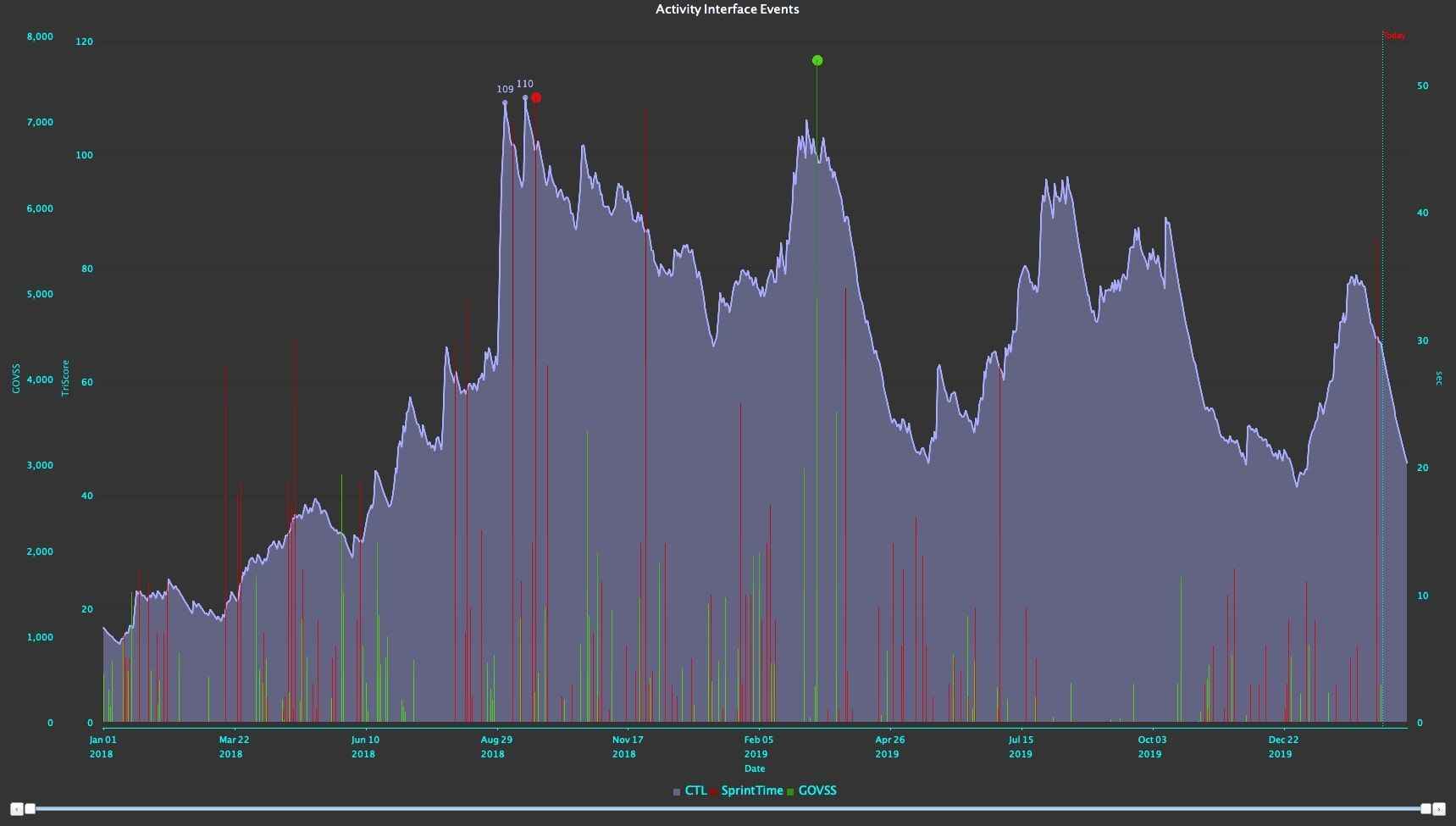,width=16cm}}
  \caption{\textbf{Exercise Interface Events.} For the years of 2018-2019, CTL is given in purple, GOVSS is given in green, HPA Load by day is given in red. These specific interface events are used to determine impact to the CRF health state.}
    \label{fig:activityinterface}
\end{figure}

\begin{figure}[H]
  \centering \centerline{\epsfig{figure=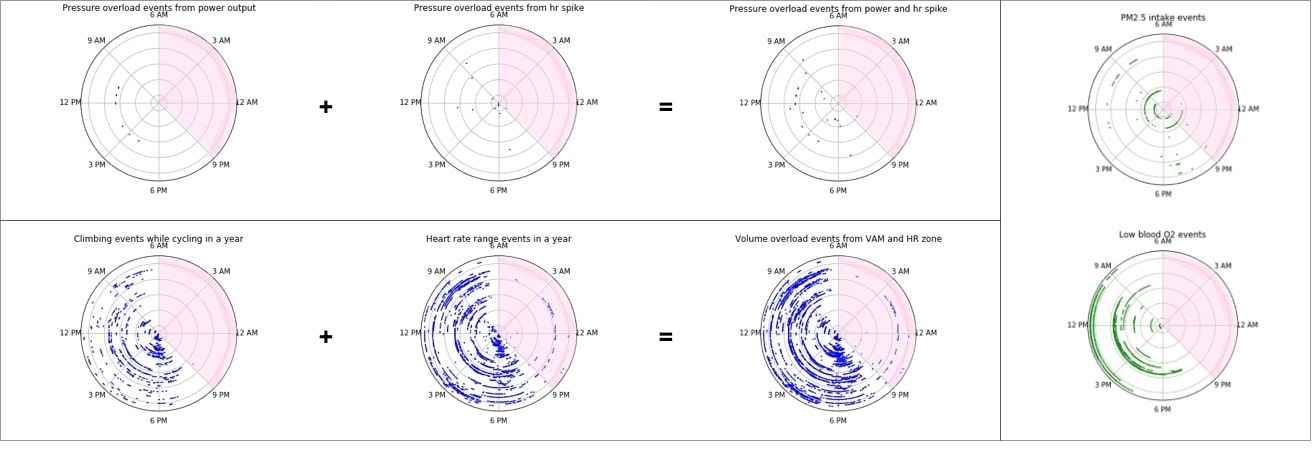,width=16cm}}
  \caption{\textbf{Circadian Rhythm of Interface Events.} This figure shows different interface events retrieved from the test subject data. Different days of the year are represented as concentric circles, and various sectors of the circle represent different times of the day. Events in a day are represented as colored arcs on that circle. Volume and pressure overload events that occur in the left ventricle are retrieved as inputs towards changing the health state. This figure also shows the environmental interface events. High PM2.5 intake events are recognized over one week. These are the instances where per minute PM2.5 intake was greater than 0.7 $\mu$g. Low blood oxygen events are recognized over one year, where blood oxygen saturation goes below 95\%. The highlighted sectors roughly represent the time between sunrise and sunset. Thus we can see how different events overlap with circadian patterns.}
    \label{fig:cardiovascular-interface-events}
\end{figure}

``In computing, an interface is a shared boundary across which two or more separate components of a computer system exchange information" \cite{Hookway2014Interface}. In the context of our work in HSE, the two sides that exchange information are the events in our life with the state of our health. Interface events bridge this gap by describing how to transform a particular event attribute into an input value for nodes within the HSE GNB structure. For events used in this experiment, we use validated aggregation metrics that connect the events to how they impact the health state of CRF. For activities, we use three algorithms of Chronic Training Load (CTL) \cite{Evans1985CardiovascularTraining.}, Gravity Ordered Velocity Stress Score (GOVSS) \cite{Wallace2014AResponses}, and High Power Activity (HPA). In a nutshell, CTL captures the long-term (42-day rolling average) of cardiac volume overload output  \cite{Evans1985CardiovascularTraining.}. GOVSS captures the work done against gravity and inertia to produce motion \cite{Wallace2014AResponses}. HPA captures high-intensity activity that stimulates muscle fiber growth and motor unit training adaptations. HPA interface events are any 5 second periods with over 10 watts per kg of power. The extraction of these features from activity data is shown for the two years of test subject data in Figure \ref{fig:activityinterface}. Using the time-stamps from the event stream, we can also inspect the circadian pattern of when particular interface events occur during the day. We show this for both activity events and environmental events in Figure \ref{fig:cardiovascular-interface-events}.

\begin{figure}[H]
  \centering \centerline{\epsfig{figure=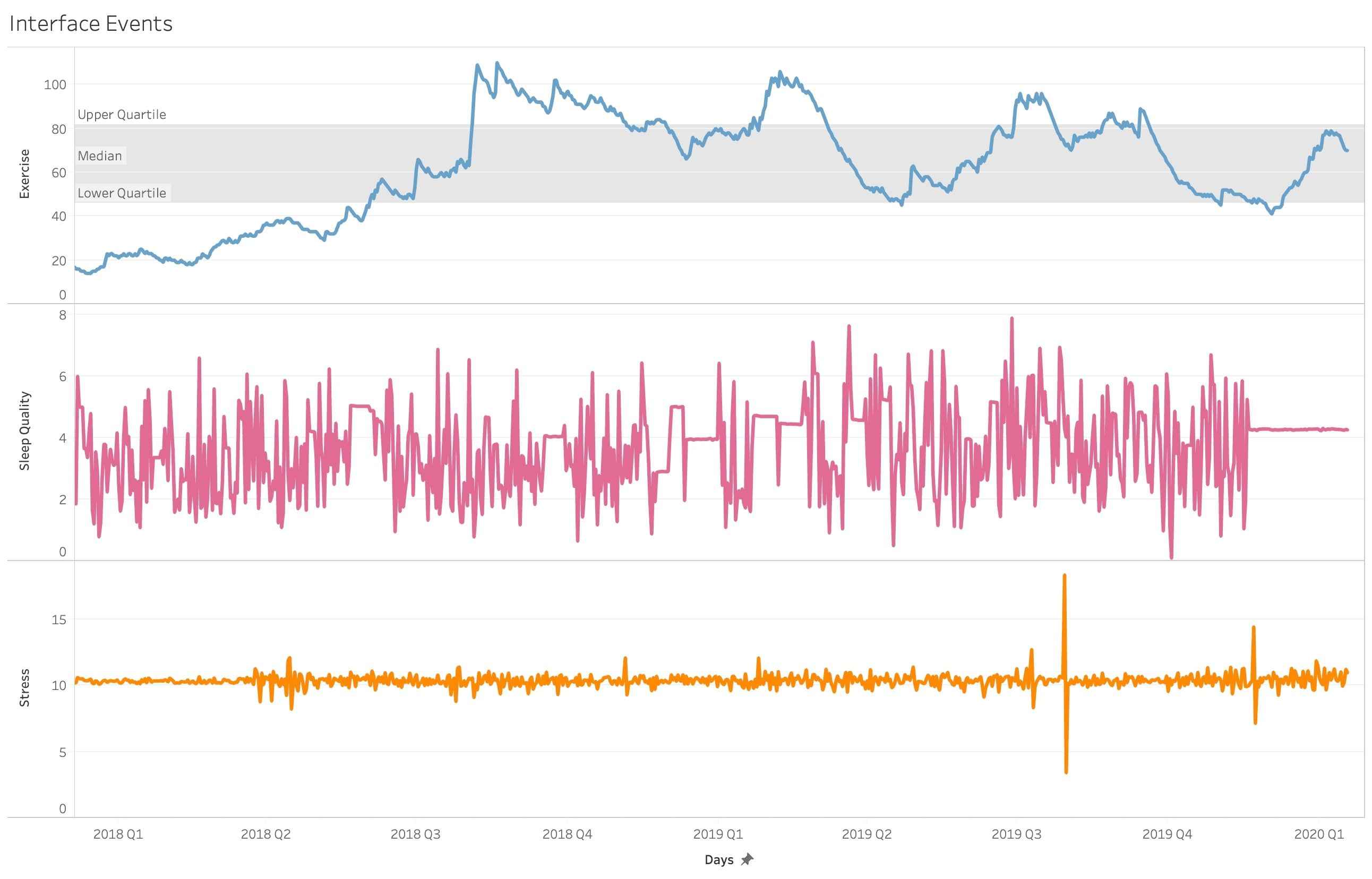,width=16cm}}
  \caption{\textbf{Interface Event Summary.} Total Exercise impact, sleep quality impact and stress impact by day for years 2018-2019 in the test subject. Exercise events are shown in quartiles showing periods of higher impact in the upper quartile, and minimal impact in the lower quartile. High variance within sleep and stress occurs daily.}
    \label{fig:ievent1}
\end{figure}

In this experiment, we summarize the impact score of exercise (Figure \ref{fig:activityinterface}), sleep as a total quality metric, food through relevant nutrients, stress through a combined score of objective and subjective data, and environmental pollution as quantity per day. The algorithms for these computations are given in the appendix. The CTL, sleep, and stress scores per day are presented in Figure \ref{fig:ievent1}, preparing the system to now propagate the effects of these events into the GNB.

\newpage
\subsection{Observation Based HSE Updates}
\begin{wrapfigure}{l}{0.35\textwidth}
    \includegraphics[width=0.3\textwidth]{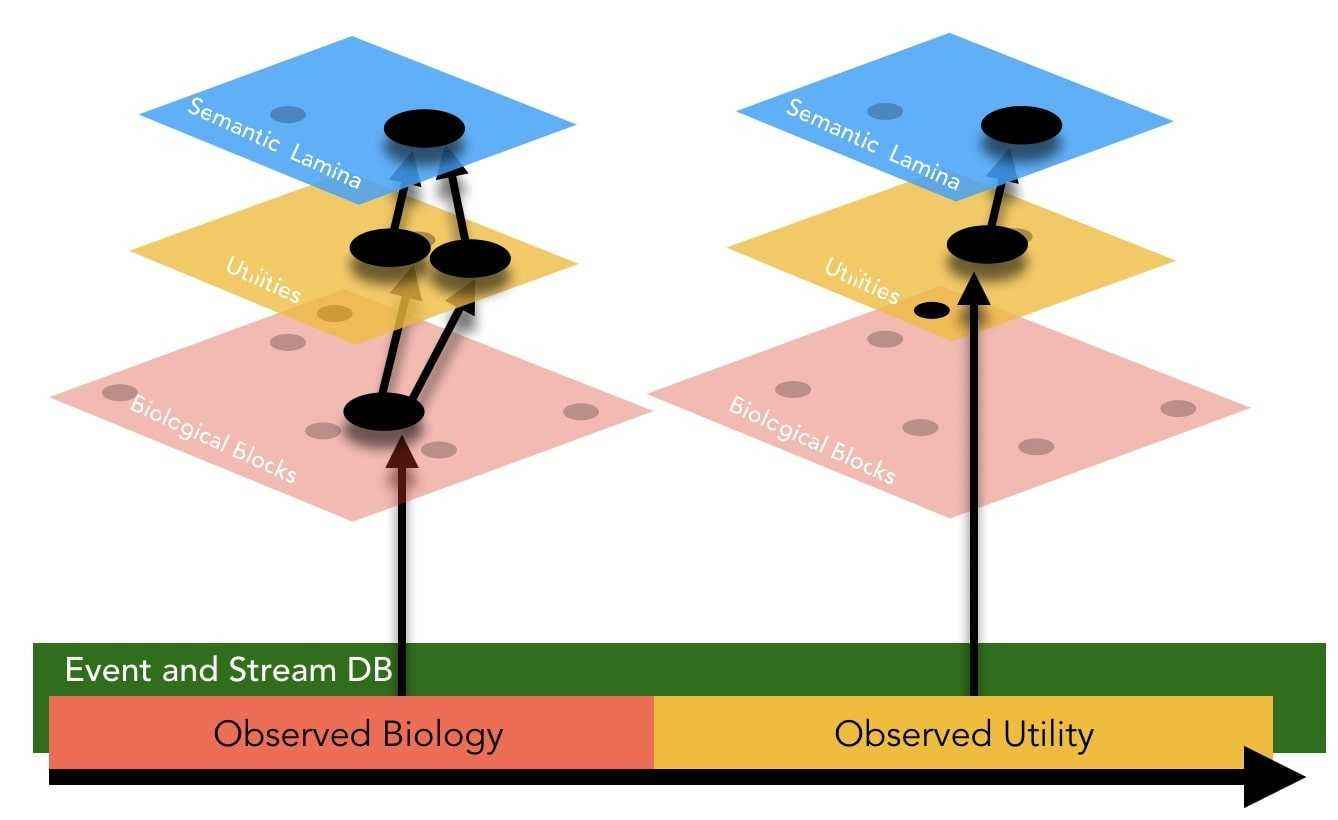}
      \label{fig:observed-update}
\end{wrapfigure} 

When we have instances of directly measuring the utility or biological values, we use these values in priority over the estimated values provided from event-based inputs. For the biological layers of the GNB, these values are provided at the organ level, or system level through the sensor collection. Examples of these values include body weight and heart rate. The node that is measured may have parent nodes that can be updated. In this experiment, we use resting heart rate and weight to derive stroke volume of the left ventricle, and thus cardiac output. For the utility states, we directly measure CP and weight during instances that indicate maximal power output. These measurements can be retrieved through looking at power intervals where heart rate is within ten beats of age-defined maximal heart rate.

\subsection{Propagating Updates}
\begin{figure}[H]
  \centering \centerline{\epsfig{figure=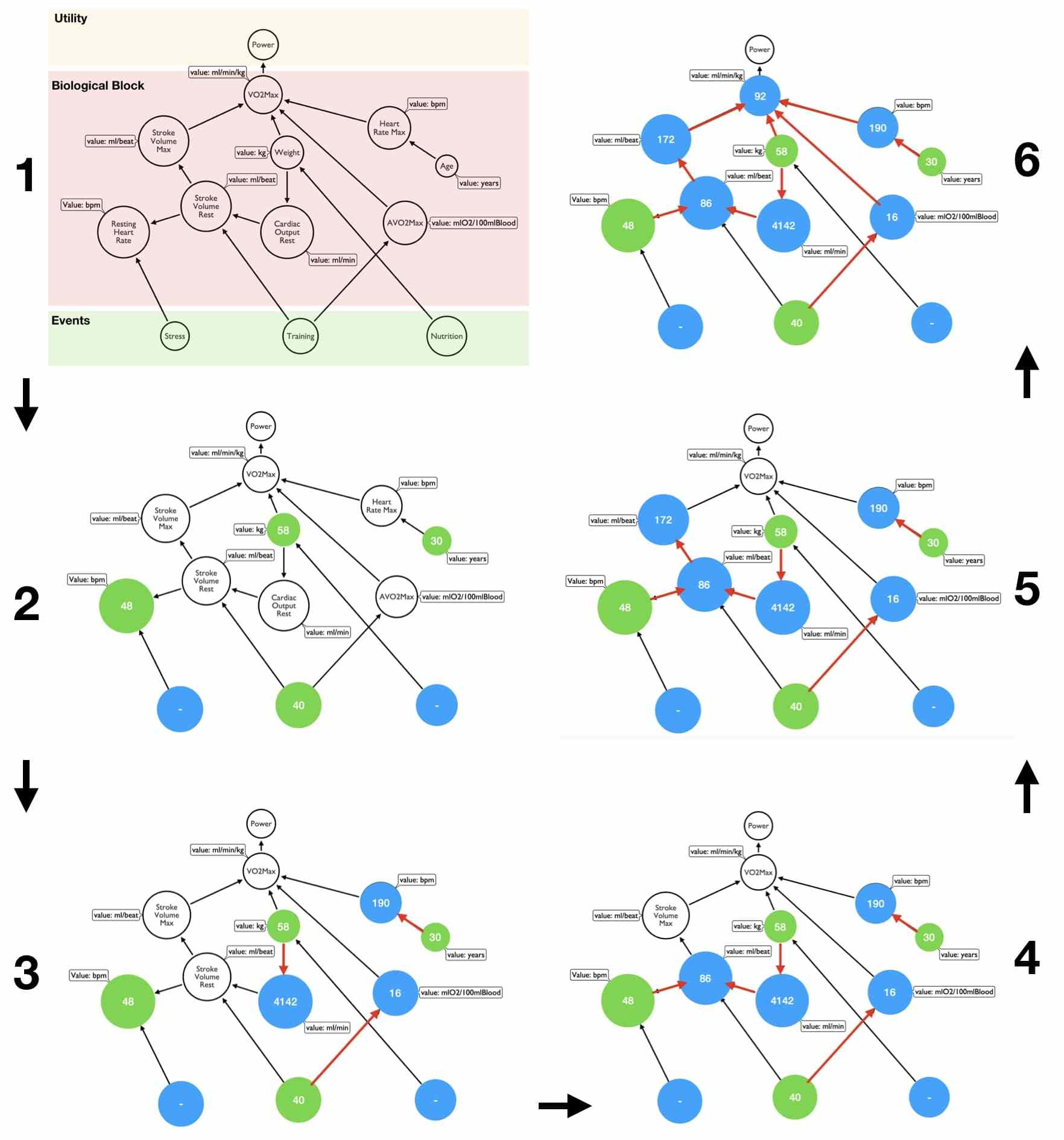,width=16cm}}
  \caption{\textbf{GNB Update Cycle.} This illustration shows the time-lapse of steps within the update cycle for a GNB system. Step 1 colors the layers of the HSE GNB structure from the initialization. Step 2 shows how the value updates are given for inputs (green) consisting of an exercise CTL of 40, age of 30, weight of 58kg, and resting heart rate of 48. The estimated nodes are blue.}
    \label{fig:update-cycle}
\end{figure}

Using the data streams of these events and direct observations, we feed new input values into the connected nodes as described in Algorithm \ref{alg:gnbupdate}. An example time-lapse of this update cycle for the CRF case is given in Figure \ref{fig:update-cycle}. We use this update system to compute the health state of the individual on a daily cycle interval for the two years of data collection (2018-2019) on the test subject's GNB structure. The results in this section show updated values for all the layers in the GNB. Through the story of queries by the user, domain expert, or interfacing computing agent may have, we describe the health state update results in various levels of depth.

\newpage
\subsubsection{Q1 User: How ready am I for the race?}
\begin{figure}[H]
  \centering \centerline{\epsfig{figure=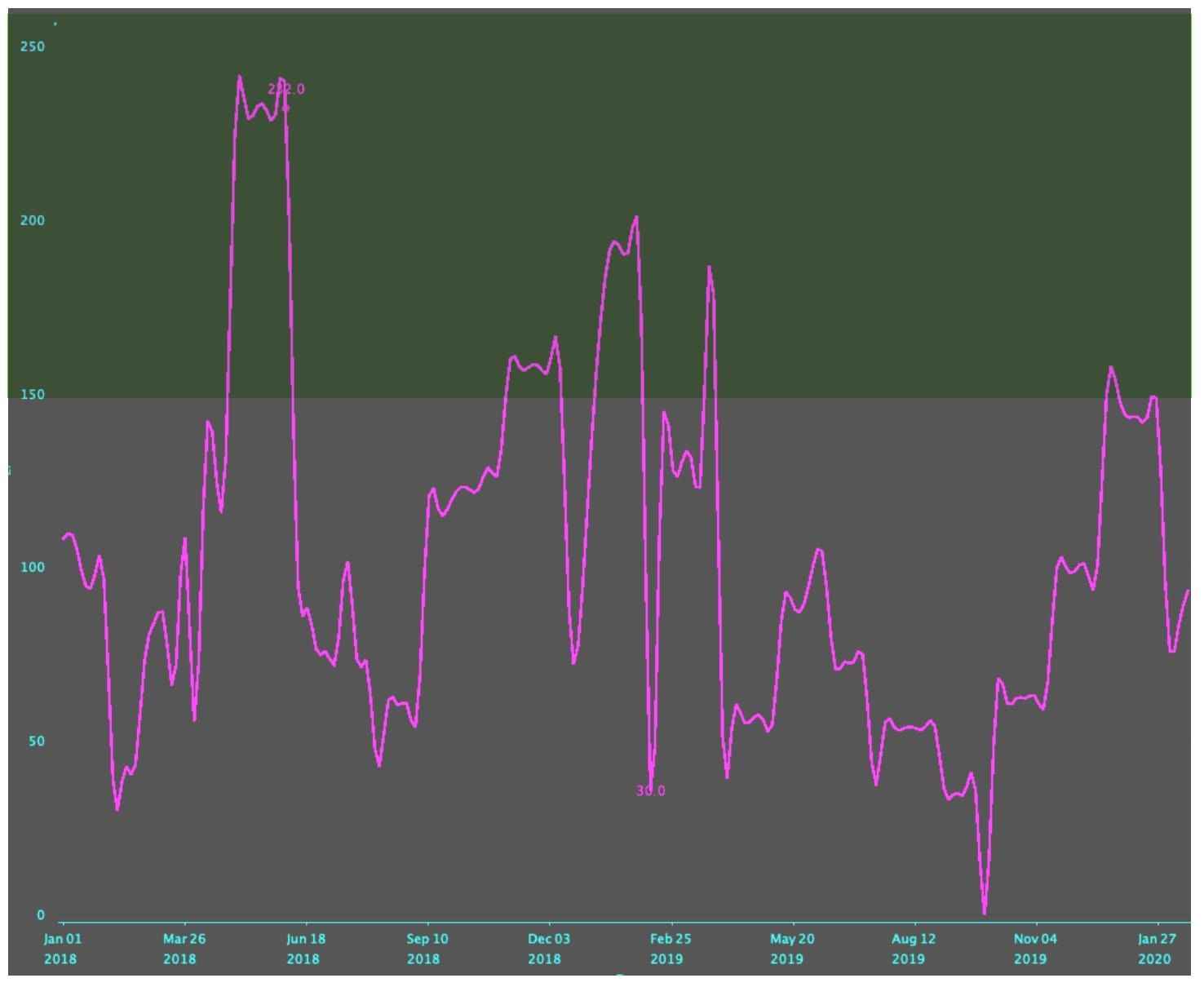,width=12cm}}
  \caption{\textbf{Single-Dimension Health State.} Here is a single dimension of one hour CP value for the subject over the course of time.}
    \label{fig:cp-history}
\end{figure}

\begin{figure}[H]
  \centering \centerline{\epsfig{figure=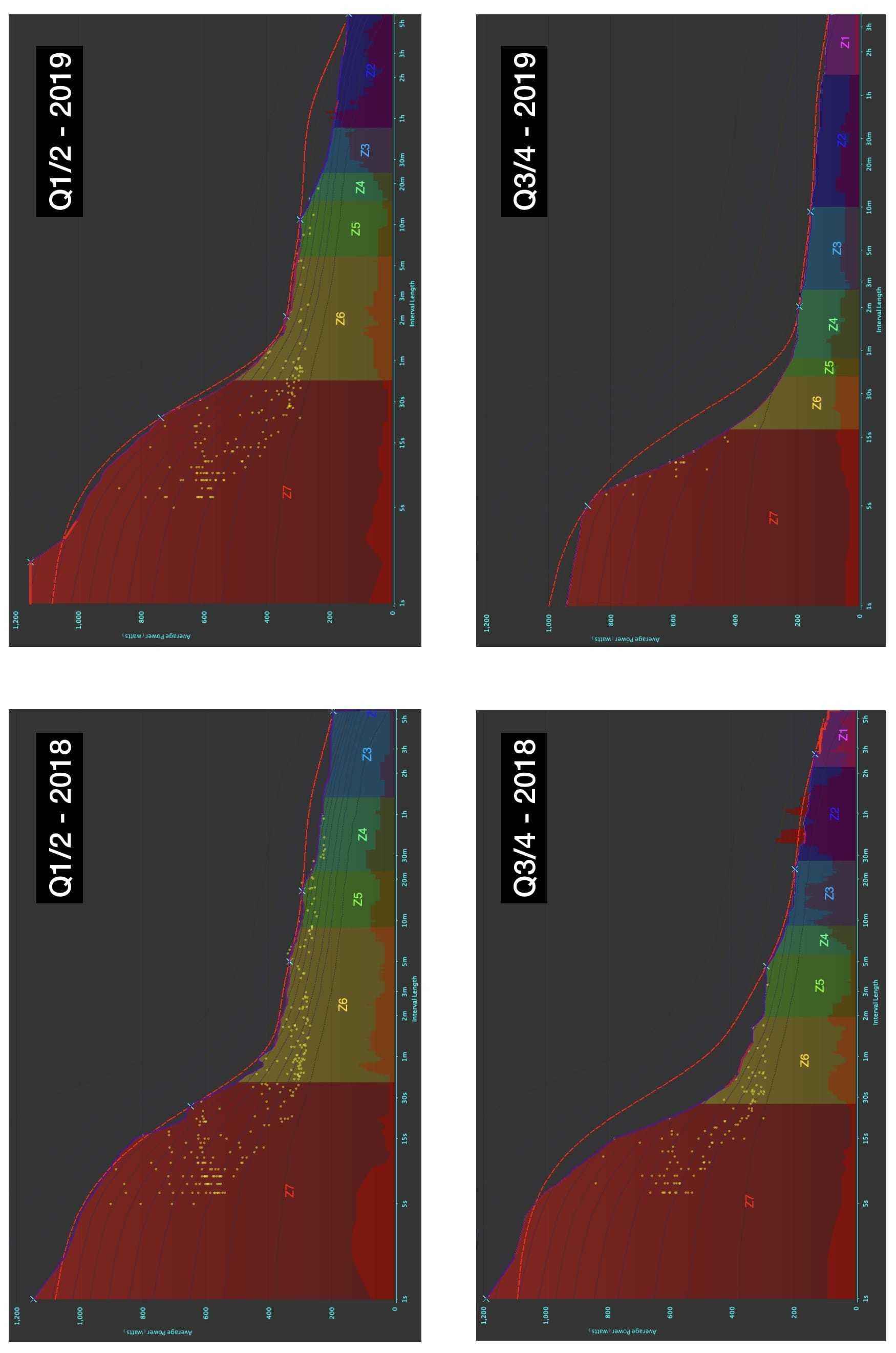,width=14cm}}
  \caption{\textbf{Multi-Dimension Health State.} Here we see all 18,000 utilities of CP at six month intervals.}
    \label{fig:stateraw}
\end{figure}

The original intent of the user was to be ready for enduro and gravel cycling. Based on the user weight, the power to produce at minimum for these events is 2.5 watts per kg for sustained periods of over 1000 seconds, equating to 152 Watts of power for this particular CP dimension. If we can estimate the utility dimension of this particular value over time, we can see when the user is capable of participating in the races they desire to do. Figure \ref{fig:cp-history} shows this particular utility dimension, with the green top section showing if the user is in the health state to fulfill their desire. We see the user has a total of 5 periods in the last two years where the user was within the appropriate utility state. Figure \ref{fig:stateraw} shows the values of the entire 18,000 utility dimensions of CP (x-axis) at four different time points throughout the last two years. Figure \ref{fig:cp-history} shows just a summary daily value of the HSE for one of these dimensions over time.

\newpage
\subsubsection{Q2 User: How does the air I breathe during a workout impact my health?}
\begin{figure}[H]
  \centering \centerline{\epsfig{figure=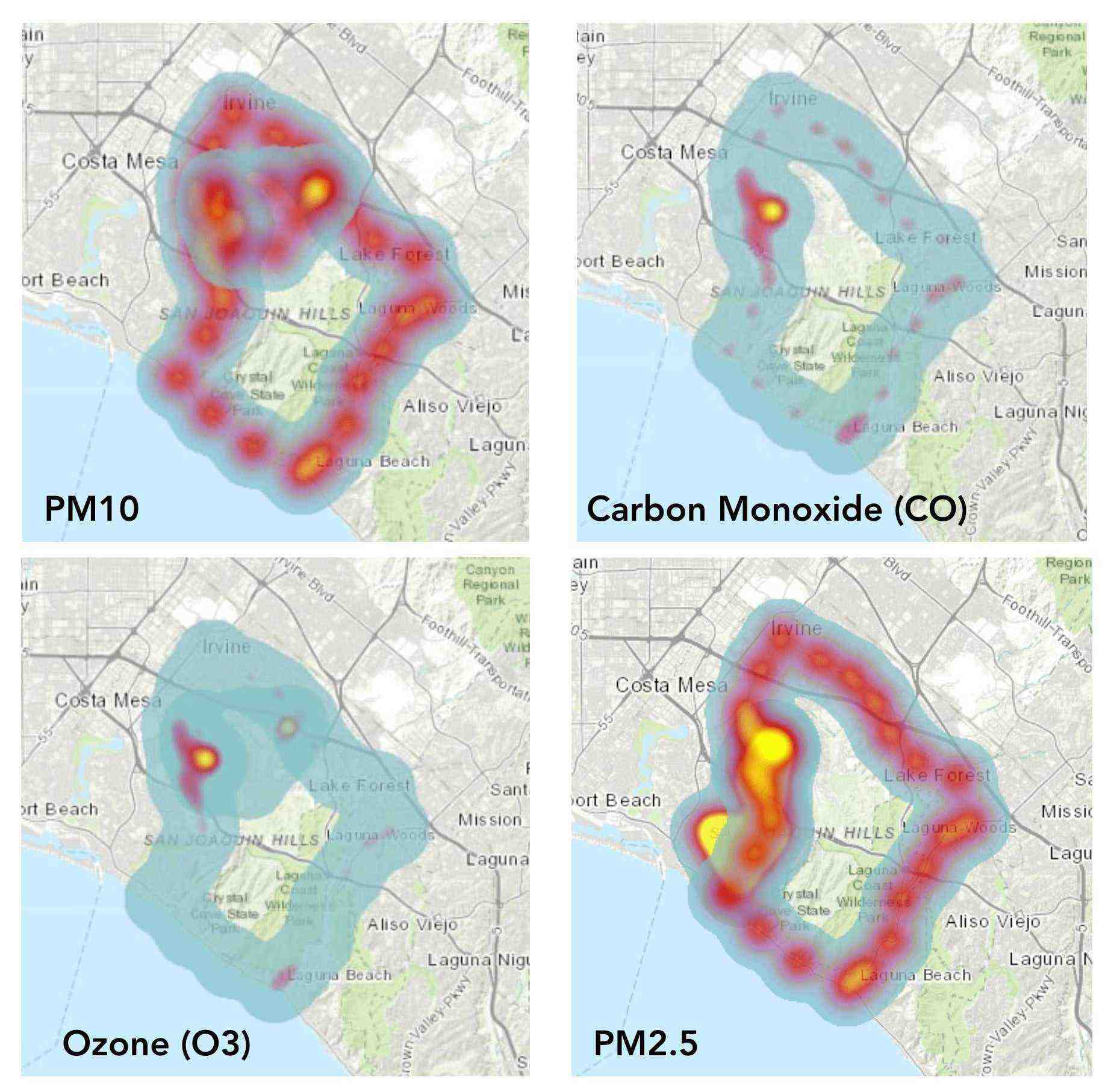,width=16cm}}
  \caption{\textbf{Location Impact on Health State.} Here we can see how four different pollutants impact a single bike ride activity. Blue is low exposure, red is moderate exposure, and yellow signifies high exposure to the particular pollutant during the unique time-points and location during the bike ride. The EPA defines these levels. The levels can be used to inform the user, health experts, civil engineering, and recommendation systems of health state impact to aid in planning.}
    \label{fig:location-impact}
\end{figure}

Our test subject wants to know how the environment interacts with their health state. GIS visualizations can give readily accessible methods by which to answer this query. Figure \ref{fig:location-impact} shows how pollution parameters like CO, O3, PM2.5, and PM10 are monitored for understanding where respiratory and cardiovascular health is changing during outdoor exercise events. This visualization uses the location tracked activities synchronized to the public data sets generated personal exposome. For improved accuracy in providing this GIS estimation, we use heart rate data to understand breathing volume mapped with the local air quality sensors via constant GPS tracking \cite{Nag2018SurfaceActivities}. Breathing rate is derived from heart rate streams generated during exercise \cite{Gastinger2010AExercise.}. Tidal volume is the volume of air inhaled during a normal breath. It is computed from gender, weight, and height of an individual, which are sourced from the respective GNB nodes. Estimation of the total volume of air intake would be the product of breathing rate and tidal volume. The total volume of air multiplied by the air particulate content gives total exposure to the particulate of interest. In summary, this query uses both the environmental data and the biological streams in conjunction to provide the result.

\newpage
\subsubsection{Q3 Doctor: How is my patient's heart health?}
\begin{figure}[H]
  \centering \centerline{\epsfig{figure=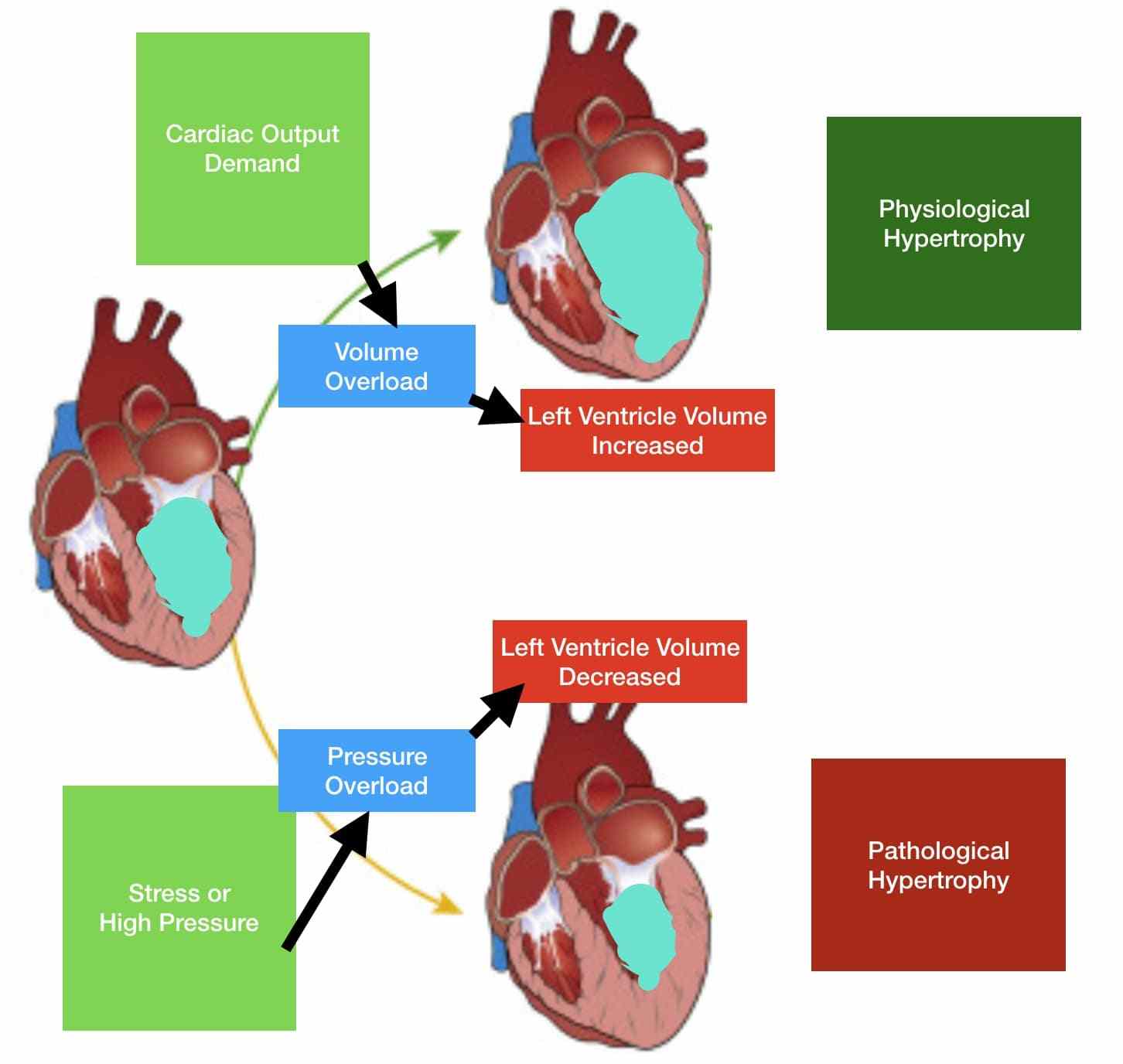,width=16cm}}
  \caption{\textbf{Interface Events for Physiological and Pathological Changes.} This figure demonstrates the difference in response between physiological and pathological heart changes resulting from various types of events. Volume overload is when the heart needs to pump at a high volume flow rate. This causes physiological hypertrophy which strengthens the stroke volume of the heart. Pressure overload causes the heart to push against higher forces which increases wall thickness, therefore reducing stroke volume and reducing cardiovascular function.}
    \label{fig:volumevspressure}
\end{figure}

\begin{figure}[H]
  \centering \centerline{\epsfig{figure=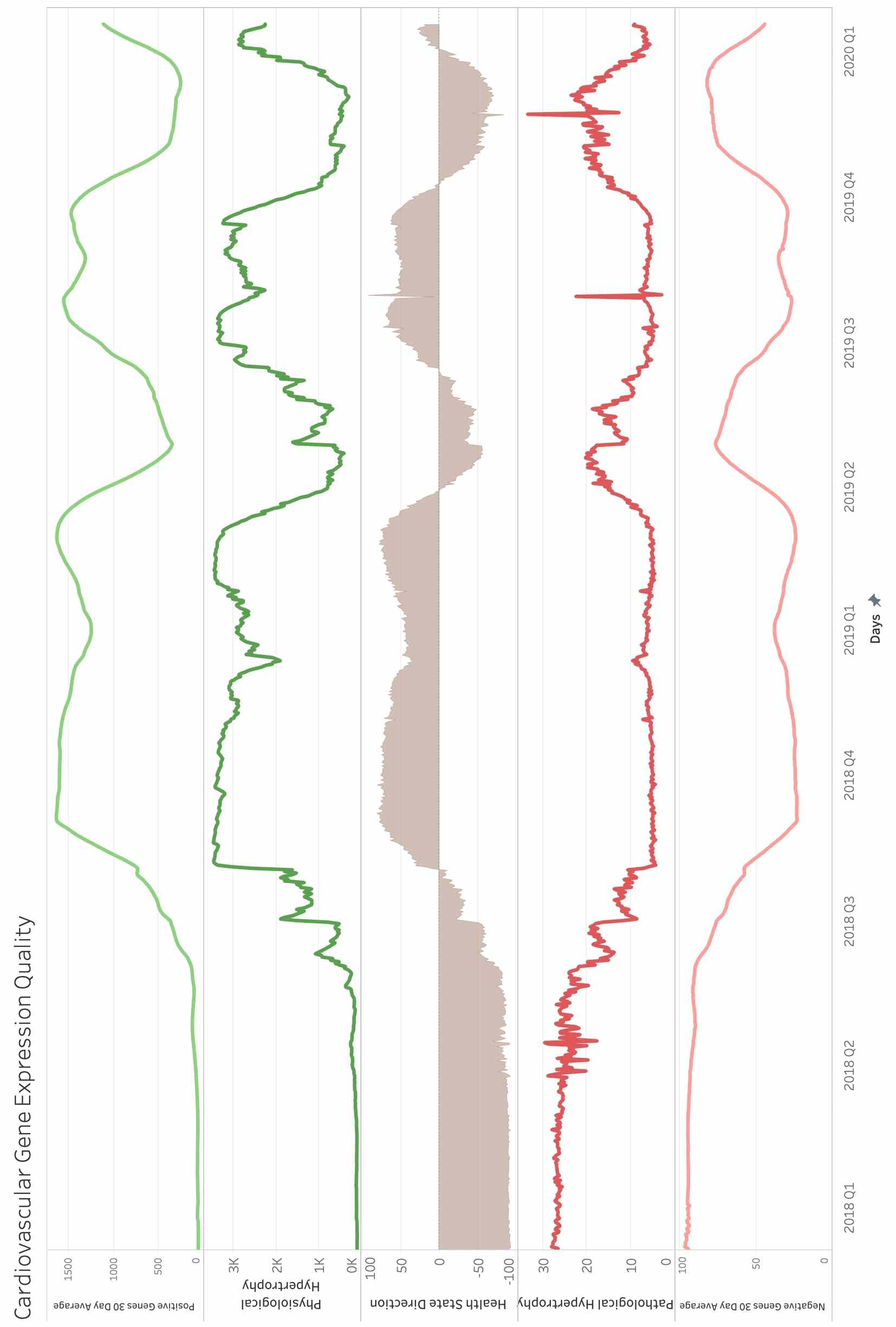,width=14cm}}
  \caption{\textbf{Daily changes in Cardiovascular Biological Health.} This figure demonstrates the balance between healthy and pathological heart function on a day-by-day basis computed through the graph structure in Figure \ref{fig:pi3k2019}.}
    \label{fig:cvgene}
\end{figure}

Our test subject visits the doctor and wants to understand if they are healthy enough to continue riding bicycles and how their health is changing. For clinical use, organ and tissue level analysis is most common. A cardiologist specialist wants to know how the left ventricular wall thickness is changing based on various life events as shown in Figure \ref{fig:volumevspressure}. The physician can use the GNB to compute the summary of the heart node global attributes. This state is shown in Figure \ref{fig:cvgene}. The subject can see there are periods where heart health is improving, and heart health is hampered. The top row of the figure shows if healthy genes are expressed in the heart, and the bottom row shows the unhealthy genes. The second row shows how much positive physiological hypertrophy is occurring. This is when the heart's left ventricle expands in its interior volume, thus allowing for an increased stroke volume. The fourth row shows the pathological hypertrophy, where fibrosis and left ventricular wall thickness may be increasing, causing the left ventricle interior volume to decrease and leads to unhealthy states such as heart failure. The middle row shows the balance of healthy and pathological processes to see if the global attributes of the heart are improving or degrading. Based on these observations, the clinician can discuss best actions towards improving heart health.

\newpage
\subsubsection{Q4 Doctor: How is my patient's left ventricular function?}
\begin{figure}[H]
  \centering \centerline{\epsfig{figure=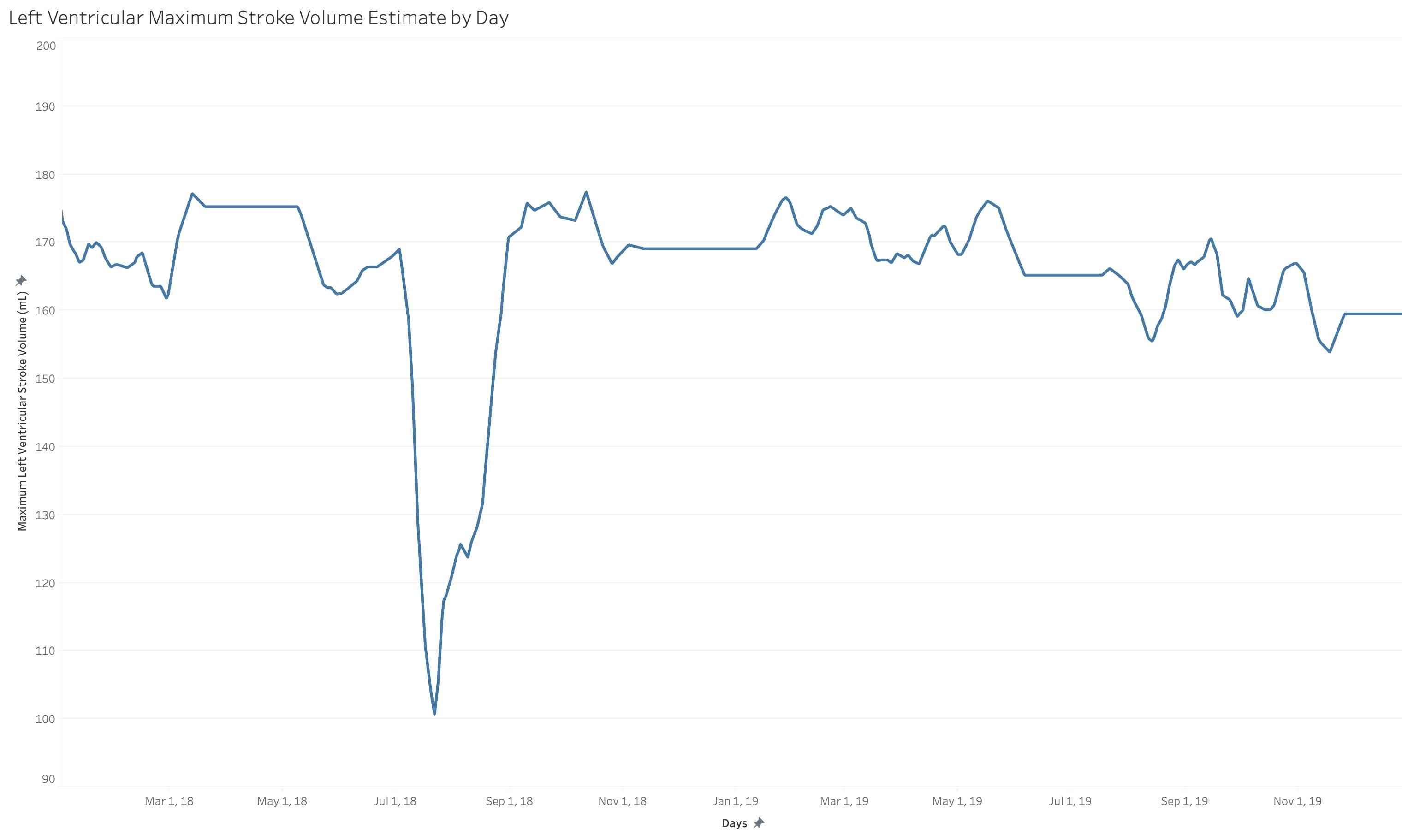,width=16cm}}
  \caption{\textbf{Daily Changes in Left Ventricular Stroke Volume.} This figure demonstrates the balance between healthy and pathological heart function on a day-by-day basis computed through the graph structure in Figure \ref{fig:pi3k2019}.}
    \label{fig:lvsv}
\end{figure}

Our test subject visits the doctor and wants to understand how their heart health is changing in regards to the left ventricular function. Again clinical use, organ and tissue level analysis is most common. The physician can use the GNB to see how the left ventricular volume is being updated on a daily basis. This state is shown in Figure \ref{fig:lvsv}. The subject can see there are periods where left ventricular health is improving or decreasing. A clinician can verify the stroke volume using ultrasound if they want a higher accuracy measurement.

\newpage
\subsubsection{Q5 Scientist: What are the molecular changes in this particular heart?}
\begin{figure}[H]
  \centering \centerline{\epsfig{figure=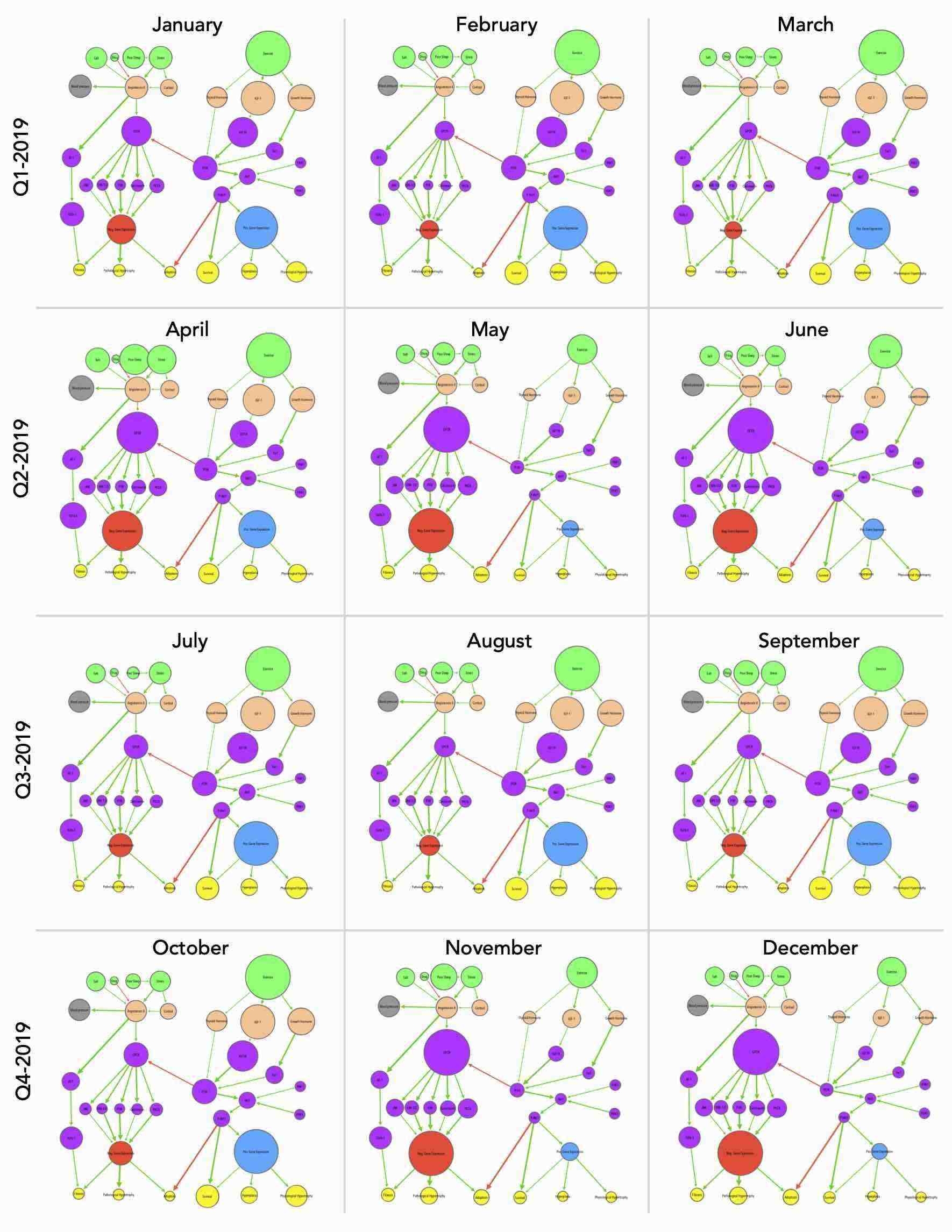,width=16cm}}
  \caption{\textbf{Health State Graph Updates.} This figure demonstrates the estimated signaling pathway nodes for cardiovascular health change at the molecular gene expression level. An average for each month is shown in the year 2019. Legend is given in Figure \ref{fig:cardiacnetwork}.}
    \label{fig:pi3k2019}
\end{figure}

The test subject participates in a research study where scientists are trying to understand what molecular pathways are potentially activated and deactivated in an individual. The scientist would like to know how cycling events are possibly changing molecular pathways that are related to heart health for potential further research and therapeutic targets. The scientist can zoom into the GNB level of the heart node molecular signaling and gene expression, as shown earlier in Figure \ref{fig:cardiacnetwork}. To see how the molecular signaling molecules are changing in the heart, a graph-based visualization is produced showing the pathway in each month of the year in 2019 (Figure \ref{fig:pi3k2019}).

The various queries above demonstrate the usability of the health state estimation at different levels of abstraction. This outlines the framework by which any HSE system can be built. Concluding this section, we see how the health state updates can be used readily by the user, healthcare provider, and scientist to inquire how and why the health state is changing. Next, we discuss altering the HSE GNB structure to use incoming data to model the user better.


\newpage
\section{Personal Model Update}
\begin{wrapfigure}{l}{0.5\textwidth}
    \includegraphics[width=0.48\textwidth]{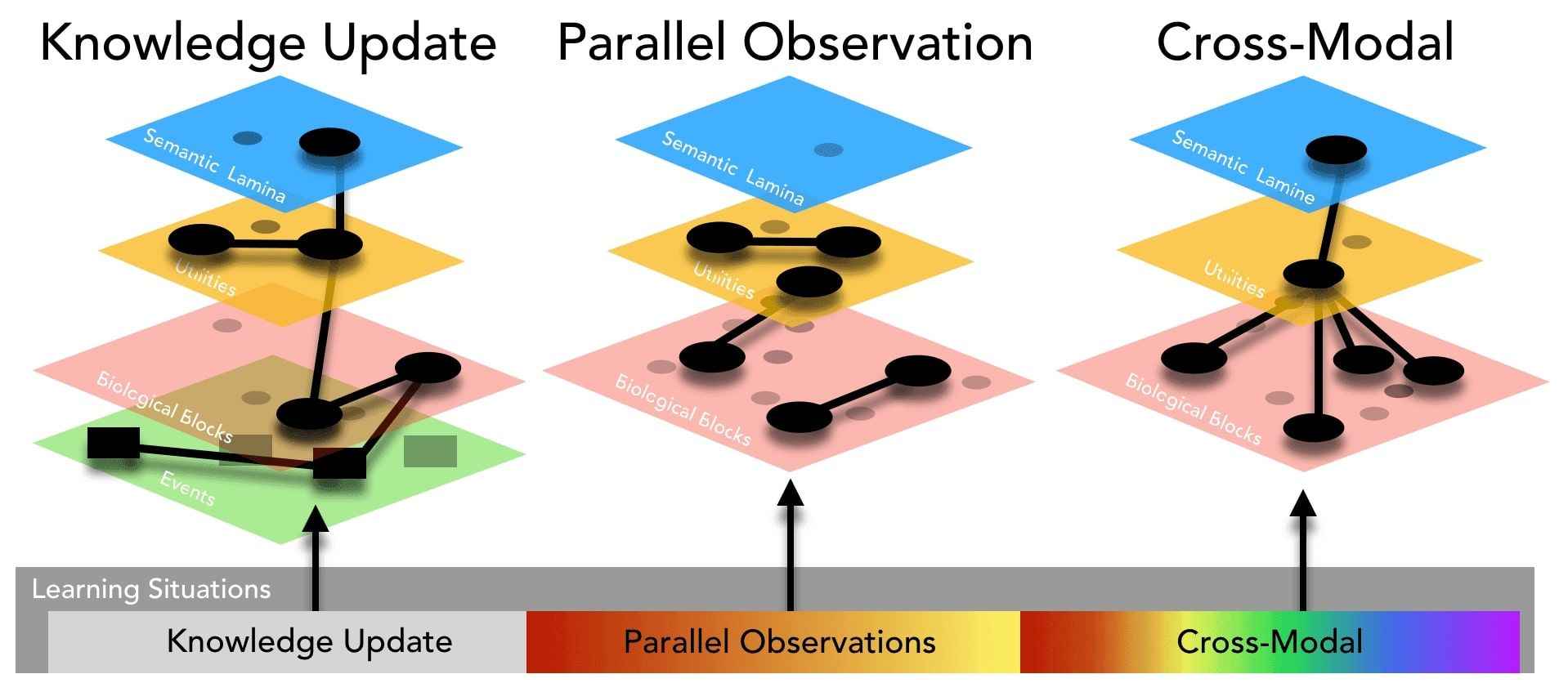}
      \label{fig:imodelupdate}
\end{wrapfigure}

To update the GNB model for the individual, we take both a bottom-up and a top-down approach on the system. We start by using a top-down approach as a ``Knowledge Update" on the model. Following this, we take two bottom-up approaches to better model the GNB. These data-driven approaches use two different methods of learning. One uses data incoming from two sources that are linked together to modify the edge. We call this parallel observation learning. The other uses many sources of incoming data to one node, to better estimate the values for that node. We call this approach, cross-modal learning.

\subsection{Knowledge Update}
\begin{wrapfigure}{l}{0.35\textwidth}
    \includegraphics[width=0.3\textwidth]{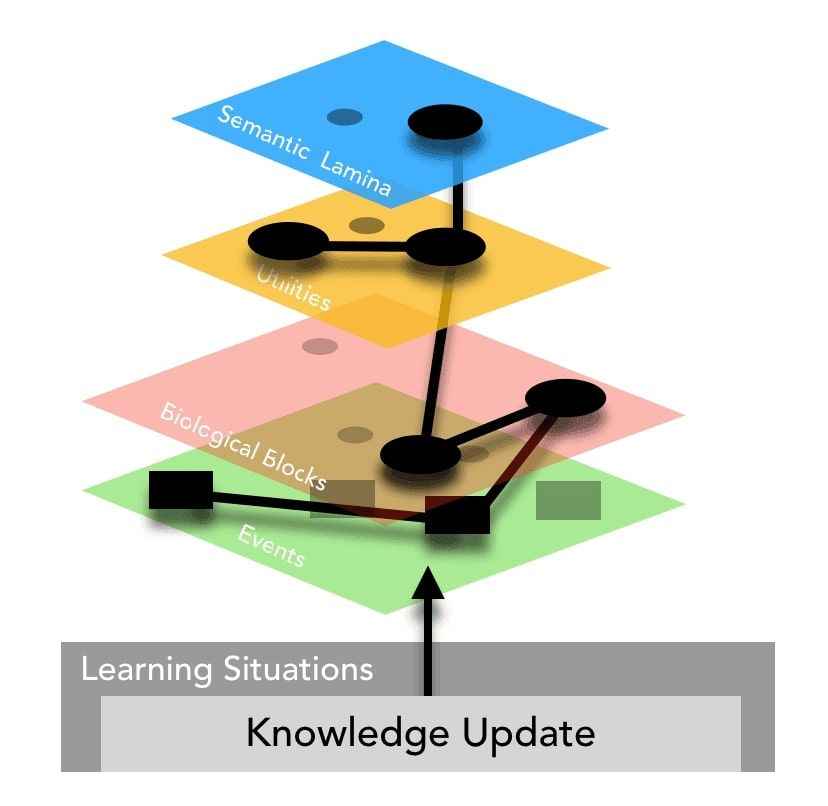}
      \label{fig:know-update}
\end{wrapfigure} 

\begin{figure}[H]
  \centering \centerline{\epsfig{figure=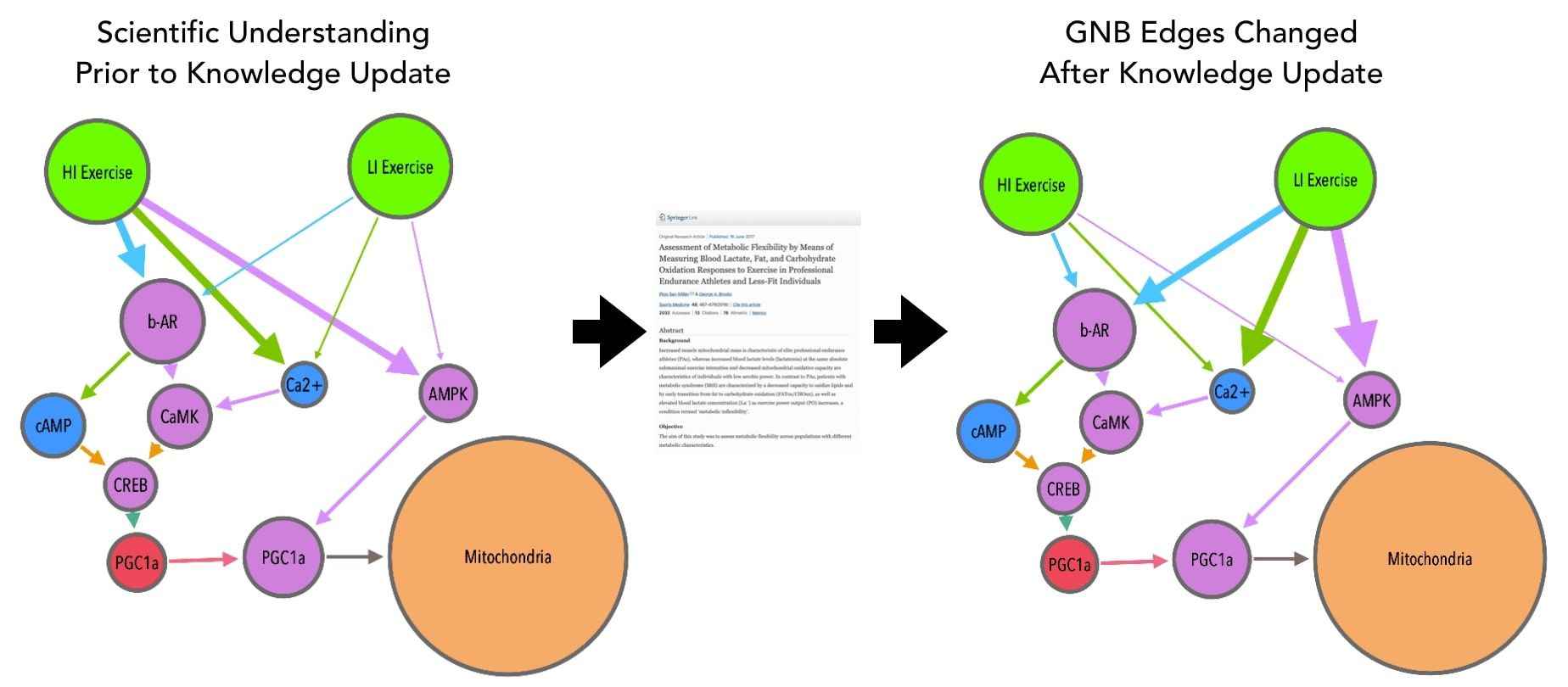,width=16cm}}
  \caption{\textbf{Updating Model of Event to Biological Node.} This figure demonstrates the difference in the model between the original knowledge-based edge weights and after the knowledge update from the literature. High Intensity (HI) exercise was thought to be the primary causal pathway to increased mitochondrial biogenesis in the muscle and heart. New literature shows that Low Intensity (LI) exercise does this much more than HI exercise. Using this update, the edge weights on the right are changed.}
    \label{fig:mitoupdate}
\end{figure}

Two forms of knowledge can inform the HSE system to update the GNB for the individual. The first form of knowledge is an update from the scientific literature on the structure and relationships between the entities. In this example, we update the relationship between an event category and biological nodes. Recently discovered knowledge in molecular biology shows how mitochondria increase in the muscles best from low-intensity exercise \cite{Yeager2020LISSSpeed}. This finding is quite contrary to the commonly accepted research idea that high-intensity interval training (HIIT) was the most effective way to improve mitochondrial density in myocytes \cite{Hoshino2013High-intensityMuscle}. In light of this discovery, we increase the edge weights between the low-intensity exercise interface events causing mitochondrial biogenesis and show the decreased edge weights for mitochondrial biogenesis from HIIT, as presented in Figure \ref{fig:mitoupdate}.

The second approach to a knowledge update is the genetics of an individual. Most genetics studies establish the correlation or causation of a gene to either an event, another biological node, or a utility. In the case of this subject, we gathered the full 23andMe Single Nucleotide Polymorphism (SNP) of the individual and filtered out any genetics information that is pertinent to the GNB structure related to CRF. We find that the subjects have several genes associated with nutrition, exercise, and sleep that impact the graph structure, as shown in Table \ref{tab:genes}. We choose to focus the example on the exercise gene, Alpha-actinin-3 (ACTN3). ACTN3 is a binding protein that plays a role in skeletal muscle fibers. This gene regulates the fast-twitch muscle fibers, essential for regulating the coordination of muscle-fiber contraction \cite{Massidda2019GeneticsFlexibility}.

Given this genetic makeup of one C and one T variant, the strength between low-intensity exercise and high power exercise will have a balanced effect on the user's GNB biological nodes. If this user were to have two C copies of the ACTN3 gene SNP, then the edge between power exercise and muscle node attribute values would change. In this case, for a given high power interface event detected would increase the resulting change in the muscle fiber. However, for a low-intensity exercise event, the consequent change in muscle composition would be less.

\begin{table}[hbt!]
\centering
\begin{tabular}{@{}llll@{}}
\toprule
Subject Genetics   & \textbf{Characteristic} & \textbf{Gene} & \textbf{Variant/Marker} \\ \midrule
\textbf{Nutrition} & Alcohol Flush           & ALDH2         & 2G                      \\
\textbf{}          & Caffeine                & CYP1A2        & rs2472297               \\
\textbf{}          & Caffeine                & AHR           & rs4410790               \\
\textbf{}          & Lactose Intolerance     & LCT           & rs4988235, 1A 1G        \\
\textbf{}          & Saturated Fat \& Weight & APOA2         & rs5082                  \\
\textbf{Exercise}  & Muscle Composition      & ACTN3         & rs1815739, 1C 1T        \\
\textbf{Sleep}     & Deep Sleep              & ADA           & rs73598374, 2C          \\
\textbf{}          & Sleep Movement          & BTBD9         & rs3923809, 1G 1A        \\ \bottomrule
\end{tabular}
\caption{\textbf{Genetics SNP.} Genetics SNP profile of subject related to the CRF GNB.}
\label{tab:genes}
\end{table}

\begin{figure}[H]
  \centering \centerline{\epsfig{figure=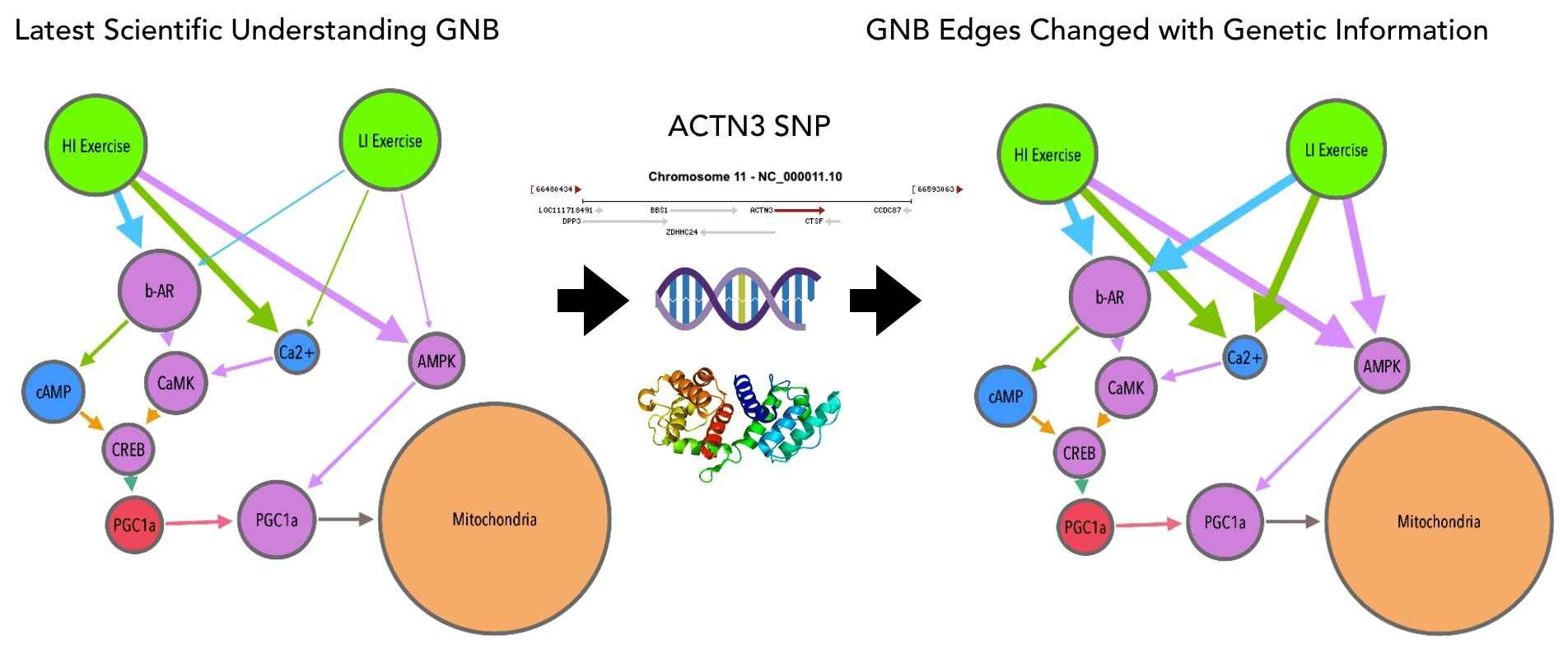,width=16cm}}
  \caption{\textbf{Genetic Update of Model of Event to Biological Node.} This figure demonstrates the difference in the model between using the most updated general knowledge and knowledge, specifically about the user's genetics. Because this user has two different types of ACTN3 gene, the impact of both HI Exercise and LI Exercise is higher.}
    \label{fig:mitoupdate2}
\end{figure}

This summarizes the knowledge-based approach to updating the graph. We continue next with data-driven model updates.

\newpage
\subsection{Parallel Observations}
\begin{wrapfigure}{l}{0.35\textwidth}
    \includegraphics[width=0.3\textwidth]{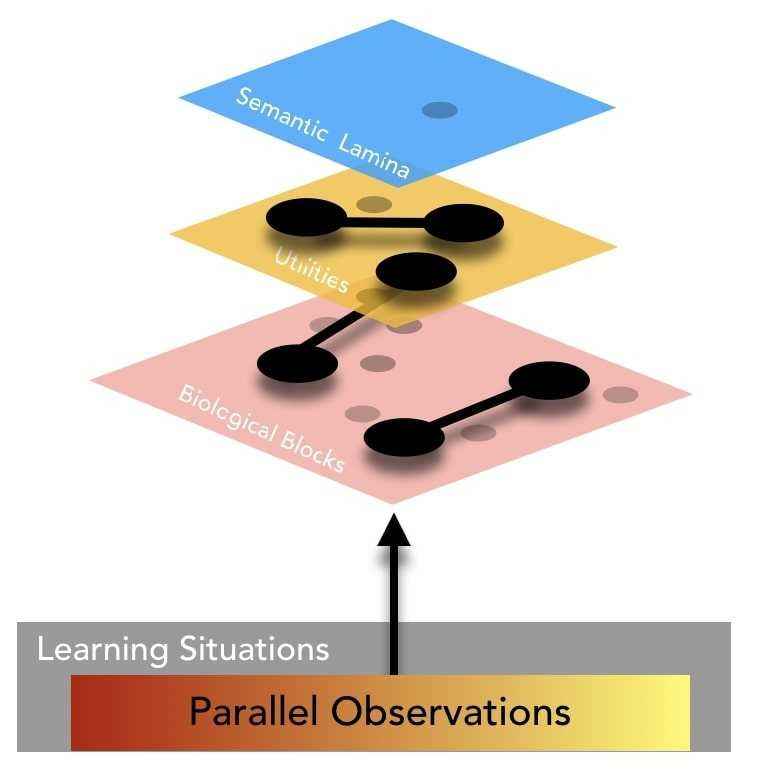}
      \label{fig:par-update}
\end{wrapfigure}
\begin{figure}[H]
  \centering \centerline{\epsfig{figure=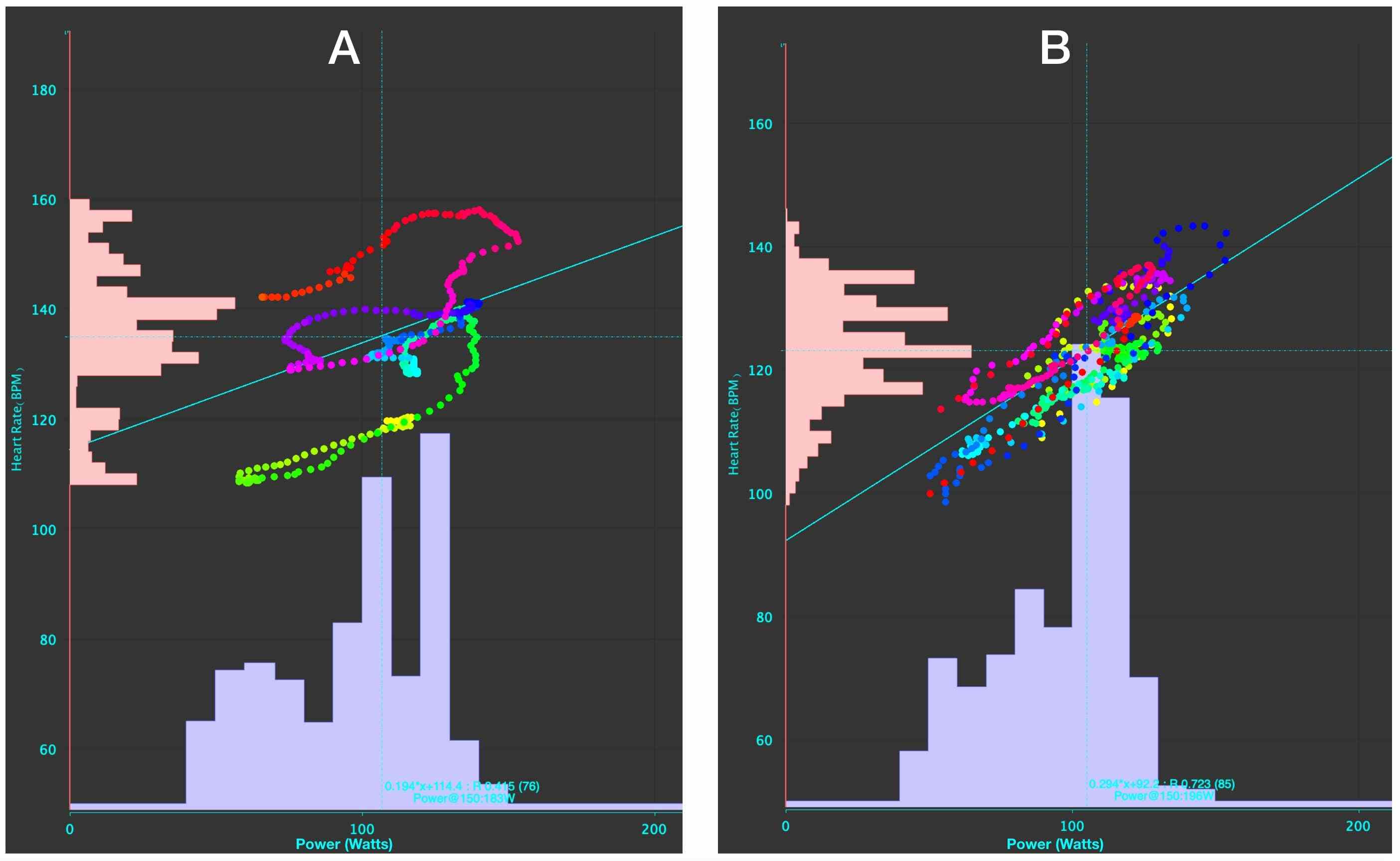,width=16cm}}
  \caption{\textbf{Parallel Data Model Learning.} The above two graphs show heart rate data related to power data in two different activities. During activity A, we can see three distinct linear bands of relationships in red, magenta, and green. These clusters are separated due to a phenomenon called aerobic decoupling. We can check the causal agent of these multi clusters in the ride data by referencing the aerobic decoupling node. In ride A, it is 40.3 percent, versus Ride B is 0.9 percent. For this reason, the graph system can causally understand that activity B is a better sample. This allows for learning the relationship with higher quality.}
    \label{fig:hrpwr}
\end{figure}

\begin{figure}[H]
  \centering \centerline{\epsfig{figure=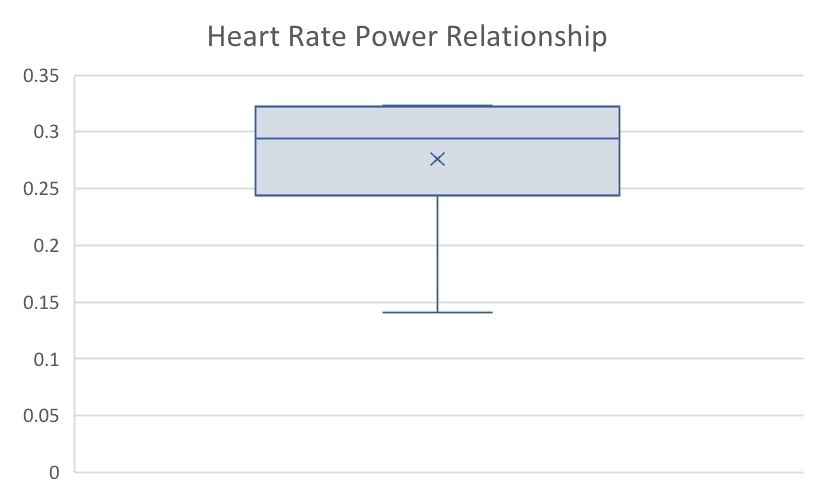,width=10cm}}
  \caption{\textbf{Heart Rate and Power Relationship.} This box plot shows the relationship between heart rate and power for the test subject from various instances within a one month period.}
    \label{fig:hrpwrelate}
\end{figure}

 
The first of two data-driven learning models we employ in this work is applicable when two nodes are observed with incoming data and are linked together. In the instance of our experiment, the two nodes that are linked are heart rate and power. Increases in heart rate result in increased power, and this is well understood both colloquially and in the biomedical literature. For each individual, this parameter is also always changing based on how their CRF state is changing. When plotting heart rate versus power, we notice the data ends up in two patterns. The first pattern is split with various linear clusters, as shown on image A in Figure \ref{fig:hrpwr}. This pattern occurs due to a physiological phenomenon called heart rate drift, also known as aerobic decoupling. Aerobic decoupling is when the heart rate dramatically shifts for the same power output due to cardiac fatigue. Instances like image A show this effect happening. The green-colored portion of Figure \ref{fig:hrpwr} shows the beginning of the ride, and the red nodes indicate the values at the end of the ride. We can see that there is a shift in this twice reflecting initial fatigue followed by another step of extreme exhaustion. Because the relationship is changing so much within one ride, this instance is not a good source of parallel observation learning. The aerobic decoupling algorithm confirms this, with a decoupling of 40.3 percent. Image B, on the other hand, shows a single linear cluster that does not have any distinct separation. The aerobic decoupling algorithm confirms this, with a decoupling of 0.9 percent. Therefore the data from this ride can be used to give a quality relationship between the two variables. We use simple linear regression for this model, but more complex learning approaches can easily be applied if desired. To teach the system to look for high-quality relationships, we can compute the aerobic decoupling factor for each ride easily with the Algorithm \ref{alg:ad}. Using the latest value of this relationship, we can update the edges between heart rate and power in the GNB. The latest value for the last month of data (December 2019) on the test subject is shown in Figure \ref{fig:hrpwrelate}.

\begin{algorithm}
\caption{Aerobic Decoupling}
\label{alg:ad}
\begin{center}
    \begin{tabular}{  p{14cm}  }
\textbf{Input: } Power and Heart Rate event streams ($P$, $HR$)

\textbf{Output: } Aerobic Decoupling Percentage ($AD$)'

\textbf{Begin}

1: for each activity 

2: \qquad if power and heart rate data present ($P$, $HR$)

3: \qquad \qquad Compute power-heart rate ratio for starting half of event, ($S$).

4: \qquad \qquad Compute power-heart rate ratio for last half of event , ($L$).

4: \qquad \qquad ($D$) = ($S$) - ($L$)

5: \qquad \qquad ($AD$)' = ($D$)*100 / ($S$)

6: \qquad end for

7: end for

8: Return ($AD$)'

\textbf{End}
    \end{tabular}
\end{center}
\end{algorithm}

\newpage
\subsection{Cross-Modal}
\begin{wrapfigure}{l}{0.35\textwidth}
    \includegraphics[width=0.3\textwidth]{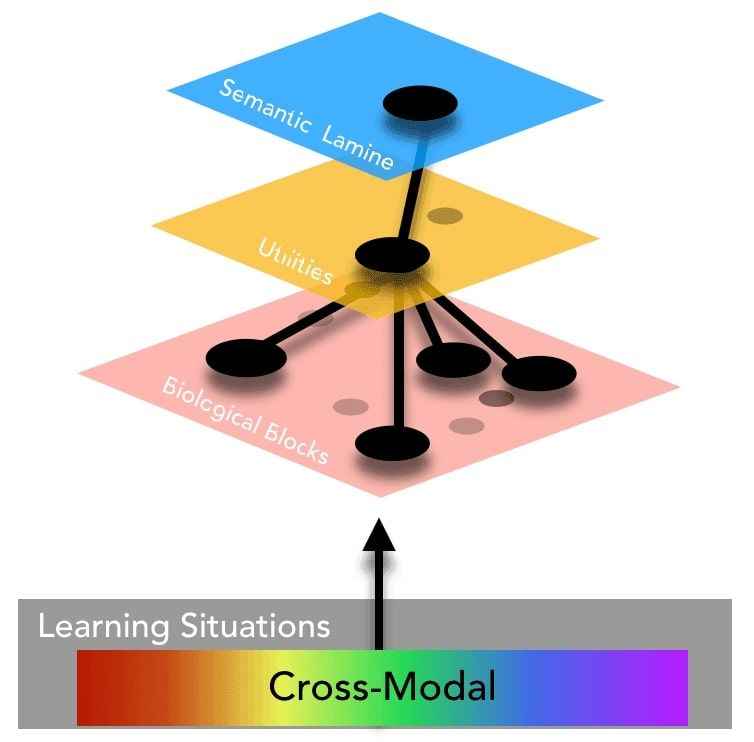}
      \label{fig:cross-update}
\end{wrapfigure} 

Cross-modal is defined as understanding of a situation from multiple \textit{types} of inputs \cite{Lalanneab2004CrossmodalAction}. In this case, cross-modal learning takes various inputs of the biological variables and events to estimate the value of a utility node. This set of experiments show that we can improve the quality of CRF estimation via this cross-modal approach. The estimations can improve by combining multiple cheap (in terms of monetary or energy costs) sensors. We use three features from the user's event and data-streams that all have a direct impact on CRF: 1) Rate of vertical ascent in meters (VAM), 2) Training Impulse (TRIMP), and 3) Cadence-detected Active Time.

\begin{figure}[H]
  \centering \centerline{\epsfig{figure=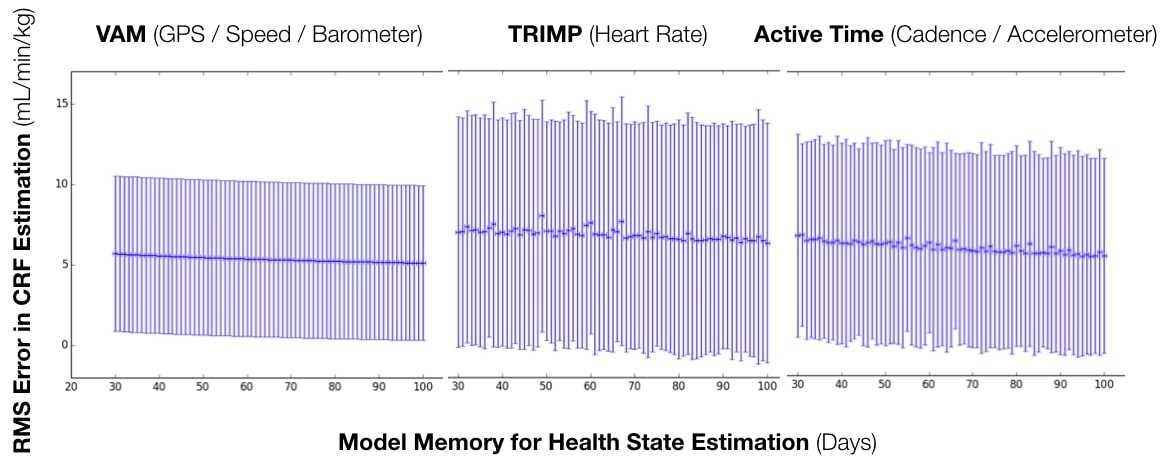,width=16cm}}
  \caption{\textbf{Memory Parameter Optimization.} For each of the three sensor modes, we tested the model to see if there was an optimal amount of days to use as a memory for the CRF prediction. Any model memory of 10-100 days gave similarly accurate learning results, demonstrating there was no optimal learning memory for the model.}
    \label{fig:parameter}
\end{figure}

Per second power output values collected from the strain gauges serve as the ground truth in our experiments for bio-variable estimation. We used a rolling average of maximum 4-minute power output per day over 42 days to generate the ground truth for our experiments as the utility node to estimate. Forty-two days is a gold standard when understanding short-term changes in CRF \cite{Brooks1996Exercise2}. Unsatisfied with this static parameter, we experimented to find out the optimum time to be considered for aggregating the metrics while estimating CRF values. We plotted the test error for the global models utilizing one metric in Figure \ref{fig:parameter}. We can see that while there is some variation in mean error, the 95\% confidence intervals overlap for all time windows, and we cannot find an optimum time window to use in our experiments based solely on the data. Therefore we have used the clinically recommended period of 42 days to aggregate the past exercise events.

We trained two sets of linear regression models for a global model and a personal model for each feature. The personal model for each individual was trained on with a 70 percent training subset of data for the subject and tested on the remaining 30 percent data.

\begin{figure}[H]
  \centering \centerline{\epsfig{figure=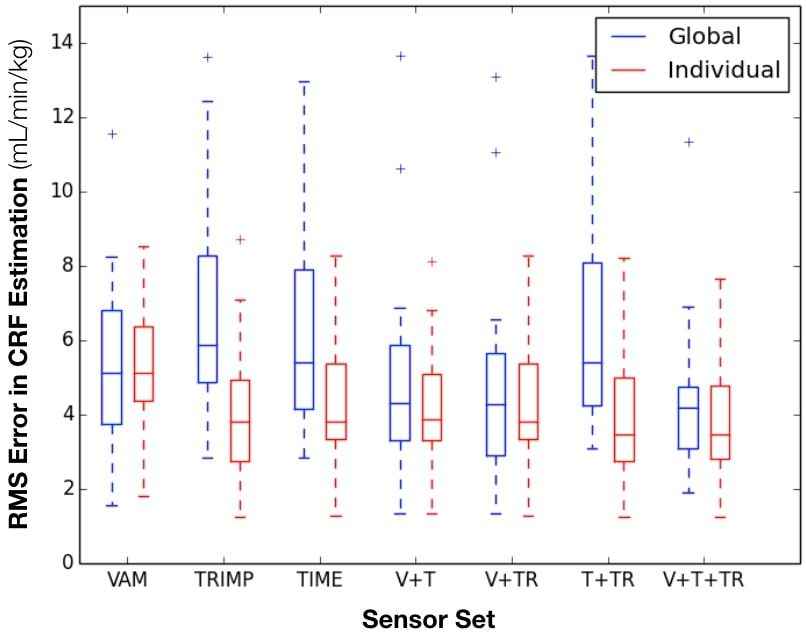,width=13cm}}
  \caption{\textbf{Cross-Modal Sensor Fusion.} By using multiple nodes in our learning model, we can see a direct improvement in the CRF utility estimation, which shows the value of having a multi-modal sensor fusion approach.}
    \label{fig:box}
\end{figure}

\textbf{VAM} models are trained to predict average power output (normalized by body weight) in 4-minute windows in an activity using the VAM in the time window, and the maximum estimated power output is then used to compute a continuous daily estimate for VO2max \cite{Cintia2013EnginePerformance}. We trained models with varying slope thresholds to identify the impact of slope on estimate accuracy. As the slope increases, the effect of other resistance factors (such as wind, rolling resistance) decreases, and the model performs better (Table \ref{table:slope_error}). We choose which model to use based on the maximum slope observed in the 4-minute windows, for example, if in a ride the maximum slope observed in a 4-minute interval is 5.3\%, we would choose the model trained on intervals where the slope is greater than 5\%. Since we are predicting body weight normalized power using VAM, none of the two metrics are greatly influenced by individual parameters. This is reflected in similar global and individual model performances for VAM (Figure \ref{fig:box}).

\textbf{TRIMP} is a method to quantify training load, and considers the intensity of exercise calculated by heart rate during exercise \cite{Halson2014MonitoringAthletes}. TRIMP captures the workload on an individual's heart in the last 42 days. We trained a linear regression model to predict an individual's VO2max value based on their total TRIMP score in the past 42 days. This metric proved to be more effective in a personal model than a global model as different individuals have different heart rate responses to the same exercise intensity, Fig \ref{fig:box}

\textbf{Active time} is the actual amount of time the individual was actively putting in effort in the past 42 days. We obtained this metric using cadence values collected at per the second resolution. We also trained a linear regression model to predict an individual's VO2max value based on their total activity time in the past 42 days. Similar to TRIMP, this metric performs better in a personalized model than in the global model as different individuals would have a different response to the same exercise volume (fig. \ref{fig:box}).

\textbf{Combination models} outperform their constituent models in all our experiments, as shown by the error plots in Fig \ref{fig:box}. The estimates from the previous models were combined using a weighted average, where weights for a model estimate are inverse of the model's training error. The error in estimates for these models is reported in Figure \ref{fig:box} and discussed in this section.
We can see from the plot that the best model in terms of average error and variance in error utilizes all available data streams.

\begin{table}[hbt!]
\centering
\caption{\textbf{Slope optimization of VAM models.}}~\label{table:slope_error}
\scalebox{0.9}{
\begin{tabular}{|l|l|l|l|l|} 
\hline
\begin{tabular}[c]{@{}l@{}}Slope~\\threshold (\%)\end{tabular} & \begin{tabular}[c]{@{}l@{}}Test Set RMSE~\\(Rel. Power)\end{tabular} & \begin{tabular}[c]{@{}l@{}}Training Set~RMSE~\\(Rel. Power)\end{tabular} & \begin{tabular}[c]{@{}l@{}}Training~\\R Squared\end{tabular} & \begin{tabular}[c]{@{}l@{}}Size of~\\training set \end{tabular}  \\ 
\hline
0+                                                             & 0.726                                                                & 0.665                                                                    & 0.381                                                        & 16810792                                                         \\
1+                                                             & 0.620                                                                & 0.537                                                                    & 0.527                                                        & 10728610                                                         \\
2+                                                             & 0.557                                                                & 0.473                                                                    & 0.593                                                        & 7952334                                                          \\
3+                                                             & 0.488                                                                & 0.424                                                                    & 0.655                                                        & 6236507                                                          \\
4+                                                             & 0.451                                                                & 0.391                                                                    & 0.695                                                        & 4884243                                                          \\
5+                                                             & 0.420                                                                & 0.363                                                                    & 0.732                                                        & 3509724                                                          \\
6+                                                             & 0.405                                                                & 0.344                                                                    & 0.760                                                        & 2294199                                                          \\
7+                                                             & 0.395                                                                & 0.328                                                                    & 0.781                                                        & 1400488                                                          \\
8+                                                             & 0.365                                                                & 0.318                                                                    & 0.793                                                        & 781841                                                           \\
9+                                                             & 0.347                                                                & 0.317                                                                    & 0.797                                                        & 423315                                                           \\
\hline
\end{tabular}
}
 \caption{\textbf{VAM Parameter Optimization.} For the VAM based model, there is a beneficial effect to choose samples in which the slope gradient of the activity is steeper, which quantifies how much weight to assign to data samples through the quality metric of slope in our learning model.}
\end{table}

In these experiments, we leverage cross-modal data to estimate a needed utility node. From our experimental results, we can see that increasing the number of data streams provide increased performance characteristics in achieving this goal.

The total health of an individual is much higher than any single node attribute, and that we need to understand better the total health state of a particular individual or organ system. We cover some actionable and semantically meaningful visualizations to see how to use the HSE.

\section{Other Visualizations of HSE}
\begin{wrapfigure}{l}{0.5\textwidth}
    \includegraphics[width=0.48\textwidth]{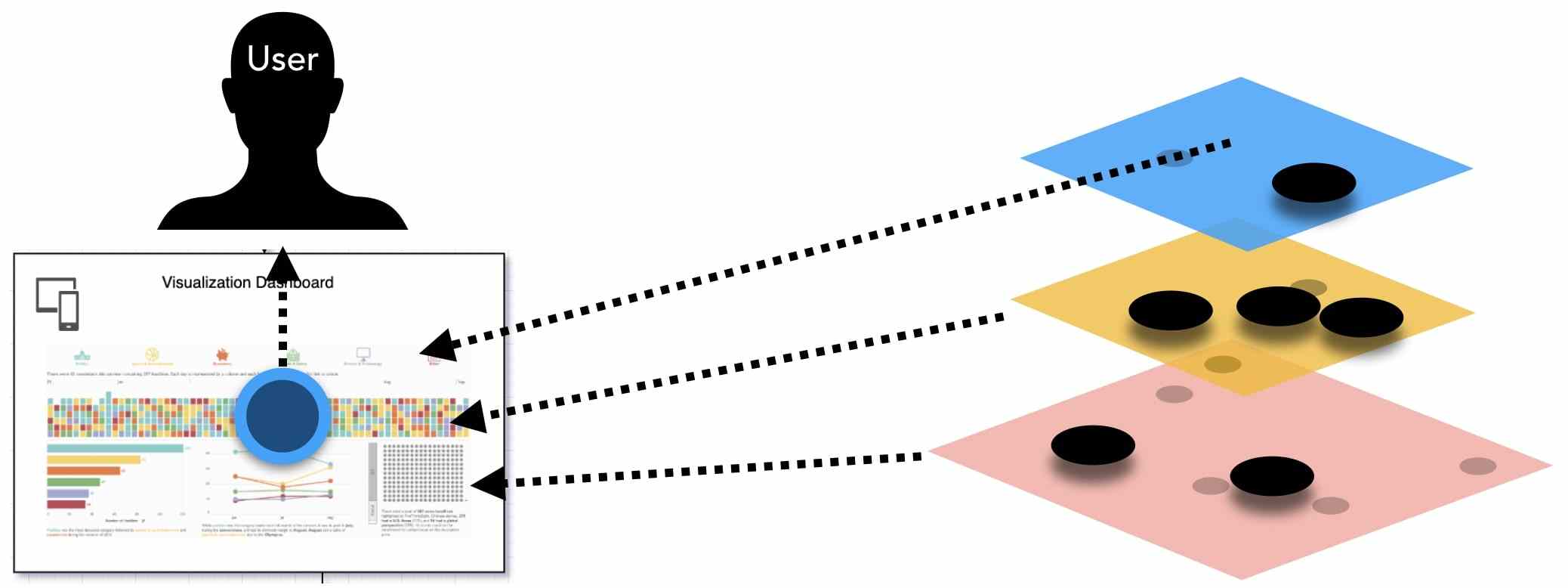}
      \label{fig:vis}
\end{wrapfigure}

\begin{figure}[H]
  \centering \centerline{\epsfig{figure=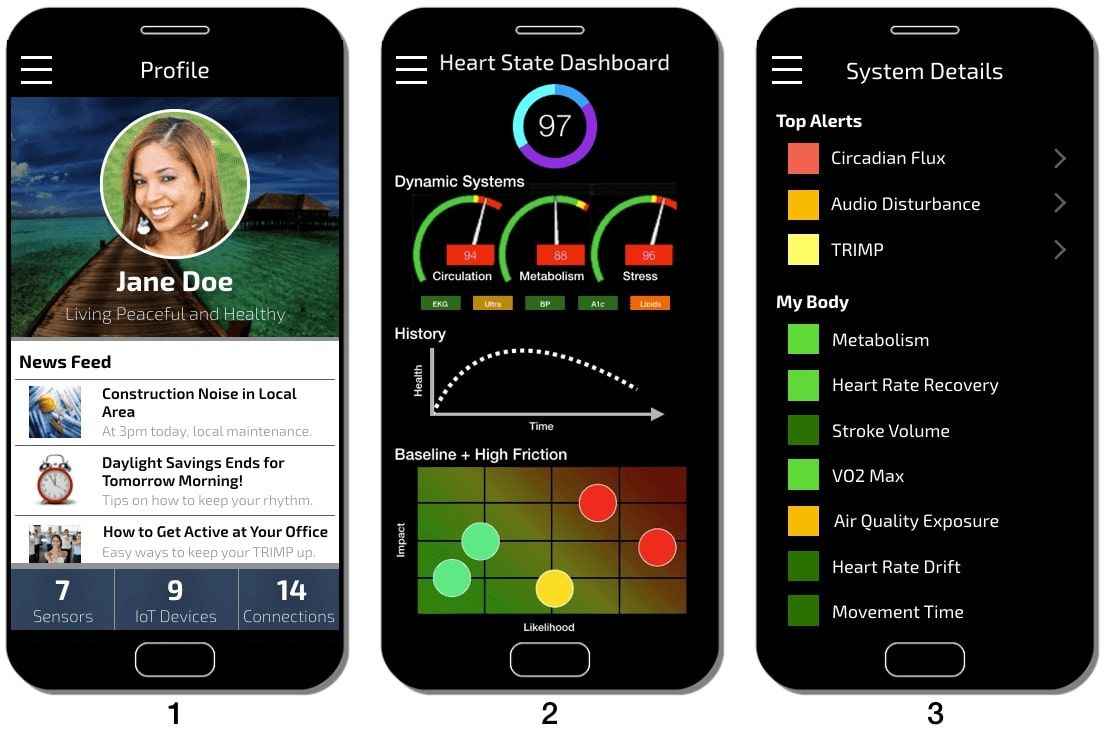,width=12cm}}
  \caption{\textbf{Personal Mobile Dashboard View.} The personal mobile dashboard view is a concept design mock-up of how a user may access their health state. Screen 1 we show how health state can fuel recommendations. Inspired by jet engine dashboards, screen 2 can give a live snapshot of the health state. Screen 3 provides a comprehensive list of all bio-variables and health states being tracked, with a ranking system to provide relevant results at the top.}
    \label{fig:mobile}
\end{figure}

How can we further assimilate HSE in a useful way for individuals and health providers? A first step is in the visual analysis of the HSE by users and health experts. Simple mobile applications such as the system mocked in Figure \ref{fig:mobile} can give a quick overview of the HSE, similar to checking the financial market state or weather state. Expanding this idea to a population provides insight in a detailed manner that can be useful to scientists, researchers, and health experts. We go over an experiment to show such a system in the next section.

\subsection{Multi-Person View}
\begin{figure}[H]
  \centering \centerline{\epsfig{figure=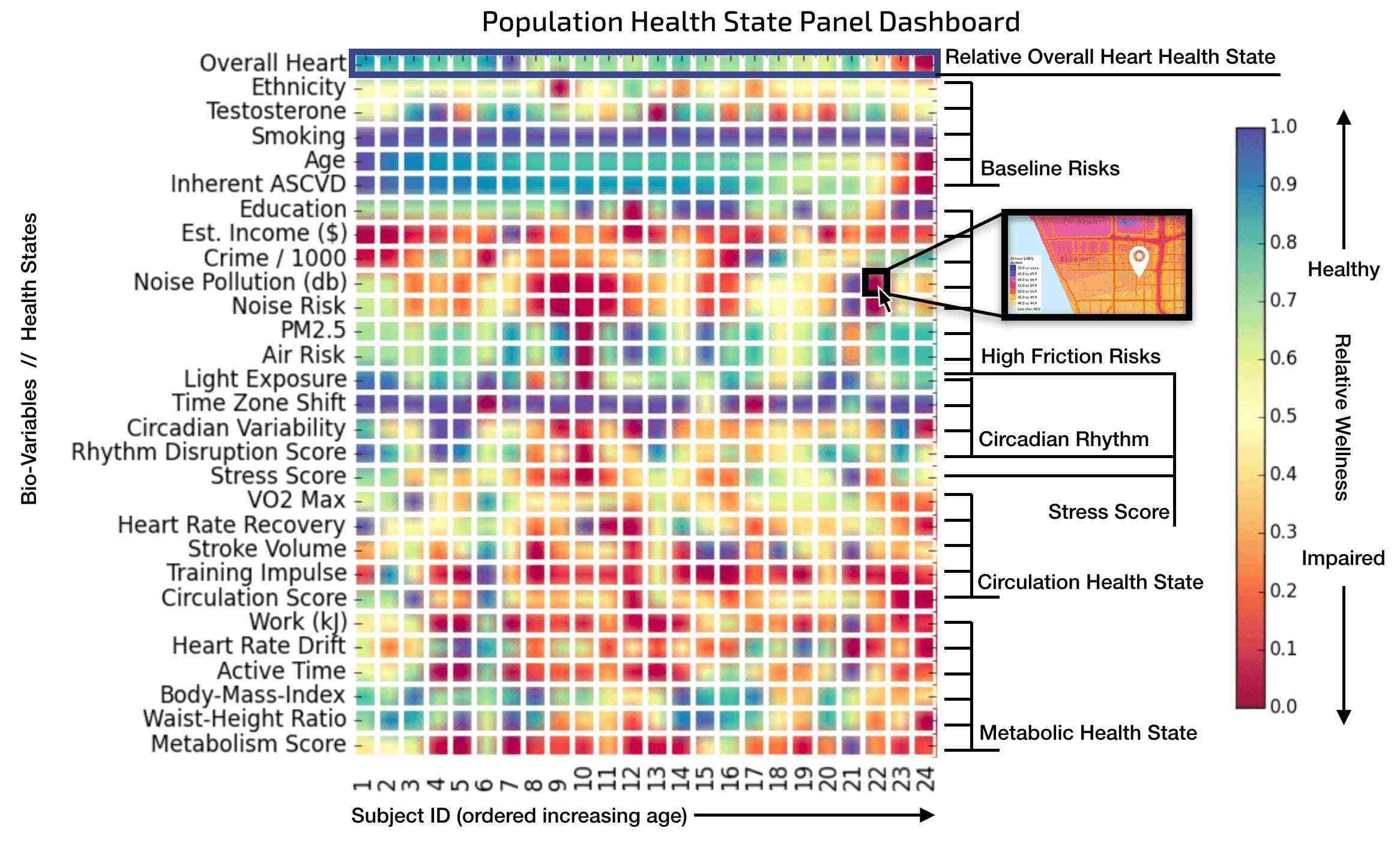,width=14cm}}
  \caption{\textbf{Patient Dashboard View.} The patient dashboard view is a heat map of bio-variables and summary scores that affect each subject. Variables are normalized to a specific color for each box. The variables are graded on a scale ranging from blue being most healthy and red being least healthy. This type of visualization integrates cross-modal data in a manner that a clinician, hospital, public health agency, or any domain expert can use to monitor the health of a patient panel or population. Clicking on a particular box would pull up further insights and details. This highlighted box in this image shows the noise pollution insight for patient 22.}
    \label{fig:multi}
\end{figure}

Using the same HSE approach for the test subject, we applied this to multiple subjects to produce a health state overview for population health analysis. The dataset used in these experiments includes the same activity sensor data streams at one-second resolution collected on 24 male cycling athletes over an average of 5 years in the United States. Athletes also had strain gauges installed on their bicycles to measure true physiologic power output. The total dataset includes 31,776 activities and 70,178 hours of exercise data (252.6 million data intervals). Social media outlets of Instagram and LinkedIn were also used to gather an image dataset of 50 images per athlete and general demographic background information. From this, we assimilate various CRF biological and environmental nodes through the health state estimation using this data for each of the 24 subjects, as shown in Figure \ref{fig:multi}. We find that even though most of these subjects are all cycling athletes, they have a wide range in these nodes. Modern-day primary care doctors would not be able to see this when a patient visits. Looking at this data panel across the subject pool, we discover some interesting trends in Figure \ref{fig:multi}. As expected, as the age increases the overall heart health state decreases, since age is a large factor of cardiac health (age range in the panel is 18-57). As age increases, we also see a reduction in crime and noise pollution, suggesting that older individuals can afford to live in safer and quieter neighborhoods. We also see circadian rhythm disruption maximally in the middle ages (20-29), suggested a more erratic lifestyle for those in their twenties. Circulation and metabolism scores also trend (including VO2max / CRF) lower as age increases. Although this is a small sample size to make any strong conclusions, we can begin to see the usefulness of using this cross-modal analysis for clinical applications. This work was published at Association for Computing Machinery's (ACM) Multimedia 2018 in the Brave New Idea track \cite{Nag2018Cross-modalEstimation}.

\begin{figure}[H]
  \centering \centerline{\epsfig{figure=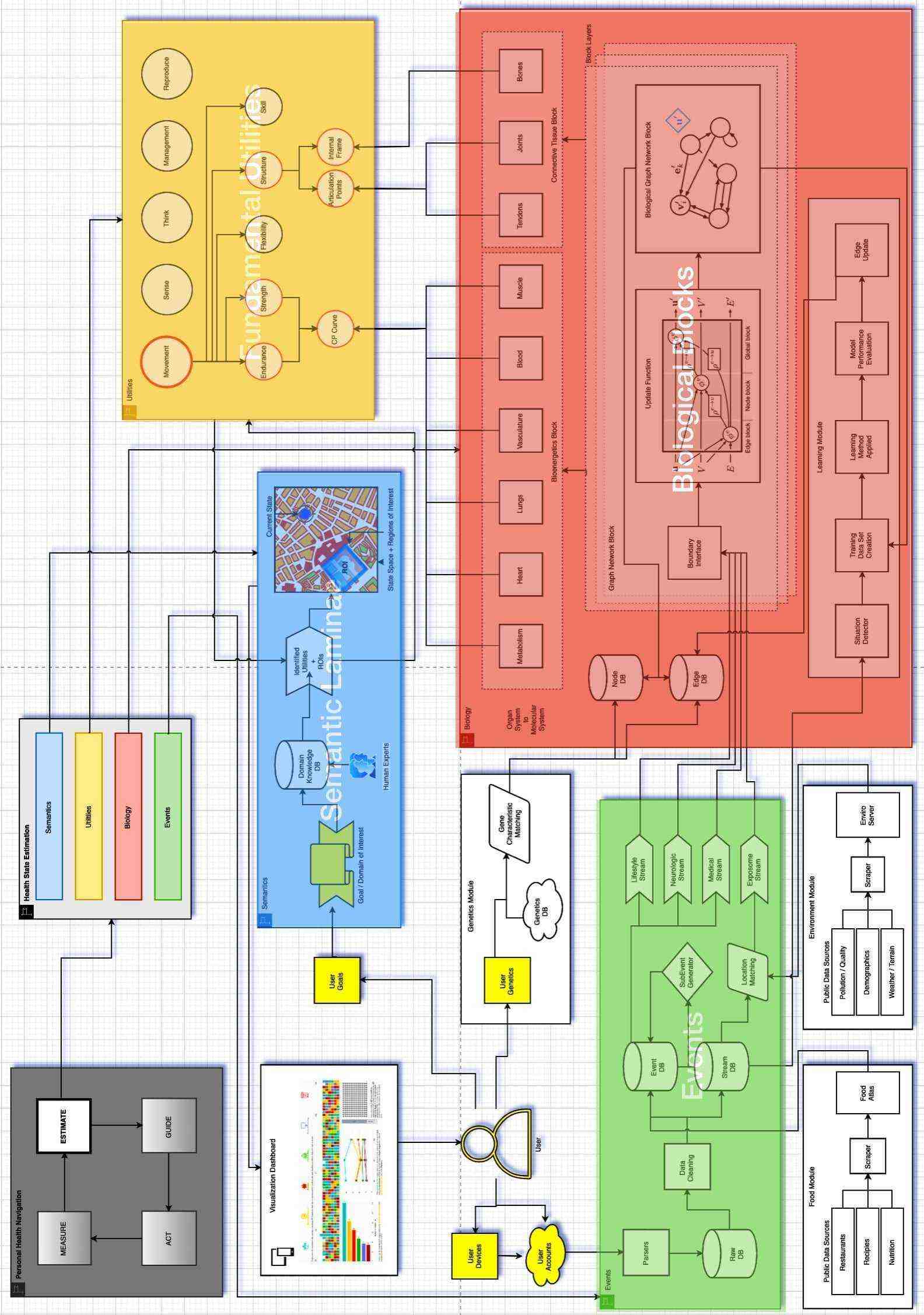,width=14cm}}
  \caption{\textbf{Total Health State Estimation System.} Here we can see all the blocks of Chapter 5 in a system overview.}
    \label{fig:system}
\end{figure}

Concluding this chapter, we see how the framework in Chapter 4 can be implemented at a detailed level for estimating the health state of an individual, and how this can scale over time and across more individuals. This working framework is shown in Figure \ref{fig:system}. The next chapter explores ideas within the original inspiration of using HSE for navigation and control. This exploration is intended to extend HSE beyond this dissertation.

\chapter{Personal Health Navigation}

\epigraph{Knowing is not enough; we must apply. Willing is not enough; we must do.}{\textit{Johann Wolfgang von Goethe}}
\vspace{12pt}

Knowing the health state provides information about one's health situation; however, individuals need to take action with this information. Akin to GPS for position, the possibilities to take action with this information are vast. Making decisions about events in one's life allow individuals to change their state towards their wishes. These chosen events become experiences, which are valuable for determining a person's quality of life. If we define quality of life as a function of positive experiences and time, then we want to optimize those experiences and extend their duration. As discussed, health is a dynamic state that is continually changing, based on biology, the environment, and lifestyle. Technology that senses a change in a user's position on the health continuum can drastically improve human health, especially once it is actionable. If people become aware of their biology, they could see how daily life is affecting their health and make informed decisions accordingly. Health decisions are too important to be episodic; they must be an intrinsic part of daily life. We propose the navigation perspective, to make actionable guidance for improved quality of life \cite{Nag2019ALife}. This chapter discusses in  detail the guidance and action components of navigation.

\begin{figure}[t!]
\begin{infobox}[A Navigational Approach to Health]
\begin{center}
\includegraphics[trim={0 -0.2cm 0 0.4cm},width=0.95\textwidth]{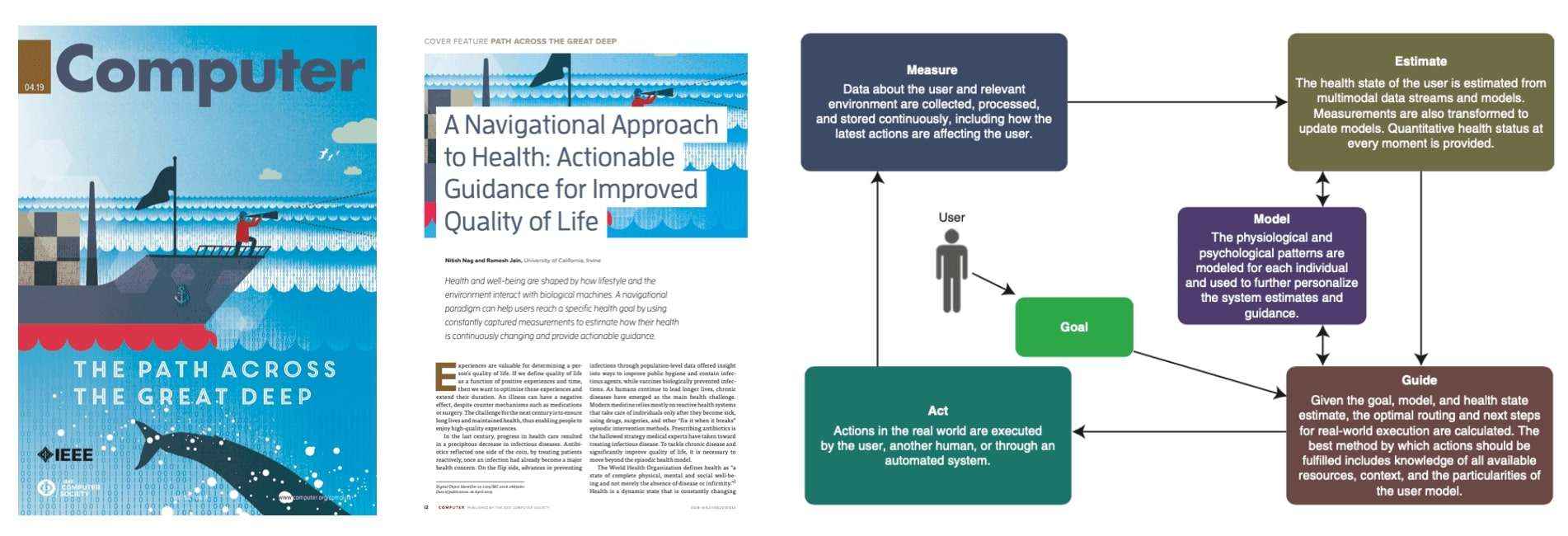}
\end{center}

Health and well-being are shaped by how lifestyle and the environment interact with biological machines. A navigational paradigm can help users reach a specific health goal by using constantly captured measurements to estimate how their health is continuously changing and provide actionable guidance. \textit{Cover Issue IEEE Computer April 2019} \cite{Nag2019ALife}
\begin{description}[noitemsep]
\item[Measure]: Data about the user and relevant environment are collected, processed, and stored continuously, including how the latest actions are affecting the user.
\item[Estimate]: The health state of the user is estimated from multi-modal data streams and models. Measurements are also transformed to update models. Quantitative health status at every moment is provided.
\item[Guidance]: Given the goal, model, and health state estimate, the optimal routing and next steps for real-world execution are calculated. The best method by which actions should be fulfilled includes knowledge of all available resources, context, and the particularities of the user model.
\item[Action]: 
Actions in the real world are executed by the user, another human, or through an automated system.
\item[Model]: The physiological and psychological patterns are modeled for each individual and used to further personalize the system estimates and guidance.
\end{description}

\label{box:navpaper}
\end{infobox}
\end{figure}

\begin{figure}[t]
  \centering
  \includegraphics[width=0.8\textwidth]
  {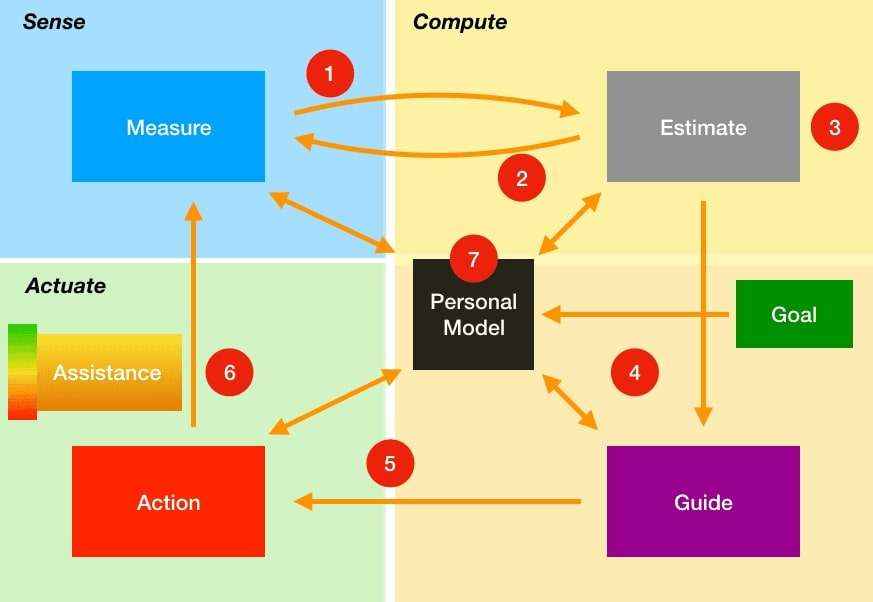}
  \caption{\textbf{Navigation Vision for Health.} Navigation for health has four key components: Measure, Estimate, Guide, Action. Step 1: Take measurements in the real world through a variety of sensors. Step 2: Additional measurements may be needed if they are not sufficient for a reasonable estimate. Step 3: From these measurements, we can produce an Estimate of the current health state. Step 4: Guide the user based on a reference goal and the user's model. Step 5: Take actions in the real world through a variety of mechanisms. Step 6: Assisted mechanisms may contribute to the executing of these actions. Step 7: Retake measurements to understand how the person's health state has changed, and this updates the personal model perpetually.}
  \label{fig:mega}
\end{figure}

Navigation is different from recommendation systems and purely cybernetic control. This goal-based guidance system perpetually estimates the current state, computes the best route through intermediate states, and guides actions that lead to health goals. For success, this system requires robust continuous sensing, accurate estimation of the current state, control systems, actuation mechanisms, understanding of the outside world and context, long-term planning, and goal decomposition at various levels of detail. Navigation must also manage stochastic, chaotic, and noisy environments effectively. This goal-driven, closed-loop sense-compute-then-act cycle combines the pioneering elements from cybernetic systems (well described in electromechanical systems and biology by Wiener et al. and the general problem solver by Newell et al. \cite{Wiener1962CyberneticsMachine,Newell2013GPSThought}.

The purpose of personal health navigation (PHN) is to help an individual reach and maintain their desired health state. Systems that provide health navigation are broad, and humans may be integral components. Examples include a personal device such as a mobile phone and an entire Internet of Things (IoT) ecosystem within a building, a human trainer at a gym, or an advanced cancer center. PHN starts with a user-specified health goal. From there, the system starts to collect measurements about the person’s health that relate to his or her goal to estimate current health states. To most optimally reach the desired goal, the system breaks the steps from the current state and provides the next step to an actuation mechanism. All actions, whether advised by the system or not, are continuously measured to provide a new estimation of the individual’s health state. A change in the state updates the next action. A cycle of these actions moves the user’s current health state closer to the target. Upon reaching the destination, the system continuously ensures minimized deviation from that state. The vast unknown nature of health means that we will obtain more diverse multi-modal data in the future and that this phenomenon requires combining these data sources.

While navigation is not the core of this dissertation work, it establishes the purpose of the health state estimation and how to best guide the better health of an individual. In this chapter, we will discuss the step by step process of personal health navigation and work that has been done to address certain challenges \cite{Nag2019ALife}.


\section{General Health State Space}
\begin{figure}[t]
  \centering
  \includegraphics[width=.85\textwidth]
  {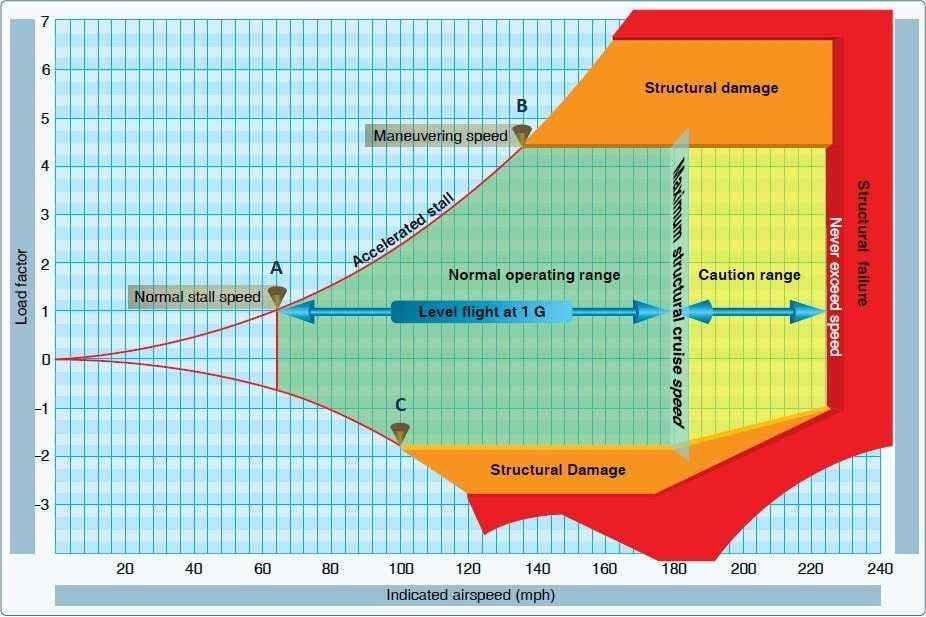}
  \caption{\textbf{Flight Envelope.} This graph shows the area of total possible states in which the plane is able to fly given its unique parameters.}
  \label{fig:envelope}
\end{figure}

The first step of navigation is establishing the general health state space, as described in Chapter 4. The ``flight envelope" of an aircraft is the inspiration for the state space concept. The envelope encompasses measurements (load factor, atmospheric density, altitude, etc.), which determines the space in which it is safe for the plane to fly in, as shown in Figure \ref{fig:envelope}. Similarly, the state space defines all possibilities and capabilities to understand the scope in which the navigation will occur. This state space includes all possible health states a human can exist in. For example, if the state space is the cardiovascular state, all possible measurements for the cardiovascular state will be considered, including all  components of fitness (with specific measurements such as VO2max), cardiovascular disease, and more.  

\begin{figure}[t]
  \centering
  \includegraphics[width=1\textwidth]
  {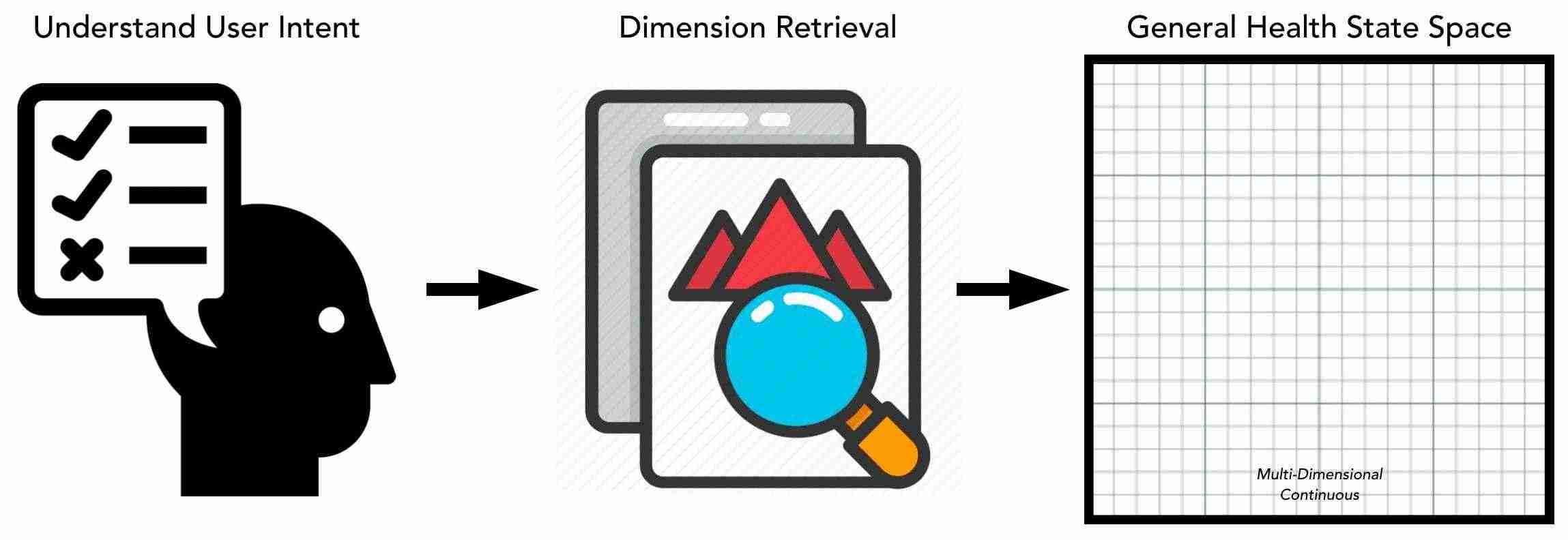}
  \caption{\textbf{General Health State Space.} The state space includes all the possible states that all humans can exist, with several dimensions. In order to interact with this state space effectively, a user may provide a particular intent, which can reduce the dimensionality to the relevant regions to visualize the general health state space.}
  \label{fig:nav1}
\end{figure}

\section{Personal Health State Space}
When we apply the state space to an individual, we create a personal health state space, which includes all the possibilities for an individual given their situation. This is by definition a subset of the general health state space. Each person has a unique and finite possibility of biological states. Classic examples such as height, eye color, or fingerprints have a small set of states after reaching adulthood. However, many aspects of biology have an wide range of possible states, which change based on daily life. Consider how exercise changes the mitochondrial content of cells, how nutrition alters our metabolic processes, how medications modulate our blood pressure, and how ultraviolet light damages DNA. The changes in these states are what determines the movement in the state space. Mapping the personal health state space (PHSS) in totality is a complex, multidimensional challenge. We can apply this concept to what we observe in flight envelopes, in which specific models of planes have unique state spaces. Each type of plane will have its possibilities by which it can fly. Narrowing the state space to the individual's possibilities give a more specified scope best set up for personalized navigation.

\begin{figure}[t]
  \centering
  \includegraphics[width=1\textwidth]
  {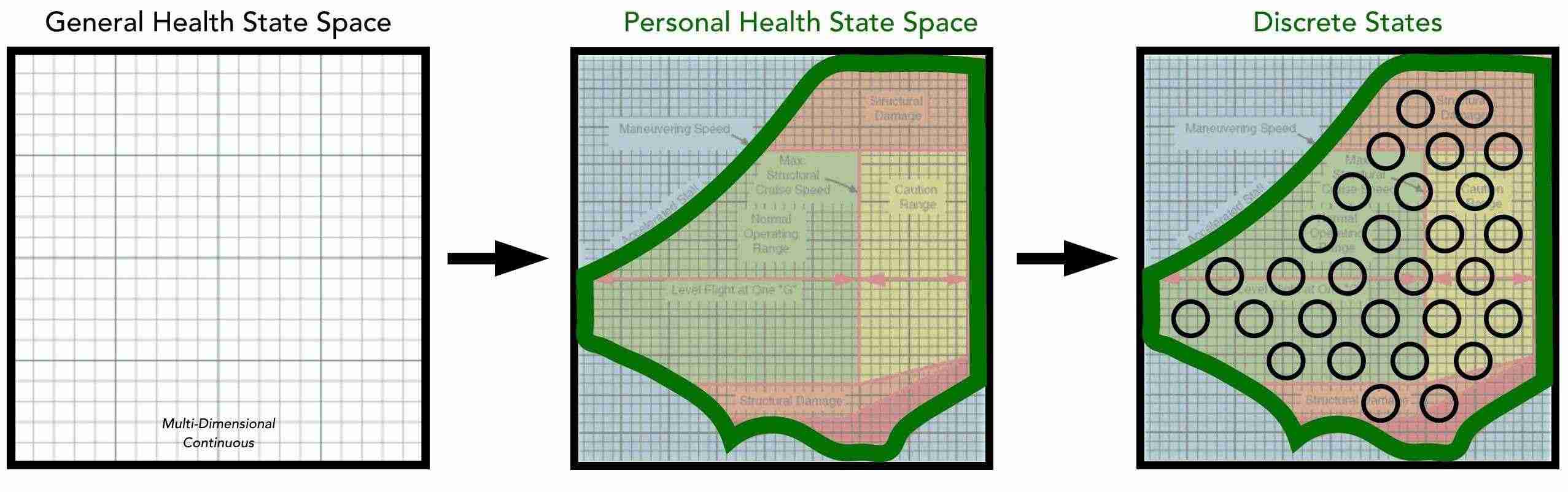}
  \caption{\textbf{Personal Health State Space.} The PHSS includes all the possible states that a specific human can exist at the current moment and at future time points. This state space is a subset of the general health state space. To convert this space to a computationally efficient world, we may apply a discrete layer on top of this state space, allowing for graph-based approaches for further applications.}
  \label{fig:nav2}
\end{figure}

\section{State Transition Network}
    Within the personal health state space, there are connections made between each of the individual state possibilities. The connections themselves are the state transition network. Each edge within the network represents transitions that take the person from one node to another. With the cardiovascular state, for example, we determine which inputs will transition the heart state. The state transition takes into consideration all the inputs that will lead to the next state, therefore making connections or relationships between states. If one were to exercise, experience stress, or have a high salt meal, they are transitioning between these states in their PHSS. To understand the PHSS in more detail, we must identify the knowledge layers and regions of interest within the space relevant to the topic of interest. Connections within a PHSS between two nodes will be unique based on the individual. For example, for person A and B to grow one pound of muscle, the inputs needed for A will be different than person B. Therefore the state transition network is also a unique part of the PHSS.

\begin{figure}[t]
  \centering
  \includegraphics[width=1\textwidth]
  {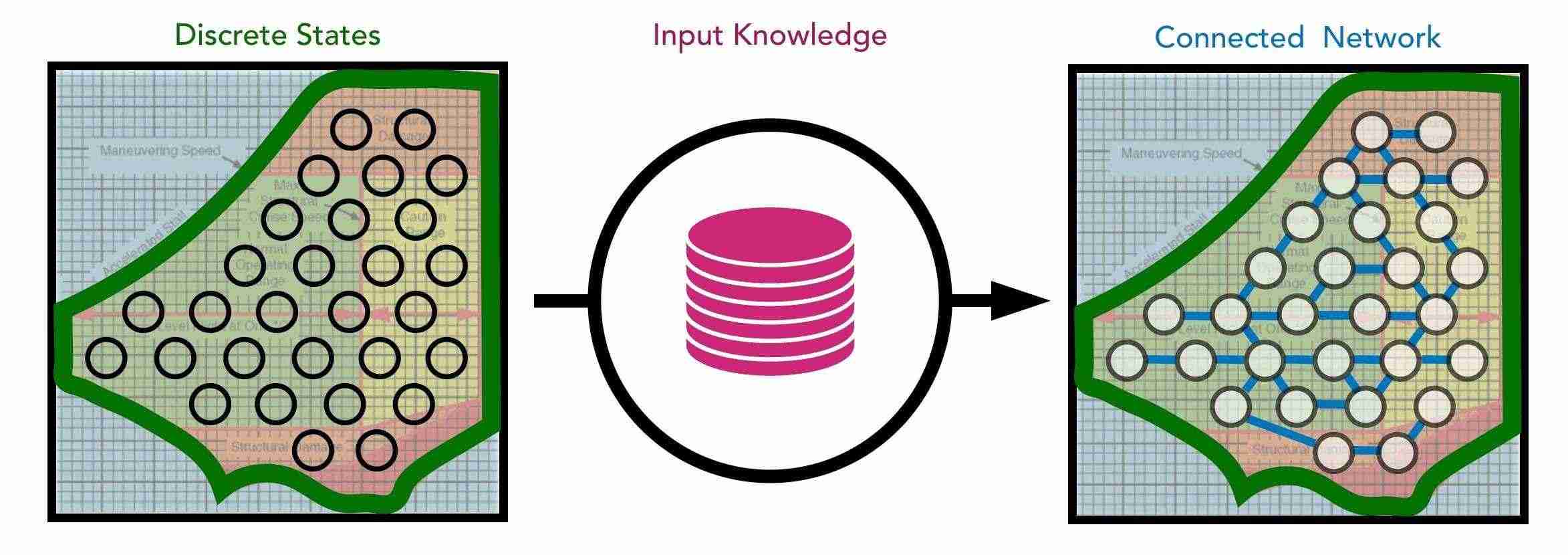}
  \caption{\textbf{Connected State Space Network.} The state space network includes all the possible transitions from one state to another for the specific individual. The connections contain the information of what inputs the matrix will cause a particular state to convert to the next state. The edges here are directed (although not visually shown). These connections are not static and are repopulated, given a reflection of the current situation.}
  \label{fig:nav3}
\end{figure}


\section{Knowledge Layers and Regions of Interest}
Layers on top of the PHSS contains all relevant health domain knowledge, similar to physical maps that use latitude, longitude, and altitude to describe the globe independently of any knowledge layers. Information layers such as roads, oceans, country borders, and satellite imagery allow for navigation within the space, depending on the context (driving requires roads and traffic layers, while flight requires air class, way-points, and airport layers). Knowledge layers represent the real world, in which humans can make sense of their state concerning their interests. A semantic translation to the relevant PHSS will depend on the user's region of interest. For example, mapping the coordinates of a ``cyclist" in the PHSS requires matching relevant coordinates within the desired region of interest in the state space. This region of interest is a particular area within the state space relative to where the individual currently is in their health status. A defined distance allows the individual can see how close or far they are from being in this region and the various nodes or states that would cause one to get there. Similarly, one can determine their risk for heart disease based on their current state's distance from an undesirable region. Mapping current location distance from regions of poor health brings about a new approach to quantify one's relationship to disease risk. 

\begin{figure}[t]
  \centering
  \includegraphics[width=1\textwidth]
  {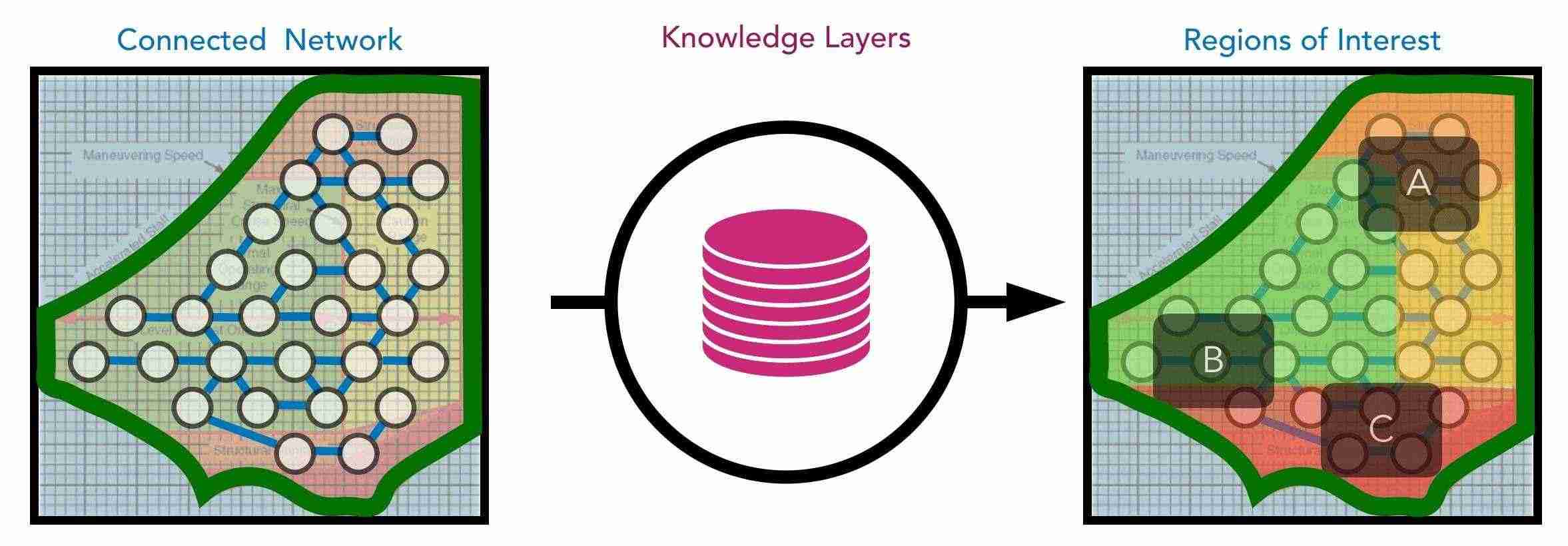}
  \caption{\textbf{Knowledge Layers.} Layering information on the state space allows for semantic interactions. Domain knowledge specifies these layers from the relevant database. We can gather specific regions of interest from knowledge sources that can label the state space so a user can understand the possibilities for their health in natural language or a relevant visualization. In this example, specific regions of interest are labeled with the colors of green, orange, yellow, and red. Red is an example of an undesirable region, while blue represents regions that are not within the person state space possibilities of the user. The black boxes represent points of interest, which is essentially a more confined region of interest.}
  \label{fig:nav4}
\end{figure}

\section{Current State Estimation}
It is important to note that there is constant flux within an individual's PHSS based on one's location in the state space. For success, navigation must be able to assign an accurate location within the PHSS, just as GPS provides physical world navigation. Understanding this location is the HSE (all previous chapters). The same HSE tool may be useful for many applications. For example, monitoring a cardiovascular health state is useful to both endurance athletes and heart disease patients. Estimation techniques have been of great interest in designing many applications, but health applications will require increasingly deep biological knowledge layers to define and estimate health states. This estimation could indicate its proximity to biological dysfunction or its capacity for the desired goal.

\section{Goal Decomposition}
For navigation to occur, an individual's requires specifying a goal state. Goal states can be specified as ROI or a specific point. If an individual has a region of interest for becoming a cyclist, they can more specifically choose a pointed goal such as competing in a 100-mile road cycling race. Goal decomposition is needed to translate the goal state into short-term goals. This process occurs through breaking down the goal to identify the unique utilities relevant for the specified goal. 

First, in this process, the system will need to translate the semantic goal to the PHSS. Because the PHSS is large, we must break down a macro health goal into several intermediate states and subgoals that the system can focus on reasonably. There may be a set of short-term goals at any given moment, each with its own set of intermediate states and subgoals. At the lowest atomic level for an application, PHN lists fulfillment tasks in priority order so that users can execute them in the real world to attain a subgoal or the next intermediate state.

To become a road cyclist, for example, there will need to be a process broken down into  the key utilities (i.e., develop muscle strength, improve cardiovascular health, ensure body alignments). Each of these intermediate steps will have various subgoals required for the optimization process. If the individual's utility is to develop bone strength, they will need adequate nutrition and weight training. Each subgoal translates to a smaller state space. In the case of nutrition, models will list macro (carbohydrates, protein, fat) and micro (vitamins, minerals) nutrients. These components may then be the chosen atomic-level intermediate states for applying nutrition.

\subsection{Timescales}
Goals and corresponding subgoals occur at various timescales. Take for example the cyclist whose race is in six months. The cyclist must maintain daily subgoals such as establish cardiovascular endurance (i.e., a two-hour bike ride, four days a week). We can then break it down into minute by minute subgoals to maintain a particular power output for interval training, to build a higher power to weight ratio. When looking at the biological level, the subgoals may be positive gene expression for mitochondrial biogenesis and improved metabolism, which has a second by the second timescale.

\subsection{Recurring Goals versus Achievement Goals}
There are two types of goals. One is a recurring goal that is an ongoing goal, such as a habit or way of living. For example, brushing teeth would be considered a recurring goal that an individual maintains over time. This requires constant navigation till the recurring goal is terminated by the user. There are also achievement-based goals, which are defined situations that once achieved terminate the navigation. Once the user meets the achievement (such as winning a race), they must then set a new goal.
    
\subsection{Synergistic Goals versus Competing Goals}
We must note that there may be multiple goals that are either working in synergy or competing with one another. For example, if an individual has the goal for high performance in a road race and also would like to be an enduro mountain bike racer, there may be overlap in synergistic goals, such as the cardiovascular training required to perform in both types of events. However, as enduro mountain bike racing relies on high skill in demanding terrain and upper body strength (therefore increased weight), it will conflicting with the training for high endurance activities, such as the 100-mile road race which require being at a lighter weight.
    
\subsection{Identifying User Motivations for Goals}
Identifying user motivations is critical for goal setting and, ultimately, the effectiveness of any lifestyle change. Engagement is dependent on providing recommendations that primarily fuel intrinsic motivation while using extrinsic motivation for further support. This understanding builds the foundation of sustainable behavioral adaptation for optimal personalized health benefits. \textbf{Intrinsic motivation} arises from internal enjoyment or satisfaction in a particular task, while \textbf{extrinsic motivation} refers to motivating factors from the outside world, such as rewards or punishment. While both can lead to lifestyle activities that benefit health, intrinsic motivation drives long term user engagement. 
    
We can use multi-modal data to learn if a user is executing an action from intrinsic motivation or external factors. For instance, if a user repeats an activity without prompting and without external influence, this gives us an increased weighting that the motivation is intrinsic. If we detect themes and concepts in their social media consumption, and one takes part in an activity when a reward prompts them, we can consider this as more weighting towards the extrinsic motivation \cite{Nag2018IntrinsicSystems}.

\begin{figure}[t!]
\begin{infobox}[Intrinsic and Extrinsic Motivation Modeling Essential for Multi-Modal Health Recommender Systems]
\begin{center}
\includegraphics[trim={0 -0.2cm 0 0.4cm},width=0.95\textwidth]{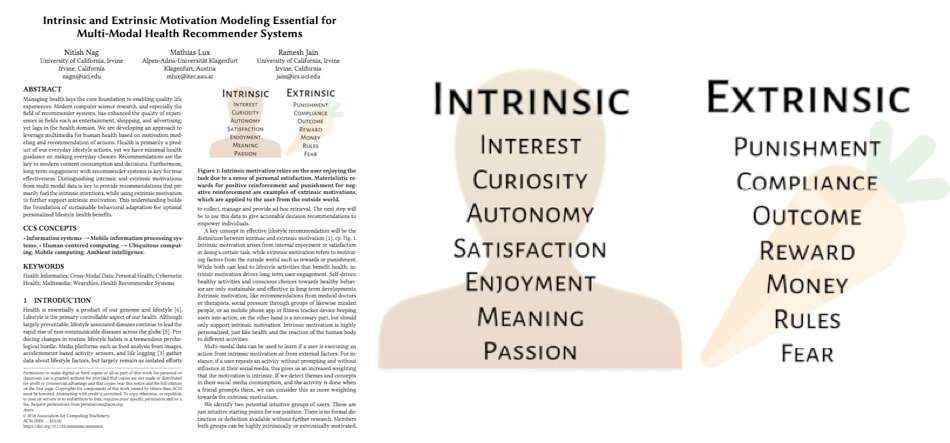}
\end{center}
Distinguishing intrinsic and extrinsic motivations from multi-modal data is key to provide recommendations that primarily fuel the intrinsic intentions, while using extrinsic motivation to further support intrinsic motivation. This understanding builds the foundation of sustainable behavioral adaptation for optimal personalized lifestyle health benefits. This paper was written in collaboration with Mathias Lux of Alpen-Adria-Universität Klagenfurt, Austria. \textit{Arxiv, August 2018} \cite{Nag2018IntrinsicSystems}
\begin{itemize}[noitemsep]
\item Understand what components are intrinsic and extrinsic motivators for an individual are key to develop more intelligent recommendations to guide a person towards good health.
\item Using intrinsic motivations as a basis for lifestyle change is much more effective for long term engagement.
\item Extrinsic motivations are useful to engage in short bursts to push behaviors, but should be used judiciously.
\end{itemize}

\label{box:intrinsic}
\end{infobox}
\end{figure}

\begin{figure}[t]
  \centering
  \includegraphics[width=1\textwidth]
  {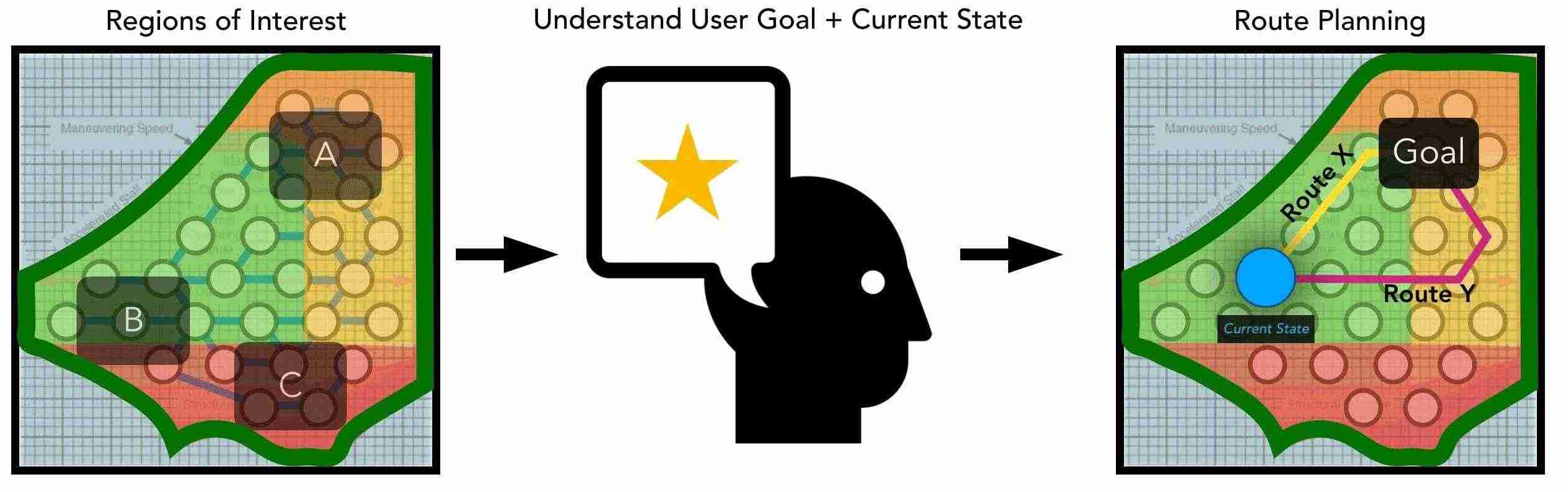}
  \caption{\textbf{Goal Selection and Route Planning.} Given a particular interaction layer, a user can select a goal of choice. We compare this goal to the current state of the user in order to determine routes that can lead to the goal state. Multiple routes may be generated based on optimization criteria that the user or the system can weigh.}
  \label{fig:nav5}
\end{figure}

\section{Guidance}
The guidance process towards the defined goal includes creating a route and recommending the next steps along the way. Computers play a role in providing both short-term guidance and longer-term guidance; however, to predict the future, provide guidance, and understand the preferences and particularities of an individual, personal models are needed. These models, which establish the premise that each individual is a unique system, are needed to best estimate the user’s health state and how various inputs uniquely move them in their PHSS. Models are then accessed to provide precise guidance towards a goal state. The personal model can build knowledge and predict many aspects of an individual’s life, such as how a person reacts to different stimuli under specific conditions or physiological change from an intervention. To model a person, a combination of long- and short-term information must interact. Long-term models of an individual can come from the genome or the history of event patterns and behavior. Biological or high-resolution continuous sensors can capture short-term models. These models are not static; they continuously change with an individual’s age, life events, and other life parameters, which makes model-building a dynamic and causal understanding process. 

After measuring, estimating,  modeling the individual, and receiving a goal the user needs to receive guidance on the next step needed to reach that goal. For an individual to make decisions about events leading up to their goal state, the system routes through intermediate steps through the PHSS. At each moment in time instructions are given for the next appropriate actions.

\subsection{Routing}
Having a map, location, and goal (or destination) now sets the stage for routing. As we cannot go back to previous time points, we must move forward through intermediate states over time towards a desired goal state. Making a route on a map requires not only knowing the start and endpoints but also all the layers of roads and traffic. In the case of PHN, each set of intermediate states and sub-goals will have its layer of information, which is relevant for mapping, along with costs and constraints to transition among intermediate states. Interactions in PHN will be extremely complex due to its large dimensionality. We must handle competing user goals, through methods such as prioritization or weighting. Means–ends analysis or other problem-solving techniques, along with appropriate routing algorithms, will reveal the best intermediate states for the user to reach the goal. There may be multiple routes to get to the desired goal. However, route selection may be made by various optimization criteria that include user preferences, efficiency, speed, and resources.

For the example case of food routing for pregnancy. It is necessary to map available food options, nutrition, and locations, along with how each choice can fulfill the tasks demanded over a period of time of nine months. Consider models for breakfast, lunch, snack, and dinner in the daily sub-goal of a nutritional state. There must be coordination of total components to reach the sub-goal appropriately. For example, if breakfast does not provide enough iron, then it must be compensated for in other meals. Researchers must further consider the logistics of adding a dish into the equation. Hence, solving routing for various PHN components remains an open opportunity.

\subsection{Actions and Cybernetic Control}
Actions of the person help lead them to a specified goal. Cybernetic control pairs the individual user and digital assistance to enact these real-world changes to transition the health state. Control mechanisms steer individuals towards the state transition on various timescales. Therefore, recommendation serves as a tool for the cybernetic control in a particular moment in time. Recommendation may help plan for a single moment in time, but it does not consider routing through the state space. This is a key difference between navigation and recommendation.

When applied to health, combining sub-population and individual data dynamically into a system gives real-time personalized evidence-based actions unique for each individual through sensors and specific feedback guidance loops. In the case of the cyclist, the control mechanism for ensuring they are on target for getting to their next state is determined by the power out for each training interval from data collected from the power meter. If their power is too high or too low, the recommendation may be to adjust for the optimal power output in the live moment of training. Based on their biology and the effort during the training ride, they can have actionable next steps on what food they should eat next for optimal recovery \cite{Nag2017CyberneticHealth}.

\begin{figure}[t!]
\begin{infobox}[Cybernetic Health]
\begin{center}
\includegraphics[trim={0 -0.2cm 0 0.4cm},width=0.95\textwidth]{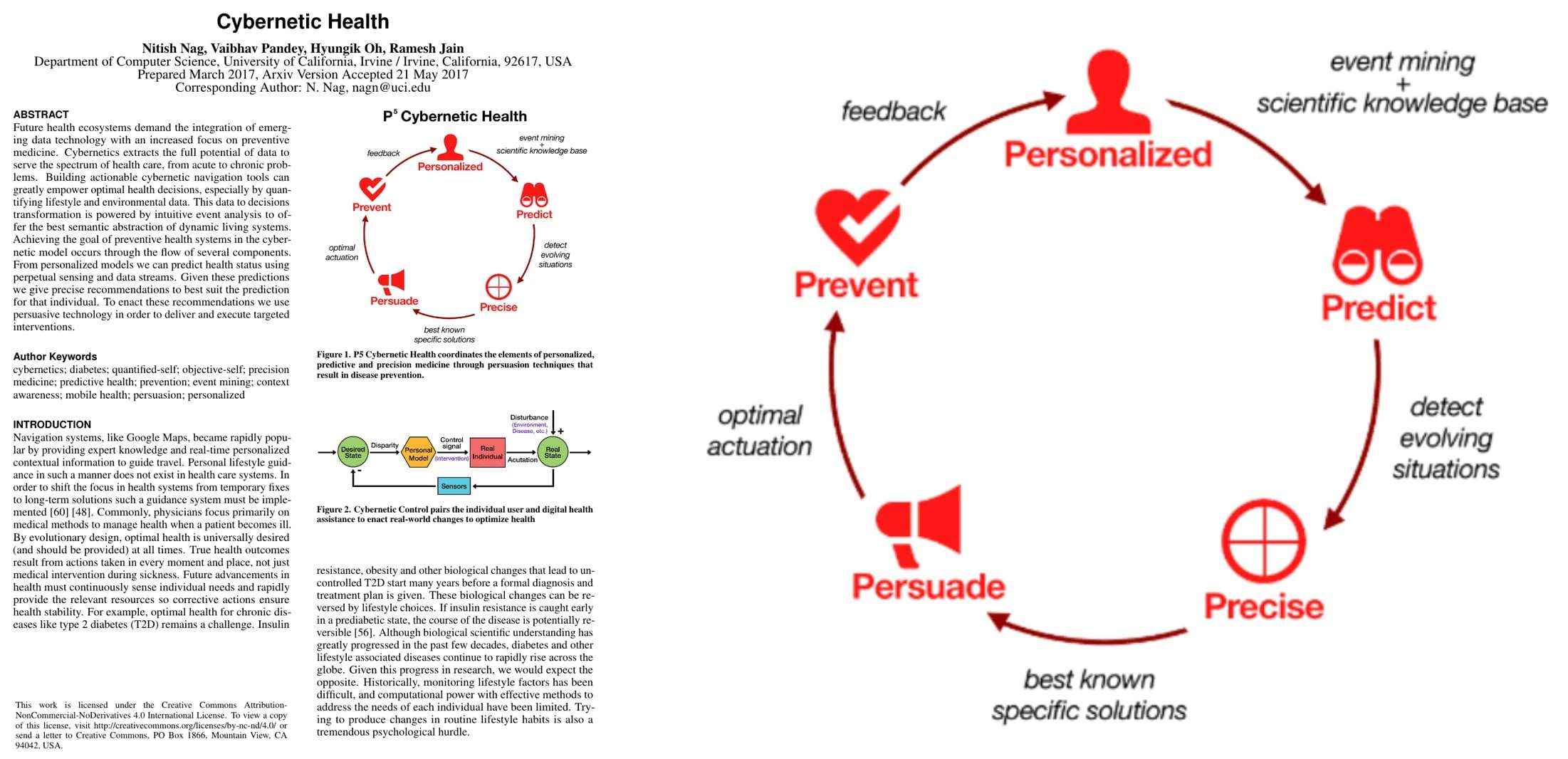}
\end{center}
Cybernetics extracts the full potential of data to help control steps along the navigation pathway. This paper presents five connected components to enhance control.
\textit{Arxiv, May 2017} \cite{Nag2017CyberneticHealth}
\begin{itemize}[noitemsep]
\item Personalization must take into account the individual model.
\item Prediction is possible with this model.
\item Precise guidance can be given based on the goal set in the cybernetic system.
\item Persuasion of the individual to execute the guided steps is key for lifestyle based intervention.
\item Prevention of undesirable regions-of-interest further teaches the model to improve in the next iterative cycle.
\end{itemize}

\label{box:boxcybernetic}
\end{infobox}
\end{figure}

\begin{figure}[t]
  \centering
  \includegraphics[width=1\textwidth]
  {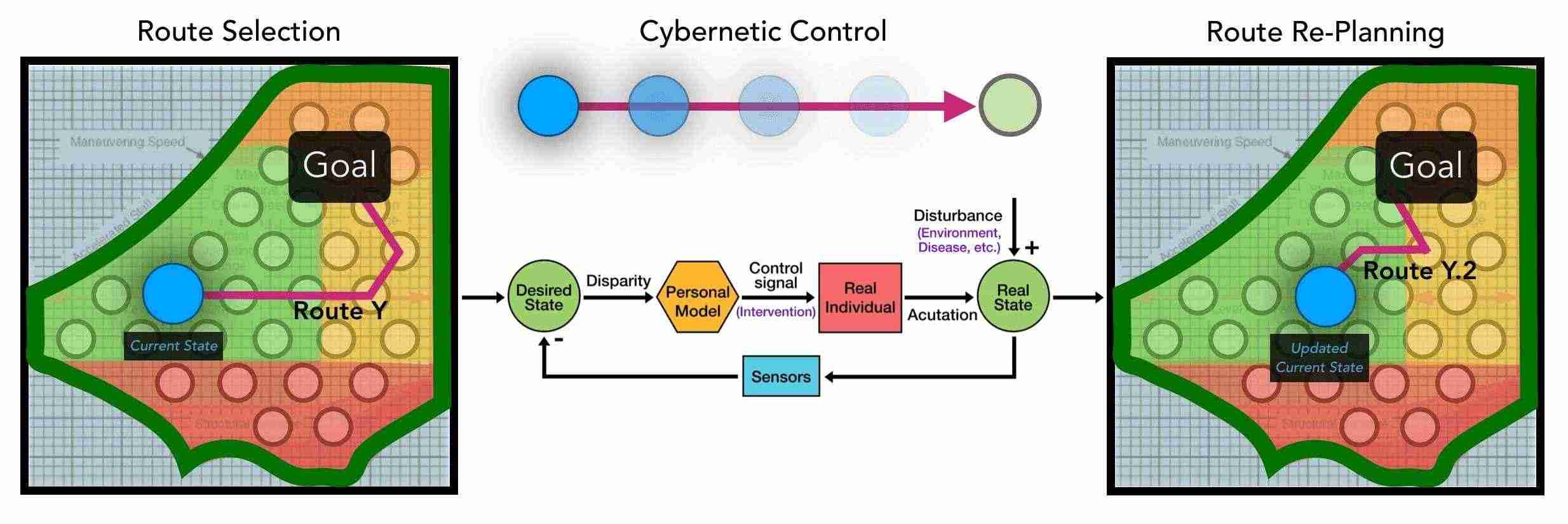}
  \caption{\textbf{Control and Re-Planning.} Once a specific route is selected, the user and system only need to worry about controlling the state transition for the current time iteration. Cybernetic loops are employed to ensure that the state transition is controlled successfully in the presence of disturbances and noise. Once the state has transitioned to a new state, the system re-plans the most optimal route again. Due to changes in resources, health state, or other factors, the route may be different at the next time iteration. Here we see route Y selected at time \textit{k}, and at the next time iteration, the route plan has changed to Y.2 based on the updated state and resources at time \textit{k+1}.}
  \label{fig:nav6}
\end{figure}

Computing optimal control mechanisms, however, are not enough. For meaningful impact, we must implement the cybernetic control principles in the real world which includes noise and disturbances. For the system to be effective in the physical work, we must consider the spectrum of actuation ranging from fully-automated systems to human-fulfilled actions.

\subsubsection{Automated Actions}
Many machines automate tasks such as laundry or aircraft control, which are instances of completely automated control mechanisms that can make an action in the real world control a task. For a type 1 diabetic patient, continuous glucose monitoring through the artificial pancreas, paired with insulin pumps, can control blood sugar far better than humans can and acts in a minute by minute timescale. Heart pacemakers have electrodes that detect the heart's electrical activity and send data to generate electrical pulses to the heart to control and monitor the heartbeat at a second by the second timescale. In these closed-loop systems, the executed actions are entirely automated, and no human intervention or uncertainty is involved, so these cases can be considered deterministic actuators in PHN.

\subsubsection{Partial Assistance}
Automation can assist humans in taking action. For example, building a Smart Home System helps modulate an individual's environment for circadian control, which ultimately assists in sleeping. Automated biological circadian rhythm control can be via connected lights, digital screen color variation, thermostats that mimic natural temperature fluctuations, and motorized curtains with smart white noise. While this may assist in circadian control and developing habits for sleep, the human needs to take action to ensure a consistent sleep schedule.

\subsubsection{User Actions}
Humans are not always good at understanding complex instructions or remembering them at the right time and in the right situations. They also are easily influenced and distracted, which makes them non-deterministic actuators. Even when the user fully intends to reach a health goal, adhering to healthy routines can be difficult. The primary challenging task is to understand how to influence people at the right time and with the best medium to produce the desired action. For each individual, actions must be persuasive in the given context, practically feasible, and encourage participation. Richard Thaler won the 2017 Nobel Prize for understanding the psychological underpinnings and control of the decision-making process in humans \cite{Thaler2016BehavioralFuture}. Specifically, the following three high-impact contributions interweave directly with health navigation.

\begin{enumerate}
    \item Limited rationality: Simplified decision-making focuses on the narrow impact of each decision rather than on its overall effect. Computing these complex effects can offload these tasks from the user.
    \item Lack of self-control: A planner–doer model describes the relationship between long-term planning and short-term actions. Short-term temptations to make healthier lifestyle choices often fail. By quantifying the health cost or benefit in any situation, users can make more controlled and informed decisions.
    \item Nudging and Probabilistic Nature of Inputs: By using subtle cues, users are pushed to make a confident decision without force, thus avoiding the pitfalls of self-control and limited rationality. Systems that employ these tactics can unobtrusively steer choices, leading to the goal state.
\end{enumerate}
      
As PHN is able to model better the behavior of an individual, we can increase the probability of executing a given action using the considerations listed. The system explains why it suggests the nudge or action to users, and it offers alternate options to embolden the feeling of choice. Another avenue for human actuation will be to reduce the decision burden by eliminating the tediousness involved in being healthy. Even professional athletes do not measure every gram or determine every ingredient in their food due to the information and calculation burden. Translated into simple, step-by-step, context-aware guidance, users can better fulfill PHN actuation steps. PHN can start to apply control to the health state, either with fulfillment by a machine or with user participation, to move the health state toward the goal and close the loop. 

Many of these actions are driven by daily decisions. Some of these decisions stem from hedonistic tendencies programmed in our biology from prehistoric times. Reward mechanisms were biologically encoded based on survival in a low resource environment with high uncertainty of the future. In the domain of food, these resources included factors such as caloric energy, water, and salt. Thus, humans have been optimized through evolution to accumulate these scarce resources in their body. Unfortunately the current era of civilization makes these types of problems largely irrelevant and much of our programmed biology is actually maladaptive in this context. In order to move our health towards a positive direction, we need better actionable insight about our health than just our naturally encoded tendencies.

When a user is a final actuator in the system, there must be a probabilistic approach to knowing if a user will take a specific planned step. One way to increase the chance is to have high-quality recommendation systems.

\subsection{Recommendation}
Recommendation systems play a role as a subset of cybernetic control. While the guidance is a full path to a goal, the recommendation systems increase the probability that a step in the process will take place for a particular state transition. A recommender system is also a filtering system that seeks to predict the rating or preference a user would give to an item. For example, services like Spotify have playlists generated based on user preferences. In the context of health, we use computed guidance and recommendation systems to guide an individual in lifestyle, nutrition, and other contextual components. Recommender systems have three components: 1) the user, 2) items to recommend, and 3) the context at a specific time and point that changes the relationship between the user and the items.

\subsubsection{User Profile}
The user health state is primarily a product of our everyday lifestyle actions, yet we have minimal access to recommendations on making everyday choices. Recommendations are the key to current content consumption and decisions. Cybernetic navigation principles that integrate health media sources can power dynamic recommendations to improve health decisions dramatically. Cybernetic components give real-time feedback on health status, while the navigational approach plots the health trajectory. These two principles coalesce data to enable personalized, predictive, and precise health knowledge that can contextually disseminate the right actions to keep individuals on their path to wellness \cite{Nag2017HealthObservations}.

\begin{figure}[t!]
\begin{infobox}[Health Multimedia: Lifestyle Recommendations Based on Diverse Observations]
\begin{center}
\includegraphics[trim={0 -0.2cm 0 0.4cm},width=0.95\textwidth]{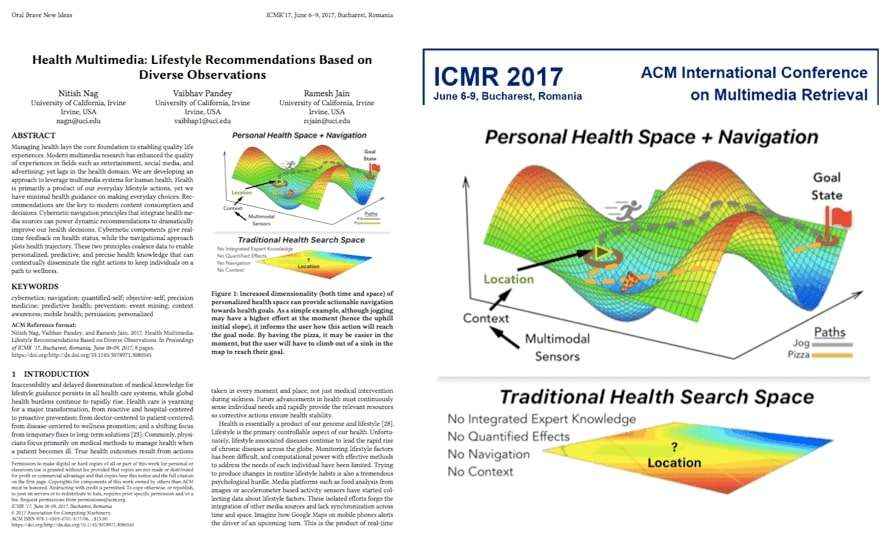}
\end{center}
Cybernetic components give real-time feedback and control on immediate health status, while the navigational approach plots health trajectory for long-term planning.
\textit{Published in ACM International Conference on Multimedia Retrieval, June 06-09, 2017, Bucharest, Romania.} \cite{Nag2017HealthObservations}
\begin{itemize}[noitemsep]
\item Cybernetic loops and navigation systems need to understand user health state.
\item Integration of health media sources can power dynamic recommendations to dramatically improve our health decisions.
\end{itemize}

\label{box:boxhealthmulti}
\end{infobox}
\end{figure}
    
Unique individual \textit{preferences} are critical for determining the user profile, as preferences determine whether or not a user will proceed with the recommendation. Understanding a user's model allows us to build a more precise and personalized recommendation system. We have investigated the relationship between cuisines and their inherent taste by collecting data on an individual's food photos. We used image metadata and computer vision tools to extract distinct insights from each user to build their food profile (unpublished).

\begin{figure}[t!]
\begin{infobox}[Personalized Taste and Cuisine Preference Modeling via Images]
\begin{center}
\includegraphics[trim={0 -0.2cm 0 0.4cm},width=0.95\textwidth]{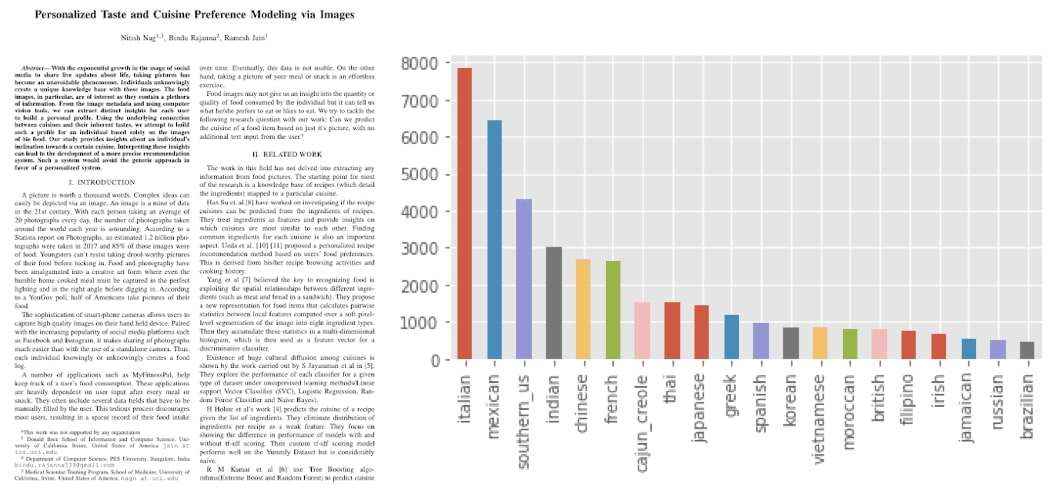}
\end{center}
From the image metadata and using computer vision tools, we can extract distinct insights for each user to build a personal taste and cuisine profile.
\textit{Arxiv, 2020} 
\begin{itemize}[noitemsep]
\item Development of rank system for cuisines based on user's preferences.
\item Weighting preferences based on event mining nutrition events.
\item Identification of ingredients commonly preferred, allowing the system to generate recommendations that are novel and exciting to the user.
\end{itemize}

\label{box:boxpersonalized}
\end{infobox}
\end{figure}

 A combination of preferences for physical activity and conditions can also be applied in a recommender system to give an individual the most optimal plan for their exercise. For example, we have used a method for using GPS data from a human-powered cyclist to classify if the user is on a paved road or unpaved trail. Because road cyclists may prefer smooth roads, versus mountain bike riders who prefer dirt roads, we can begin to understand user preferences for bicycle travel paths. We can draw other relevant insights from the data we collect through Strava. We can collect social media platforms to determine the weather preference of the user, the kind of roads they prefer to cycle on, the time during which they regularly perform the activity, and much more. These can be combined in order to recommend suitable activities to improve the health state \cite{Nag2018SurfaceActivities}.
 
 \begin{figure}[t!]
\begin{infobox}[Surface Type Estimation from GPS Tracked Bicycle Activities]
\begin{center}
\includegraphics[trim={0 -0.2cm 0 0.4cm},width=0.95\textwidth]{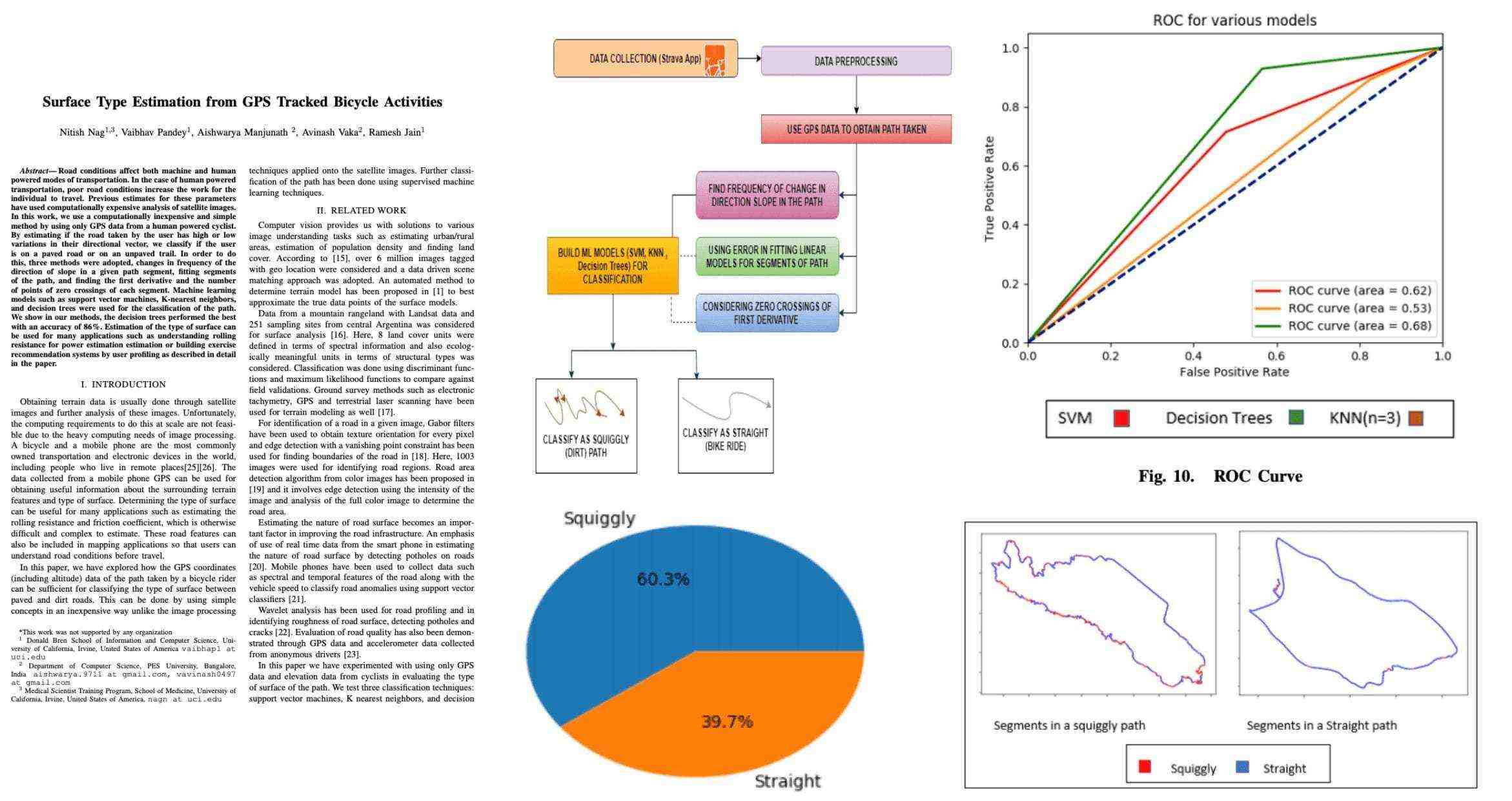}
\end{center}
By estimating if the road taken by the user has high or low variations in their directional vector, we classify if the user is on a paved road or on an unpaved trail without accessing expensive satellite imagery or databases. 
\textit{Arxiv, September 2018} \cite{Nag2018SurfaceActivities}
\begin{itemize}[noitemsep]
\item Providing exercise recommendations by profiling the users path choices.
\item Understanding rolling resistance for power estimation for cardiovascular applications in VO2max estimation.
\item In the above figure, the orange slice highlights smooth roads, and the blue slice shows undulating terrain as part of the model regarding the test subjects preferences.
\end{itemize}

\label{box:boxsurface}
\end{infobox}
\end{figure}

Just as \textit{motivation} is critical for goal setting, it is also vital for one's willingness to follow through with a recommendation. Strong recommendations need to be set based on the intrinsic preferences of the user. Combining the right triggers with context, we can maximize the effectiveness of sustainable positive health recommendations. Over time, the recommender system should adapt to the user and identify triggers that are needed to support their motivations. For example, exercising every day should become habitual rather than a constant suggestion from the recommendation engine. In addition to one's health, \textit{user enjoyment} is also a critical component of determining the user profile. Enjoyment leads to a higher quality of living, so it is critical to incorporate the best preferences.
	
\subsubsection{Items}
We must capture in detail the items or characterization of specific objects that change the health state. By knowing the user profile, we can then match a user model with items. 

For food recommendations, we propose a mechanism to determine the taste profile of items. We incorporate into a recommendation engine an empirical method to determine the flavor of food. Such a system has advantages of suggesting food items that the user is more likely to enjoy based upon matching with their flavor profile through the use of the taste biological domain knowledge. This preliminary intends to spark more robust mechanisms by which flavor of food can apply to recommendation systems \cite{Nag2019FlavourRecommendation}. 

\begin{figure}[t!]
\begin{infobox}[Flavour Enhanced Food Recommendations]
\begin{center}
\includegraphics[trim={0 -0.2cm 0 0.4cm},width=0.95\textwidth]{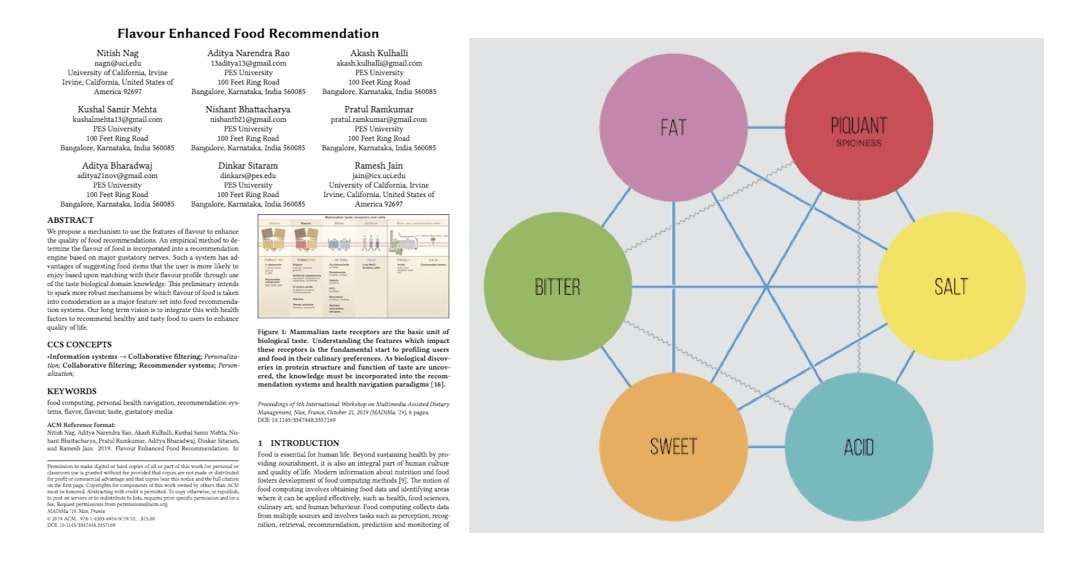}
\end{center}

 An empirical method to determine the flavour of food is incorporated into a recommendation engine based on major gustatory nerve signals.
\textit{MADiMa '19: Proceedings of the 5th International Workshop on Multimedia Assisted Dietary Management, October 2019} \cite{Nag2019FlavourRecommendation}
\begin{itemize}[noitemsep]
\item Readily available data about nutrition can help in identifying item flavor.
\item Our long term vision is to integrate health factors to recommend healthy and tasty food to users for a high quality dining experience.
\end{itemize}
\label{box:boxflavour}
\end{infobox}
\end{figure}

\begin{figure}
  \centering
  \includegraphics[width=.75\textwidth]
  {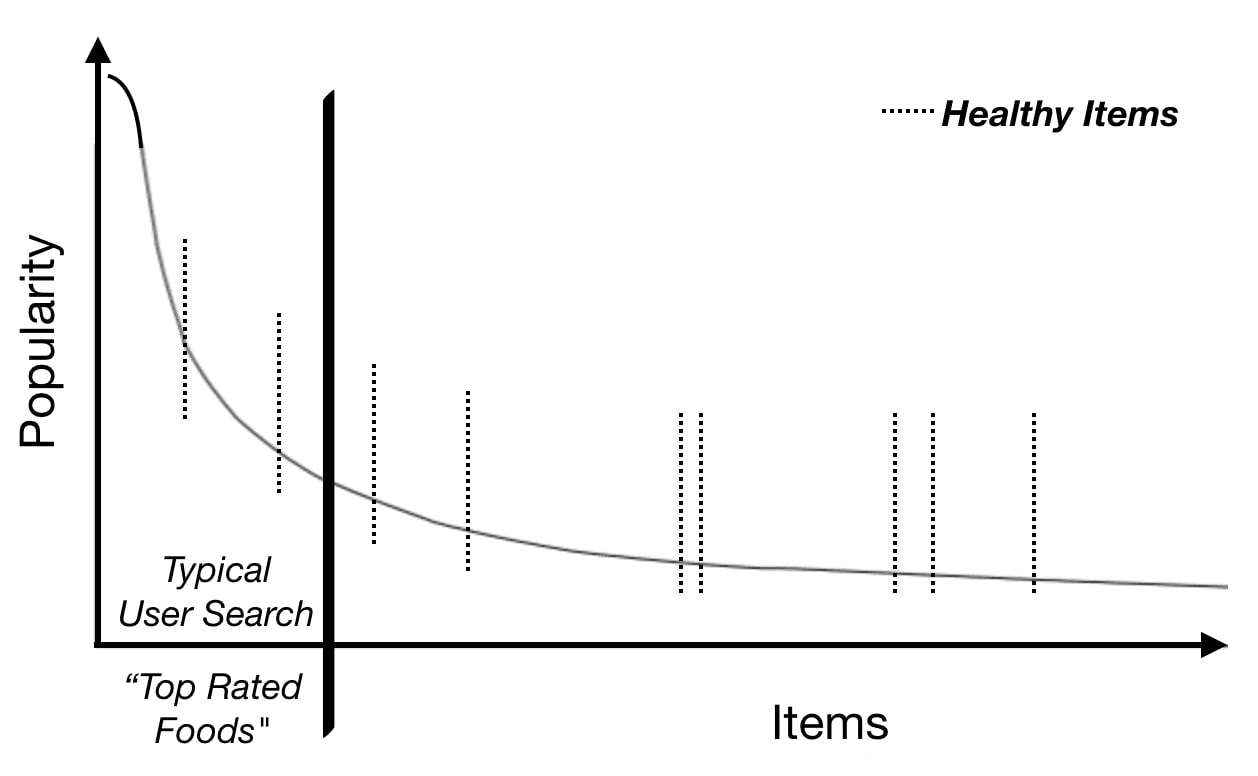}
  \caption{\textbf{The Long Tail Problem.} Usually users only see items that are top rated by other users. This leaves items that are highly relevant for the user far down in the search results, such as items that are healthy.}
  \label{fig:tail}
\end{figure}

Along with the taste profile, we propose that food items should also have a health profile associated. Recommending items based on unique preferences comes challenges. For example, the Long Tail Problem highlights items that are not popular but highly relevant to the user, shown in Figure \ref{fig:tail} \cite{Park2008TheIt}. However, by understanding the health status of the user, we can match health-relevant items that a user would not usually browse through. This matching process allows the user to discover healthy items that are relevant but would have been difficult to find through popularity based rating systems. Thus, a recommender system should provide relevantly matched items to reduce browsing burden. The goal of this system is to capture the healthy items in the long tail and match them to the personal needs of a user based on, for example, sodium needs. \cite{Nag2018EndogenousHealth,Nag2017PocketLocation,Nag2017LiveEngine}.

\begin{figure}[t!]
\begin{infobox}[Pocket Dietitian: Automated Healthy Dish Recommendations by Location]
\begin{center}
\includegraphics[trim={0 -0.2cm 0 0.4cm},width=0.95\textwidth]{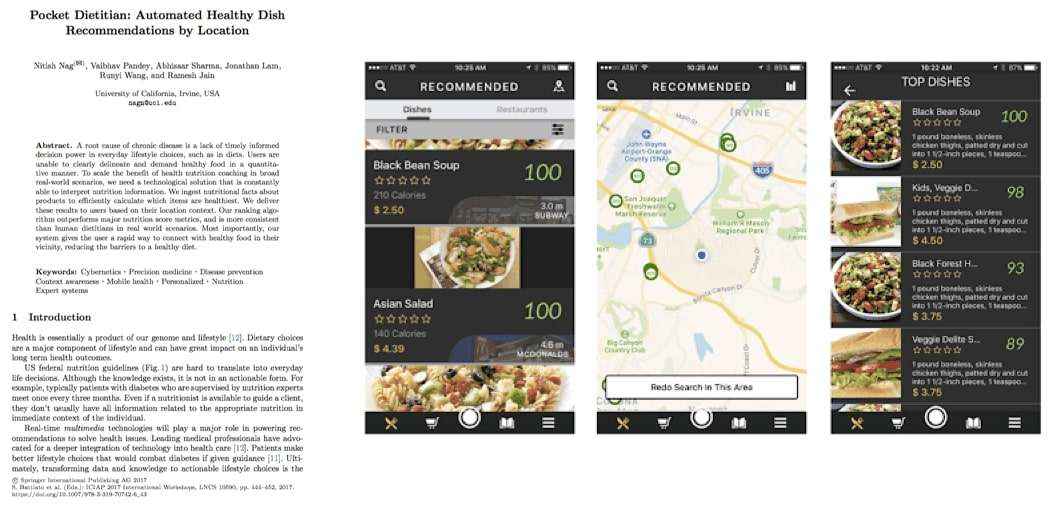}
\end{center}
Pocket dietitian refers to the application built that can rapidly locate healthy dishes on a mobile application.
\textit{2017} \cite{Nag2017PocketLocation}
\begin{itemize}[noitemsep]
\item Algorithms are needed to constantly interpret nutrition information.
\item We have calculated nutritional facts about products to determine which products are healthiest for several types of patients.
\item Our system is designed to give the user a rapid way to connect with healthy food in their vicinity, reducing barriers to taking a step in their health actions.
\end{itemize}

\label{box:boxpocket}
\end{infobox}
\end{figure}

People want to know what foods they can consume at a specific place, alongside the taste, ingredients, allergens, nutrients, and visual and other sensory characteristics. Combining both taste profiles and health profiles for food items give us a basic \textit{food atlas} \cite{WorldFoodAtlas.org2020WorldExperience}. This \textit{food atlas} is a location-based database that has a live data base of nutritional items of food along with all parameters about each physical entity within inventory. The information regarding each food type can be globally available and creates standardization for food computing purposes. This idea has just recently been launched to the public as an open source project. ``The World Food Atlas project is an open source effort to prepare a live atlas of specific food at a specific geographic location.
World Food Atlas (WFA) aspires to provide information relevant to all dishes and food items for health and enjoyment that foods offers us, wherever you are in the world. We want to make it widely available so it can be converted to actionable information for a person in specific context" \cite{WorldFoodAtlas.org2020WorldExperience}.
    
\subsubsection{Context}
The next step is understanding the context (location and resources) for a given individual situation, which then determines the relationship between the user profile and items. We can continue to improve recommender systems with in-depth context provided. The utility function is the primary ranking method to predict what a given person would explicitly prefer. There are two layers of user and context modeling, coupled with traditional recommender system approaches. One is an endogenous context, which integrates data to estimate the physiologic or innate needs of the user (health state) or preferences, and second is the exogenous context, which incorporates factors outside of the person's body (i.e., points of interest, geospatial resources, social context) \cite{Nag2018EndogenousHealth}. 

\begin{figure}[t!]
\begin{infobox}[Endogenous and Exogenous Multi-Modal Layers in Context Aware Recommendation Systems for Health]
\begin{center}
\includegraphics[trim={0 -0.2cm 0 0.4cm},width=0.95\textwidth]{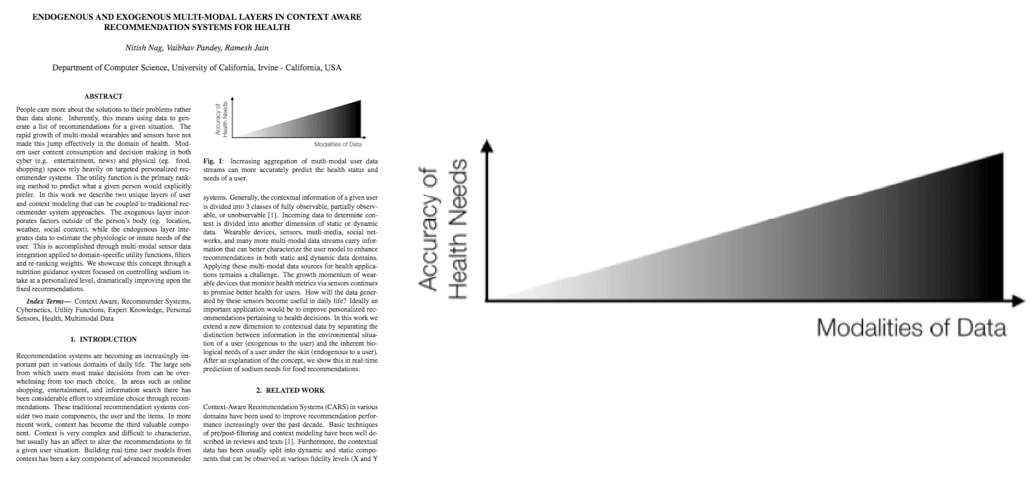}
\end{center}

Key Points
\textit{August 2018} \cite{Nag2018EndogenousHealth}
\begin{itemize}[noitemsep]
\item The utility function is the primary ranking method to predict a person's preferences.
\item We review both endogenous and exogenous layers determine user preferences, using multi-modal sensor data integration applied to domain-specific utility functions, filters, and re-ranking weights.
\item Our approach is showcased with a nutrition guidance system at a personalized level.
\end{itemize}

\label{box:boxendogenous}
\end{infobox}
\end{figure}

Through multi-modal sensor data integration applied to domain-specific utility functions, filters, and re-ranking weights, we can use context in recommendations. We showcase this concept through a nutrition guidance system focused on controlling sodium intake at a personalized level, dramatically improving upon the fixed recommendations. For a particular user, identifying contexts through continuous sensing is the best execution strategy. For one's nutrition, the system will need to decide the best meal choice to move the user's health state closer to his or her goal state based on situational constraints. This decision will require knowledge of all available meals, as well as their locations, hours, nutrition facts, and ingredients along with the personal sensor data modeling the physiologic needs.

A challenge that arises is the user's inability to delineate and demand healthy food in a quantitative manner. We have developed a technological solution that is continuously able to interpret nutrition information, ingest nutritional facts about products to efficiently calculate which items are healthiest, and deliver these results to users based on their location context. Our system gives the user a fast way to connect with healthy food in their vicinity, reducing the barriers to a healthy diet. Addressing the correct granularity of recommendations, such as meals, is essential for simple decision making. We have made a decision support system using multi-modal data relying on timely, contextually aware, personalized data to find local restaurant dishes to satisfy a user's needs. Algorithms in this system take nutritional facts regarding products, efficiently calculate which items are healthiest, then re-rank and filter results to users based on their personalized health data streams and environmental context. The primary goal of lowering barriers to a healthy personalized choice drives our recommendation engine \cite{Nag2017LiveEngine}.

\begin{figure}[t!]
\begin{infobox}[Live Personalized Nutrition Recommendation Engine]
\begin{center}
\includegraphics[trim={0 -0.2cm 0 0.4cm},width=0.95\textwidth]{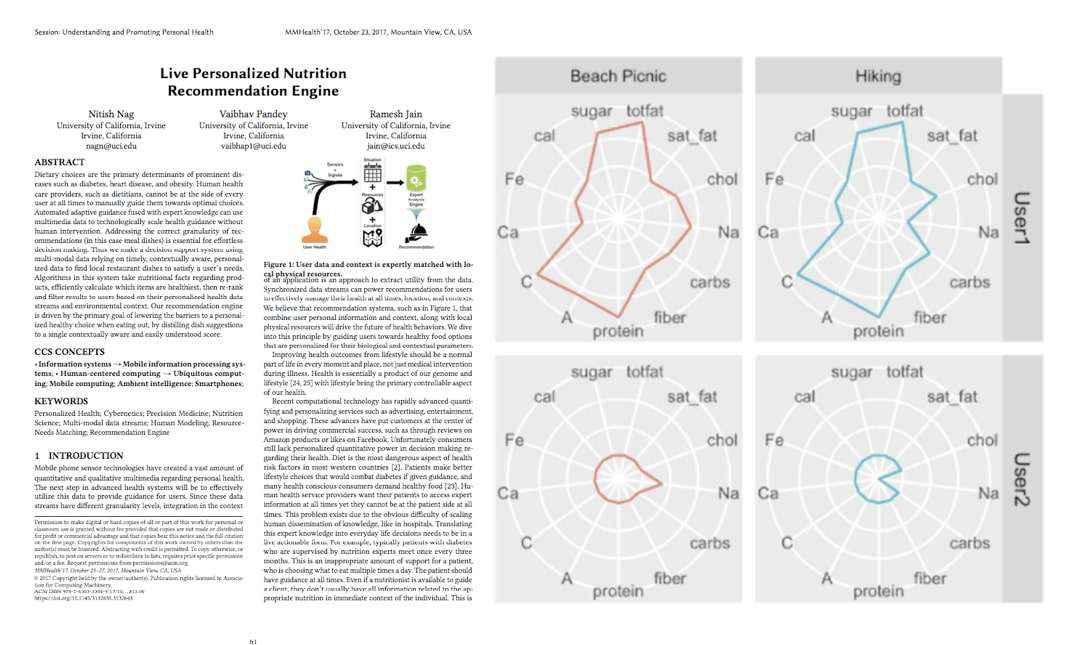}
\end{center}

Key Points
\textit{August 2018} \cite{Nag2017LiveEngine}
\begin{itemize}[noitemsep]
\item Automated adaptive guidance combined with expert knowledge can use multimedia data to technologically scale health guidance.
\item We introduce a decision support recommendation system for finding local dishes satisfying a user's personalized needs.
\item The recommendation engine aims to reduce barriers to healthy eating by suggesting dishes with a health score.
\end{itemize}

\label{box:boxlive}
\end{infobox}
\end{figure}

\subsubsection{Expert Guidance}
There may be guidance mechanisms beyond the scope of computing. Human experts may be utilized for recommending change to an individual's health state in person or through various forms of telecommunication. These forms of guidance can be synchronous in time (real-time), or they can be asynchronous through various forms of communication and multi-media.

Through multi-modal sensor data integration applied to domain-specific utility functions, filters, and re-ranking weights, we can use context in recommendations. We showcase this concept through a nutrition guidance system focused on controlling sodium intake at a personalized level, dramatically improving upon the fixed recommendations. For a particular user, identifying contexts through continuous sensing is the best execution strategy. For one's nutrition, the system will need to decide the best meal choice to move the user's health state closer to his or her goal state based on situational constraints. This decision will require knowledge of all available meals, as well as their locations, hours, nutrition facts, and ingredients along with the personal sensor data modeling the physiologic needs.

The end purpose of the PHSS is to ensure that the individual arrives closer to their end goal destination, which may mean high-quality intermediate events that determine one's experience in the navigation process. The quality of the experience provided is, therefore, one of the requirements of the system. In order to have a long-term engagement success with the navigation system, we must understand the individual's motivations, preferences, context, and other factors that may make the individual more engaged with the system.

The personal navigation system to health allows for concrete actions to be taken for the future performance of one's health. Future systems can leverage modern computing power with rich multi-modal data to understand movements on the health state space continuously and to contextually guide users toward wellness. The nexus of biomedical knowledge, sensor technology, computing power, mobile networks, and artificial intelligence will fuel significant growth in PHN systems.
\chapter{Conclusions and Future Work}

\epigraph{If you're walking down the right path and you're willing to keep walking, eventually you'll make progress.}{\textit{Barack Obama}}
\vspace{12pt}


Advancing the understanding of our health state is essential for improving human health. Throughout time, humans have primarily taken an episodic approach to interact with health. The next evolution in health will depend upon a continuous and high-resolution understanding of our health state to better control the future of our well-being.

This thesis introduces an open framework and concept to use a diverse array of data streams and knowledge to enhance the health state estimation of an individual. The product is graph modeled a digital twin that parallels the user's physical life. The first main research contribution provides a consolidation of defining health and it's role within the influence of control systems (Chapter 2). The second research contribution builds a computational framework to represent this health. The representation is based on connecting the user's life events, biology, and general utilities through a graph network block approach (Chapter 4). This work has the dual coupling of bottom up data-driven modeling and top-down domain knowledge. By consolidating systems into smaller blocks that can continuously be updated with new data or knowledge, the graph network can flexibly evolve and improve its representation of the user's health state. Demonstration of this framework with diverse high-resolution personal data is conducted with a focus on bioenergetics, cardiovascular disease, and performance (Chapter 5). Given this approach to health state estimation, we can enable personal health navigation systems that continuously guide an individual to a desired health state using a combination of planning intelligence and cybernetic control (Chapter 6). We hope individuals may be more motivated to track their health state, especially if it will be used in professional clinical decision making or influencing their daily actions through such navigational systems. Instant feedback on the health state can give insight on how lifestyle modification, medicine, environment and other inputs are causally changing the person's overall quality of life. The highest motivation of this work is to catalyze the research community to advance these ideas of health state estimation and personal health navigation. There remain significant challenges and opportunities in this endeavour. An non-exhaustive set of these challenges is given below.

\section{Technical Challenges}
Improvements in various technical areas are necessary to bring further fruition of this framework to individuals. Many of the challenges need to be understood in depth.

\subsection{Building Rich Knowledge Atlases}
The work presented here focuses only on one domain of cardiovascular and athletic performance. Experts must also address areas beyond cardiovascular health. To establish the best cold-start graph representation of an individual, we need accessible domain knowledge. This accessibility needs to cooperate with both human and computational interfaces. There have been substantial efforts in both of these interfaces in general domains as well as health-specific fields. Examples of knowledge bodies that are widely used today include Wikipedia, PubMed, and Google Scholar \cite{Wikipedia2020Wikipedia,NCBI2020HomeNCBI,Google2020GoogleScholar}. For physicians and healthcare providers, the most commonly used atlas of knowledge is UptoDate \cite{UpToDate2016Evidence-BasedUpToDate}. Most of these knowledge atlases store information in static natural language, which is very useful to interface with humans. The computational interface with these knowledge sets will require robust Natural Language Processing techniques to produce a virtual representation of knowledge. Efforts have been made in specific domains to encode the knowledge in a more computable format. MDCalc is a website that allows domain experts to define formulas by which other users or applications can run domain knowledge-based calculations for precise biological estimations of physiology, risk, or medication dosing \cite{MDCalc2020MDCalcGuidelines}. One step further for computational representation is the entire field of knowledge graphs. Open source platforms such as Cytoscape and HetNets have shown promise in taking the current domain knowledge and producing a connected system that can incorporate within further applications \cite{Shannon2003Cytoscape:Networks}. The ultimate ideology behind the knowledge atlases would be a dynamic reference database that applications can use to access the latest scientific understanding to build systems of navigation, estimation, and other services. These atlases would require the coordinated representation of both highly complex data and human knowledge structure and taxonomy.

\subsection{Hardware}
Hardware improvements for health state estimation largely depend upon sensors that can unobtrusively gather data about events, biological metrics, and utility measures. Event sensors that can capture more attributes in detail can give insight on how the event is impacting the health state, and exactly what aspects of the event are relevant. Improved biological sensors are needed to know what the observed values of biological health state components are to better estimate an all encompassing health state. State estimation may also become computationally demanding depending on the case. There remains a need to use computing hardware to include local, edge, and cloud based approaches for high performance health estimation. In various contexts, there will be a need for both online and offline methods to digest and compute the information. The online approaches need to synchronize at appropriate times to update models in the offline approaches.

\subsection{Computing}
On the mathematical side, computing through modern modeling techniques can increase the computing efficiency and capability to handle higher loads of computational complexity with more efficiency. Black box artificial intelligence techniques to improve data driven relationships need to transition into explainable systems. If any health estimation is done on a battery powered device, energy management also becomes a key issue for constant updates of the health state. Given the high cost of accuracy and complexity, there needs to be better optimization of computing resources.
        
\subsection{Accuracy and Precision}
Precision in estimation will be best suited for measuring relative change in an individual, as it refers to the closeness of measurements to one another. Accuracy refers to the closeness in measurements of to a specific value, giving a definite health measurement. Estimation in its initial iteration may not be accurate, but it is assumed to improve over time with refining of equations, algorithms, feature extraction methods, learning methods, as well as with improvements in hardware technology. We use basic methods in this paper as a starting point for further advancement with modern learning and predictive techniques. Because of the relative large range in accuracy, wearable devices and similar low cost sensors are promoted for use as a screening tool to identify if a user is at risk or has a change in their health state \cite{IEEEComputerSociety2019ComputerSociety}. Clinical gold standard testing, a more expensive alternative, may be used to confirm the true state of the user if the state estimation passes a certain screening threshold. More work will need to be done to ensure the robustness of the estimations are good enough for key decision-making, such as ensuring validated clinical and biomedical research. 

\subsection{N=1 Statistics}
  Baseline comparisons are also difficult when studying individual subjects, and will require statistical methods for n of 1 studies \cite{McQuay1994DextromethorphanDesign,Chapple2011FindingTrials}. Even while referencing established clinical science research we still have no way to validate ground truth about an individual until decades into the future when a person becomes verified in their disease or die. Therefore, comparisons are not the best way to experimentally validate the estimation model. Performance comparisons for this type of experiment will need to be validated through large scale data collection and monitoring in prospective studies as mentioned in related works. Parallel models may compete for the best representation of the individual through an adversarial and recurrent network that is temporally dynamic. Ultimately, high quality validation techniques using only a single individual need to be advanced in order to definitively improve and verify the state estimation quality.

\subsection{Event Mining}
Efficiently extracting knowledge from events and log data for an individual remains a challenge. Tracking of individual data may be limited, and barriers to efficient extraction of useful information exist. We see problems with the accessibility of data, more adequate sensors developed for determining events, and event descriptions. There are also issues with semantic variability in the real-world, including multilevel event semantics, implicit semantics facets, and the influence of context \cite{Xie2008EventStreams}. The field of behavioral modeling can benefit greatly by having more data-driven personal models that can be fed into a navigation system to provide more engaging routing for the individual. Event mining can also help understand causal chain of events and discover edges in the graph that are unknown.



\section{Real-World Adoption}
For real-world adoption of this work into tangible services for society and individuals, we must consider various usability, economics, legal, ethical, and protection issues. For health state estimation to occur, users must have some reason to engage in collecting and sharing data about themselves. This estimation must be made useful through services back to the individual that is part of the broader health economy. All the sharing of information must also occur safely. Privacy refers to the rights to control personal information by the individual and how authorized agents use it. Security, on the other hand, refers to how personal information will be protected and defended against unauthorized and malicious access. Since health data inherently contains sensitive information about an individual, we must address these at both individual and societal levels.

\subsection{Engagement}
One of the greatest struggles in the health field is individual engagement with their health. In the current healthcare model, individuals predominantly interact with their health when there is a problem or urgent need (a reactive approach). Individuals must see value in state estimation and proactively engage in collecting their health data. We see a few challenges:

1) Health state estimation relies heavily on computing mechanisms and may provide complexities for humans to understand and use effectively. Humans need to be technologically savvy in interacting with sensors that are collecting their health information. Sensors must provide ease of use, and individuals must trust that the sensors are not harming them in any capacity.

2) It is not very easy to visualize the health state as it is so robust that a person may not be able to comprehend all the various components and relationships. There will be challenges with simplifying information to the semantic understanding of the individual.

3) Currently, individuals interact with their health through service providers (i.e., doctors, health professionals, insurance companies). Because of health's interconnectedness with external systems, service providers must be involved in the health estimation in order to help the individual in both understandings and taking steps towards better health.

4) Individuals may be disinclined to engaging with their health. It is necessary to consider behavioral psychology to see where there are gaps in engaging with health and having motivation for improving future health. There is the problem of temporal discounting, in which ``value of a reward decreases as the time delay until its receipt increases" \cite{Goldstein2011EncyclopediaDevelopment}. Humans are less inclined to make changes in the present day if the outcomes are far in the future.

\subsection{User Autonomy, Authority, and Ownership}
As more personal health data is acquired to estimate the health state, there will arise issues with data ownership and utilization by external sources, introducing legal and ethical barriers and ambiguity in who will ultimately make use of the data. The privacy and security issues regarding sensitive health data can be at an individual level, organizational level, or national level for civilian protection. Emerging threats and vulnerabilities will continue to grow as data becomes expansive, and it will be a challenge to balance innovation in health estimation with the regulation of health data.

\subsection{Privacy}
A challenge with user health data is ownership of the information and the ability to release this data to external agents such as healthcare service providers. As the information is sensitive and robust, there must be controls and policies in place to ensure ownership of data by the individual and appropriately used. There are regulations in place today that help with data privacy issues such as the Health Insurance Portability and Accountability Act (HIPAA) passed by Congress in 1996. This law created rules to address disclosure of ``protected health information" by organizations and created privacy rights standards for individuals to control the use of their health information \cite{HHS2013SummaryHHS.gov}. Similarly, The European Union has established the General Data Protection Regulation (GDPR) in May 2018 to strengthen individual privacy rights by clarifying rules for companies and public bodies in the digital market \cite{EuropeanCommission2016DataCommission}.

We must carefully address information sharing as there can be external threats and harm to the individual with improper data management.

\subsection{Security}
Security is a significant concern for personal health data as from breaches by external threats, hackers and ransomware are highly prevalent. Mishandling of information can stem from insider information misuse and lack of security for mobile devices. Healthcare providers may fail to evaluate risk when working with outside vendors. There becomes an issue with individuals sharing information voluntarily on the web or via apps, in which case attackers can take disparate information pieces from many sources, including social networks, public records, and search engines \cite{Li2011NewPrivacy}.

Some methods for addressing security are being taken, such as blockchains used for storing and sharing information that is secure because of its transparency. Blockchain systems are more commonly used in healthcare to protect information but are susceptible to attacks \cite{Huang2019BlockchainModel}. Encryption methods take the data and convert it to encoded text. Data is unreadable unless an individual has the necessary key or code to decrypt it. 

With the sensitivity of the information at hand, we must ensure further advancements to ensure all individual data is secure for the protection of the individual.

\subsection{Health Economics}
Determining the widespread adoption of the health state estimation relies on external stakeholders and the nature of health economics. Healthcare today heavily relies on an infrastructure set by hospital systems, insurance agencies, and other health service providers. These agents are primarily focused on consumer demand, cost, and profitability, while individual health improvement may not be at the forefront. The incorporation of the health state estimation concept will depend on stakeholder alignment and building a model to incentive both providers and individuals. As the health economy continues to expand, there will inevitably arise challenges managing multiple stakeholders.

\subsection{Standards and Interoperability}
Information and Communication Technologies (ICTs) play notable roles in the improvement of patient care via the exchange of information among healthcare providers. For complex systems to work together in an open way, the private and public sectors to collaborate and build frameworks for building health tools and services, technical specifications, and content. It becomes a challenge to unite multiple stakeholders. The Standards and Interoperability Framework within the Office of the National Coordinator for Health Information Technology has made progress to take input from various stakeholders and create health IT specifications for us throughout the U.S. \cite{HHS2020StandardsFunctions}. The Fast Healthcare Interoperability Resources (FHIR) is an international standard for exchanging healthcare information electronically,  advancing interoperability in the healthcare community \cite{NHS2019FastDigital}. FHIR allows for the sharing of information in a standard way regardless of ways local electronic health records represent or store data, allowing for more access to real-time data. However, there remain challenges with the seamless exchange of patient information across different healthcare systems.

\subsection{Research Community}
Collaboration between many different fields of medicine, bioscience, social science, computer science, statistics, and engineering must come together to have an inter-disciplinary approach towards these challenges. HealthMedia can be viewed as one response from the multimedia community to rise to this challenge. There is an increasing number of research work that shows how core multimedia research is becoming an essential enabler for solutions with applications and relevance for the societal health questions. Within this workshop, we continue to explore the relevance, contribution, and future directions of multimedia to health care and personal health. This workshop brings together researchers from diverse topics such as multimedia, tracking, life-logging, accessibility, HCI, but also health, medicine, and psychology to address challenges and opportunities of multimedia in and for health. To forward this mission through my contributions to the research community, I have led several projects, including leading a workshop committee chair for HealthMedia at the ACM Multimedia 2019 Conference in Nice, France \cite{Boll2019HealthMedia19}, founding the Institute for Future Health at the University of California, Irvine, and leading a session during an international workshop for Food Computing alongside Professor Ramesh Jain.


\section{Final Words}
Health State estimation is just one piece of the total health continuum concept. Ultimately, there must be advancements to ensure the development of reliable, practical, and useful systems to guide people towards better health through a navigational paradigm. Additionally, privacy and security methods must evolve concurrently for such systems to function in the real world. This research invites others to participate in building the foundations of health state estimation and navigational approaches to health. This starting point sets the aim by which people can enjoy peak health by utilizing the combined advancements in computer science, medical research, and technology at large. We hope this builds research momentum towards providing a high-quality, healthy life for every person.

\clearpage
\phantomsection

\bibliographystyle{authordate1}
\bibliography{thesis}

\captionsetup[table]{list=no}


\myappendix
\section{Abbreviations}

\begin{longtable}[c]{@{}ll@{}}
\caption{Abbreviations}
\label{tab:abb}\\
\toprule
\textbf{Abbreviation} & \textbf{Expansion}                         \\* \midrule
\endfirsthead
\endhead
\bottomrule
\endfoot
\endlastfoot
ABMS                  & American Board of Medical Specialties      \\
ACM                   & Association for Computing Machinery          \\
AALM                  & All Ages Lead Model                        \\
ABP                   & Athlete Biological Passport                \\
ACC                   & American College of Cardiology             \\
ACTN3                 & Alpha-actinin-3             \\
AHA                   & American Heart Association                 \\
API                   & Application Programming Interface          \\
ASCVD                 & Atherosclerotic Cardiovascular Disease     \\
ATP                   & Adenosine Triphosphate                     \\
ATP                   & Association of Teachers of Preventive Medicine Foundation   \\
BAC                   & Blood Alcohol Content                      \\
CDC                   & Centers for Disease Control and Prevention \\
CEDH                  & Center for Educational Development Health  \\
cMRI                  & Cranial Magnetic Resonance Imaging         \\
CO                    & Cardiac Output                             \\
COX                   & Cycloxygenase                              \\
CP                    & Critical Power                             \\
CRF                   & Cardio-respiratory Fitness                 \\
CTL                   & Chronic Training Load                      \\
CVD                   & Cardiovascular Disease                     \\
DALY                  & Disability Adjusted Life Year              \\
DB                    & Database                                   \\
DEXA                   & Dual-energy X-ray absorptiometry          \\
EEG                   & Electroencephalogram                       \\
EKG                   & Electrocardiogram                          \\
EMA                   & Ecological Momentary Assessment            \\
EPA                   & Environmental Protection Agency            \\
FHIR                  & Fast Healthcare Interoperability Resources \\
fMRI                  & Functional Magnetic Resonance Imaging      \\
GDPR                  & General Data Protection Regulation        \\
GIS                   & Geographical Information Systems           \\
GNB                   & Graph Network Block                        \\
GOVVS                 & Gravity Ordered Velocity Stress Score      \\
GPS                   & Global Positioning System                  \\
GSR                   & Galvanic Skin Resistance                   \\
HCI                   & Human-Computer-Interaction                 \\
HIPAA                   & Health Insurance Portability and Accountability Act Regulation \\
HIIT                  & High-Intensity Interval Training           \\
HPA                   & High Power Activity                        \\
HR                    & Heart Rate                                 \\
HRV                   & Heart Rate Variability                     \\
HSE                   & Health State Estimation                    \\
ICD                   & International Classification of Disease    \\
ICTs                  & Information and Communication Technologies \\
ICU                   & Intensive Care Unit                        \\
IoT                   & Internet of Things                         \\
MRI                   & Magnetic Resonance Imaging                 \\
NIH                   & National Institute of Health               \\
NLP                   & Natural Language Processing                \\
NSAID                 & Non-Steroidal Anti Inflammatory Drugs      \\
O2                    & Oxygen                                     \\
PHSS                  & Personal Health State Space                \\
QoL                   & Quality of Life                            \\
ROI                   & Region of Interest                         \\
SCD                   & Sudden Cardiac Death                       \\
SDK                   & Software Development Skit                  \\
SNP                   & Single Nucleotide Polymorphism             \\
SV                    & Stroke Volume                              \\
TCA Cycle             & Tricarboxylic Acid Cycle                   \\
TRIMP                 & Training Impulse                           \\
VAM                   & Vertical Ascent in Meters                  \\
VO2max                & Maximum Volume of Oxygen                   \\
WHO                   & World Health Organization                  \\
YLL                   & Years of Life Lost                         \\* \bottomrule
\end{longtable}

\section{Resources}

See next page for Table.

\begin{landscape}
\begin{longtable}[c]{@{}lll@{}}
\caption{Hardware, Software, Knowledge Sources, Data Tools, and Visualization Resources}
\label{tab:hardware_software}\\
\toprule
\textbf{Hardware}                & \textbf{Description}                                      & \textbf{Website}                         \\* \midrule
\endfirsthead
\endhead
\bottomrule
\endfoot
\endlastfoot
Garmin                           & GPS multi-sport smartwatch                                 & https://www.garmin.com/en-US/            \\
Garmin                           & Heart rate monitor with chest strap                       & https://www.garmin.com/en-US/            \\
Garmin                           & Pedaling cadence wireless sensor                          & https://www.garmin.com/en-US/            \\
23andMe                          & Health and ancestry genetic reports                       & https://www.23andme.com/                 \\
Concept2                         & Rowing ergometer, indoor rowing machine                   & https://www.concept2.com/indoor-rowers   \\
Stages                           & Bicycle crank with strain gauge                        & https://stagescycling.com/               \\
Elite                            & Cycling trainer that includes a power meter.          & https://www.elite-it.com/it              \\
AccuFitness                      & Measuring tape                                            & https://www.accufitness.com/             \\
Elite Medical                    & Protractor Goiniometer                                    & https://www.elitemedicalemi.com/         \\
Plumb Bob                        & A perpendicular reference line to gravity                 & https://en.wikipedia.org/wiki/Plumb\_bob \\
                                 &                                                           &                                          \\
{\ul \textbf{Software}}          &                                                           &                                          \\
Strava                           & Tracking physical activity using GPS data                 & https://www.strava.com/                  \\
Zwift                            & Online cycling/running physical training program          & https://zwift.com/                       \\
Sleep Cycle                      & Sleep pattern analysis and REM-based alarm clock          & https://www.sleepcycle.com/              \\
Nomie                            & Mood tracking app                                         & https://nomie.app/                       \\
Balance Health                   & Blood pressure data analysis                              & https://greatergoods.com/                \\
Weight Gurus                     & Weight tracking app connecting data from weight scale     & https://weightgurus.com/                 \\
Kinsa Smart                      & Smart body temperature readings and tracking              & https://www.kinsahealth.co/              \\
Nuvoair                          & Lung function data collection and analysis                & https://www.nuvoair.com/                 \\
Life Cycle                       & Tracking time on phone automatically                      & http://www.northcube.com/lifecycle/      \\
Apple Health Kit & Health informatics mobile application and integration & https://developer.apple.com/healthkit/ \\
Sensor Push                      & Monitors humidity, temperature, etc.                      & https://www.sensorpush.com/              \\
Lumen                            & Measuring metabolism through gas analyzer                 & https://www.lumen.me/                    \\
RescueTime                       & Tracking time on phone and computer                       & https://www.rescuetime.com/              \\
Concept2                         & Rowing data collection application                        & https://www.concept2.com/                \\
Amazon Alexa                     & Virtual assistant AI technology                           & https://www.amazon.com                           \\
Nest Thermostat                  & Smart thermostat                                          & https://nest.com/                        \\
Nest Protect                     & Smart smoke detector                                      & https://nest.com/                        \\
IFTTT                            & Chains of simple conditional statements                   & https://ifttt.com/                       \\
Phillips Hue                     & Wireless LED lamps                           & https://www2.meethue.com/                \\
Amazon                           & WiFi smart plugs                                          & https://www.amazon.com                           \\
Calorie Mama                     & Food image recognition app                                & https://www.caloriemama.ai/              \\
Google Maps                      & Satellite mapping and routing by GPS                      & https://www.google.com/maps              \\
Epic                             & Medications, tests, appointments, medical bills, pricing  & https://www.mychart.com/                 \\
BAC Track                        & Estimated blood alcohol level                             & https://www.bactrack.com/                \\
Hypoxico                         & Pulse oximeter with Bluetooth connectivity                 & https://hypoxico.com/                    \\
One Touch                        & Blood glucose monitor                                     & https://www.onetouch.com/                \\
Withings Health Mate             & Tracking and analysis of activity, sleep, weight, etc.    & https://www.withings.com/                \\
Time Out                         & Time tracking software                                    & https://cws-software.com/                \\
                                 &                                                           &                                          \\
{\ul \textbf{Knowledge Sources}} &                                                           &                                          \\
Cytoscape                        & Open source platform for visualizing complex networks     & https://cytoscape.org/                   \\
UptoDate                         & Point-of-care medical, evidence-based clinical resource   & https://www.uptodate.com/home            \\
Nature Reviews                   & Publishing Agency                                         & https://www.nature.com/                  \\
PubMed                           & Accessing database references/abstracts on life sciences. & https://www.ncbi.nlm.nih.gov/pubmed/     \\
DarkSky                          & Information on weather                                    &                                          \\
In the Sky                       & Information on daily sunset times                         & https://in-the-sky.org/sunrise.php       \\
Google Scholar                   & Web search engine of scholarly literature                 & https://scholar.google.com/              \\
Mendeley                         & Reference Manager                                         & https://www.mendeley.com/                \\
Aqicn                            & World-wide air quality monitoring data coverage           & https://aqicn.org/sources/               \\
EPA                              & Agency of the United States federal government            & https://www.epa.gov/                     \\
                                 &                                                           &                                          \\
{\ul \textbf{Data Analysis}}     &                                                           &                                          \\
QGIS                             & Open Source Geographic Information System                 & https://www.qgis.org/en/site/            \\
Golden Cheetah                   & Open Source Activity Data Analysis Platform               & https://www.goldencheetah.org/           \\
Google Sheets                    & Spreadsheet program                                       & https://www.google.com/sheets/about/     \\
Apple Photos                     & Photo download, editing platform                          & https://www.apple.com/ios/photos/        \\
Calorie Mama                     & Food image recognition and analysis                       & https://www.caloriemama.ai/              \\
Insight Maker                    & Creating dynamic simulation models                        & https://insightmaker.com/                \\
                                 &                                                           &                                          \\
{\ul \textbf{Visualization}}     &                                                           &                                          \\
Tableau                          & Data visualization software                               & https://www.tableau.com/                 \\
Gephi                            & Open-source network visualization software                & https://gephi.org/                       \\
Draw.io                          & Network drawing software                                  & https://www.draw.io/                     \\
Keynote                          & Drawing and Presentation Software                         & https://www.apple.com/keynote/           \\* \bottomrule
\end{longtable}
\end{landscape}

\end{document}